\documentclass[times, preprint, 10pt]{elsarticle}

\journal{Information Sciences}

\usepackage[numbers]{natbib}
\usepackage{float} % use H on floats
\usepackage[english, ruled, linesnumbered]{algorithm2e} % Algorithms
\usepackage{subcaption}
\usepackage{listings}
\usepackage[table]{xcolor}

\usepackage{makecell} % Table cells
\usepackage{multicol}
\usepackage[flushleft]{threeparttable}
\usepackage{pifont} % Checkmark

\usepackage{longtable,xtab,booktabs}
\usepackage[breaklinks=true]{hyperref}

\usepackage{xtab}

\usepackage{multirow}

\usepackage{amsmath}

\usepackage{comment}

\usepackage{rotating}

\def\tsc#1{\csdef{#1}{\textsc{\lowercase{#1}}\xspace}}
\tsc{WGM}
\tsc{QE}
\tsc{EP}
\tsc{PMS}
\tsc{BEC}
\tsc{DE}
\SetKw{KwTo}{to}
\SetKwFor{For}{for}{do}{} % disables "end"

\begin{document}
\let\WriteBookmarks\relax
\def\floatpagepagefraction{1}
\def\textpagefraction{.001}

%\shorttitle{A Feature %Selection Layer Approach for Multi-Class Classification on Neural Networks}

%\shortauthors{J. P. Silveira e Silva et~al.}
%{CV Radhakrishnan et~al.}}

\begin{frontmatter}

%\title [mode = title]{A Feature Selection Layer Approach for Multi-Class Classification on Neural Networks} 
\title{A Feature
Selection Layer Approach for Multi-Class Classification on Neural Networks} 



% \tnotetext[2]{The second title footnote which is a longer text matter
%    to fill through the whole text width and overflow into
%    another line in the footnotes area of the first page.}


% https://www.elsevier.com/researcher/author/policies-and-guidelines/credit-author-statement

\author[1]{João Pedro Silveira e Silva}%[orcid=0009-0004-4357-2268]
\ead{jpedross1999@gmail.com}
%\credit{Software, Investigation, Writing - Review and Editing, Visualization}

\author[1,2,3]{Márcio Dorn}%[orcid=0000-0001-8534-3480]
\ead {mdorn@inf.ufrgs.br}
%\credit{Conceptualization of this study, Resources, Writing - Review and Editing, Supervision, Project Administration, Funding Acquisition}

\author[1]{Bruno Iochins Grisci}%[orcid=0000-0003-4083-5881]
%\cormark[4]
\ead{bigrisci@inf.ufrgs.br}
%\credit{Conceptualization of this study, Resources, Methodology, Writing - Review and Editing, Project Administration, Funding Acquisition}




%\ead[url]{www.cvr.cc, cvr@sayahna.org}


\affiliation[1]{organization={Institute of Informatics}, 
    addressline={Federal University of Rio Grande do Sul}, 
    city={Porto Alegre},
    % citysep={}, % Uncomment if no comma needed between city and postcode
    %postcode={1043 NX}, 
    state={RS},
    country={Brazil}}

% \author[2,4]{Han Theh Thanh}[style=chinese]

% % Third author
% \author[2,3]{CV Rajagopal}[%
%    role=Co-ordinator,
%    suffix=Jr,
% \ead{cvr3@sayahna.org}
% \ead[URL]{www.sayahna.org}

\affiliation[2]{organization={Center for Biotechnology},
    addressline={Federal University of Rio Grande do Sul}, 
    city={Porto Alegre},
    % citysep={}, % Uncomment if no comma needed between city and postcode
    state={RS},
    country={Brazil}}

\affiliation[3]{organization={National Institute of Science and Technology - Forensic Science},
    %addressline={Mepukada}, 
    city={Porto Alegre},
    % citysep={}, % Uncomment if no comma needed between city and postcode
    state={RS},
    country={Brazil}}

% \author%
% [1,3]
% {Rishi T.}
% \cormark[2]
% \fnmark[1,3]
% \ead{rishi@stmdocs.in}
% \ead[URL]{www.stmdocs.in}

%\cortext[4]{Corresponding author}

\begin{abstract}
In the age of data-driven decision-making, deep learning models have become powerful tools for uncovering patterns and making predictions that influence many aspects of society.
However, their broader adoption in scientific and business applications is often limited by a lack of interpretability.
Feature weighting methods, which assess the relevance of individual features, are among the most widely studied approaches for improving interpretability.
While post hoc techniques have advanced significantly, especially in modeling nuanced relationships between features and labels in multiclass problems, embedded methods typically fall short in this regard.
Embedded techniques integrate feature relevance learning directly into the model training process, but they are often unable to capture label-specific relationships.
To address the limitation, we propose multiple extensions to the Feature Selection Layer, an embedded feature weighting mechanism for neural networks.
Through extensive experimentation, we identify the most effective extension and benchmark it against established feature weighting methods.
Our results show that the proposed method matches existing techniques in identifying globally relevant features and outperforms them in capturing label-specific relevance.
By leveraging the internal dynamics of the trained neural network, it reveals how individual features contribute to the prediction of specific labels, requiring no post-processing beyond the training process.
\end{abstract}

% Keywords

% \includegraphics[width=\linewidth]{Graphical_Abstract.pdf}

\end{frontmatter}

{\let\thefootnote\relax\footnotetext{\copyright~2027. This manuscript version is made available under the CC-BY-NC-ND 4.0 license. \\
\textbf{Cite as:} Silveira e Silva, J. P., Dorn, M., \& Iochins Grisci, B. (2027). A feature selection layer approach for multi-class classification on neural networks. \textit{Information Sciences}, 763, Article 124273. \url{https://doi.org/10.1016/j.ins.2026.124273}}}

%\maketitle

\section{Introduction}
\label{sec:introduction}
Over the past decade, the exponential growth of data driven by the Internet has increased the successful application of deep neural networks across diverse fields, including computer vision, natural language processing, robotics, and healthcare \cite{wason_deep_evolution:2018}.
However, state-of-the-art models often operate as ``black boxes'', lacking interpretability.  
In critical domains like healthcare or human resources, high predictive accuracy alone is insufficient \cite{miotto_deep_health_care:2018}. 
Hidden biases, such as a model inappropriately relying on demographic attributes like gender, can severely compromise a model's reliability and exacerbate social issues \cite{understanding_deep_learning:2023}. 
Furthermore, neural networks are prone to learning ``shortcuts'', achieving high classification performance by exploiting data artifacts rather than capturing the true underlying relevance of the features \cite{shortcut:2020}.
Regulations like the EU General Data Protection Regulation (GDPR) increasingly mandate a ``right to explanation'' for automated decisions, making interpretability not just a scientific preference, but a legal necessity \cite{emre_the_right_to_context:2022}.

These challenges have catalyzed extensive research in Interpretable Machine Learning (IML).
IML methods generally fall into two categories: local methods, which explain individual predictions, and global methods, which elucidate the overall behavior of the model across a dataset.
Among global methods, feature weighting is a widely adopted approach that assigns importance scores to input features \cite{molnar_interpretable:2020}.
When applied to neural networks, feature weighting clarifies the decision-making process and can enhance generalization by mitigating the influence of noisy features.
This is particularly relevant for tabular data. 
Unlike images or text, tabular data lacks spatial or sequential structure, which poses unique challenges for representation learning.
Consequently, standard neural networks often underperform on these datasets due to their high sensitivity to uninformative attributes \cite{ANIPS:2022}.

\textbf{Problem:} Formally, in a multi-class tabular dataset, a neural network learns a mapping from an input matrix $X \in \mathbb{R}^{N \times M}$ ($N$ samples, $M$ features) to a set of discrete labels $y \in \{1, \dots, L\}$ ($L$ possible labels or classes).
Most global feature weighting methods generate a single global feature relevance vector $w \in \mathbb{R}^{M}$, assigning one score per feature.
However, in multi-class tasks, feature importance frequently varies across different labels, a feature highly discriminative for one label may act as noise for another \cite{relevance_aggregation:2021}.
An ideal approach should instead produce a label-specific relevance matrix $W \in \mathbb{R}^{M \times L}$, where $W_{m,l}$ indicates the relevance of a feature $m$ specifically for a label $l$.
A robust label-specific method must filter general noise, highlight unique label-defining features, maintain  predictive performance, and provide granular interpretability without requiring multiple separate models.

\textbf{Research gap:} A prominent method to handle feature weighting within neural networks is the Feature Selection Layer (FSL) \cite{figueroa_feature_selection_layer:2021}, an embedded technique that autonomously learns feature relevance from the model's internal dynamics during training.
However, the current FSL architecture is fundamentally constrained by its design: it only learns a single general weight per feature (a global relevance vector).
Consequently, it cannot discern whether a feature is universally important or only relevant to a specific subset of labels, limiting its utility in uncovering complex, label-dependent patterns.

\textbf{Proposed method:} To bridge this gap, we propose an architectural extension to the FSL that explicitly captures label-specific relevance.
Instead of passing input features through a single one-dimensional weight vector, our extension embeds a two-dimensional matrix directly into the network.
During training, the network dynamically adjusts these weights via gradient descent, learning to route and scale features differently depending on the specific label pathway.
Consequently, the network simultaneously classifies the data and builds a transparent matrix mapping exactly which features drive the prediction of each specific label.
Evaluations on both synthetic and real-world high-dimensional tabular datasets demonstrate that our approach performs on par with established techniques in identifying general features and significantly outperforms them in capturing label-specific relevance, all while maintaining or improving overall predictive performance.

\textbf{This study addresses the research question:} 
\emph{``Can an embedded FSL be capable of learning label-specific feature relevance?''} 
To answer this, we address the following complementary questions: 
\begin{enumerate} 
\label{list:introduction_research_questions} 
    \item ``How to evaluate a label-specific feature weighting algorithm?'' (Section \ref{sec:methods_and_concepts})
    \item ``How to adapt FSL to learn label-specific feature relevance?'' (Section \ref{sec:proposed_extensions}) 
    \item ``How does this approach compare to other algorithms?'' (Section \ref{sec:experiments_and_results})
\end{enumerate}

Guided by these research questions, and to provide a comprehensive roadmap of our proposed approach, the main contributions of this paper are:
\begin{itemize}
\item We introduce a novel label-specific extension to the Feature Selection Layer (FSL), expanding its learnable parameters to map features to specific labels dynamically during training.
\item We consolidate and adapt evaluation methodologies to rigorously assess the quality of both general and label-specific feature weighting techniques.
\item We validate the proposed approach on synthetic and real-world high-dimensional tabular data, demonstrating its efficacy against established state-of-the-art baselines.
\end{itemize}

The remainder of this work is structured as follows: Section \ref{sec:methods_and_concepts} reviews interpretability and evaluation methodologies; Section \ref{sec:related_work} outlines related studies; Section \ref{sec:proposed_extensions} details multiple possible label-specific FSL extensions; Section \ref{sec:experiments_and_results} presents experiments and results; and Section \ref{sec:conclusios} concludes the paper, discussing limitations and future work.

%%%%%%%%%%%%%%%%%%%%%%%%%%%%%%%%%%%%%%%%%%
\section{Methods and concepts}
\label{sec:methods_and_concepts}
This section outlines the theoretical foundation of this work, covering interpretability (Section \ref{subsec:interpretability_and_fairness}), feature weighting as a tool for interpretability (Section \ref{subsec:feature_weighting_and_selection}), and evaluation metrics for validating feature weighting techniques with emphasis on multi-class problems (Section \ref{subsec:feature_weighting_evaluation}). 

\subsection{Interpretability}
\label{subsec:interpretability_and_fairness}
Interpretability is a broad and challenging concept to define. \citet{murdoch_definitions_interpretably_ml:2019} describe interpretability in Machine Learning (ML) as the extraction of relevant knowledge from a model, relating to relationships either contained in the data or learned by the model. 
Interpretable Machine Learning (IML) methods can be used to discover knowledge, to debug or justify the model and its predictions, and to improve it \cite{molnar_interpretable:2020}.
IML methods can be local or global. 
Local methods explain individual predictions, such as Shapley values.
Global methods, on the other hand, explain the overall model behavior and how it behaves on average for a given dataset. 
One popular type of global explanation is feature weighting, which weights features based on their relevance to the predictions \cite{molnar_interpretable:2020}.
In this context, \citet{barbieri_analysis_feature_selection:2024} and \citet{relevance_aggregation:2021} highlight the importance of considering multi-class problems while ranking feature relevance. 
When dealing with different labels, the relevance of features can vary, and acknowledging these variations enhances model interpretability.
Most state-of-the-art machine learning algorithms are ``black-box'' models, with complex structures that make their learned patterns difficult to explain. 
This lack of interpretability can lead to distrust and conceal bias.

\subsection{Feature weighting and selection}
\label{subsec:feature_weighting_and_selection}
One effective approach to enhancing interpretability is feature weighting, which involves algorithms that assign continuous relevance scores to features based on their importance for  predictions.
Feature selection is a subset of this approach, using binary weights to retain or discard features \cite{tahir_simultaneous_selection_and_weighting:2007}.
Both are widely used in machine learning, pattern recognition, statistics, and data mining for identifying relevant features, improving accuracy, reducing computation, and enhancing interpretability \cite{miao_survey_feature_selection:2016}. 
Unlike feature extraction, these methods preserve the original input space, resulting in more interpretable models \cite{barbieri_analysis_feature_selection:2024}.
According to \citet{miao_survey_feature_selection:2016}, feature weighting algorithms are usually grouped into five categories: filter, wrapper, embedded, ensemble and hybrid categories.
Another approach uses post-hoc techniques, which define feature weights using trained models by understanding the relevance that they gave to each feature based on one or multiple instances.
In this work, we choose five consolidated algorithms to evaluate our new approach: Kruskal–Wallis filter, ReliefF, Decision Trees, Random Forests and Least Absolute Shrinkage and Selection Operator (LASSO).
Further details on these algorithms are provided in \ref{sec:feature_weighting_algorithms}.

\subsection{Feature weighting evaluation} \label{subsec:feature_weighting_evaluation}
To assess and compare feature weighting algorithms, we adapted the evaluation framework proposed by \citet{barbieri_analysis_feature_selection:2024}, originally designed for binary classification, to handle multi-class problems by including additional metrics. 
Our evaluation follows four complementary strategies.

(1) \textbf{Identification of informative features:} 
In real-world, high-dimensional datasets, the exact mapping of relevant features to specific labels is often unknown, making it difficult to objectively evaluate feature weighting algorithms. 
To overcome this, we utilize synthetic datasets where the ground-truth informative features for each label are explicitly defined. 
This precise knowledge allows us to explicitly track how effectively an algorithm recovers these relevant features while ignoring noise. 
To quantify this, we evaluate top-$n$ feature subsets (ranked by their assigned weights) using two complementary metrics.
\textbf{Percentage of Informative Features Selected (PIFS)} acts as a measure of recall (or sensitivity). 
It calculates the proportion of the total explicitly informative features that are successfully included in a given selected subset (see \autoref{eq:percentage_of_informative_features_selected}). 
A higher PIFS score indicates a better recovery of the underlying relevant features. 
A score of 1.0 signifies that all genuinely informative features for that label have been successfully identified and included in the subset.
\textbf{Percentage of Selected Features that are Informative (PSFI)} acts as a measure of precision. 
It evaluates the purity of the selected subset by calculating the proportion of its features that are genuinely informative (see \autoref{eq:percentage_of_selected_features_that_are_informative}).
A higher PSFI score indicates that the algorithm is highly accurate in its weighting, assigning high ranks strictly to relevant features rather than noisy or irrelevant ones. 
These metrics are uniquely appropriate for evaluating label-specific feature selection because they can be calculated individually against the ground-truth feature set of each specific label. 
By calculating PIFS and PSFI across incrementally larger subset sizes, we can observe the exact threshold where an algorithm stops selecting label-specific informative features and begins incorporating global noise.

(2) \textbf{Feature erasure and predictive power:} 
Following established perturbation analysis for tabular data \cite{samek_evaluating_w_heatmap:2016}, we evaluate feature relevance by masking specific inputs. 
This is achieved by setting feature values to zero after normalization, and measuring the impact on the model performance.
The computational simplicity of this approach is critical, as feature erasure evaluation can be prohibitive for the high-dimensional datasets used in our experiments.
Furthermore, we evaluate the predictive power of different feature subsets by training models as SVMs using the most relevant features identified.

(3) \textbf{Weighted t-SNE and Silhouette Coefficient \cite{WTSNE:2025}:} An extension of t-Distributed Stochastic Neighbor Embedding (t-SNE), a widely used technique for visualizing high-dimensional data in 2D. 
Weighted t-SNE modifies the standard approach by incorporating feature relevance scores (normalized between 0 and 1) into the Euclidean distance calculation.
Consequently, features with higher relevance exert a greater influence on the spatial arrangement of points in the final visualization, while less relevant features contribute less to the distance between points.
To quantify the cluster separation quality, the weighted t-SNE results are presented alongside the Silhouette Coefficient. 
This metric ranges from -1 to 1.
A value close to 1 indicates that a point is well-matched to its assigned cluster, while 0 implies the point is near a decision boundary and -1 suggests the point may be assigned to the wrong cluster.
Compared to other visualization techniques, t-SNE has some important advantages that make it preferable for feature weighting, such as its ability to capture non-linear data relationships and its robust performance on clustering separation \cite{revisiting_dimensionality_reduction_techniques:2022}.
The primary goal of weighted t-SNE is to provide a global visual inspection tool for comparing the impact of different feature weighting techniques on data clustering, being already applied in different works as \cite{relevance_aggregation:2021},
\cite{interpretability_with_relevance_aggregation:2022} and \cite{evaluating_the_impact_of_data_transformation:2023}.
Therefore, the analysis is comparative. 
The visualization of the weighted dataset should be contrasted with the original (unweighted) t-SNE projection.
Regarding implementation, we adhere to the guidelines by \citet{the_art_of_using_tsne:2019} and \citet{WTSNE:2025}. 

(4) \textbf{Algorithm stability:} Given the importance of reproducibility, particularly in fields such as Bioinformatics, we assess stability via two methods.
First, we apply data perturbation using K-Fold Cross-Validation as proposed by \citet{awada_review_feature_selection_stability:2012}. 
We train each feature weighting algorithm on different fold combinations and compare the generated feature weights using the Jaccard index, Spearman’s rank correlation, and Pearson correlation.
Secondly, we employ the $m-1$ approach, being $m$ the number of features.
After initially training the algorithm, we modify the dataset by removing the top-ranked feature. 
We then execute the algorithm again and compare the newly generated weights against the original results using the same stability metrics.
More details are provided in \ref{sec:feature_selection_metrics}.

\section{Related work}
\label{sec:related_work}
This section reviews key research on extracting feature weights, focusing on interpretability.
The majority of existing methods capable of assigning label-specific feature weights rely on post-hoc analysis.
LIME is a popular post-hoc, model-agnostic method for explaining individual predictions of black-box models \cite{ribeiro_lime:2016}. 
It works by locally approximating the model around a specific input using a simple interpretable model.
This involves perturbing the input features and observing the impact in the model's output, estimating feature relevance \cite{slack_fooling_shap_lime:2020}.
Though designed for local explanations, LIME can also support global feature weighting by aggregating feature weights from multiple representative inputs \cite{shin_feasibility_lime_feature_selection:2023}.
Another widely used post-hoc technique is SHAP, which encompasses multiple local approximation methods using Shapley values as a measure of feature relevance.
Here, we focus on DeepSHAP. 
While the original SHAP is a model-agnostic method, DeepSHAP leverages the inherent structure of a neural network to enhance computational efficiency \cite{lindberg_shap:2017}.
Relevance Aggregation is another post-hoc algorithm designed to improve the interpretability of neural networks trained on tabular data based on the Layer-wise Relevance Propagation (LRP) \cite{bach_lrp:2015}.
LRP was designed primarily for local interpretability on image datasets.
Unlike images, where the order of pixels and their spatial relationships are crucial, tabular data features are often independent of their order.
Additionally, the same information in an image, as a pixel, can have varying meanings, a characteristic absent in tabular data.
By leveraging this fixed representation, Relevance Aggregation combines the feature relevance weights computed from multiple samples, producing global label-specific feature weights \cite{relevance_aggregation:2021}.
Unlike other post-hoc techniques, FeatureX integrates model perturbation and feature correlation analysis to quantify feature importance \cite{featurex:2025}. 
Perturbation measures each feature’s impact on performance, while correlation highlights associations and redundancies. 
Compared to LIME and SHAP, FeatureX achieved slight improvements in prediction and up to 9 times faster execution, mainly on complex models where SHAP performance degrades.
SAGE is a post-hoc, model-agnostic feature weighting algorithm designed to estimate global feature importance.
It quantifies predictive power while accounting for complex feature interactions.
Like SHAP, SAGE derives from game-theoretic Shapley values, however, it is specific for global rather than local feature importance.
This approach makes SAGE orders of magnitude faster than naive SHAP.
It also demonstrates superior performance compared to methods such as Feature Ablation, Mean Importance, and Permutation Tests \cite{SAGE:2020}.
Unlike post-hoc methods, embedding methods integrate feature weighting during training, improving computational efficiency.
However, they typically lack the ability to define label-specific feature relevance.
Being designed for high-dimensional biological data, FsNet is a nonlinear general feature selection network.
It comprises a selection layer composed of concrete random variables, which are continuous relaxations of one-hot vectors.
A reconstruction layer is also included to stabilize the training process.
During training, FsNet selects features using the selection layer while simultaneously maximizing classification accuracy and minimizing reconstruction error \cite{singh_fsnet:2023}.
By also incorporating a selection layer, Feature Selection Layer (FSL) \cite{figueroa_feature_selection_layer:2021}, Algorithm \ref{alg:fsl}, is another embedded approach for general feature weighting which adds a layer without bias after the input layer where each neuron has a one-to-one correspondence with a feature.
Unlike FsNet, FSL does not employ a reconstruction or any other additional layer, resulting in a simplified implementation with lower impact on model training.
During prediction, before joining in the network, each feature is multiplied by its correspondent weight that is trained jointly with the rest of the network's parameters and represents the feature relevance.
FSL demonstrated superior performance when compared to other methods like Random Forest, ReliefF, Mutual Information and the Attention-based Feature Selection (AFS) \cite{figueroa_feature_selection_layer:2021}.
AFS uses neural networks with attention modules to assign general feature weights by formulating feature-label correlations as binary classification problems supported by a shallow attention network for each feature.
For small datasets, weight assignment may be limited, so a hybrid initialization using normalized weights from other methods was proposed by the authors \cite{gui_afs:2019}.
LassoNet represents a distinct approach among embedded methods, capable of inferring label-specific feature importance.
Unlike the standard Lasso, which is limited to linear models, LassoNet extends feature weighting to neural networks with multiple hidden layers.
It achieves this by incorporating skip (residual) layers and enforcing a hierarchical constraint where a feature may only participate in any non-linear hidden layer if its skip-layer representative is active.
Furthermore, the method employs a modified objective function to integrate feature selection directly into the training process \cite{LassoNet:2021}.
Unlike a generic embedded method, TabNet is an interpretable deep learning architecture designed for tabular data that employs a sequential attention mechanism to select salient features at each decision step. 
By jointly performing feature selection and output mapping, the model provides both local and global interpretability regarding feature importance.
Empirical evaluations demonstrate that TabNet yields superior predictive performance compared to established tree-based ensembles like XGBoost, LightGBM and CatBoost, while offering more robust feature weighting than post-hoc methods such as LIME and Integrated Gradients \cite{Tabnet:2021}.
DOvE \cite{aguilar_class:2024}, an extension of One-versus-Each (OvE) methods, address the limitations of label-independent selection by employing a relevance matrix, named ``three-layer label-specific classification'', where classification models can be used to discriminate between different combinations of labels given the selected features to generate the final output.
LSFSFI \cite{li_label_specific_multi_label:2024} addresses the limitations of global feature selection in multi-label datasets, where each sample can have multiple labels simultaneously.
According to the authors most feature selection methods used for multi-label datasets consider a single feature space for all labels, leading to suboptimal solutions since different labels can depend on different features.
LSFSFI addresses this issue by using a partial fuzzy mutual implication granularity information to model feature-label dependencies, integrating this as prior knowledge into an embedded multi-label feature selection model.

\begin{algorithm}[t]
\caption{
\textbf{Feature Selection Layer (FSL) training.}
A single feature weighting vector $W$ is learned jointly with the downstream classifier $f_\phi$: at each step, $W$ is passed through a ReLU to produce a non-negative gate $W^+$, which element-wise multiplies the input features before they reach the classifier. An $\ell_1$ penalty on $W$, scaled by $\lambda$, drives the gate toward sparsity, while the classifier is trained to minimize a class-frequency-weighted cross-entropy loss. After training for $E$ epochs, the final activated gate $W^+ = \mathrm{ReLU}(W)$ is returned as the learned per-feature importance scores.
}
\label{alg:fsl}
\KwIn{Training set $\mathcal{D}_{\text{train}} = \{(\mathbf{x}_i, y_i)\}_{i=1}^{n}$, regularization strength $\lambda$, learning rate $\eta$, epochs $E$}
\KwOut{Learned feature weights $W^+ \in \mathbb{R}^d$}
Initialize feature weighting vector parameters $W \gets \mathbf{1}_d$\;
Initialize classifier parameters $\phi$ randomly\;
\For{$\textit{epoch} = 1$ \KwTo $E$}{
    \ForEach{mini-batch $(\mathbf{X}, \mathbf{y}) \in \mathcal{D}_{\text{train}}$}{
        $W^+ \gets \mathrm{ReLU}(W)$ \tcp*{non-negative feature gate}
        $\mathbf{X}' \gets \mathbf{X} \odot W^+$ \tcp*{gate the input features}
        $\hat{\mathbf{y}} \gets f_\phi(\mathbf{X}')$ \tcp*{shared classifier}
        $\mathcal{L} \gets \mathrm{CE}_{\text{balanced}}(\hat{\mathbf{y}}, \mathbf{y}) + \lambda \lVert W \rVert_1$ \tcp*{loss + regularization}
        Update $(W, \phi)$ via backpropagation on $\mathcal{L}$ (AdamW, learning rate $\eta$)\;
    }
}
\KwRet{$W^+ = \mathrm{ReLU}(W)$}
\end{algorithm}

\section{Proposed extensions}
\label{sec:proposed_extensions}
As discussed in Section \ref{subsec:feature_weighting_and_selection}, feature weighting is essential for interpreting models and handling high-dimensional, low-sample datasets. 
While global feature weighting suits single-class tasks, it often fails in multi-class problems, especially under label imbalance, harming minority class performance.
We propose four extensions to FSL that enable label-specific feature weighting.
These enhance both multi-class accuracy and interpretability while preserving FSL's core properties by leveraging internal model dynamics.
Although these are essentially feature weighting models, we chose to retain the name ``Feature Selection Layer'' to explicitly reference the adapted method.

\subsection{Listener extension}
\label{subsec:listener_extension}
\begin{figure*}[h]
    \centering
    \includegraphics[width=0.9\textwidth]{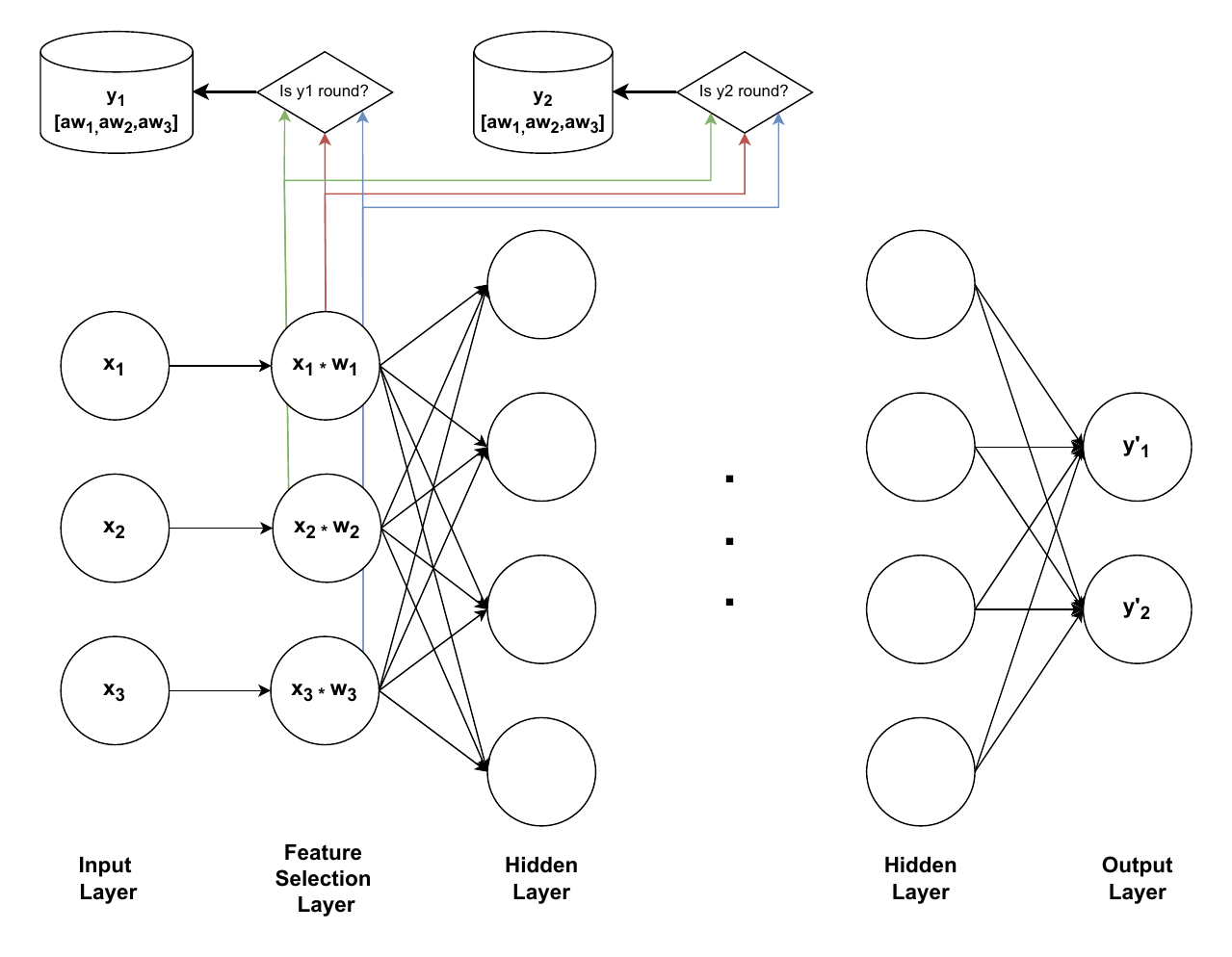}
    \caption[Representation of Listener extension.
    Representation of Listener extension.
    A neural network is shown with three inputs, followed by a FSL, multiple hidden layers, and two output neurons.
    Above the neural network, two databases are displayed, each associated with a single label and containing an accumulator for each weight in the Feature Selection Layer.
    During the backpropagation process, the change in the FSL weights is stored in one of the databases based on the label of the training data used.
    In the end of the process those databases will contain the feature relevance given by the FSL for each label.]{\textbf{Representation of Listener extension.} 
    In this example, two listeners, one per label, accumulate weight updates for each of the three features ($aw_1 \ldots aw_3$) based on the label in the current batch. The FSL remains unchanged. For $M$ features and $L$ labels, $y_L[{aw}_1,\ldots,{aw}_M]$ stores the accumulated weights per feature-label pair.
    }
    \label{fig:feature_selection_layer_extension_listener}
\end{figure*}

We begin by introducing the Listener, with the objective of gaining deeper insights into the feature weight learning process within the original FSL implementation.
Unlike the other extensions, which intervene directly in the forward pass to compute label-specific weights during training, Listener leaves the forward pass unchanged from the base FSL: a single shared weight vector $w$ gates the input for every batch, regardless of label. 
Its label-specific signal is instead obtained passively, by observing how that shared vector moves while training is restricted to one label at a time. 
Each training step processes a batch containing samples from a single class. 
Between each training forward pass, Listener records the weight vector's current value, and after the optimizer updates it, it accumulates the resulting shift into a per-label accumulator dedicated to that batch's class composed of a matrix $W_{L \times M}$, where $L$ is the number of labels and $M$ is the number of features (\autoref{fig:feature_selection_layer_extension_listener}). 
Over the course of training, each label's accumulator stores the cumulative updates derived specifically from its own data. 
Rather than acting as a FSL extension, it functions as an observer, inferring a label's influence on a single, shared FSL by measuring how much it perturbs the model.
While it remains relevant to understand how individual labels influence the FSL weights, it is crucial to distinguish this tool's specific role. 
It acts strictly as an analyzer of feature weight updates to the original FSL, rather than a true extension that modifies the underlying architecture, unlike the methods detailed in the following sections.

\subsection{Residual extension}
\label{subsec:residual_extension}
\begin{figure*}[h]
    \centering
    \includegraphics[width=0.9\textwidth]{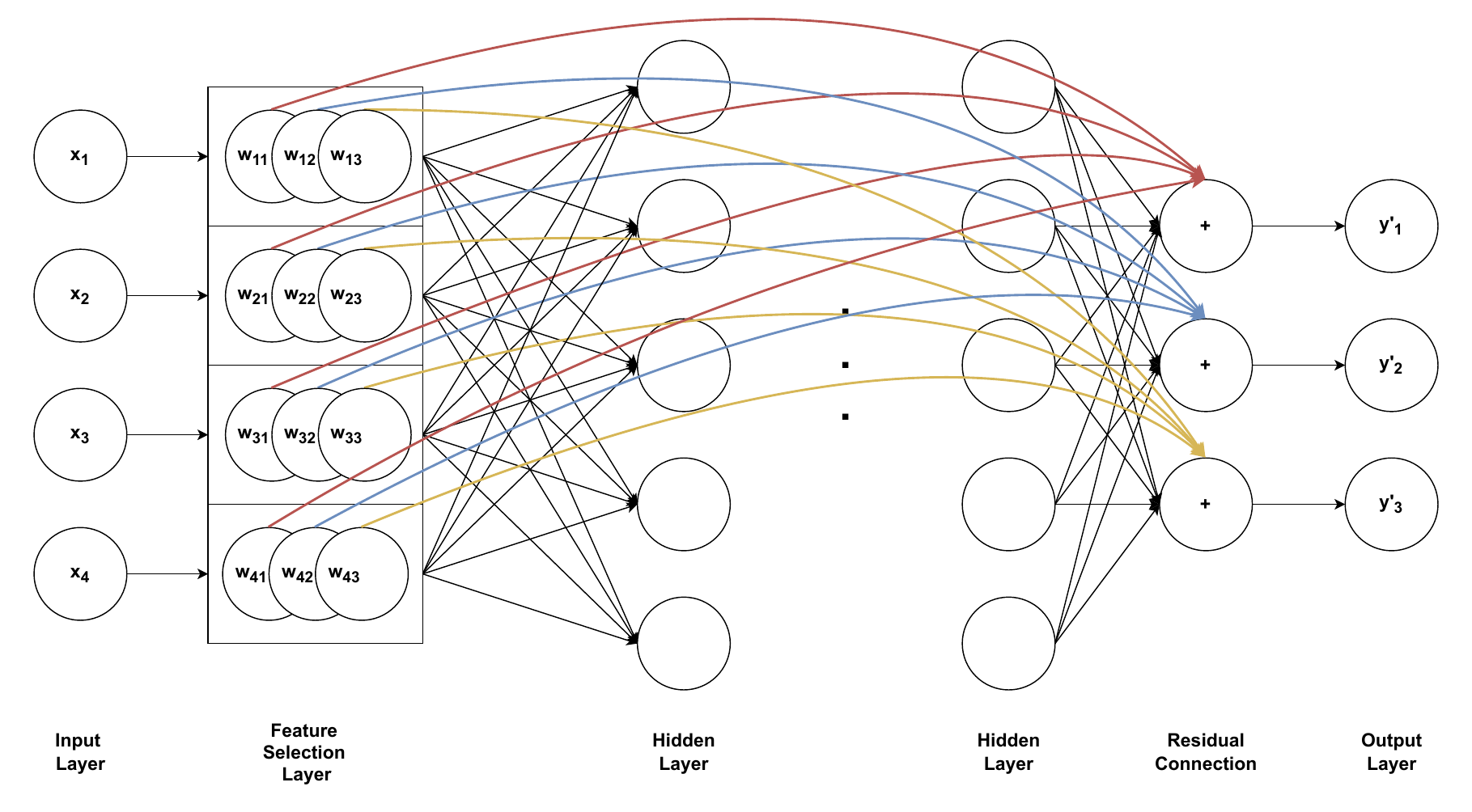}
    \caption[Representation of Residual extension.
    A neural network is shown with four inputs, followed by a FSL, multiple hidden layers, and three output neurons.
    Unlike the original implementation, this FSL consists of a matrix where each row contains a weight for each label in the dataset.
    For example, for input x1, there are three neurons with weights w11, w12, and w13 corresponding to outputs y1, y2, and y3, respectively.
    Each neuron in the FSL is connected via a residual connection to the final layer of the network, where it is added to the network's output for its corresponding label.
    During the backpropagation process, the weights are updated along with the other parameters of the neural network.
    ]{\textbf{Representation of Residual extension.} In this example, three weights were added for each feature on FSL, one per label. The colored lines represents the connection from the feature weights, already multiplied by the feature value, to their specific label output.}
    \label{fig:feature_selection_layer_residual}
\end{figure*}

Residual learning addresses the vanishing gradient problem commonly encountered in deep neural networks, by incorporating ``shortcut connections'' between layers, which allows gradients to flow more easily through the network \cite{he_residual_networks:2016}.
We extend the FSL by introducing a weight matrix $W_{L \times M}$, where $L$ is the number of labels and $M$ is the number of features (\autoref{fig:feature_selection_layer_residual}). 
Each row should learn the feature relevance for a single label.
To enable the model to learn different feature weights for each label, we incorporated a shortcut connection that computes the sum of the FSL and the last hidden layer outputs to generate the model's output for each label.
To ensure compatibility between the dimensions of the outputs, we employed a linear projection as specified by \citet{he_residual_networks:2016}.
The residual connection adds, for each label, the label-specific weight vector $W_l$ multiplied by the input $X$ to the output $l$ of the last hidden layer, where $l$ is one of the $L$ labels.
This process influences the output values for each specific label and the backpropagation process as the gradients can flow directly back through the residual connection to the Feature Selection Layer. 
Since Residual exclusively records label-specific feature weights, general feature weights are determined by averaging the corresponding label-specific weights, the same happens to Output Aware (Section \ref{subsec:output_aware_extension}) and Multiple Feature Selection Layer (Section \ref{subsec:multiple_feature_selecion_layer_extension}).

\subsection{Output Aware extension}
\label{subsec:output_aware_extension}
\begin{figure*}[t]
    \centering
    \includegraphics[width=0.9\textwidth]{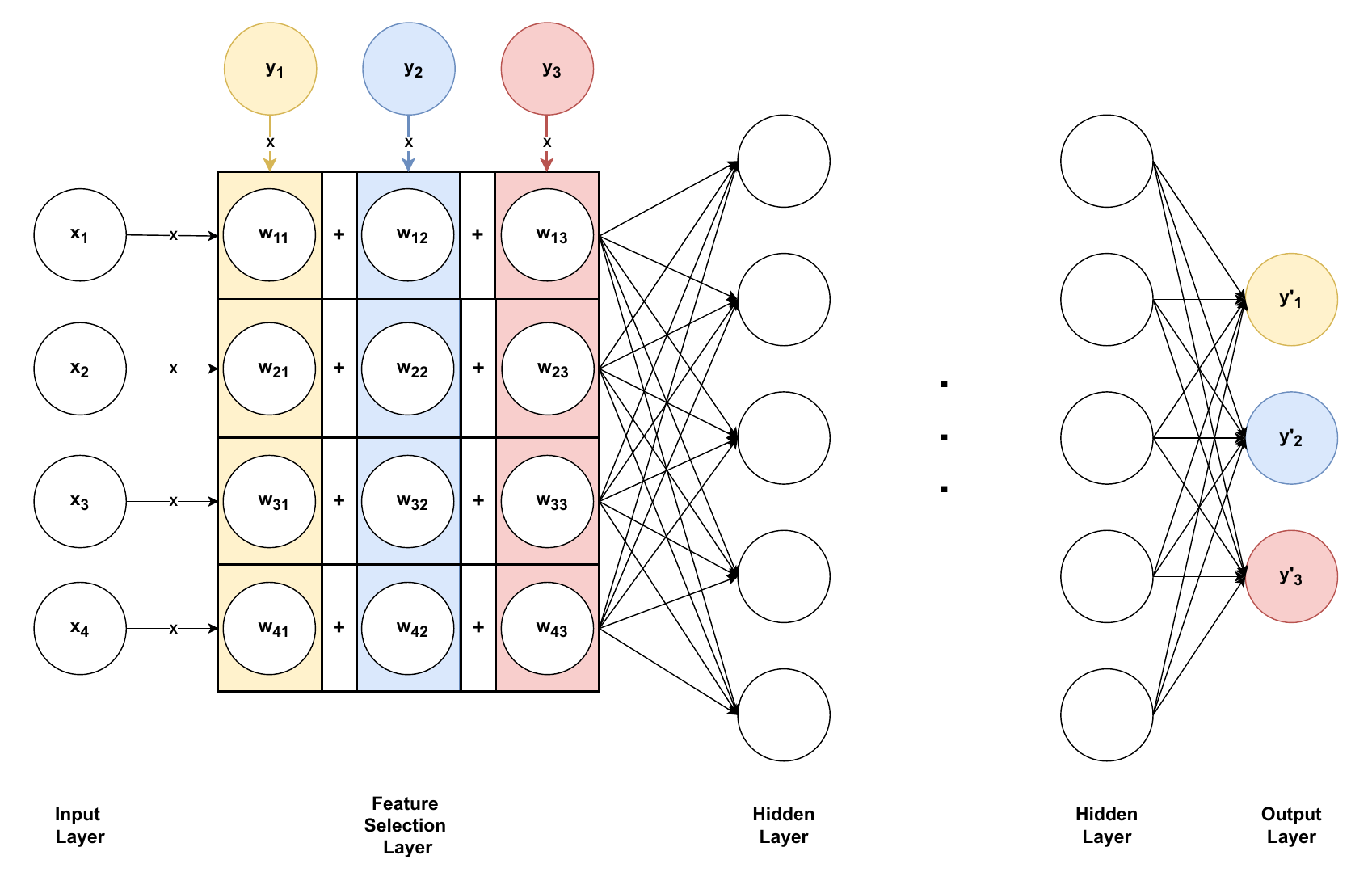}
    \caption[Representation of Output Aware extension.
    A neural network is shown with four inputs, followed by three FSL, multiple hidden layers, and three output neurons.
    Each FSL is created according to the original implementation.
    However, during the training phase, the neural network has knowledge of the output for each data sample and selects only the corresponding FSL during the feedforward process.
    Above each FSL, the expected output value for each class is shown, with the output encoded using one-hot encoding.
    The FSL's weight vectors are multiplied by the expected output value of their corresponding label, causing the FSLs unrelated to the expected output to have their weight values zeroed out.
    After this multiplication, all FSLs are summed, and the result is multiplied by the input, feeding into the rest of the network.
    Each FSL has a distinct color, yellow, blue, and red.
    These colors correspond to the label each one is associated with.
    The output neurons of the network share the same colors.
    During the backpropagation process, the weights are updated along with the other parameters of the neural network.]
    {
    \textbf{Representation of Output Aware extension.} 
    FSL is implemented with multiple weight columns, each corresponding to a specific label (yellow, blue, and red). During training, a one-hot encoded vector ($y_1$, $y_2$, $y_3$) indicates the expected label, which will have value $1$, while the others will be $0$. 
    Each one-hot encoded vector multiplies the weight matrix, effectively zeroing out weights not related with the current input's expected label.
    After being trained, the input features are multiplied by the sum of its feature weights.
    }
    \label{fig:output_aware_feature_selection_layer}
\end{figure*}

The proposed extension also leverages a feature weight matrix, $W_{L \times M}$ (\autoref{fig:output_aware_feature_selection_layer}),
where each row represents the feature weights associated with a specific label.
During training, the expected output $y$ is utilized to dynamically select the row that corresponds to the respective label from $W_{L \times M}$ to be multiplied with the input before it is fed to the hidden layers.
The weights are updated alongside all other network parameters during backpropagation.
Given a batch of $N$ training samples the process has the following steps: 
(1) The expected output vector $\overrightarrow{y_N}$ for the batch of $N$ samples is converted into a one-hot encoded matrix $Y_{N \times L}$. 
Each row in $Y_{N \times L}$ represents a sample, with the value one in the column that corresponds to the expected label and the value zero elsewhere.
(2) The one-hot encoded matrix $Y_{N \times L}$ is multiplied by the feature weight matrix $W_{L \times M}$ to produce a new matrix $V_{N \times M}$ (\autoref{eq:output_aware_matrix_equation}) which will be composed by $N$ rows, one for each sample, and $M$ columns with a single weight for each feature that corresponding to sample's label.
(3) The matrix $V_{N \times M}$ is then multiplied element-wise with the input matrix $X_{N \times M}$, weighting the input features based on their relevance to the expected label.
(4) After training, to keep using the feature relevance information for future predictions, all columns of the weight matrix $W_{L \times M}$ are summed together to form a single, general weight vector which will be used for predictions.
\begin{equation} 
    \label{eq:output_aware_matrix_equation}
    \resizebox{0.8\textwidth}{!}{
        $
        \begin{bmatrix}
           y_{11} & y_{12} & \cdots & y_{1L}\\
           y_{21} & y_{22} & \cdots & y_{2L}\\
           \vdots & \vdots & \ddots & \vdots\\ 
           y_{N1} & y_{N2} & \cdots & y_{NL}\\ 
        \end{bmatrix}
        \times
        \begin{bmatrix}
           w_{11} & w_{12} & \cdots & w_{1M}\\
           w_{21} & w_{22} & \cdots & w_{2M}\\ 
           \vdots & \vdots & \ddots & \vdots\\ 
           w_{L1} & w_{L2} & \cdots & w_{LM} 
        \end{bmatrix}
        =
        \begin{bmatrix}
           v_{11} & v_{12} & \cdots & v_{1M}\\
           v_{21} & v_{22} & \cdots & v_{2M}\\ 
           \vdots & \vdots & \ddots & \vdots\\ 
           v_{N1} & v_{N2} & \cdots & v_{NM} 
        \end{bmatrix}
        $
    }
\end{equation}

\subsection{Multiple Feature Selection Layer extension}
\label{subsec:multiple_feature_selecion_layer_extension}

\begin{figure*}[!b]
    \centering
    \includegraphics[width=0.9\textwidth]{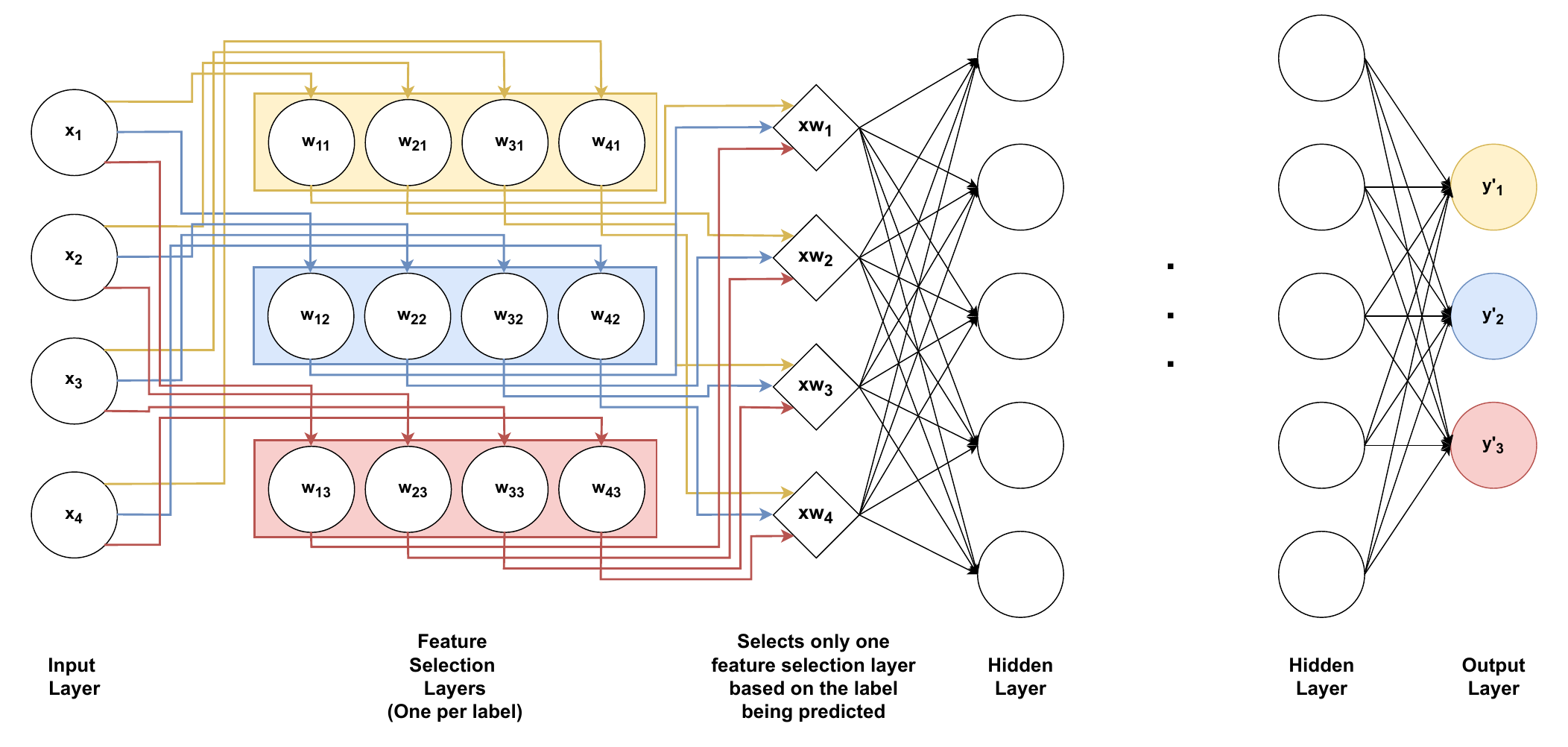}
    \caption[Representation of Multiple Feature Selection Layer extension.
    A neural network is shown with four inputs, followed by three horizontally displayed FSLs, a layer representing the multiplication of each input by a weight from one of the FSLs, multiple hidden layers, and three output neurons.
    Each FSL is created according to the original implementation.
    However, during the training phase, the neural network performs the feedforward process multiple times.
    In each execution, it uses the weights from only one of the FSLs, which are multiplied by the input, passed through the hidden layers, and produce an output solely for the corresponding label.
    At the end of the process, each output is generated using only the FSL associated with its label.
    Each FSL has a distinct color, yellow, blue, and red.
    These colors correspond to the label each one is associated with.
    The output neurons of the network have the same colors.
    During the backpropagation process, the weights are updated along with the other parameters of the neural network.]
{\textbf{Representation of Multiple Feature Selection Layer extension.} 
    Multiple selection layers are created, one for each label described by colors yellow, blue and red. 
    The internal network is executed three times before backpropagation, one for each possible label.
    Only the feature weights related to the label being estimated are selected at each execution.
    }
    \label{fig:multiple_feature_selection_layer}
\end{figure*}

The Multiple Feature Selection Layer (Multiple FSL) extends the base model with one dedicated FSL gate per label, while keeping the underlying hidden and output layers shared across every label (\autoref{fig:multiple_feature_selection_layer}). 
Concretely, each label $l \in \{1, \dots, L\}$ owns a learnable weight vector $\theta_l \in \mathbb{R}^M$, collected into a matrix $\Theta = [\theta_1; \dots; \theta_L] \in \mathbb{R}^{L \times M}$, where $M$ is the number of input features and $L$ the number of labels. 
As defined in Section \ref{subsec:regulatization_and_activation_functions}, a ReLU activation clips each weight at zero,
\begin{equation}
    w_l = \mathrm{ReLU}(\theta_l), \qquad l = 1, \dots, L,
    \label{eq:multiple_fsl_gate}
\end{equation}
so that $w_l$ acts as a non-negative, per-label feature gate: a feature $k$ with $w_{l,k} = 0$ is entirely excluded from label $l$'s prediction, regardless of its relevance to any other label.
Given an input $\mathbf{x} \in \mathbb{R}^M$, every gate is applied independently and propagated through the shared network $f : \mathbb{R}^M \rightarrow \mathbb{R}^L$ (the hidden and output layers common to every label):
\begin{equation}
    \mathbf{z}_l = f(\mathbf{x} \odot w_l), \qquad l = 1, \dots, L,
    \label{eq:multiple_fsl_propagation}
\end{equation}
where $\odot$ denotes the element wise (Hadamard) product. Each $\mathbf{z}_l \in \mathbb{R}^L$ is a full output vector, but only its $l$-th entry (the one computed from label $l$'s own gated view of $\mathbf{x}$) is kept:
\begin{equation}
    \hat{y}_l = [\mathbf{z}_l]_l = [f(\mathbf{x} \odot w_l)]_l, \qquad l = 1, \dots, L.
    \label{eq:multiple_fsl_output}
\end{equation}
The model is trained end-to-end with the classification loss (cross-entropy) plus an $\ell_1$ penalty (Section \ref{subsec:regulatization_and_activation_functions}) over $\Theta$,
\begin{equation}
    \mathcal{L} = \mathcal{L}_{\mathrm{CE}}(\hat{\mathbf{y}}, y) + \lambda \lVert \Theta \rVert_1,
    \label{eq:multiple_fsl_loss}
\end{equation}
which drives each label's gate toward using only the features that remain informative for that label.

Because \autoref{eq:multiple_fsl_propagation} evaluates $f$ once per label, a single prediction requires $L$ full propagations through the shared network instead of one, increasing both training and inference cost linearly in $L$. 
This sequential evaluation, Multiple FSL as defined above, is the less memory consuming implementation, since only one label's propagation is ever held in memory at a time, and is therefore useful when GPU memory is limited. 
However, all experiments reported in this article instead use a mathematically identical but substantially faster implementation which issues the $L$ propagations as a single batched matrix operation rather than $L$ sequential ones. 
Concretely, it stacks every label's gated view of $\mathbf{x}$ into one matrix,
\begin{equation}
    \tilde{X} = \left(\mathbf{1}_L \mathbf{x}^\top\right) \odot W \in \mathbb{R}^{L \times M}, \qquad W = \begin{bmatrix} w_1^\top \\ \vdots \\ w_L^\top \end{bmatrix} \in \mathbb{R}^{L \times M},
    \label{eq:multiple_fslp_stack}
\end{equation}
where $\mathbf{1}_L \in \mathbb{R}^L$ is the all-ones vector, so that row $l$ of $\tilde{X}$ is exactly $\mathbf{x} \odot w_l$. The shared network is then evaluated once, applied row-wise to $\tilde{X}$,
\begin{equation}
    Z = f(\tilde{X}) \in \mathbb{R}^{L \times L}
    \label{eq:multiple_fslp_propagation}
\end{equation}
and the final prediction is recovered by reading off the diagonal of $Z$, which is identical to \autoref{eq:multiple_fsl_output}. \autoref{eq:multiple_fslp_propagation} performs the same $L$ forward passes as the sequential formulation, but as one matrix multiplication over $L$ rows instead of $L$ separate calls to $f$, which GPUs parallelize far more efficiently, at the cost of holding all $L$ rows of $\tilde{X}$ (and their intermediate activations) in memory simultaneously.

Unlike Output Aware, once training converges the weight gates $w_1, \dots, w_L$ are fixed and reused unchanged for every subsequent prediction.
The predicted class is therefore never computed from a single, generic representation of $\mathbf{x}$, but from $L$ label-specific ones, each shaped by the feature weighting that training assigned to that label. 
This holds regardless of which class ends up being predicted: every candidate label's score is produced using its own dedicated, already-learned gate, so the label-specific feature weighting learned during training carries over intact to inference, for every class, on every sample.

\subsection{Regularization, activation functions and initial weights}
\label{subsec:regulatization_and_activation_functions}
\begin{figure*}[h]
    \centering
     \begin{subfigure}[b]{0.49\textwidth}
         \centering
         \includegraphics[width=\textwidth]{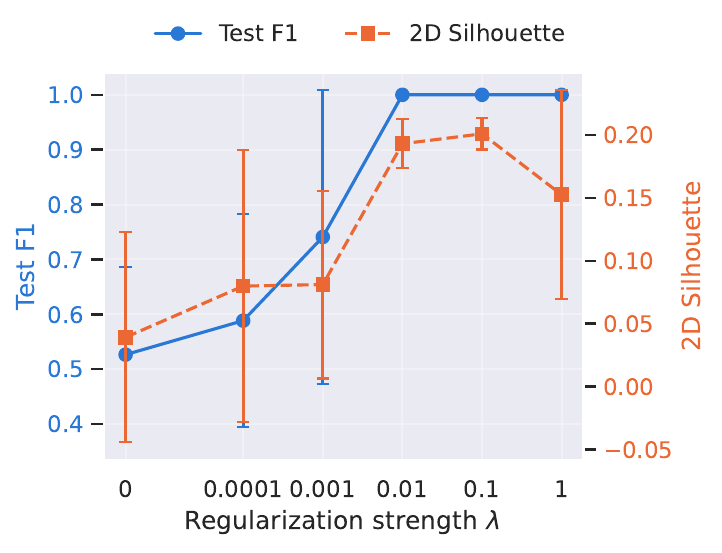}
         \caption{XOR}
         \label{subfig:ablation_regularization_f1_silhouette_xor}
     \end{subfigure}
     \begin{subfigure}[b]{0.49\textwidth}
         \centering
         \includegraphics[width=\textwidth]{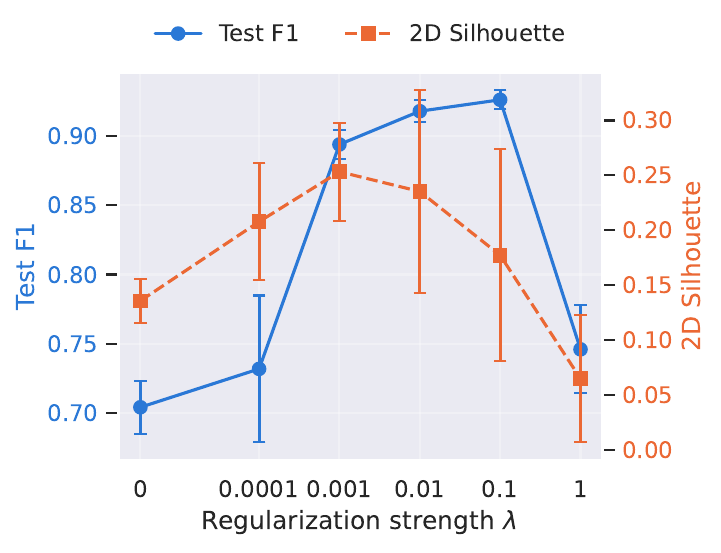}
         \caption{SynthA}
         \label{subfig:ablation_regularization_f1_silhouette_synthA}
     \end{subfigure}
     \begin{subfigure}[b]{0.49\textwidth}
         \centering
         \includegraphics[width=\textwidth]{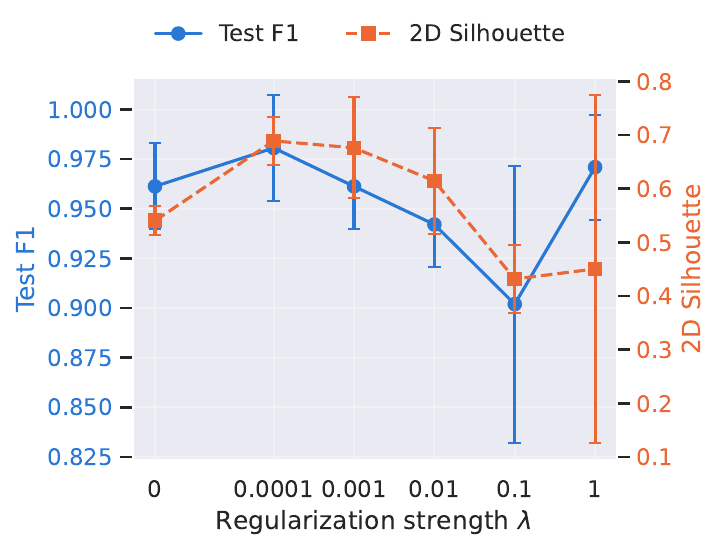}
         \caption{Breast}
         \label{subfig:ablation_regularization_f1_silhouette_isolet}
     \end{subfigure}
     \begin{subfigure}[b]{0.49\textwidth}
         \centering
         \includegraphics[width=\textwidth]{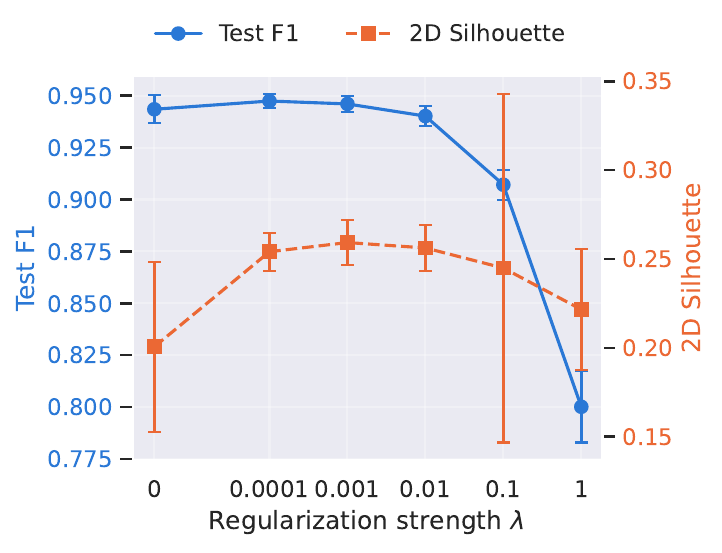}
         \caption{Isolet}
         \label{subfig:ablation_regularization_f1_silhouette_breast}
     \end{subfigure}
     \caption[
        The image presents a panel of four line graphs—labeled (a) XOR, (b) SynthA, (c) Breast, and (d) Isolet, illustrating the effect of regularization strength on machine learning model performance. 
        Each graph plots two separate metrics against regularization strength on a logarithmic x-axis, with values marked at 0, 0.0001, 0.001, 0.01, 0.1, and 1.
        The two metrics shown in each graph are the test F1 score represented by a solid blue line with circular markers, corresponding to the left y-axis, and the 2D silhouette score represented by a dashed orange line with square markers, corresponding to the right y-axis.
        Both lines include vertical error bars indicating the standard deviation over five executions.
        On XOR the Test F1 score steadily increases and plateaus at a perfect 1.0 at a regularization strength of 0.01. 
        The 2D Silhouette score also trends upward but peaks later at a strength of 0.1 before dipping slightly.
        For SynthA the Test F1 score peaks around a regularization strength of 0.1 before sharply declining. 
        Conversely, the 2D Silhouette score peaks earlier at 0.001 and declines steadily thereafter.
        On Breast both scores start high and peak at a very low regularization strength of 0.0001. 
        As regularization increases to 0.1, both scores generally trend downward, though the Test F1 score sharply recovers at a strength of 1.
        Finally, on Isolet the Test F1 score remains consistently high and stable up to a regularization strength of 0.01, after which it drops significantly. 
        The 2D Silhouette score reaches a mild peak at 0.001 and then steadily decreases.
    ]
    {\textbf{Effect of regularization strength.}
        Test F1 (left axis, solid) and high-dimensional silhouette score (right axis,
        dashed) as a function of regularization strength $\lambda$, per dataset (mean $\pm$ std over
        5 executions).
        Where the two curves diverge, the $\lambda$ that
        maximizes classification accuracy is not the $\lambda$ that maximizes feature-space
        separability.
    }
    \label{fig:ablation_regularization_f1_silhouette}
\end{figure*}

The original FSL incorporate an adapted $l_1$ regularization term exclusively for its weights. 
In this work, we employ the default $l_1$ regularization without adaptations.
This encourages sparsity and feature weights to become zero or near-zero, effectively eliminating irrelevant features \cite{sun_design_lasso:2016}.
\autoref{tab:ablation_regularization} details a study evaluating Multiple FSL on datasets with a wide range of scales (500–7,797 samples, 50–54,677 features, 2–26 classes) across various regularization strengths. 
The range $\lambda \in \{10^{-3}, 10^{-2}\}$ was the only one statistically tied for the best test F1-score on every dataset, all other values suffered significant performance drops on at least one dataset. 
The extremes proved uniformly detrimental for different reasons. 
Without regularization ($\lambda=0$), the absence of selection pressure caused sharp F1-score drops on XOR (0.53 vs. 1.00) and SynthA (0.70 vs. 0.93). 
Conversely, excessive regularization ($\lambda=1$) collapsed the weight vector toward a constant, degrading performance on SynthA (0.93 to 0.75) and Isolet (0.95 to 0.80). 
Additionally, \autoref{fig:ablation_regularization_f1_silhouette} highlights a tradeoff between predictive accuracy and feature-space separability. 
On SynthA, the highest silhouette score is achieved at $\lambda=10^{-3}$, while the highest F1-score is achieved at $\lambda=10^{-1}$.
Isolet, on the other hand, exhibits greater agreement between the test F1 and 2D silhouette scores, which both reach their peak values between $\lambda=10^{-4}$ and $\lambda=10^{-2}$. 
On XOR, both curves rise together until $\lambda=10^{-2}$, where the F1-score saturates at 1.00, whereas the silhouette score continues to increase at $\lambda=10^{-1}$ before decreasing at $\lambda=1$. 
For the Breast dataset, the metrics track each other closely, with the exception of a significant divergence at $\lambda=1$. Ultimately, while there is no universally optimal value, these findings support $\lambda \in \{10^{-3}, 10^{-2}\}$ as the recommended default.
For the activation function we adopted ReLU, which sets negative values to zero, setting all activated feature weights to be zero or positive.
In our experiments, this configuration yielded superior performance.
However, as can be noticed on \autoref{tab:ablation_activation} other activations such as Tanh can achieve similar results, for this reason the activation function remains modular allowing different configurations. 
For instance, allowing negative weights can be valuable for analyzing negative contribution.
In our experiments, we observed that an initial weight of $1$ consistently achieved an equal or better prediction score and feature weighting than the original $1/m$ initialization used by FSL, where $m$ is the number of features (as shown in \autoref{tab:ablation_1m}). Consequently, we adopt a fixed initial weight of 1 for our experiments.

\begin{table}[h]
\centering
\caption{
\textbf{Test F1-score, mean $\pm$ standard deviation over 5 executions, for Multiple FSL across varying regularization strengths $\lambda$.}
In bold is the best average per dataset.
Best results with statistical significance ($p < 0.01$) according to the Kruskal-Wallis and Dunn’s test are highlighted in orange.
}
\label{tab:ablation_regularization}
\begin{tabular}{lllll}
\toprule
$\lambda$ & XOR & SynthA & Isolet & Breast \\
\midrule
$0.0000$ & 0.526 $\pm$ 0.160 & 0.704 $\pm$ 0.019 & \cellcolor{orange!20}0.943 $\pm$ 0.007 & \cellcolor{orange!20}0.961 $\pm$ 0.022 \\
$0.0001$ & 0.588 $\pm$ 0.194 & 0.732 $\pm$ 0.053 & \cellcolor{orange!20}\textbf{0.947 $\pm$ 0.004} & \cellcolor{orange!20}\textbf{0.981 $\pm$ 0.026} \\
$0.0010$ & \cellcolor{orange!20}0.741 $\pm$ 0.268 & \cellcolor{orange!20}0.894 $\pm$ 0.010 & \cellcolor{orange!20}0.946 $\pm$ 0.004 & \cellcolor{orange!20}0.961 $\pm$ 0.022 \\
$0.0100$ & \cellcolor{orange!20}\textbf{1.000 $\pm$ 0.000} & \cellcolor{orange!20}0.918 $\pm$ 0.008 & \cellcolor{orange!20}0.940 $\pm$ 0.005 & \cellcolor{orange!20}0.942 $\pm$ 0.021 \\
$0.1000$ & \cellcolor{orange!20}\textbf{1.000 $\pm$ 0.000} & \cellcolor{orange!20}\textbf{0.926 $\pm$ 0.007} & 0.907 $\pm$ 0.007 & \cellcolor{orange!20}0.902 $\pm$ 0.070 \\
$1.0000$ & \cellcolor{orange!20}\textbf{1.000 $\pm$ 0.000} & 0.746 $\pm$ 0.032 & 0.800 $\pm$ 0.017 & \cellcolor{orange!20}0.971 $\pm$ 0.026 \\
\bottomrule
\end{tabular}
\end{table}

\begin{table}[h]
\centering
\caption{\textbf{Effect of the activation function on Multiple FSL across datasets calculated over 5 executions.}
In bold is the best average per dataset and metric, among activation functions.
Best results with statistical significance ($p < 0.01$) according to the Kruskal-Wallis and Dunn’s test are highlighted in orange.}
\label{tab:ablation_activation}
\begin{tabular}{lllll}
\toprule
Dataset & Activation & Test F1 & Test Loss & 2D Silhouette \\
\midrule
\multirow{3}{*}{XOR} & ReLU & \cellcolor{orange!20}\textbf{1.0000 $\pm$ 0.0000} & \cellcolor{orange!20}\textbf{0.0004 $\pm$ 0.0007} & \cellcolor{orange!20}0.1931 $\pm$ 0.0195 \\
 & Sigmoid & 0.4235 $\pm$ 0.0746 & 0.7011 $\pm$ 0.0083 & 0.0028 $\pm$ 0.0024 \\
 & Tanh & \cellcolor{orange!20}0.9980 $\pm$ 0.0045 & \cellcolor{orange!20}0.0029 $\pm$ 0.0061 & \cellcolor{orange!20}\textbf{0.1974 $\pm$ 0.0177} \\
\midrule
\multirow{3}{*}{SynthA} & ReLU & \cellcolor{orange!20}0.9168 $\pm$ 0.0111 & \cellcolor{orange!20}\textbf{0.4945 $\pm$ 0.0733} & \cellcolor{orange!20}0.1656 $\pm$ 0.0240 \\
 & Sigmoid & 0.7090 $\pm$ 0.0119 & 0.7701 $\pm$ 0.0168 & \cellcolor{orange!20}0.1472 $\pm$ 0.0178 \\
 & Tanh & \cellcolor{orange!20}\textbf{0.9187 $\pm$ 0.0097} & \cellcolor{orange!20}0.5199 $\pm$ 0.0428 & \cellcolor{orange!20}\textbf{0.1971 $\pm$ 0.0693} \\
\midrule
\multirow{3}{*}{Isolet} & ReLU & \cellcolor{orange!20}0.9461 $\pm$ 0.0038 & \cellcolor{orange!20}0.1956 $\pm$ 0.0124 & \cellcolor{orange!20}\textbf{0.2592 $\pm$ 0.0127} \\
 & Sigmoid & 0.9197 $\pm$ 0.0039 & 0.3021 $\pm$ 0.0154 & -0.1497 $\pm$ 0.0275 \\
 & Tanh & \cellcolor{orange!20}\textbf{0.9494 $\pm$ 0.0061} & \cellcolor{orange!20}\textbf{0.1885 $\pm$ 0.0153} & \cellcolor{orange!20}0.2534 $\pm$ 0.0140 \\
\midrule
\multirow{3}{*}{Breast} & ReLU & \cellcolor{orange!20}\textbf{0.9807 $\pm$ 0.0265} & \cellcolor{orange!20}\textbf{0.0948 $\pm$ 0.1350} & \cellcolor{orange!20}0.6937 $\pm$ 0.0494 \\
 & Sigmoid & \cellcolor{orange!20}0.9613 $\pm$ 0.0216 & \cellcolor{orange!20}0.3730 $\pm$ 0.3584 & 0.5106 $\pm$ 0.0614 \\
 & Tanh & \cellcolor{orange!20}0.9710 $\pm$ 0.0265 & \cellcolor{orange!20}0.1326 $\pm$ 0.1241 & \cellcolor{orange!20}\textbf{0.7137 $\pm$ 0.0423} \\
\bottomrule
\end{tabular}
\end{table}

\begin{table}[h]
\centering
\caption{\textbf{Comparison of Multiple FSL with weight init of $1$ and $1/m$ ($m$ is the number of input features) across four synthetic datasets and the real world Breast dataset.}
In bold is the best average per number of features.
Best results with statistical significance ($p < 0.01$) according to the Kruskal-Wallis and Dunn’s test are highlighted in orange.
}
\label{tab:ablation_1m}
\begin{tabular}{lllll}
\toprule
Features & Init. Weight & Test F1 & Test Loss & 2D Silhouette \\
\midrule
\multirow{2}{*}{50} & $1$ & \cellcolor{orange!20}\textbf{1.000 $\pm$ 0.000} & \cellcolor{orange!20}\textbf{0.000 $\pm$ 0.000} & \cellcolor{orange!20}0.648 $\pm$ 0.021 \\
 & $1/m$ & \cellcolor{orange!20}0.667 $\pm$ 0.456 & \cellcolor{orange!20}0.440 $\pm$ 0.602 & \cellcolor{orange!20}\textbf{0.682 $\pm$ 0.027} \\
\midrule
\multirow{2}{*}{500} & $1$ & \cellcolor{orange!20}\textbf{1.000 $\pm$ 0.000} & \cellcolor{orange!20}\textbf{0.000 $\pm$ 0.000} & \cellcolor{orange!20}0.725 $\pm$ 0.008 \\
 & $1/m$ & \cellcolor{orange!20}0.833 $\pm$ 0.373 & \cellcolor{orange!20}0.220 $\pm$ 0.493 & \cellcolor{orange!20}\textbf{0.726 $\pm$ 0.007} \\
\midrule
\multirow{2}{*}{5000} & $1$ & \cellcolor{orange!20}\textbf{0.998 $\pm$ 0.001} & \cellcolor{orange!20}\textbf{0.003 $\pm$ 0.002} & \cellcolor{orange!20}0.600 $\pm$ 0.038 \\
 & $1/m$ & \cellcolor{orange!20}0.997 $\pm$ 0.002 & \cellcolor{orange!20}0.016 $\pm$ 0.013 & \cellcolor{orange!20}\textbf{0.601 $\pm$ 0.072} \\
\midrule
\multirow{2}{*}{50000} & $1$ & \cellcolor{orange!20}\textbf{0.976 $\pm$ 0.018} & \cellcolor{orange!20}\textbf{0.086 $\pm$ 0.038} & \cellcolor{orange!20}\textbf{0.350 $\pm$ 0.216} \\
 & $1/m$ & 0.946 $\pm$ 0.013 & \cellcolor{orange!20}0.162 $\pm$ 0.057 & \cellcolor{orange!20}0.099 $\pm$ 0.076 \\
\midrule
\multirow{2}{*}{54675 (Breast)} & $1$ & \cellcolor{orange!20}\textbf{0.981 $\pm$ 0.026} & \cellcolor{orange!20}\textbf{0.095 $\pm$ 0.135} & \cellcolor{orange!20}\textbf{0.694 $\pm$ 0.049} \\
 & $1/m$ & \cellcolor{orange!20}0.942 $\pm$ 0.021 & \cellcolor{orange!20}0.187 $\pm$ 0.063 & \cellcolor{orange!20}0.648 $\pm$ 0.041 \\
\bottomrule
\end{tabular}
\end{table}

\section{Experiments and results}
\label{sec:experiments_and_results}
This section presents experiments to evaluate the proposed FSL extensions compared with existing feature weighting algorithms.
We start by introducing the utilized datasets (Section \ref{subsec:datasets}).
Then we compare the extensions with the original FSL implementation, evaluate their capacity to select label-specific feature weights (Section \ref{subsec:comparative_analysis_between_proposed_extensions}), and select the best-performing one (Section \ref{subsection:proposed_extensions_results}).
Next, we compare it with consolidated and established feature weighting techniques (Section \ref{subsec:comparison_with_existent_feature_weighting_algorithms}). 
Finally, we discuss the obtained results (Section \ref{subsection:results}).
All experiments were conducted using Python on a workstation equipped with an Intel Core i5-14600K CPU, 128 GB of RAM, and an NVIDIA GeForce RTX 5070 Ti GPU with 16 GB of VRAM.
Decision Tree, Random Forest and Lasso were implemented and optimized using the Scikit-Learn library. 
Kruskal Wallis and ReliefF implementations were based on \citet{barbieri_analysis_feature_selection:2024}'s framework.
LIME, DeepSHAP, LassoNet, TabNet and SAGE were implemented with their respective author’s libraries and recommendations, being optimized for each dataset.
Further details on the classification and mechanics of each feature weighting method are available in \autoref{tab:selector-comparison}.

Weighted t-SNE evaluations were performed following the author's recommended parameters.
Neural network models were trained employing AdamW optimizer and label-weighted cross-entropy loss.
For all neural networks, the same model architecture was used, differing only by the inclusion of the feature weighting algorithm.
Input data was normalized when needed.
For all experiments, we ensured a separation between learning and evaluation. 
Feature weights were computed using only the training subsets, while the test subsets, comprising $20\%$ of synthetic data and $15\%$ of real-world data, remained unseen.
This procedure was applied across all methods ensuring the use of the same data partitions.
More details about the experiments parameters are provided in \ref{sec:complementary_experiments}.

\subsection{Datasets}
\label{subsec:datasets}
To evaluate and compare the proposed FSL extensions against existing feature weighting algorithms, we utilized both synthetic and real-world datasets.
While the final objective of any feature weighting method is to demonstrate effectiveness on real-world datasets, validation on synthetic data is a common practice in this field \cite{identification_of_full_and_partial_class_relevant_genes:2010} and serves as a critical first step. 
It enables controlled experiments where experimental conditions, such as the proportion of irrelevant features, can be systematically varied, and provide ground truth knowledge of the optimal feature set, allowing for a rigorous and confident assessment of the method's accuracy \cite{synthetic_data_for_feature_selection:2023}.
In this work we propose four synthetic datasets.
The first is a binary XOR dataset, with only two relevant features.
It tests the algorithm's ability to capture strong non-linear feature interactions.
To evaluate performance on multi-class problems, we generate three synthetic datasets (\autoref{tab:synthetic_used_datasets}) using the algorithm from \citet{guyon_design_experiments_make_classification:2003} adapted by Scikit-Learn.
Each of the three datasets comprises 30 relevant features.
SynthA provides a common set of relevant features for all labels.
SynthB utilizes a label-specific feature approach, assigning 10 different features for each label. 
SynthC incorporates both methodologies, assigning 15 relevant features per label.
In addition to these synthetic datasets, we also generated simpler datasets using the algorithm proposed by \citet{guyon_design_experiments_make_classification:2003}, varying only the number of features and labels for ablation studies (Tables \ref{tab:ablation_regularization}, \ref{tab:ablation_1m}, \ref{tab:ablation_time} and \ref{tab:ablation_memory}).
For real-world evaluation, we use three microarray datasets from CuMiDa \cite{feltes_cumida:2019}, one binary and two multi-class, characterized by high dimensionality and limited sample sizes, and two multi-class benchmark datasets from UCI Machine Learning Repository, Letter Recognition \cite{letter_recognition:1991} and Isolet \cite{isolet:1991}, with larger sample sizes and lower-to-moderate dimensionality (\autoref{tab:cumida_label_number}).
For simplicity, labels were replaced with numerical identifiers.
The default t-SNE results for each dataset are provided in Figures \ref{fig:synthetic_tsne} and \ref{fig:cumida_tsne} of \ref{sec:complementary_experiments}.

\begin{table}[b]
\centering
\caption[Synthetic datasets with defined number of relevant features (``Relevant'' column) and relevant features per label (``Per label'' column).] {Synthetic datasets with defined number of relevant features (Relevant) and relevant features per label (Per label).}
\label{tab:synthetic_used_datasets}
\begin{tabular}{lllllll}
\hline
Name   & Feature Type & Labels & Features & Relevant & Per label & Samples \\ \hline
XOR    & Binary       & 2       & 50       & 2                & 2                          & 500     \\
SynthA & Numerical    & 3       & 100     & 30                & 30                          & 3000     \\
SynthB & Numerical    & 3       & 300     & 30                & 10                          & 3000     \\
SynthC & Numerical    & 3       & 300     & 30               & 15                          & 3000     \\ \hline
\end{tabular}
\end{table}

\begin{table}[h]
\centering
\caption[
Real-world datasets selected from CuMiDa and the UCI Machine Learning Repository. Liver, Isolet, and Letter Recognition are class-balanced, whereas Colorectal and Breast exhibit greater imbalance across classes.
The table presents the labels per dataset, the identifier that will be used as numbers and the number of samples for each label.
] 
{\textbf{Real-world datasets selected from CuMiDa and the UCI Machine Learning Repository.} Liver, Isolet (298-300 samples per class), and Letter Recognition (734-813 samples per class) are class-balanced, whereas Colorectal and Breast exhibit greater imbalance across classes.
}
\label{tab:cumida_label_number}
\begin{tabular}{llllll}
    \hline
    Dataset         & Features      & Labels        & Identifier     & Samples \\ 
    \hline
    \multirow{2}{*}{Liver 22405} & \multirow{2}{*}{22284}       
                    & normal                        & 0 & 24 \\
                    & & heptocarcinoma                & 1 & 24 \\
    \hline
    \multirow{4}{*}{Colorectal 41657} & \multirow{4}{*}{33468}
                    & normal                        & 0 & 12  \\
                    & & high\_grade\_dysplasia          & 1 & 28\\
                    & & low\_grade\_dysplasia           & 2 & 21 \\
                    & & adenocarcinoma                & 3 & 25 \\
    \hline
    \multirow{6}{*}{Breast 45827} & \multirow{6}{*}{54676}
                    & Basal                         & 0 & 41\\
                    & & HER                           & 1 & 30\\
                    & & cell\_line                     & 2 & 14 \\
                    & & normal                        & 3  & 7\\
                    & & luminal\_A                     & 4  & 29\\
                    & & luminal\_B                     & 5  & 30 \\
    \hline
    Isolet                 & 617           & Letters from A to Z & 0-25   & 7797 \\
    \hline
    Letter Recognition      & 16            & Letters from A to Z & 0-25  & 20000 \\
\hline
\end{tabular}
\end{table}

\subsection{Comparative analysis between proposed extensions} \label{subsec:comparative_analysis_between_proposed_extensions}
This section presents a series of experiments to evaluate the proposed extensions and the original FSL implementation.
To validate an ideal FSL extension, we aim to answer two key questions: (1) Do the general feature weights contribute meaningfully to model predictions as in the original FSL implementation? (2) Do the label-specific feature weights significantly impact mostly individual label predictions?

We start by evaluating the feature weighting capabilities of each extension on the XOR dataset.
Since the FSL is integrated and trained alongside the neural network, its capacity to learn relevant features is expected to be directly correlated with the predictive performance.
On our tests all algorithms consistently achieved the highest performance on the XOR dataset, achieving precision, recall, and F1 scores of 1.0 on test data.
When analyzing the Percentage of Informative Features Selected (PIFS) and the Percentage of Selected Features that are Informative (PSFI), both metrics reached a perfect score of 1 across all algorithms and executions considering only the two highest-weighted features. 
Meaning that in the XOR dataset, all methods consistently prioritized the two key features.
\autoref{tab:comparative_stability_on_extensions_for_xor} presents the Pearson and Spearman scores computed over all features. 
It also reports the Jaccard score for the top-$2$ features, a threshold chosen to match the number of informative features in the dataset.
All algorithms achieved the highest correlation value on Jaccard, confirming that all of them were able to assign the highest weights to the two relevant features in all 11 executions consistently.
The observed values near to $1$ on Pearson means a high correlation on the feature weights given to all features, while the values near to $0$ on Spearman means a low correlation between the ranking generated from the feature weights.
Apart from the relevant features, almost all the other features had their weights close to zero, which explains the high correlation values given by Pearson.
Additional experiments involving the XOR datasets are provided in Figures \ref{fig:comparative_informative_features_graph_on_extensions_for_xor}, \ref{fig:heatmap_on_extensions_for_xor}, \ref{fig:feature_position_per_execution_on_extensions_for_xor} and \ref{fig:weighted_tsne_on_extensions_for_xor} of \ref{sec:complementary_experiments}.

\begin{table}[h]
\centering
\caption[Stability comparison between algorithms feature weights on XOR dataset.] {\textbf{Stability comparison between algorithms feature weights on XOR dataset.}
Jaccard similarity measures set overlap, ranging from 0 (no overlap) to 1 (identical sets), being used to assess the similarity between the top-n highest-weighted features.
Spearman accesses ranking similarity, ranging from -1 (perfect inverse correlation) to 1 (perfect correlation). 
Pearson measures linear correlation between the feature weights, ranging from -1 (perfect negative correlation) to 1 (perfect positive correlation) where 0 means no correlation.
In bold are the best values for each metric.
}
\label{tab:comparative_stability_on_extensions_for_xor}
\begin{tabular}{l|lll}
\hline
Algorithm    & Stability metric   &   Features &    Score \\ 
\hline
\multirow{3}{*}{FSL} 
& Pearson            &                             50 &  .990 \\ 
& Spearman           &                             50 &  .098 \\ 
\cline{2-4}
& Jaccard            &                             2 &  \textbf{1}  \\ 
\hline
\multirow{3}{*}{Listener}  
& Pearson            &                             50 &  .987   \\ 
& Spearman           &                             50 &  \textbf{.150} \\  
\cline{2-4}
& Jaccard            &                             2 & \textbf{1} \\ 
\hline
\multirow{3}{*}{Residual}
& Pearson            &                             50 &  \textbf{.994} \\ 
& Spearman           &                             50 &  .106 \\ 
\cline{2-4}
& Jaccard            &                             2 &  \textbf{1}       \\ 
\hline
\multirow{3}{*}{Output Aware}
& Pearson            &                             50 &  .981 \\ 
& Spearman           &                             50 &  .145 \\ 
\cline{2-4}
& Jaccard            &                             2 &  \textbf{1}       \\ 
\hline
\multirow{3}{*}{Multiple FSL}
& Pearson            &                             50 &  .974 \\ 
& Spearman           &                             50 &  .103 \\ 
\cline{2-4}
& Jaccard            &                             2 &  \textbf{1}       \\ 
\hline
\end{tabular}
\end{table}

To obtain deeper insights into feature relevance, all extensions introduce additional processing time and memory overhead during training. As shown in \autoref{tab:ablation_time}, which details an experiment using synthetic datasets with varying numbers of features and labels, the Listener method generally exhibited the highest average execution time, followed by Output Aware. FSL, Residual, and Multiple FSL achieved the best results, with a slight advantage for FSL when considering the overall averages.
Regarding memory usage on the same datasets (\autoref{tab:ablation_memory}), all models showed an expected increase in peak GPU memory as the number of features grew, since the size of the feature weight vector scales accordingly. 
Multiple FSL also demonstrated a linear increase in memory usage relative to the number of labels. 
This occurs because the method processes all labels' feature weight vectors simultaneously in a single combined batch. 
While this memory footprint could be reduced by processing each label sequentially, doing so would negatively impact execution time. 
Notably, we generated specific synthetic datasets with varying numbers of features and labels for this experiment and deliberately employed a lightweight classifier architecture, since maximizing prediction scores was not the primary objective.
This setup ensures that the observed memory growth predominantly reflects the overhead of the feature selection layer itself.

\begin{table}[t]
\centering
\caption{
\textbf{Execution time (s) of each proposed extension across varying numbers of features and labels.} 
Values represent the mean $\pm$ standard deviation over 5 executions.
In bold is the best average per dataset.
Best results with statistical significance ($p < 0.01$) according to the Kruskal-Wallis and Dunn’s test are highlighted in orange.
}
\label{tab:ablation_time}
\resizebox{\textwidth}{!}{%
\begin{tabular}{lllllll}
\toprule
Features & Labels & FSL & Listener & Output Aware & Residual & Multiple FSL \\
\midrule
\multirow{4}{*}{50} & 2 & \cellcolor{orange!20}\textbf{15.90 $\pm$ 1.50} & 21.32 $\pm$ 1.09 & \cellcolor{orange!20}19.95 $\pm$ 1.43 & \cellcolor{orange!20}16.54 $\pm$ 1.29 & \cellcolor{orange!20}16.44 $\pm$ 1.47 \\
 & 5 & \cellcolor{orange!20}\textbf{14.60 $\pm$ 0.42} & 19.37 $\pm$ 0.80 & 17.60 $\pm$ 0.51 & \cellcolor{orange!20}15.88 $\pm$ 0.38 & \cellcolor{orange!20}15.64 $\pm$ 0.44 \\
 & 10 & \cellcolor{orange!20}\textbf{17.03 $\pm$ 1.57} & 27.15 $\pm$ 1.73 & 24.62 $\pm$ 1.95 & \cellcolor{orange!20}18.25 $\pm$ 1.79 & \cellcolor{orange!20}17.95 $\pm$ 1.76 \\
 & 20 & \cellcolor{orange!20}\textbf{15.30 $\pm$ 1.20} & 26.43 $\pm$ 2.71 & 24.37 $\pm$ 2.84 & \cellcolor{orange!20}17.08 $\pm$ 1.60 & \cellcolor{orange!20}16.62 $\pm$ 1.44 \\
\midrule
\multirow{4}{*}{500} & 2 & \cellcolor{orange!20}\textbf{17.83 $\pm$ 1.44} & 39.24 $\pm$ 2.33 & 35.86 $\pm$ 2.13 & \cellcolor{orange!20}18.73 $\pm$ 1.29 & \cellcolor{orange!20}18.38 $\pm$ 1.15 \\
 & 5 & \cellcolor{orange!20}15.17 $\pm$ 2.08 & 33.12 $\pm$ 3.42 & \cellcolor{orange!20}29.87 $\pm$ 3.78 & \cellcolor{orange!20}14.94 $\pm$ 2.04 & \cellcolor{orange!20}\textbf{14.66 $\pm$ 2.42} \\
 & 10 & \cellcolor{orange!20}\textbf{12.14 $\pm$ 0.19} & 31.47 $\pm$ 0.67 & 27.17 $\pm$ 0.43 & \cellcolor{orange!20}12.82 $\pm$ 0.29 & \cellcolor{orange!20}12.62 $\pm$ 0.10 \\
 & 20 & \cellcolor{orange!20}\textbf{12.09 $\pm$ 0.51} & 32.56 $\pm$ 0.45 & \cellcolor{orange!20}27.78 $\pm$ 0.60 & \cellcolor{orange!20}12.56 $\pm$ 0.35 & \cellcolor{orange!20}12.11 $\pm$ 0.48 \\
\midrule
\multirow{4}{*}{5000} & 2 & \cellcolor{orange!20}\textbf{12.91 $\pm$ 2.00} & 126.38 $\pm$ 7.54 & 121.74 $\pm$ 4.47 & \cellcolor{orange!20}13.26 $\pm$ 2.43 & \cellcolor{orange!20}13.06 $\pm$ 2.06 \\
 & 5 & \cellcolor{orange!20}\textbf{12.40 $\pm$ 0.11} & 121.33 $\pm$ 3.78 & 118.62 $\pm$ 3.39 & \cellcolor{orange!20}13.14 $\pm$ 0.52 & \cellcolor{orange!20}13.24 $\pm$ 0.48 \\
 & 10 & \cellcolor{orange!20}\textbf{11.69 $\pm$ 1.19} & 124.43 $\pm$ 5.29 & 120.03 $\pm$ 7.36 & \cellcolor{orange!20}12.20 $\pm$ 1.80 & \cellcolor{orange!20}12.02 $\pm$ 1.66 \\
 & 20 & \cellcolor{orange!20}\textbf{9.89 $\pm$ 0.30} & 118.86 $\pm$ 1.10 & 118.73 $\pm$ 0.99 & \cellcolor{orange!20}10.61 $\pm$ 0.42 & \cellcolor{orange!20}10.84 $\pm$ 0.42 \\
\bottomrule
\end{tabular}%
}
\end{table}

\begin{table}[h]
\centering
\caption{
\textbf{Peak GPU memory usage (MB) of each proposed extension across varying numbers of features and labels.}
Standard deviation is omitted because the variation in peak GPU allocation is minimal, given the same architecture and dataset, the memory footprint remains nearly identical across all executions.
All peak GPU memory allocations were measured using an NVIDIA GeForce RTX 5070 Ti (16 GB VRAM).
In bold is the best average per dataset and metric.
}
\label{tab:ablation_memory}
\resizebox{\textwidth}{!}{%
\begin{tabular}{lllllll}
\toprule
Features & Labels & FSL & Listener & Output Aware & Residual & Multiple FSL \\
\midrule
\multirow{4}{*}{50} & 2 & \textbf{18.1} & \textbf{18.1} & \textbf{18.1} & \textbf{18.1} & 18.4 \\
 & 5 & \textbf{18.1} & 18.2 & 18.2 & \textbf{18.1} & 19.0 \\
 & 10 & \textbf{18.1} & 18.2 & 18.2 & 18.2 & 20.2 \\
 & 20 & \textbf{18.1} & 18.2 & 18.2 & 18.2 & 22.4 \\
\midrule
\multirow{4}{*}{500} & 2 & \textbf{24.8} & 25.7 & 25.7 & \textbf{24.8} & 26.1 \\
 & 5 & \textbf{24.8} & 25.9 & 25.9 & 24.9 & 29.9 \\
 & 10 & \textbf{24.8} & 25.8 & 25.9 & 24.9 & 36.2 \\
 & 20 & \textbf{24.8} & 25.9 & 26.1 & 25.1 & 48.9 \\
\midrule
\multirow{4}{*}{5000} & 2 & \textbf{89.3} & 113.6 & 113.6 & 89.5 & 101.0 \\
 & 5 & \textbf{89.3} & 113.0 & 113.4 & 89.9 & 136.9 \\
 & 10 & \textbf{89.3} & 112.5 & 113.3 & 90.5 & 194.5 \\
 & 20 & \textbf{89.3} & 113.4 & 115.2 & 91.9 & 311.6 \\
\bottomrule
\end{tabular}%
}
\end{table}

In addition to preserving the FSL property of generating non-label specific feature weights, the proposed extensions should be capable of quantifying label-specific relevance.
We utilize the synthetic multi-class datasets SynthA, SynthB, and SynthC on this analysis.
As previously discussed, the effectiveness of feature weighting is closely tied to the model's predictive capacity. 
Models that fail to generalize their classification to new samples often exploit patterns found in noisy features, leading the FSL to assign higher feature weights to them.
The proposed extensions should therefore have a neutral or positive impact on the model's predictive capacity.
\autoref{tab:f1_score_performance_synthA_synthB_synthC_on_extensions} compares the prediction performance using F1 score on SynthA, SynthB and SynthC datasets.
In contrast to the strong performance of Multiple FSL, the Output Aware extension suffered a substantial performance degradation on SynthA and SynthB datasets.
On SynthB, Multiple FSL, FSL and Listener consistently achieved the highest and most balanced scores across all labels, while Residual demonstrated a high degradation.
For SynthC, Multiple FSL achieve the highest scores, followed by Output Aware, while Residual performed worst.
In general, Output Aware exhibited inconsistent performance, with better results on SynthC.
Its reliance on output knowledge during training negatively affected its generalization ability.
Residual consistently underperformed for all datasets.
Multiple FSL, Listener, and the original FSL consistently performed well.

\begin{figure*}[h]
    \centering
     \begin{subfigure}[b]{0.45\textwidth}
         \centering
         \includegraphics[width=\textwidth]{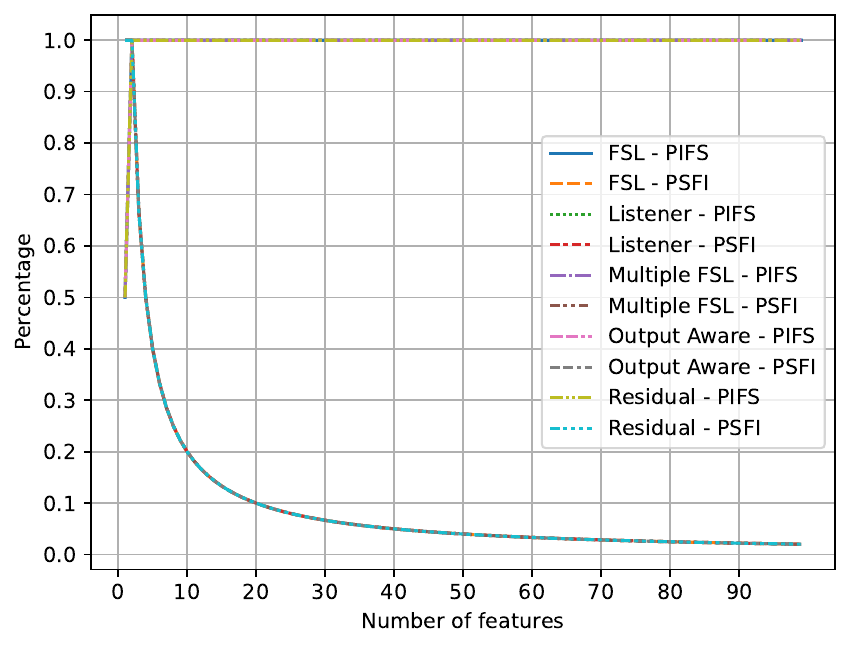}
         \caption{XOR}
         \label{subfig:pifs_psfi_analysis_xor_on_extensions}
     \end{subfigure}
     \hfill
     \begin{subfigure}[b]{0.45\textwidth}
         \centering
         \includegraphics[width=\textwidth]{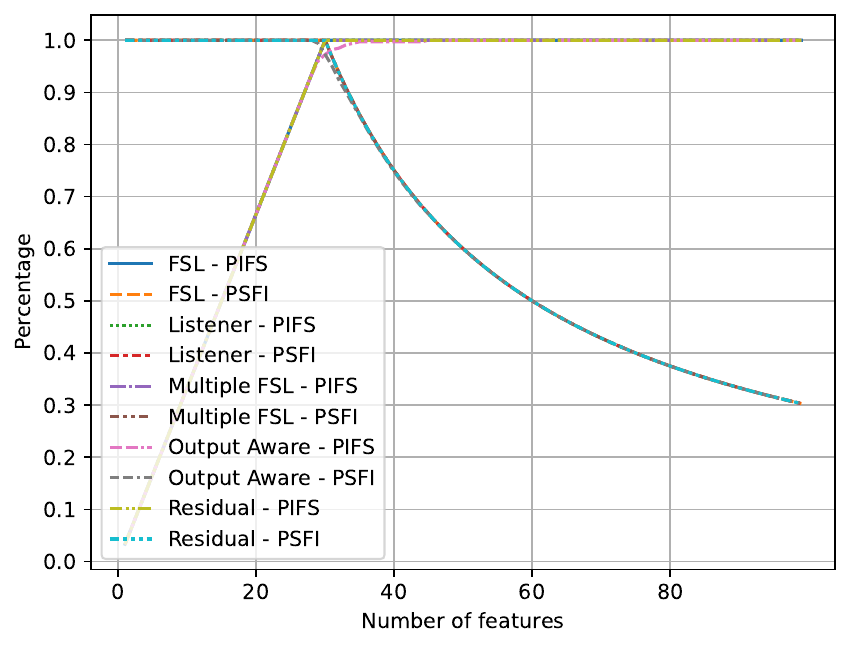}
         \caption{SynthA}
         \label{subfig:pifs_psfi_analysis_synthA_on_extensions}
     \end{subfigure}
     \hfill
     \begin{subfigure}[b]{0.45\textwidth}
         \centering
         \includegraphics[width=\textwidth]{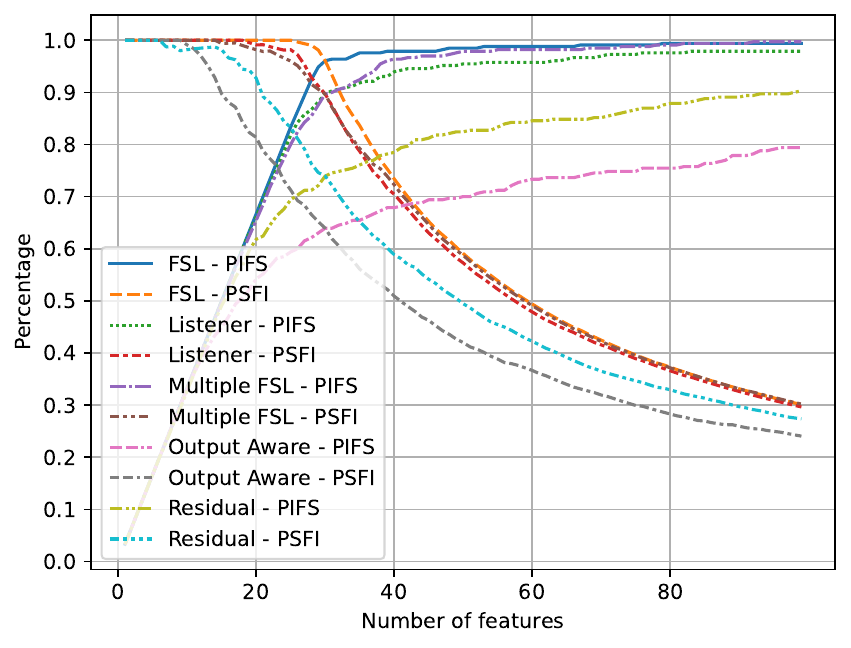}
         \caption{SynthB}
        \label{subfig:pifs_psfi_analysis_synthB_on_extensions}
     \end{subfigure}
     \hfill
     \begin{subfigure}[b]{0.45\textwidth}
         \centering
         \includegraphics[width=\textwidth]{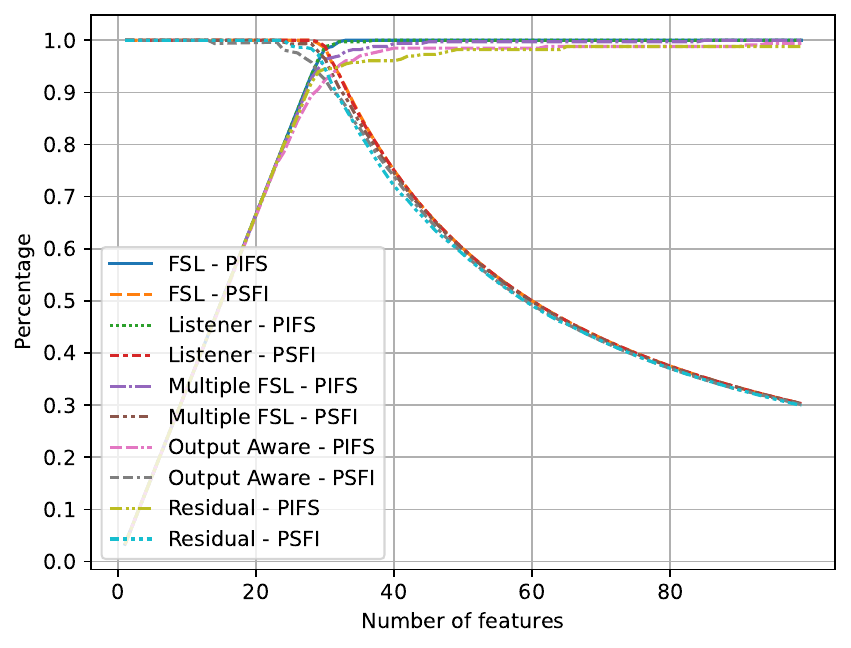}
         \caption{SynthC}
         \label{subfig:pifs_psfi_analysis_synthC_on_extensions}
     \end{subfigure}
     \caption[
        Analysis of PIFS and PSFI on the XOR, SynthA, SynthB, and SynthC datasets, comparing the original FSL and proposed extensions for increasing selection subsets. 
        Visually, the figure is composed of four line charts arranged in a two-by-two grid, labeled (a) XOR, (b) SynthA, (c) SynthB, and (d) SynthC. 
        Each chart's vertical axis represents the score percentage (constrained from 0.0 to 1.0), and the horizontal axis represents the subset size selected from the top-n highest-weighted features (ranging from 0 to 100). 
        Every chart contains ten distinct lines mapping the PIFS and PSFI metrics for five methods: FSL, Listener, Multiple FSL, Output Aware, and Residual. 
        The charts reveal distinct dataset-dependent behaviors: in (a) XOR, all methods tightly cluster and drop sharply from 1.0 to near 0 early on; in (b) SynthA and (d) SynthC, the methods closely track one another, rising to a peak of 1.0 around 30 features before smoothly declining; conversely, (c) SynthB displays high variance, with the method trajectories diverging significantly (FSL had the best result, followed by Listener and Multiple FSL).
        All plotted scores represent the average over rankings generated by 11 independent model executions.
    ]
    {
        \textbf{Analysis of PIFS and PSFI on XOR, SynthA, SynthB and SynthC datasets between original FSL and proposed extensions for increasing selection subsets.} 
        All values scores are the average over rankings generated by 11 independent models execution.
        Vertical axis represents the scores constrained from 0 to 1 as both measures are percentages.
        Horizontal axis represents the subsets size selected from the top-n features with the highest weights.
    }
    \label{fig:pifs_psfi_analysis_xor_synthA_synthB_synthC_on_extensions}
\end{figure*}

Given their strong prediction performance, it's expect that these models were able to identify relevant features.
\autoref{fig:pifs_psfi_analysis_xor_synthA_synthB_synthC_on_extensions} shows the PIFS and PSFI average scores for different feature subsets based on feature weights generated through 11 executions.
As previously discussed, all methods were able to consistently detect the two relevant features for the XOR dataset.
For SynthA, Output Aware was the only algorithm that did not assign the highest weights for all the 30 relevant features in all executions.
For SynthB and SynthC, no method constantly achieved the 30 relevant features on the top-$30$ highest weights.

\begin{table}[h]
\centering
\caption[Per label F1 scores of FSL and proposed extensions on the SynthA, SynthB, and SynthC datasets. For SynthA dataset, Multiple FSL had the highest average and also the highest result with statistical significance for all labels. For SynthB Listener had the highest average for all labels, Multiple FSL had the highest average for label 1 and FSL, Listener and Multiple FSL had the highest results with statistical significance for all labels. For Synth dataset Multiple FSL had the highest average and the highest result with statistical significance for all labels.] {\textbf{Per label F1 scores of FSL and proposed extensions on the SynthA, SynthB, and SynthC datasets.}
The scores were calculated across 11 independent executions using test dataset.
For each reported metric we present the average and standard deviation.
In bold is the highest average of each dataset and label.
Best results with statistical significance ($p < 0.01$) according to the Kruskal-Wallis and Dunn’s test are highlighted in orange.
}
\label{tab:f1_score_performance_synthA_synthB_synthC_on_extensions}
\begin{tabular}{lrlll}
    \hline
    Algorithm & Label & SynthA & SynthB & SynthC \\
    \hline
    \multirow{3}{*}{FSL} 
    & 0 & .887$\pm$.015   & \cellcolor{orange!20} .934$\pm$.013   & .878$\pm$.012 \\
    & 1 & .885$\pm$.013   & \cellcolor{orange!20} .942$\pm$.011   & .863$\pm$.016 \\
    & 2 & .888$\pm$.014   & \cellcolor{orange!20} .924$\pm$.013   & .865$\pm$.012 \\
    \hline
    \multirow{3}{*}{Listener} 
    & 0 & .888$\pm$.007   & \cellcolor{orange!20} \textbf{.945$\pm$.011}   & .890$\pm$.016 \\
    & 1 & .883$\pm$.009   & \cellcolor{orange!20} \textbf{.946$\pm$.021}   & .867$\pm$.018 \\
    & 2 & .889$\pm$.009   & \cellcolor{orange!20} \textbf{.934$\pm$.013}   & .880$\pm$.016 \\
    \hline
    \multirow{3}{*}{Residual} 
    & 0 & .845$\pm$.017   & .811$\pm$.028  & .796$\pm$.012 \\
    & 1 & .846$\pm$.014   & .833$\pm$.065  & .765$\pm$.012 \\
    & 2 & .864$\pm$.017   & .783$\pm$.049  & .767$\pm$.009 \\
    \hline
    \multirow{3}{*}{Output Aware} 
    & 0 & .742$\pm$.103   & .774$\pm$.054  & .919$\pm$.015 \\
    & 1 & .689$\pm$.142   & .762$\pm$.066  & .905$\pm$.015 \\
    & 2 & .740$\pm$.070   & .657$\pm$.259  & .911$\pm$.014 \\
    \hline
    \multirow{3}{*}{Multiple FSL} 
    & 0 & \cellcolor{orange!20} \textbf{.915$\pm$.011}   & \cellcolor{orange!20} .943$\pm$.007           & \cellcolor{orange!20} \textbf{.950$\pm$.010} \\
    & 1 & \cellcolor{orange!20} \textbf{.911$\pm$.007}   & \cellcolor{orange!20} \textbf{.946$\pm$.012}  & \cellcolor{orange!20} \textbf{.944$\pm$.011} \\
    & 2 & \cellcolor{orange!20} \textbf{.916$\pm$.011}   & \cellcolor{orange!20} .933$\pm$.011           & \cellcolor{orange!20} \textbf{.947$\pm$.005} \\
    \hline
\end{tabular}
\end{table}

The feature weights produced by each extension for SynthB (\autoref{fig:heatmap_on_extensions_for_synthB}) shows that Multiple FSL and Output Aware correctly assigns the highest weights to label-relevant features. 
In contrast, Listener assigns high weights to all relevant features regardless of their specific label association, failing to accurately prioritize features based on label specificity. 
Residual exhibits an inconsistent behavior, while it assigns the highest weights to the relevant label in some cases, it conversely assigns the lowest weights to the correct label in others.

\begin{figure*}[h]
     \centering
     \begin{subfigure}[b]{0.49\textwidth}
         \centering
         \includegraphics[width=\textwidth]{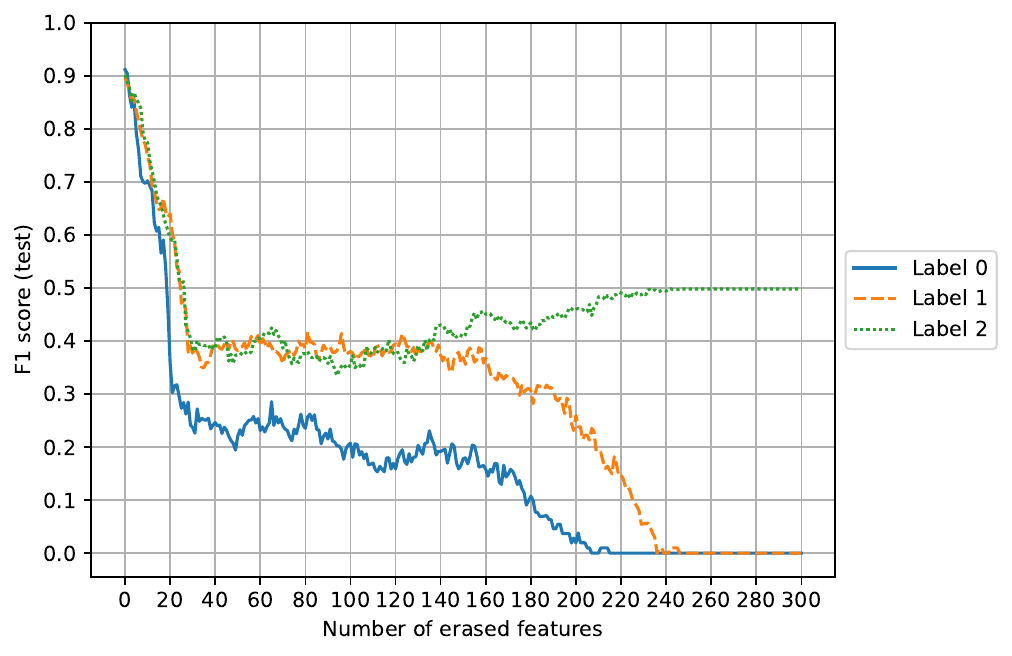}
         \caption{Listener}
         \label{subfig:occlusion_on_extensions_for_synthC_class_0_listener}
     \end{subfigure}
     \hfill
     \begin{subfigure}[b]{0.49\textwidth}
         \centering
         \includegraphics[width=\textwidth]{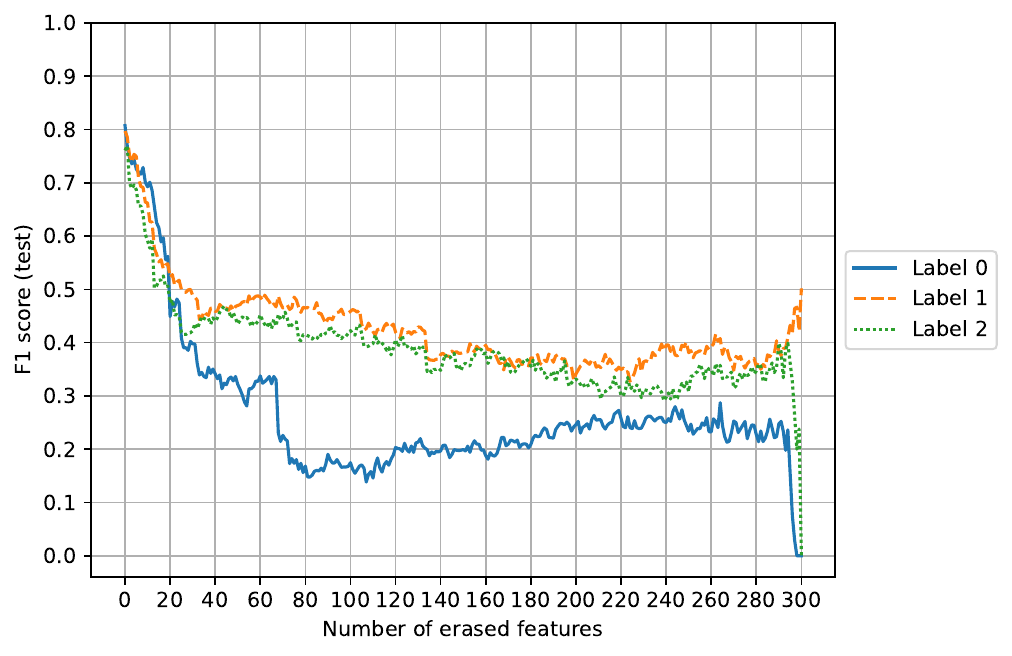}
         \caption{Residual}
         \label{subfig:occlusion_on_extensions_for_synthC_class_0_residual}
     \end{subfigure}
     \begin{subfigure}[b]{0.49\textwidth}
         \centering
         \includegraphics[width=\textwidth]{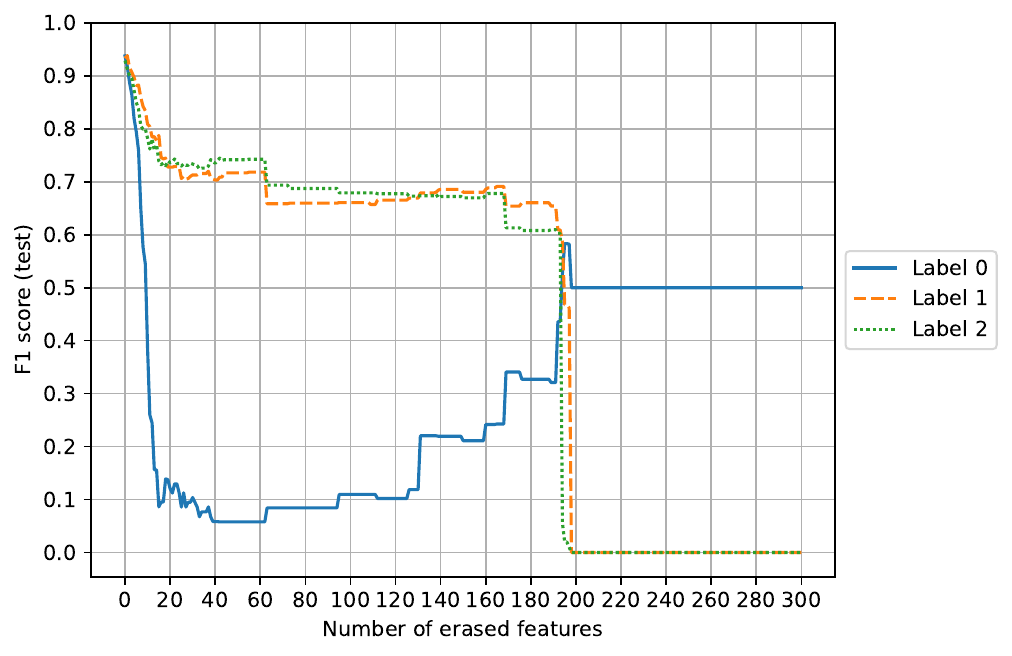}
         \caption{Output Aware}
         \label{subfig:occlusion_on_extensions_for_synthC_class_0_output_aware}
     \end{subfigure}
     \hfill
     \begin{subfigure}[b]{0.49\textwidth}
         \centering
         \includegraphics[width=\textwidth]{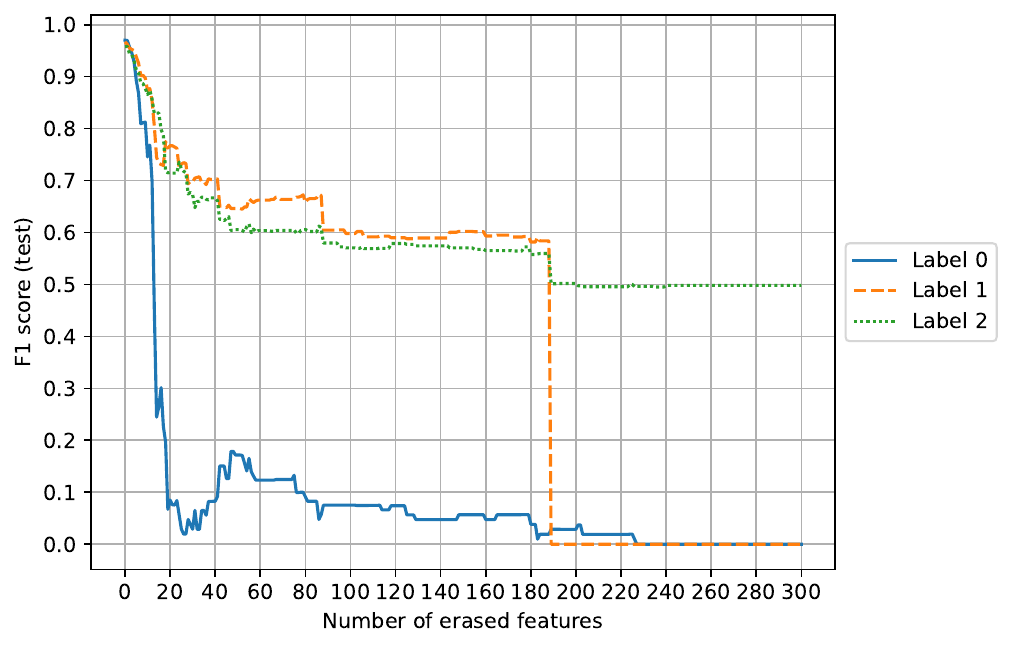}
         \caption{Multiple FSL}
         \label{subfig:occlusion_on_extensions_for_synthC_class_0_multiple_fsl}
     \end{subfigure}
     \caption[Feature erasure (occlusion) based on SynthC using feature weights related to label 0.
    Four charts are displayed, each with three lines representing the F1 score for labels 0 (blue), 1 (orange), and 2 (green), respectively.
    The first chart shows the result of applying feature erasure to a model using the Listener.
    After removing the top relevant features for label 0, a significant drop in performance is observed across all labels, although the drop for label 0 is slightly greater.
    When the top 20 most relevant features are removed, the F1 score for label 0 drops to 0.3, while the scores for labels 1 and 2 drop to around 0.65.
    Initially, with all features, the F1 score for all labels was above 0.9.
    The second chart shows the same process applied to a model using the Residual approach.
    When the top 20 features relevant to label 0 are removed, the F1 score for all labels drops to approximately 0.5, whereas with all features, the score for each label reached around 0.8.
    The third chart represents the Output Aware model, where a significant difference is observed.
    While F1 scores for all labels were above 0.9 when using all features, removing the top 20 features relevant to label 0 causes its F1 score to drop close to 0.1, while the other labels remain above 0.7.
    This huge difference of F1 score continues until around the top 120 features, when the F1 scores for all labels starts to have closer values again.
    The fourth chart represents the Multiple FSL model, with results similar to the Output Aware model.
    With all features, the F1 score for each label exceeds 0.95.
    After removing the 20 most relevant features for label 0, its F1 score drops below 0.1, while the scores for the other labels remain above 0.7.
    This difference continues until around the top 190 features where the label 1 F1 score also drops to near zero and the label 2 F1 score stays near to 0.5.]
    {\textbf{Feature erasure (occlusion) based on SynthC using feature weights related to label $0$ (blue line).}
        Each line represents the F1 score trend related to a single label as we erase the top-$n$ features based on their weights in descending order.
        All values were calculated using the test dataset.
    }
    \label{fig:occlusion_on_extensions_for_synthC}
\end{figure*}

\autoref{fig:occlusion_on_extensions_for_synthC} illustrates the effect of applying feature erasure based on SynthC using label 0 feature weights to evaluate the effectiveness of the extensions in detecting label-specific relevant features that are impacting mostly individual label's prediction.
Since the original FSL implementation lacks the ability to identify label-specific feature weights it is not included in this analysis. 
The observed higher drop in F1 scores for label 0 suggests that the removed features play a crucial role in correctly classifying samples belonging to this label on the trained model for all proposed extensions. 
However, Output Aware and Multiple FSL demonstrated a higher impact on their label 0 predictions when compared to the other labels.

\begin{figure*}[h]
     \centering
     \begin{subfigure}[b]{0.49\textwidth}
         \centering
         \includegraphics[width=\textwidth]{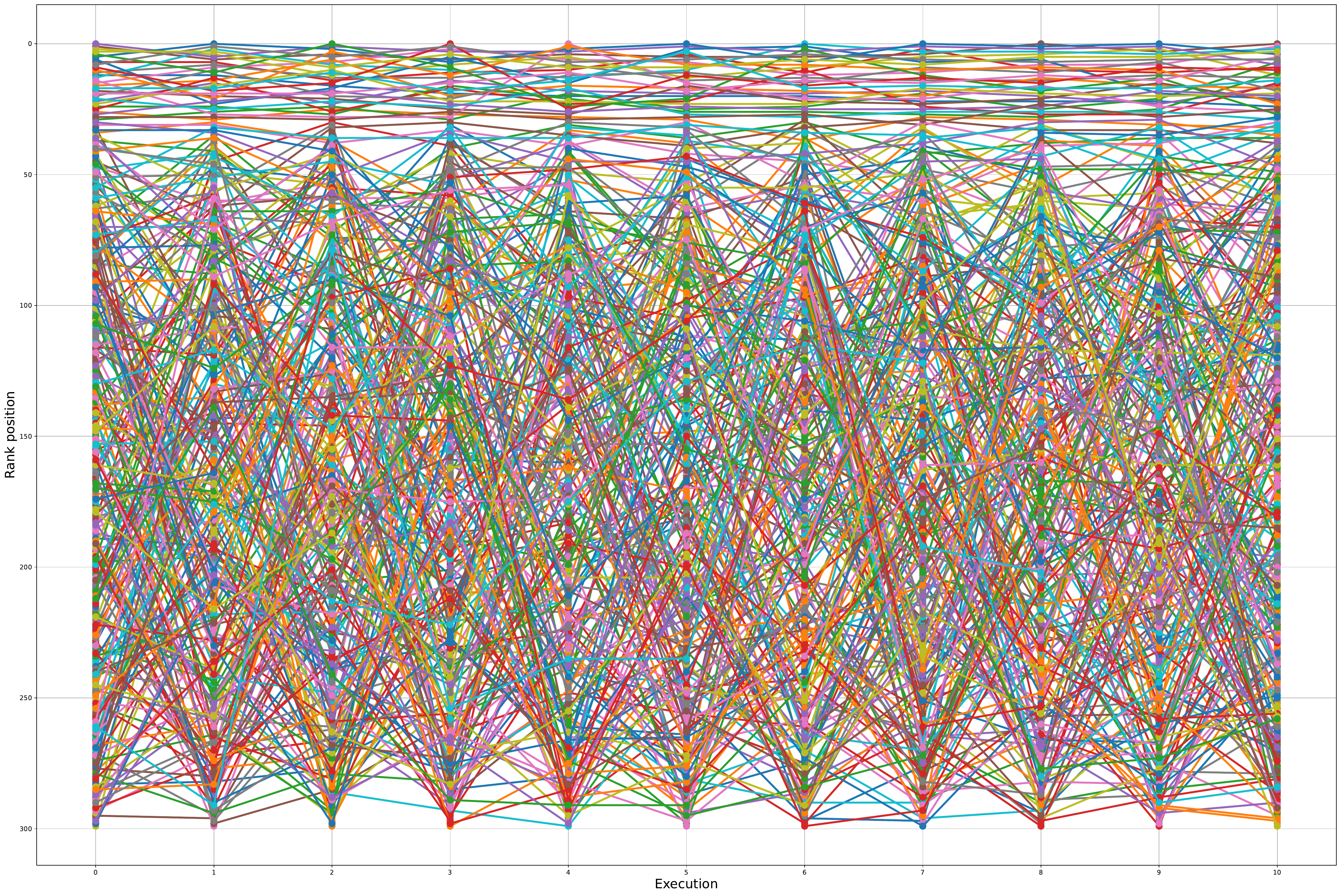}
         \caption{Listener}
         \label{subfig:listener_feature_position_per_execution_on_extensions_for_synthC_label0}
     \end{subfigure}
     \hfill
     \begin{subfigure}[b]{0.49\textwidth}
         \centering
         \includegraphics[width=\textwidth]{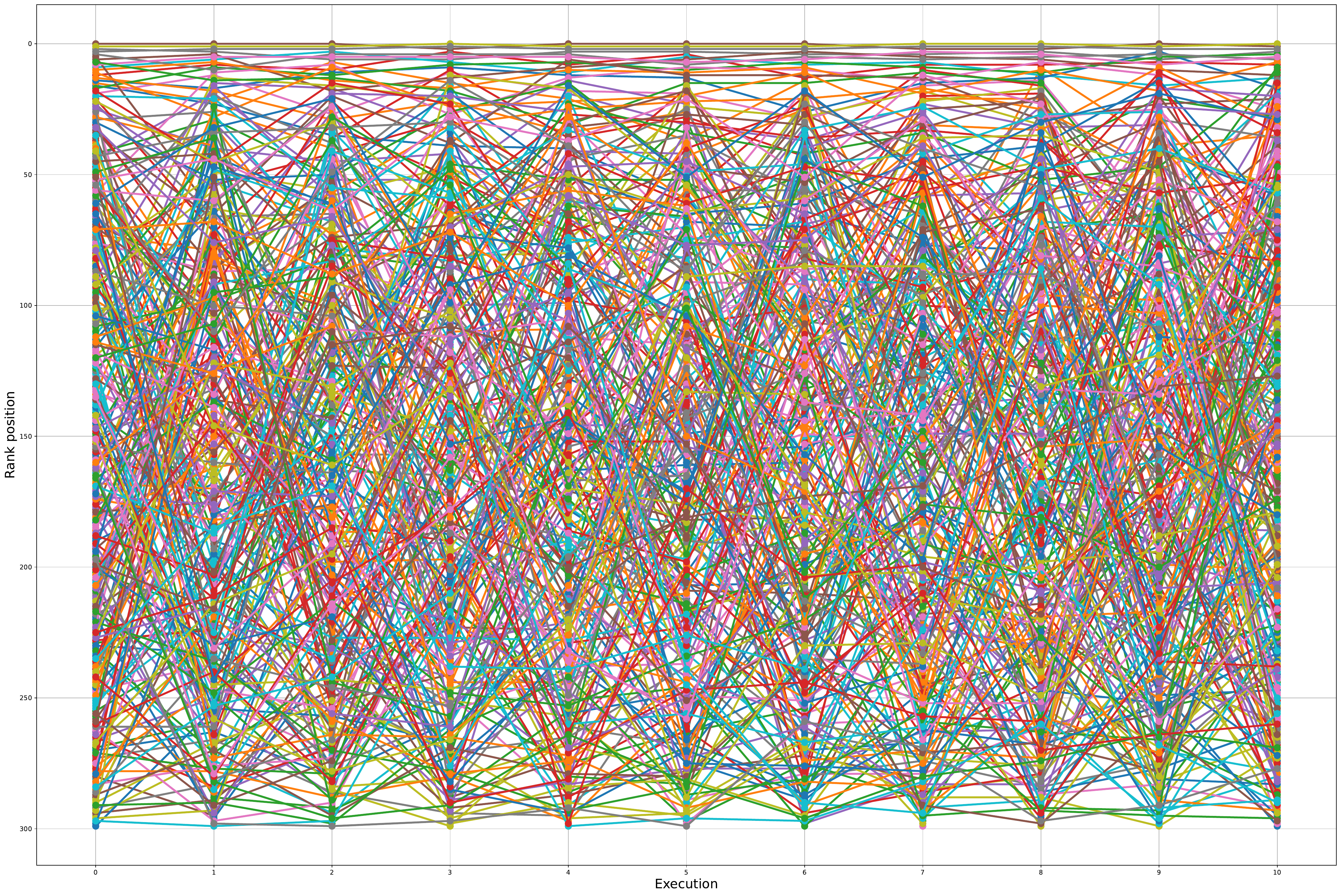}
         \caption{Residual}
         \label{subfig:residual_feature_position_per_execution_on_extensions_for_synthC_label0}
     \end{subfigure}
     \begin{subfigure}[b]{0.49\textwidth}
         \centering
         \includegraphics[width=\textwidth]{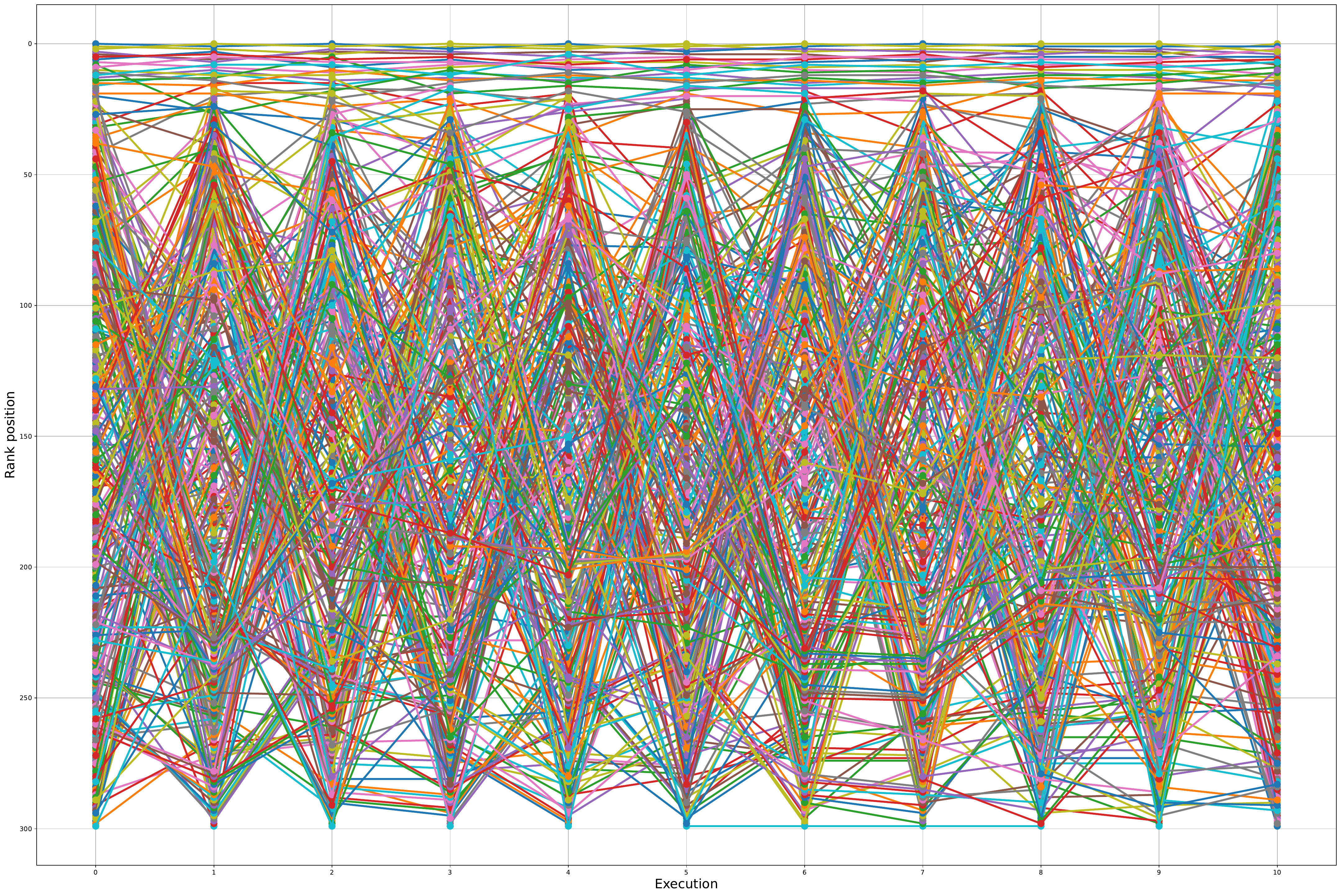}
         \caption{Output Aware}
         \label{subfig:output_aware_feature_position_per_execution_on_extensions_for_synthC_label0}
     \end{subfigure}
     \hfill
     \begin{subfigure}[b]{0.49\textwidth}
         \centering
         \includegraphics[width=\textwidth]{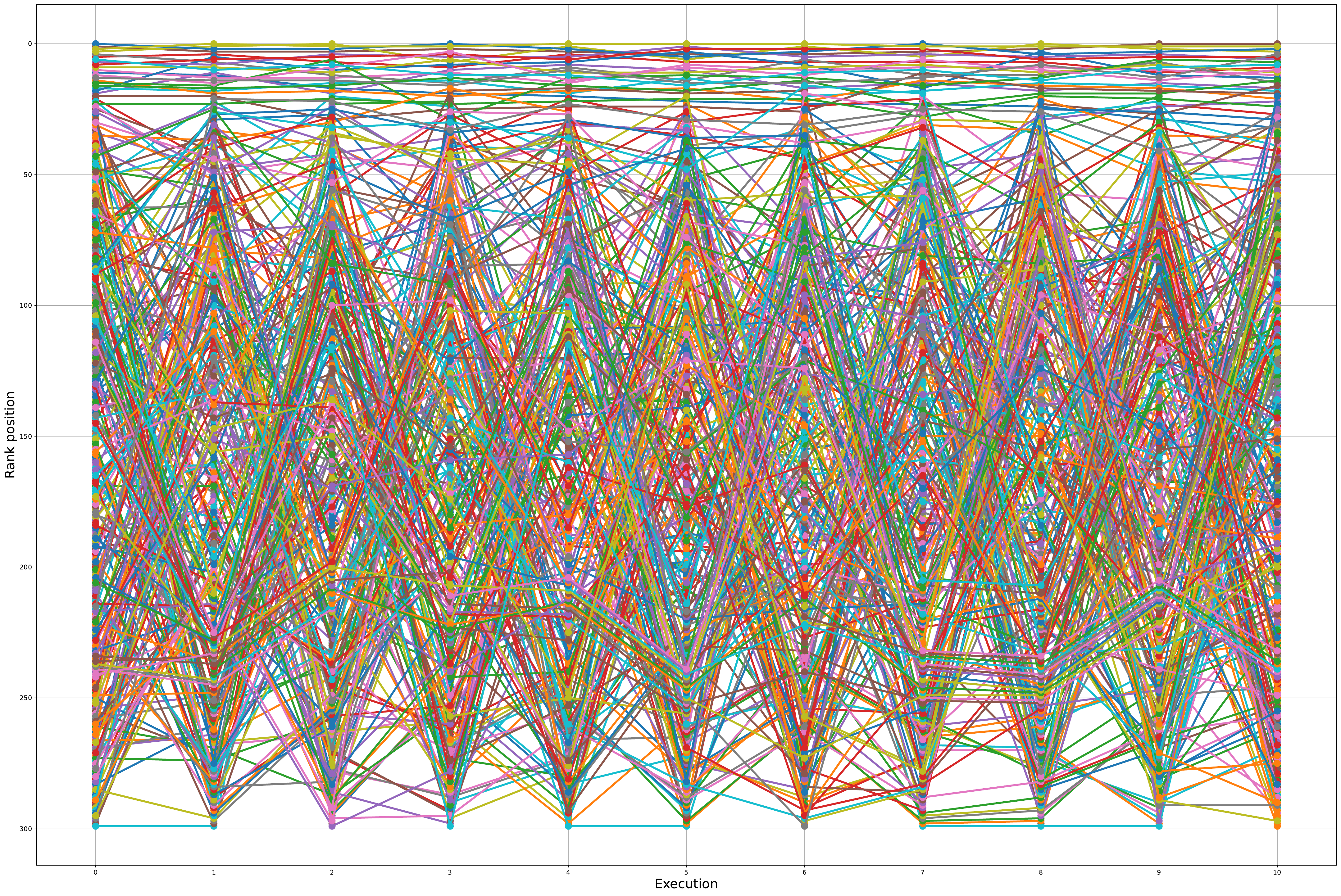}
         \caption{Multiple FSL}
         \label{subfig:mfs_feature_position_per_execution_on_extensions_for_synthC_label0}
     \end{subfigure}
    \caption[Feature ranking position related to label $0$ across proposed extensions executions on SynthC.
    Four charts are shown, each corresponding to one of the proposed extensions: Listener, Residual, Output Aware, and Multiple FSL. 
    In each plot, one hundred colored lines represent each one of the features from SynthC dataset and their ranking positions over eleven executions of the respective algorithm.
    The vertical axis indicates the ranking position of each feature by relevance, while the horizontal axis corresponds to the 11 training runs using the same technique.
    A clear separation is visible between features that consistently rank at the top, frequently appearing among the highest positions, and those that exhibit high variability, shifting between ranks near the top and the bottom of the ranking.
    For the first chart (Listener) the amount of features at the top is huger. 
    Other plots display a smaller amount of features at the top.
    Output Aware and Multiple FSL shows a huger instability of the features on the lower positions, creating large triangles as features constantly migrate from the middle of the chart toward the top, from the middle toward the bottom, and vice versa.]
    {\textbf{Feature ranking position related to label $0$ across proposed extensions executions on SynthC.}
    All rankings were created from the feature weights related to label 0 in descending order. 
    Each algorithm was executed 11 times. 
    The vertical axis represents each individual execution, while the horizontal axis shows the ranking position of each feature, from 0 (top) to 299 (bottom), within that run. 
    Different colors represent different features.
    Given the high density of lines, this visualization is not intended for tracking individual features. 
    Instead, the objective is to highlight a macro-level pattern: the clear distinction between relevant features, which form stable, straight bands consistently ranked at the top, and noisy features, which exhibit highly chaotic and fluctuating trajectories at the bottom of the graph. 
    Additionally, the results demonstrate how the positions of both informative and noisy features vary across executions within each group.
    }

\label{fig:feature_position_per_execution_on_extensions_for_synthC_label0}
\end{figure*}

\autoref{fig:feature_position_per_execution_on_extensions_for_synthC_label0} displays the positions of each feature, represented by colored lines, across 11 different executions of each method on SynthC dataset related specifically to label 0. 
The horizontal axis indicates the rank of each feature, ranging from 0 (top) to 299 (bottom), based on the score given by each FSL extension. 
The objective is not to trace the trajectory of individual features across executions, but rather to distinguish the patterns characterizing informative versus noisy features, thereby expanding our understanding of algorithm stability.

Consistent with the stability evaluation on XOR dataset (\autoref{tab:comparative_stability_on_extensions_for_xor}), a distinct pattern emerges: informative features are consistently positioned at the top of the graph, while noisy features remain at the bottom.
However, the precise positioning of relevant and irrelevant features varies between executions.
Notably, the Listener method places a higher number of features in the upper ranks. 
This aligns with the observations in \autoref{fig:occlusion_on_extensions_for_synthC}. 
Listener is not able to assign label-specific feature weights, it consistently maintains all informative features at the top, whereas Output Aware and Multiple FSL only assign higher scores to informative features related to label 0.
Conversely, Residual exhibited the highest instability, making it difficult to distinguish between the informative features at the top and the noisy ones at the bottom.
Additional experiments involving the synthetic datasets are provided in Figures \ref{fig:pifs_psfi_analysis_synthA_synthB_synthC_on_extensions}, \ref{fig:weighted_tsne_on_extensions_for_synthA} and Tables \ref{tab:weighted_tsne_on_extensions_for_synthA}, \ref{tab:informative_per_label_L_0_1_extensions} of \ref{sec:complementary_experiments}.

\begin{figure*}[!h]
     \centering
     \begin{subfigure}[b]{0.45\textwidth}
         \centering
         \includegraphics[width=\textwidth]{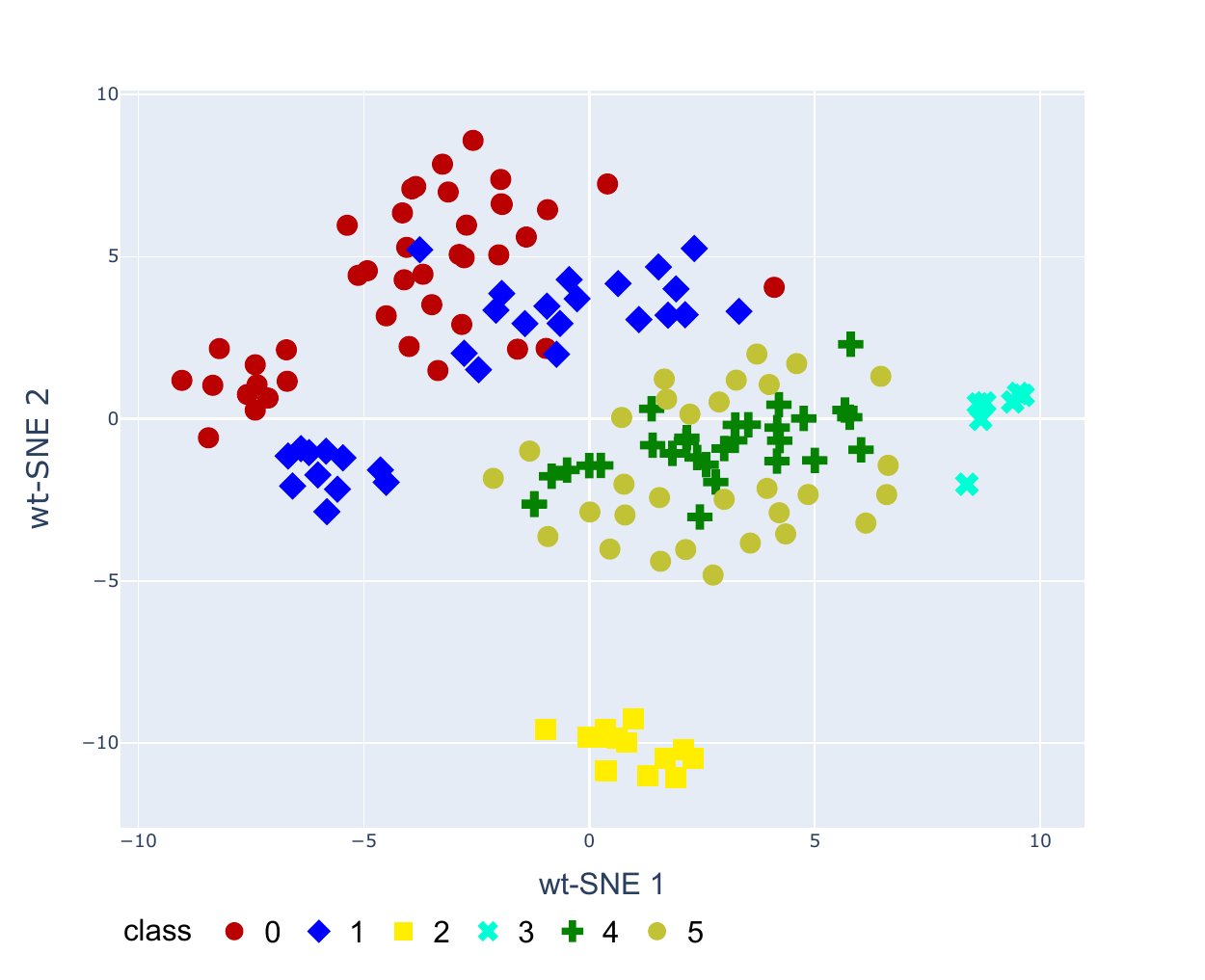}
         \caption{Without weights}
         \label{subfig:wtsne_without_weights_on_extensions_breast_label_0}
     \end{subfigure}
     \hfill
     \begin{subfigure}[b]{0.45\textwidth}
         \centering
         \includegraphics[width=\textwidth]{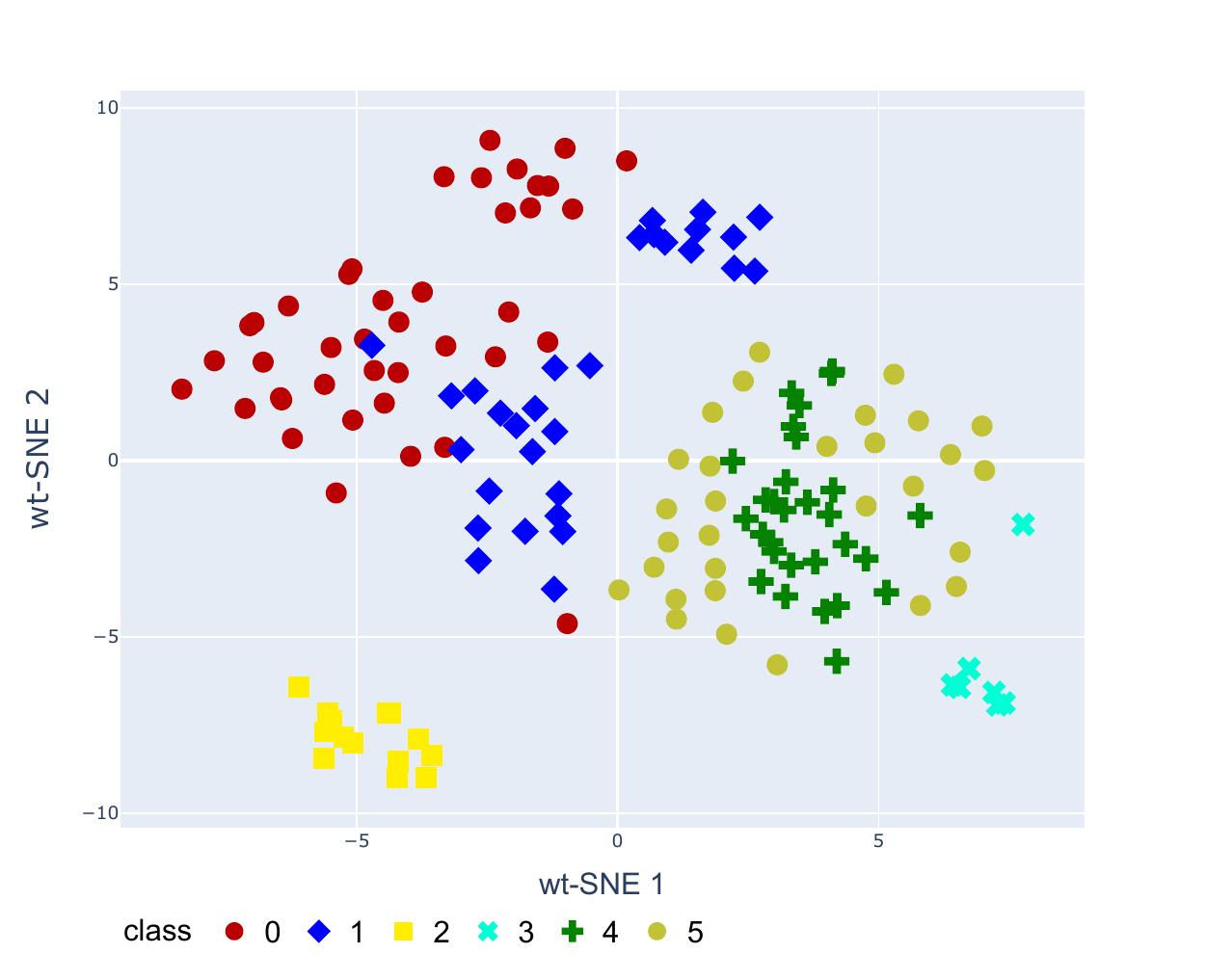}
         \caption{Listener}
         \label{subfig:wtsne_listener_on_extensions_breast_label_0}
     \end{subfigure}
     \begin{subfigure}[b]{0.45\textwidth}
         \centering
         \includegraphics[width=\textwidth]{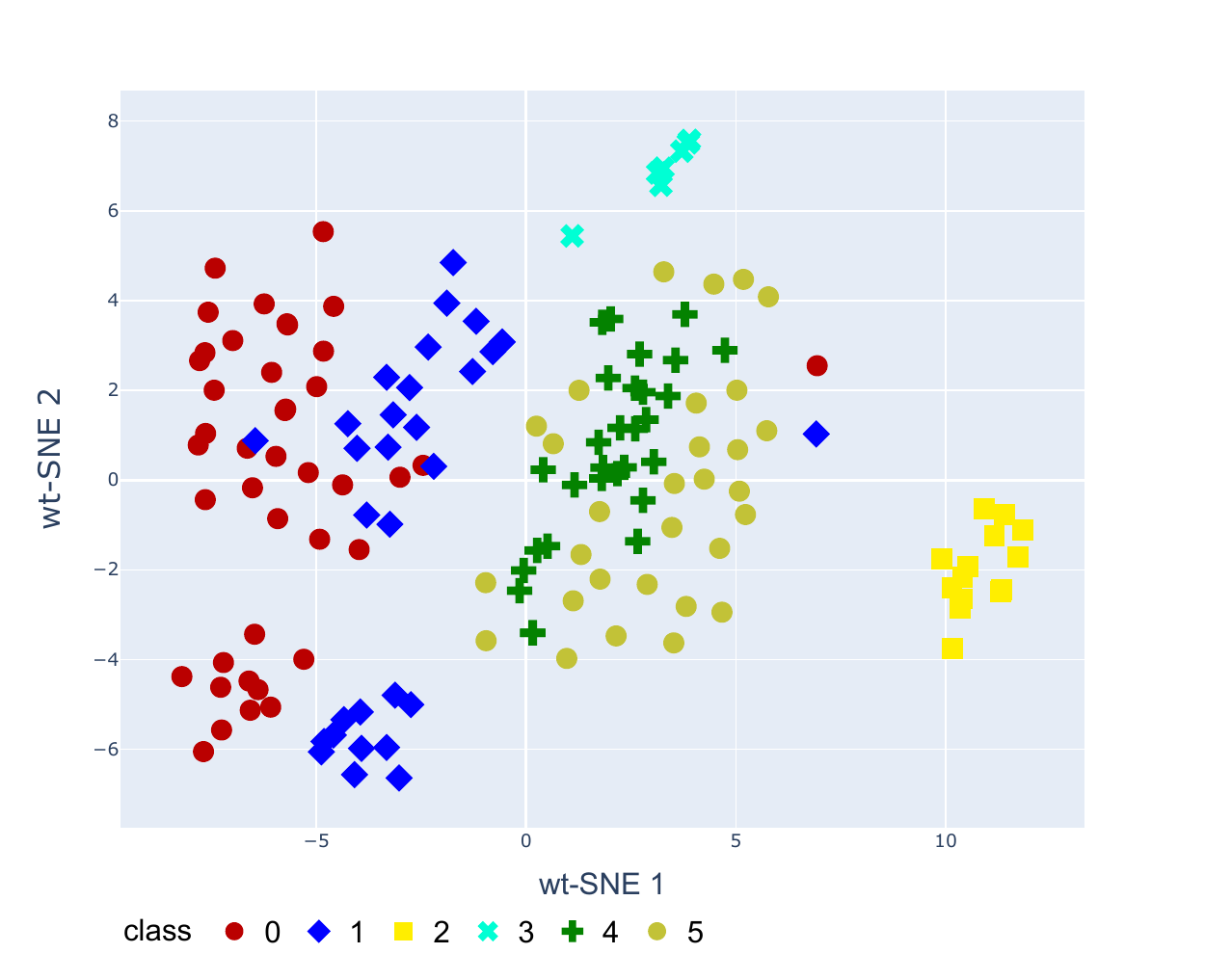}
         \caption{Residual}
         \label{subfig:wtsne_residual_on_extensions_breast_label_0}
     \end{subfigure}
     \hfill
     \begin{subfigure}[b]{0.45\textwidth}
         \centering
         \includegraphics[width=\textwidth]{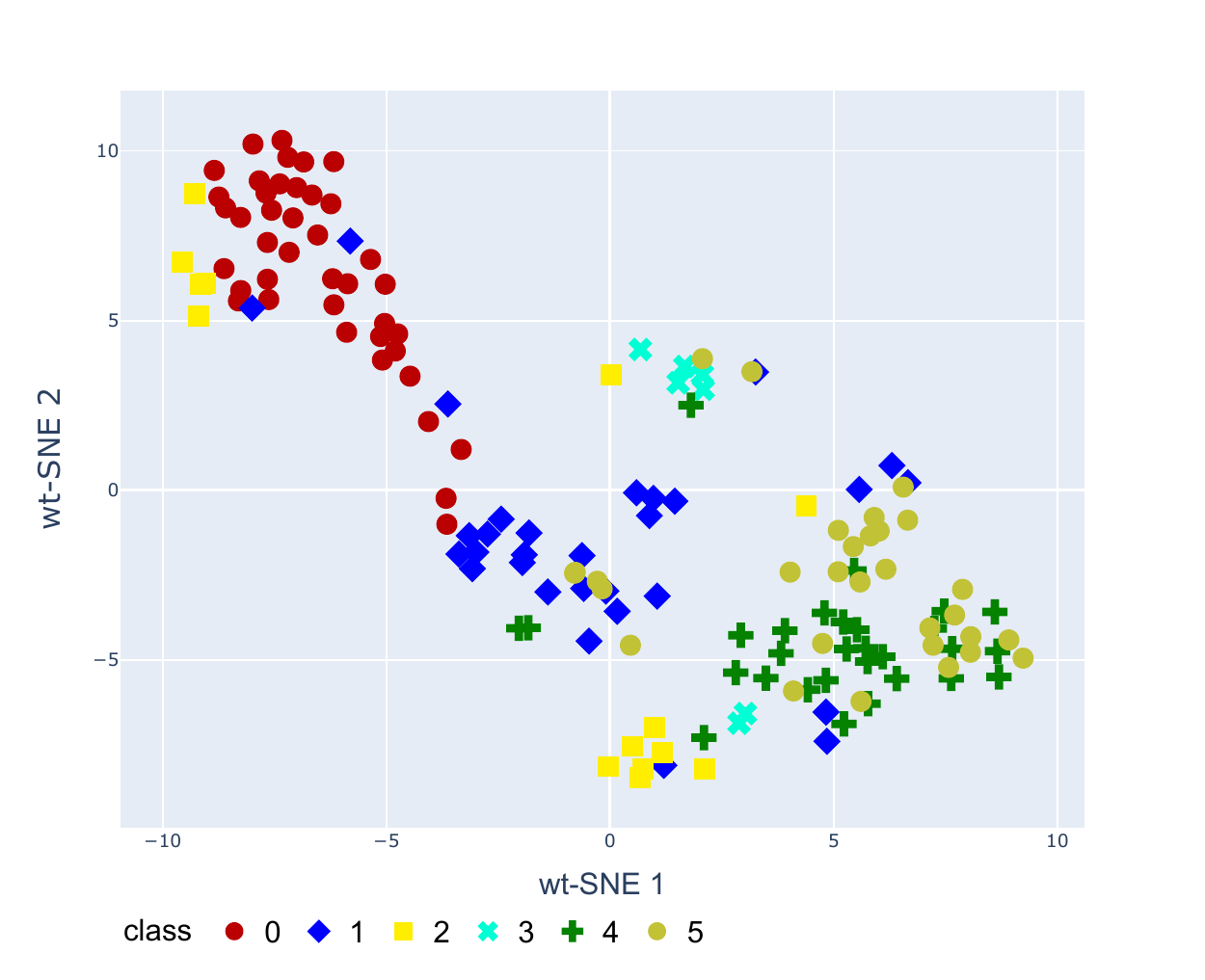}
         \caption{Output Aware}
         \label{subfig:wtsne_output_aware_on_extensions_breast_label_0}
     \end{subfigure}
     \begin{subfigure}[b]{0.45\textwidth}
         \centering
         \includegraphics[width=\textwidth]{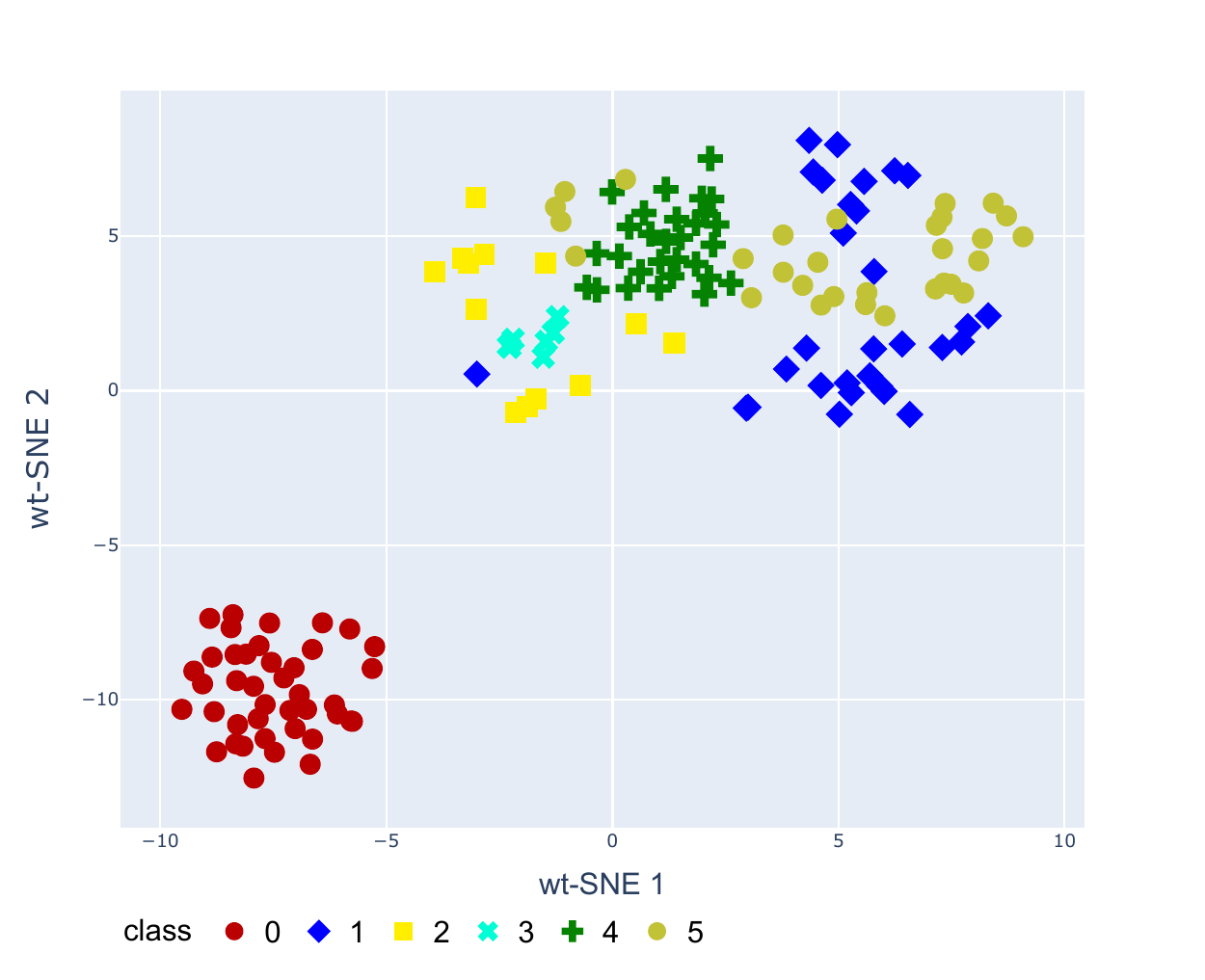}
         \caption{Multiple FSL}
         \label{subfig:wtsne_multiple_fsl_on_extensions_breast_label_0}
     \end{subfigure}
     \caption[Visualization of weighted t-SNE using label 0 feature weights from Breast dataset.
    The figure displays five plots. 
    The first shows the result of applying plain t-SNE on the Breast dataset, while the remaining four show the results of applying weighted t-SNE using the weights generated by the Listener, Residual, Output Aware, and Multiple FSL extensions, respectively, being the weights specific to label 0.
    In the first plot, a greater separation between labels 2 from the others can be observed, the same can be observed for label 3, while labels 4 and 5 appear mixed in the 2D space, as do labels 0 and 1.
    In the second plot (Listener), a slight improvement in the separation between labels 0 and 1, as well as 4 and 5, can be noticed.
    In the third plot (Residual), no clear changes are observed compared to the first plot.
    In the fourth plot (Output Aware), there is a huger change, but without improvement, leading to a mix of label 0 with both label 1 and label 2.
    In the fifth plot (Multiple FSL), a clear improvement is seen, with a significant distance formed between label 0 and all other labels. 
    However, while label 0 becomes clearly separated, the remaining labels become more mixed.]
    {\textbf{Visualization of weighted t-SNE using label 0 (red dots) feature weights from Breast dataset.}
        Each label is represented by a distinct color, being red assigned to label $0$.
        \autoref{subfig:wtsne_without_weights_on_extensions_breast_label_0} presents the original t-SNE without feature weights.
        \autoref{subfig:wtsne_listener_on_extensions_breast_label_0}, \autoref{subfig:wtsne_residual_on_extensions_breast_label_0},
        \autoref{subfig:wtsne_output_aware_on_extensions_breast_label_0} and \autoref{subfig:wtsne_multiple_fsl_on_extensions_breast_label_0} represents the weighted t-SNE for all proposed extensions using the assigned weights for each feature related to label $0$.
    }
    \label{fig:feature_wtsne_on_extensions_breast}
\end{figure*}

\begin{figure*}[h]
    \centering
     \caption[
        Comparison of Support Vector Classifier (SVC) prediction performance on the top 1000 features ranked by weight from FSL and its proposed extensions for the Breast dataset. 
        The vertical axis shows the F1-score on the test dataset, while the horizontal axis displays the number of features used (from 1 to 1000). 
        For each method, two evaluations are plotted: one using only the ranked feature subsets (Rank) and another using the weighted features (Weight). 
        Visually, the figure consists of a single line chart plotted over a grid. 
        It displays ten distinct lines representing the Rank and Weight variants for five methods (FSL, Listener, Multiple FSL, Output Aware, and Residual).
        The chart visually confirms the data trends: lines for FSL, Multiple FSL, and Listener rise steeply and plateau at an F1-score of 1.0 within the first 100 features. 
        The Output Aware method follows a similar steep trajectory, plateauing at comparable results in under 200 features. 
        In contrast, the lines for the Residual extension remain near the bottom of the graph, exhibiting a stepped, fluctuating pattern at much lower F1-scores (predominantly between 0.0 and 0.4) across the entire 1000-feature range, highlighting its significantly lower performance.
     ]
    {\textbf{Comparing SVC prediction performance on top-1000 features ranked by weight from FSL and proposed extensions for Breast dataset.}
        Vertical axis shows the F1 score on test dataset.
        Horizontal axis shows the number of features being used, from 1 to 1000. 
        Feature subsets were defined by the general feature weight given for each specific method.
        For each method, two comparisons were conducted: one using only feature subsets, and another using weighted features.
    }
     \includegraphics[width=0.75\textwidth]{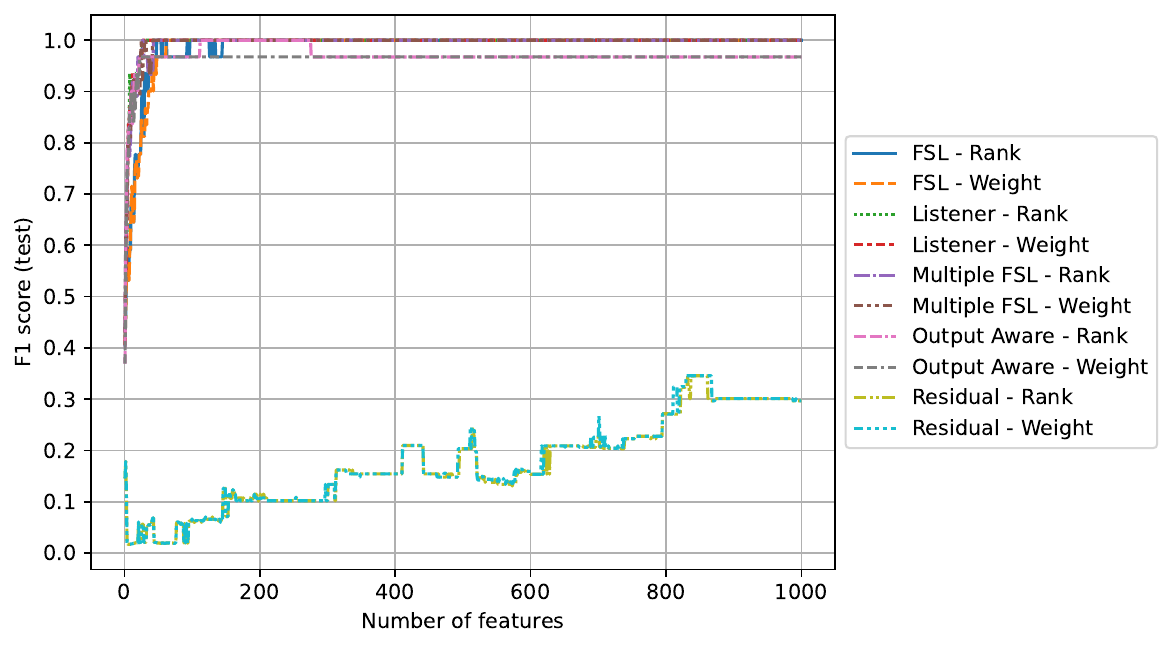}
     \label{fig:predictor_extensions_for_breast}
\end{figure*}

\begin{figure*}[h]
    \centering
     \begin{subfigure}[b]{0.45\textwidth}
         \centering
         \includegraphics[width=\textwidth]{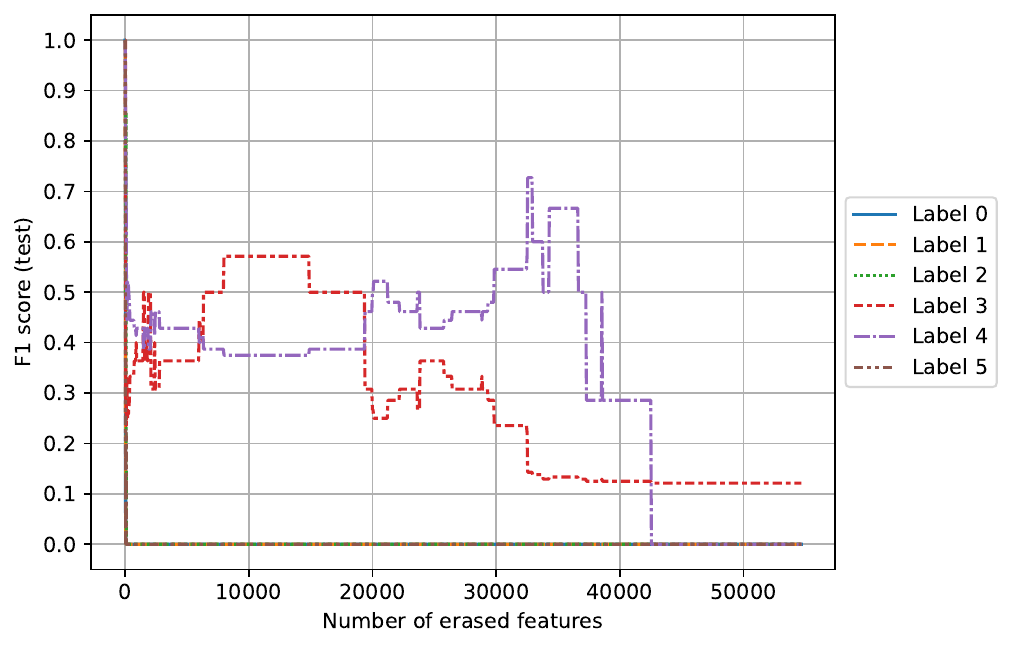}
         \caption{Listener}
         \label{subfig:occlusion_on_extensions_for_breast_label_0_listener}
     \end{subfigure}
     \hfill
     \begin{subfigure}[b]{0.45\textwidth}
         \centering
         \includegraphics[width=\textwidth]{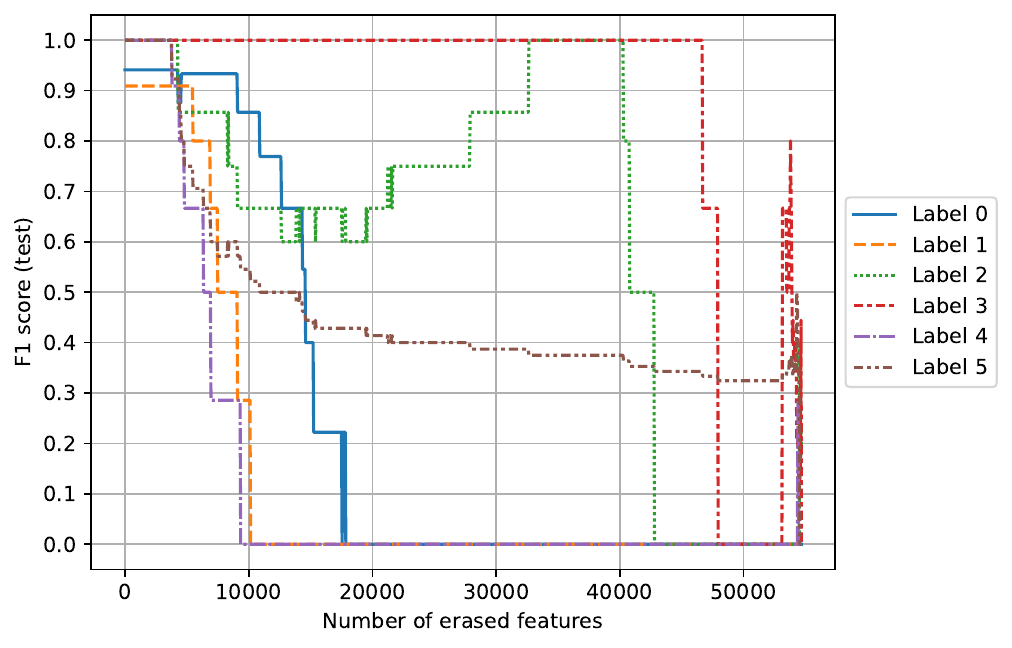}
         \caption{Residual}
         \label{subfig:occlusion_on_extensions_for_breast_label_0_residual}
     \end{subfigure}
     \hfill
     \begin{subfigure}[b]{0.45\textwidth}
         \centering
         \includegraphics[width=\textwidth]{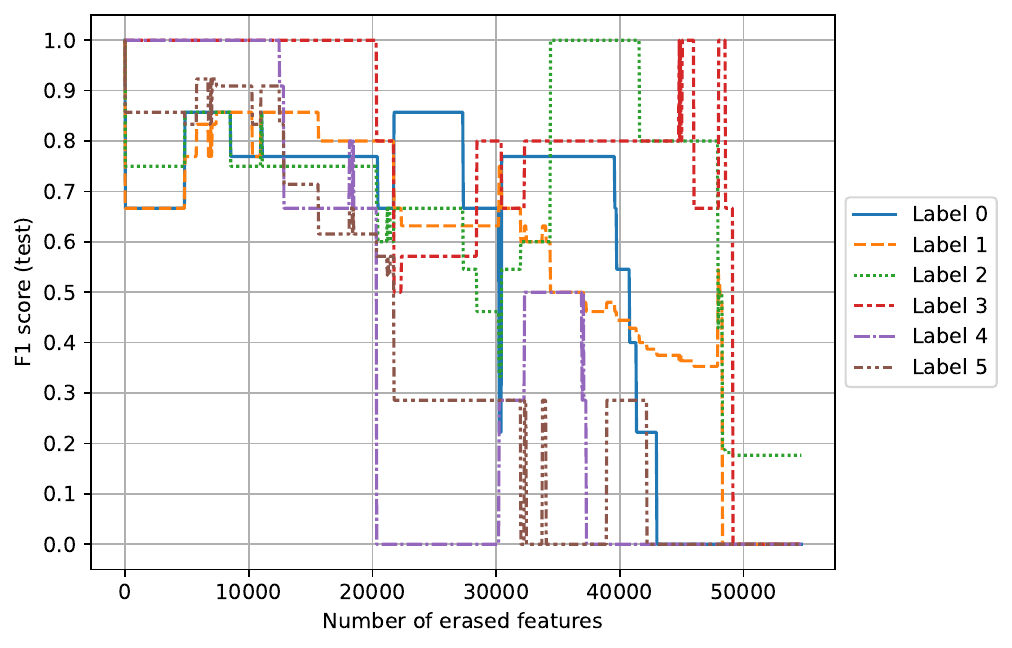}
         \caption{Output Aware}
        \label{subfig:occlusion_on_extensions_for_breast_label_0_output_aware}
     \end{subfigure}
     \hfill
     \begin{subfigure}[b]{0.45\textwidth}
         \centering
         \includegraphics[width=\textwidth]{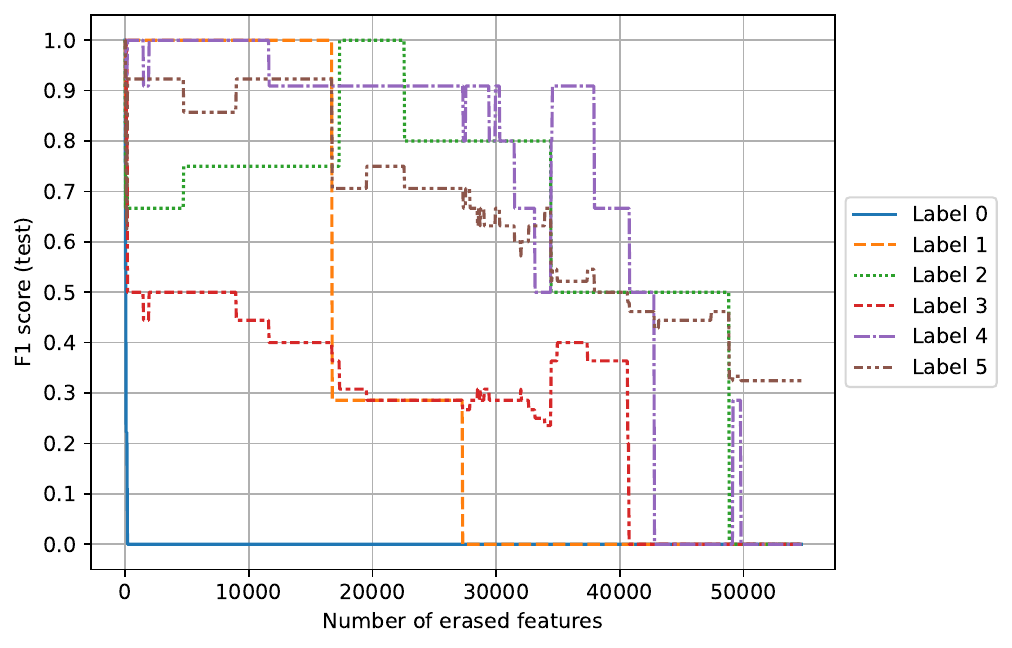}
         \caption{Multiple FSL}
         \label{subfig:occlusion_on_extensions_for_breast_label_0_multiple_fsl}
     \end{subfigure}
     \caption[
        Impact of feature erasure based on label 0 feature weights for the Breast dataset. 
        Features were erased sequentially from the dataset following a ranking derived from the weights assigned specifically to label 0. 
        The vertical axis displays the F1-score calculated from the test dataset, while the horizontal axis shows the number of erased features. 
        Visually, the figure comprises a two-by-two grid of line charts representing four methods: (a) Listener, (b) Residual, (c) Output Aware, and (d) Multiple FSL. 
        Each chart plots the F1-score (from 0.0 to 1.0) against the number of erased features (from 0 to over 50,000) using six distinct lines, one for each label (Label 0 through Label 5). 
        The charts visually confirm the analytical findings: in (d) Multiple FSL, the solid blue line (Label 0) drops precipitously to zero almost immediately upon feature erasure, demonstrating the model's heavy reliance on the top-ranked features essential for predicting that specific label. 
        A similarly steep drop for Label 0 occurs in (a) Listener; however, the F1-scores for several other labels also degrade significantly early on, aligning with its lower ability to isolate label-specific features. 
        Conversely, charts (b) Residual and (c) Output Aware exhibit a much more gradual, step-wise decline for Label 0, accompanied by high variance and fluctuation across the remaining labels as features are removed.
    ]
    {\textbf{Impact of feature erasure based on label 0 feature weights for Breast dataset.}
        The features were erased from the dataset following a ranking based on the feature weights given for label 0.
        The horizontal axis shows the number of erased features, while the vertical axis displays the F1 score calculated from the test dataset.
    }
    \label{fig:occlusion_on_extensions_for_breast_label_0}
\end{figure*}

To assess the performance of the algorithms on real-world data, we employed the Breast dataset.
When comparing the prediction scores, FSL, Listener, and Multiple FSL achieved statistically significant better performance for all labels considering precision, recall and F1 score metrics.
Residual demonstrated the lowest prediction performance, with inconsistent results for most of the labels.
The complete results with statistical significance are provided in \autoref{tab:comparative_selector_prediction_on_extensions_for_breast} of \ref{sec:complementary_experiments}.
All algorithms, except Residual, achieved high prediction scores, then we expect they are able to assign higher weights to relevant features, similar to the observations made for synthetic datasets.
\autoref{fig:predictor_extensions_for_breast} presents an analysis of the impact of subsets with the top 1000 features, ranked by weight, on the predictive performance of a Support Vector Classifier (SVC).
The analysis was limited to the top 1000 features as most algorithms needed less than 100 features to achieve the best score of 1.0.
Residual extension had the lowest performance when compared to the other feature weighting algorithms, confirming that it was not able to assign higher weights to the most relevant features compared to the other algorithms.
In contrast, FSL, Multiple FSL, and Listener achieved high performance with feature subsets of less than 100 features, while Output Aware required fewer than 200 features to achieve comparable results, which suggests that they were able to detect more relevant features by assigning higher weights to them.
In addition to assigning higher weights to the most relevant features, the proposed method must be able to identify which features are relevant to specific labels in real-world datasets.
\autoref{fig:feature_wtsne_on_extensions_breast} presents the weighted t-SNE visualizations for each proposed extension, using the feature weights specifically assigned to label 0 (red dots) of Breast dataset. When comparing the weighted t-SNE visualizations of Listener (\autoref{subfig:wtsne_listener_on_extensions_breast_label_0}) and Residual (\autoref{subfig:wtsne_residual_on_extensions_breast_label_0}) to the default t-SNE (\autoref{subfig:wtsne_without_weights_on_extensions_breast_label_0}), no significant improvements in label 0 clustering is observed. 
While Output Aware (\autoref{subfig:wtsne_output_aware_on_extensions_breast_label_0}) had a more noticeable impact on the overall visualization, it failed to improve the clustering of label 0 samples mainly. 
Among all algorithms, only Multiple FSL successfully enhanced the separation of label 0 samples specifically.
While the results for label 0 improved, the performance for other labels declined. 
This indicates that by prioritizing features specifically relevant to a label, the algorithm omitted features necessary for distinguishing the others. 
This behavior is consistent with an algorithm capable of identifying label-specific importance.
\autoref{fig:occlusion_on_extensions_for_breast_label_0} confirms that Multiple FSL final model relies heavily on the features that it classified as more relevant by applying feature erasure based on label 0 feature relevance, represented by the blue line.
Listener and Multiple FSL were significantly more affected by feature erasure compared to the other algorithms, suggesting that these algorithms gave highest weights for features essential for label 0 predictions. 
Although on Listener the performance on other labels was more affected than on Multiple FSL, this aligns with its lower ability to select label-specific relevant features, as observed in the analysis of synthetic datasets and on weighted t-SNE visualization. 
Output Aware, despite its ability to identify label-specific feature weights demonstrated with synthetic datasets, exhibited decreased performance on microarray datasets. 
In such cases, the model may prioritize noisy features over truly relevant ones, effectively ``memorizing'' the dataset using irrelevant patterns.

\subsection{Proposed extensions results}
\label{subsection:proposed_extensions_results}

When analyzing XOR and SynthA datasets, we could confirm that all extensions are able to assign highest weights to relevant features.
However, when considering label-specific feature relationships within the SynthB and SynthC datasets, only Output Aware and Multiple FSL exhibited the capability to identify such distinctions. 
All proposed extensions increase memory usage by storing label-specific feature weights, unlike the original FSL implementation. 
With $M$ features and $L$ labels, the original FSL adds $M$ new weight parameters to the model, while the proposed extensions add $M \times L$ new weights. 
For Listener, these label-specific weights are accumulated as detached statistics after each batch rather than optimized as trainable parameters, so its GPU peak memory usage remains close to that of the original FSL implementation. 
Output Aware adds a new calculation during the forward pass that causes a small increase in memory usage based on the number of features, but this increase remains independent of the number of labels. 
Residual adds a small residual connection computed separately from the main forward path, which causes no measurable impact on memory consumption compared to the original FSL.
Multiple FSL, on the other hand, increases the size of the weighted input based on both the number of features and labels, making it the only extension whose GPU memory usage scales with the number of labels. 
In our experiments, this increase in memory did not translate into longer execution times, as the hardware had sufficient memory to accommodate all network parameters. 
However, it is important to note that Multiple FSL can instead be implemented to run a separate forward pass per class rather than a single forward pass with an enlarged input, reducing memory consumption to the level of the other extensions while increasing execution time. 
On the Breast dataset (6 labels), for example, running a forward pass per class caused an average $3.8\times$ increase in execution time when compared with the original FSL.
Despite the higher memory usage, Multiple FSL consistently achieved the best results across all datasets.
Output Aware, while effective on synthetic datasets, caused a degradation impact on the model prediction metrics and underperformed on the Breast dataset, which has more features and less samples than the other datasets, yielding lower accuracy and lower capacity to identify relevant features.
Unlike Output Aware, given the multiple feed-forward executions, Multiple FSL is able to train each Feature Selection Layer with all available samples, reducing the bias and having the most consistent performance.
Similar to the original FSL implementation, the selection layers directly influence the feature values used for prediction. 
However, each layer affects only the label to which it is explicitly associated.
In the following section, we will compare the proposed Multiple FSL with other feature weighting techniques.

\subsection{Comparative analysis with existent feature weighting algorithms} 
\label{subsec:comparison_with_existent_feature_weighting_algorithms}

\begin{figure*}[h]
    \centering
    
    \begin{subfigure}[b]{0.32\textwidth}
         \centering
         \includegraphics[width=\textwidth]{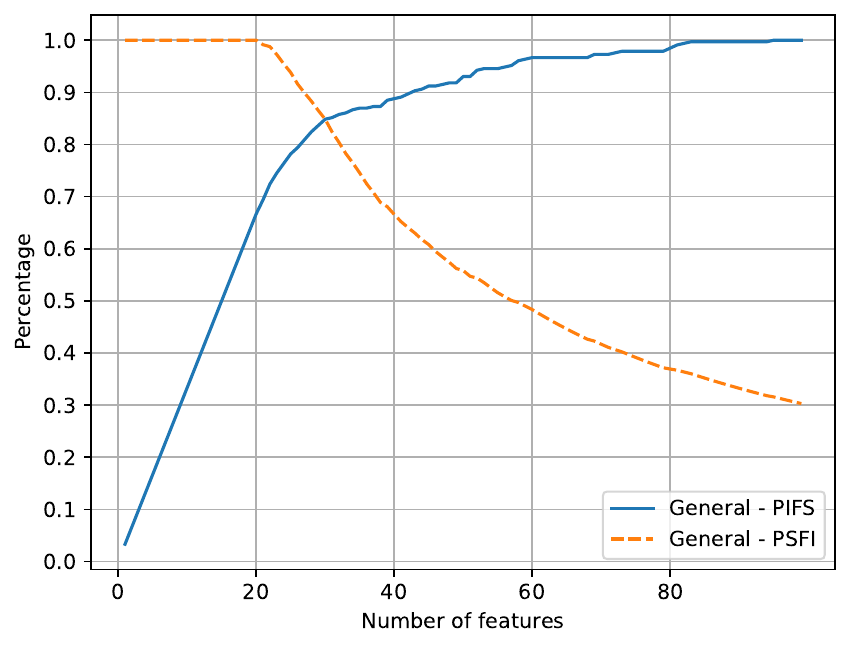}
         \caption{Decision Tree}
         \label{subfig:comparative_informative_features_graph_on_comparison_for_SynthA__decision_tree}
     \end{subfigure}
     \hfill
     \begin{subfigure}[b]{0.32\textwidth}
         \centering
         \includegraphics[width=\textwidth]{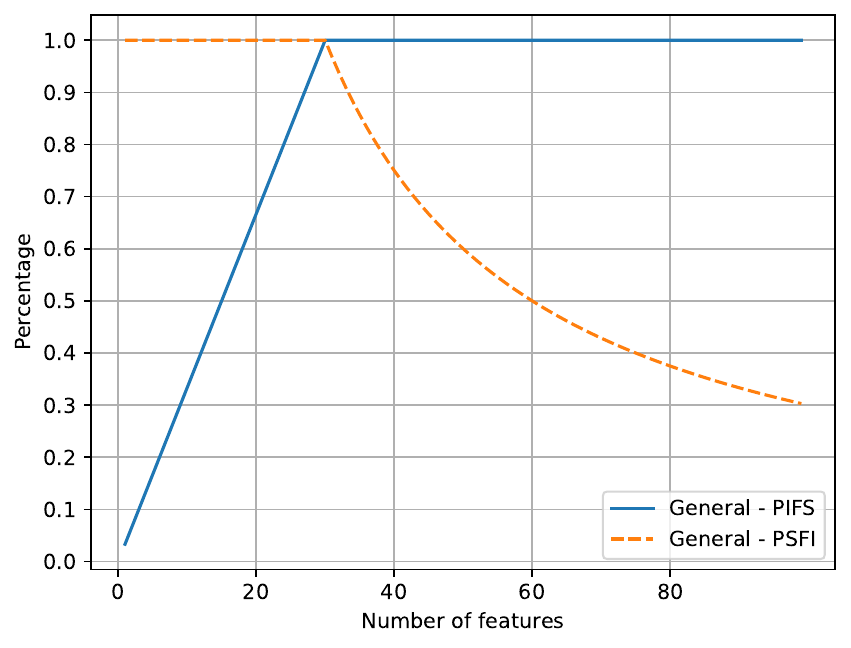}
         \caption{Random Forest}
         \label{subfig:comparative_informative_features_graph_on_comparison_for_SynthA__random_forest}
     \end{subfigure}
     \hfill
     \begin{subfigure}[b]{0.32\textwidth}
         \centering
         \includegraphics[width=\textwidth]{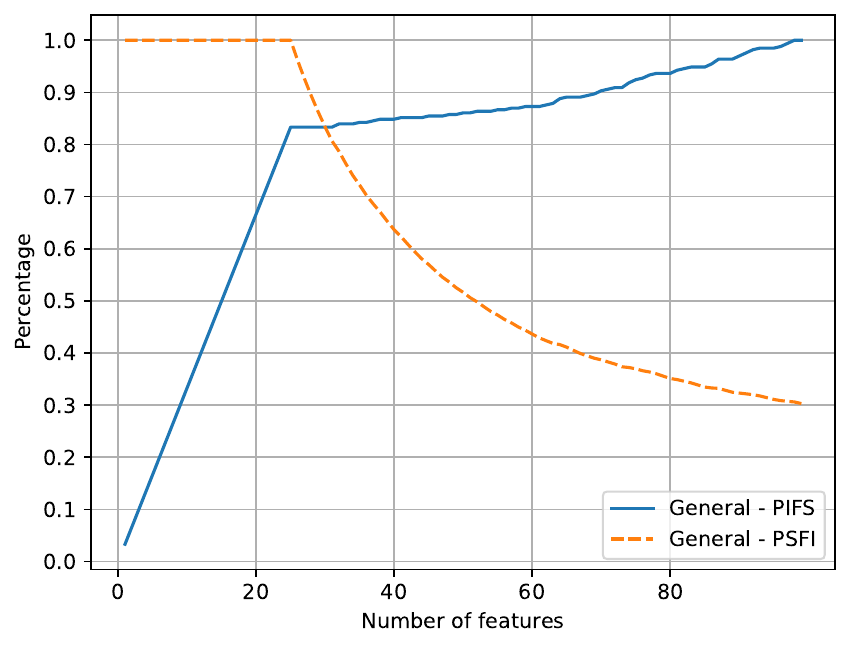}
         \caption{Kruskal Wallis}
         \label{subfig:comparative_informative_features_graph_on_comparison_for_SynthA__kruskal_wallis}
     \end{subfigure}
     \begin{subfigure}[b]{0.32\textwidth}
         \centering
         \includegraphics[width=\textwidth]{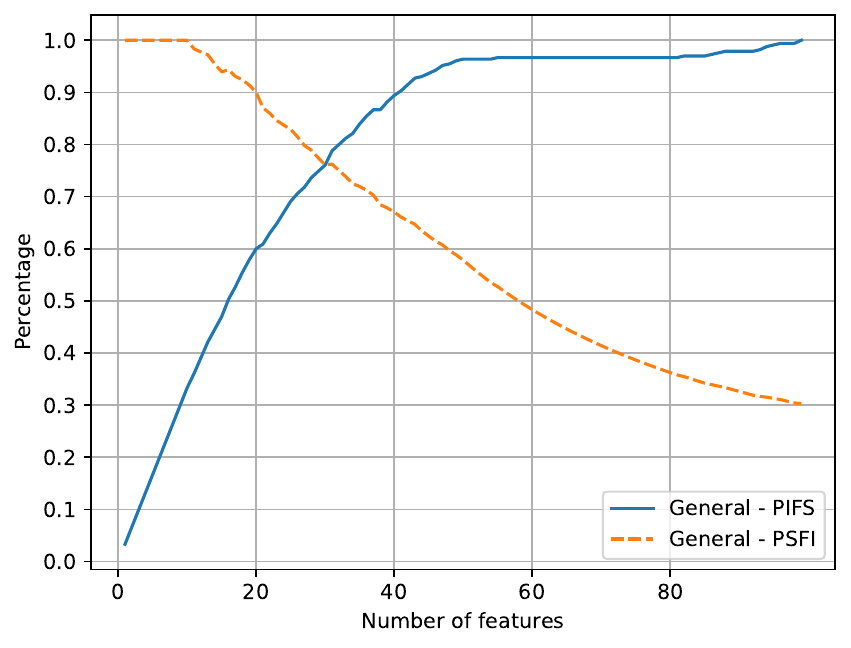}
         \caption{ReliefF}
         \label{subfig:comparative_informative_features_graph_on_comparison_for_SynthA__relieff}
     \end{subfigure}
     \hfill
     \begin{subfigure}[b]{0.32\textwidth}
         \centering
         \includegraphics[width=\textwidth]{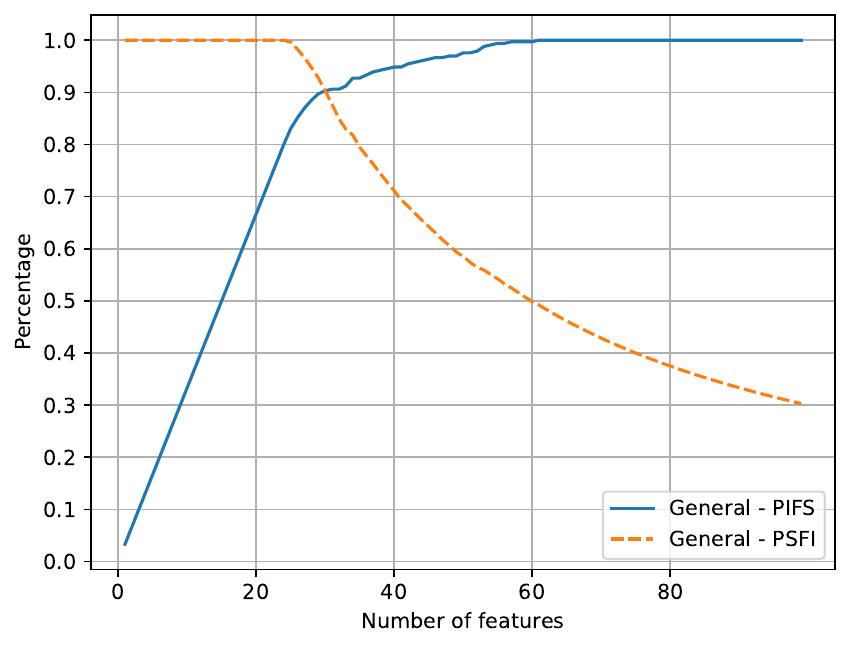}
         \caption{Lasso}
         \label{subfig:comparative_informative_features_graph_on_comparison_for_SynthA__lasso}
     \end{subfigure}
     \hfill
     \begin{subfigure}[b]{0.32\textwidth}
         \centering
         \includegraphics[width=\textwidth]{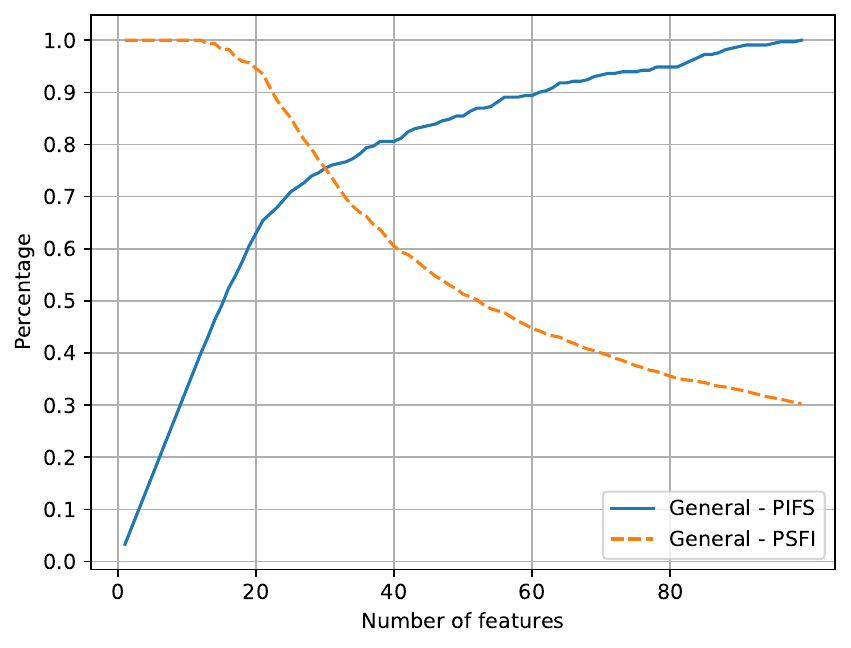}
         \caption{LIME}
         \label{subfig:comparative_informative_features_graph_on_comparison_for_SynthA__lime}
     \end{subfigure}
     \hfill
     \begin{subfigure}[b]{0.32\textwidth}
         \centering
         \includegraphics[width=\textwidth]{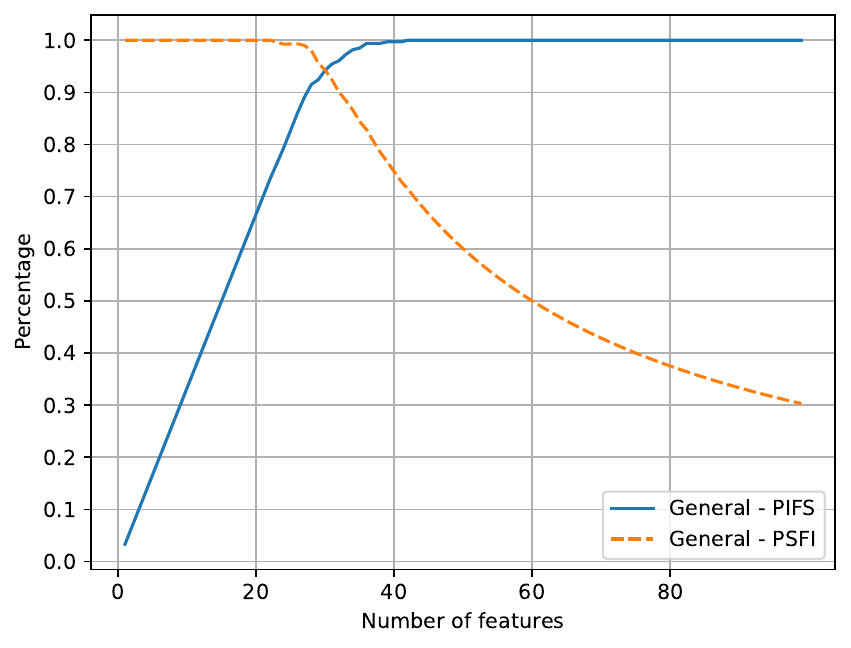}
         \caption{DeepSHAP}
         \label{subfig:comparative_informative_features_graph_on_comparison_for_SynthA__deepshap}
     \end{subfigure}
     \begin{subfigure}[b]{0.32\textwidth}
         \centering
         \includegraphics[width=\textwidth]{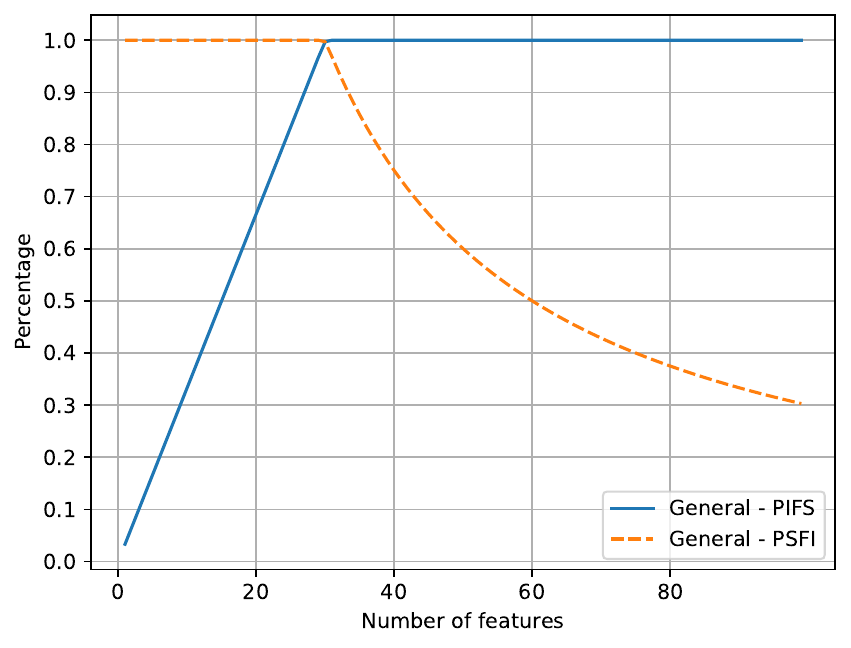}
         \caption{Multiple FSL}
         \label{subfig:comparative_informative_features_graph_on_comparison_for_SynthA__multiple_fsl}
     \end{subfigure}
     \caption[Evaluation of PIFS and PSFI on the SynthA dataset for Multiple FSL and compared feature weighting algorithms.
     Eight charts are shown.
     Each chart contains two lines moving across the horizontal and vertical axes.
     The first line represents the PSFI score, while the second line represents the PIFS score.
     Both scores range from 0 to 1.
     The vertical axis represents the score, while the horizontal axis represents the size of the feature subset considered in the score calculation.
     The subset size ranges from 1 to 100, given the dataset has 100 features, and follow the feature relevance ranking given by each algorithm in each execution.
     Understanding the meaning of PSFI and PIFS, and knowing that the dataset in question contains 30 features, it is important to note that a perfect result would be achieved by an algorithm whose average shows the PSFI value stable at 1.0 up to the top 30 features, and the PIFS value reaching 1.0 exactly at the 30th feature. 
     This means that in all executions of the algorithm, the 30 features assigned with the highest weights are exactly the 30 relevant features.
     In the first chart, we see the results for Decision Tree, where the PSFI score stays at 1.0 up to approximately the top 20 features, while the PIFS score for those same features reaches only 0.7, achieving 1.0 only when over 80 features are considered.
     In the second chart, we see the results for Random Forest, where both scores reach exactly 1.0 with 30 features, representing a perfect score for all its executions.
     In the third chart, we see the results for Kruskal-Wallis, where the PSFI score remains at 1.0 up to about 30 features, but the PIFS score only reaches 1.0 near 100 features. 
     This indicates that most relevant features were consistently ranked among the most important by the algorithm, but some were also frequently ranked among the least relevant too.
     In the fourth chart, we see the results for ReliefF, where PSFI remains at 1.0 until about 10 features, and PIFS gradually increases, reaching 1.0 only near 100 features.
     In the fifth chart, we see the results for Lasso, where PSFI stays at 1.0 up to around 25 features, while PIFS reaches 1.0 before 60 features.
     In the sixth chart, we see the results for LIME, where PSFI remains at 1.0 up to near 20 features, and PIFS gradually increases, reaching 1.0 only when nearly all 100 features are considered.
     In the seventh chart, we see the results for DeepSHAP, where PSFI stays at 1.0 up to about 30 features, and PIFS reaches 1.0 near 40 features.
     In the eighth chart, we see the results for Multiple FSL, which closely resemble the perfect results of Random Forest. 
     PSFI remains at 1.0 up to around 30 features, and PIFS also reaches 1.0 very close to 30 features.
     Although Random Forest and Multiple FSL show very similar results, it is worth noting that while Random Forest achieved perfect results in all its 11 executions, Multiple FSL had some executions with nearly perfect results and others with perfect results, resulting in a slightly lower average performance.]
    {\textbf{Evaluation of PIFS and PSFI on the SynthA dataset for Multiple FSL and compared feature weighting algorithms.} 
        All scores are averages over 11 independent runs per algorithm. 
        Vertical axis represents scores constrained from $0$ to $1$, and the horizontal axis shows subset sizes based on the top-$n$ features by descending feature weight. 
        PSFI is represented by orange lines, PIFS by blue lines.
        With 30 features, a perfect result has PSFI stable at 1.0 up to the top 30 features and PIFS reaching 1.0 at the 30th feature. 
    }.
    \label{fig:comparative_informative_features_graph_on_comparison_for_SynthA}
\end{figure*}

We start by comparing the overall performance and ability of each method to identify general relevant features by testing them on SynthA.
Given the constraints of time and the scope of the research involving multiple datasets and evaluations methods, the presented experiments use a subset of the available feature weighting algorithms: Decision Tree, Random Forest, Lasso, Kruskal Wallis, ReliefF, LIME, and DeepSHAP.
DeepSHAP was executed using a representative background of 1000 instances, as recommended by its library documentation to obtain the best performance in a suitable time.
LIME weights were based on the results for 1000 samples.
DeepSHAP, LIME, and Multiple FSL shared the same neural network architecture, except for the Feature Selection Layers on the last one.
\autoref{fig:comparative_informative_features_graph_on_comparison_for_SynthA} illustrates the results on the SynthA dataset using rank-based feature subsets. 
The blue line represents the average PIFS (Percentage of Informative Features Selected), while the orange line represents the PSFI (Percentage of Selected Features that are Informative). 
The vertical axis denotes the metric score (ranging from 0 to 1), and the horizontal axis represents the subset size (top-$n$ features) ranging from 0 to 100.
For example, in the Decision Tree results (\autoref{subfig:comparative_informative_features_graph_on_comparison_for_SynthA__decision_tree}), subsets containing the top-1 to top-20 features achieve a PSFI score of 1.0.
This indicates that all features within these ranks are informative across all executions. 
However, starting at the top-21 subset, noisy features are introduced. 
The blue line (PIFS) only reaches 1.0 at the top-80 mark, meaning all informative features are only fully captured in subsets of size 80 or larger for all executions. 
This suggests that the Decision Tree consistently ranks 20 informative features highly but assigns lower scores to the remaining 10.
Most algorithms consistently assigned high weights to at least 20 of the 30 informative features. 
Notably, Random Forest (\autoref{subfig:comparative_informative_features_graph_on_comparison_for_SynthA__random_forest}) achieved the best results for the SynthA dataset. 
It consistently identified the 30 informative features as the most relevant, evidenced by the intersection of the PIFS and PSFI curves exactly at the top-30 subset.
Multiple FSL and DeepSHAP followed closely, achieving average PIFS scores of 0.997 and 0.942, respectively, when considering the top-30 features.
As all algorithms successfully assigned higher weights to relevant features, we expect them to be able to improve the performance of a separate model given their features with highest weight only.

\begin{figure*}[h]
    \centering
     \caption[
        Comparison of Support Vector Classifier (SVC) prediction performance trained with the top-n features ranked by weight for the SynthA dataset. 
        The vertical axis shows the test F1-score (from 0.0 to 1.0), while the horizontal axis displays the number of features used (from 0 to 100). 
        For each algorithm, two comparisons are conducted: one using only the ranked feature subsets (Rank) and another using the weighted features (Weight).
        Visually, the figure consists of a single line chart plotting these two variations for eight algorithms: Decision Tree, DeepSHAP, Kruskal-Wallis, LIMESelector, Lasso, Multiple FSL, Random Forest, and ReliefF. 
        All lines rise steeply together from the start, reaching peak F1-scores between 20 and 40 features, with Random Forest, Multiple FSL, and DeepSHAP visually hitting the highest peaks (around 0.94 to 0.96). 
        After this peak, a clear divergence occurs between the two evaluation types. 
        The lines representing the standard feature subsets (Rank) generally begin to decline steadily toward 0.8 as the number of features approaches 100. 
        In contrast, several dashed lines representing the weighted features (Weight)—specifically those for Multiple FSL, DeepSHAP, Random Forest, Lasso, and Decision Tree—plateau and run horizontally near the top of the chart. 
        These top dashed lines remain high and stable across the rest of the graph, visually illustrating their resilience against the inclusion of additional noisy features.
     ]
    {\textbf{Comparing SVC prediction performance trained with top-$n$ features ranked by weight from FSL and proposed extensions for SynthA dataset.}
        Vertical axis shows the F1 score on test dataset.
        Horizontal axis shows the number of features being used. 
        Feature subsets were defined by the general feature weight given for each specific algorithm.
        For each algorithm, two comparisons were conducted: one using only feature subsets, and another using weighted features.
    }
     \includegraphics[width=0.75\textwidth]{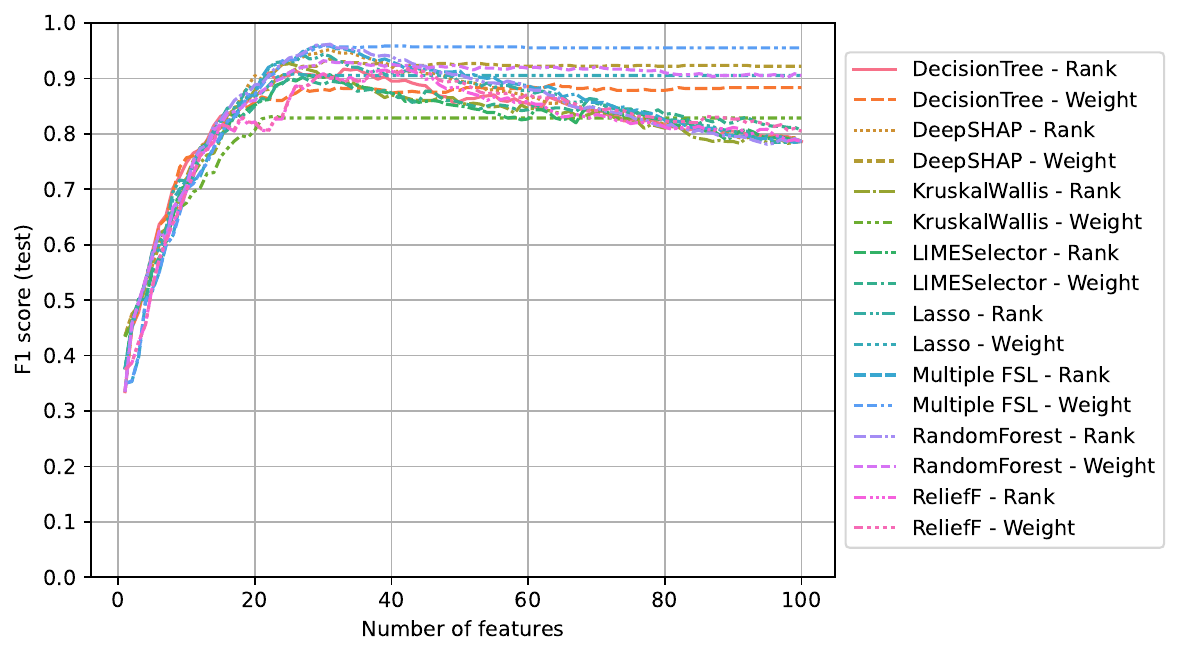}
     \label{fig:predictor_comparison_for_synthA}
\end{figure*}

\begin{figure*}[h]
    \centering
     \begin{subfigure}[b]{0.45\textwidth}
         \centering
         \includegraphics[width=\textwidth]{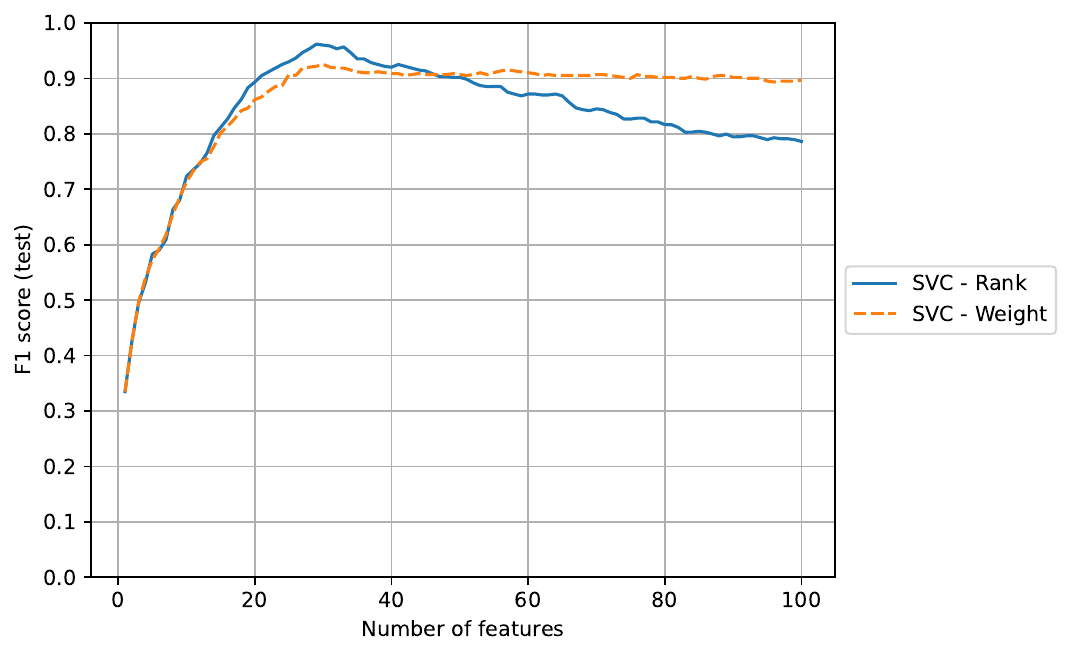}
         \caption{Random Forest}
         \label{subfig:predictor_comparison_for_synthA_specific_algorithms_random_forest}
     \end{subfigure}
     \hfill
     \begin{subfigure}[b]{0.45\textwidth}
         \centering
         \includegraphics[width=\textwidth]{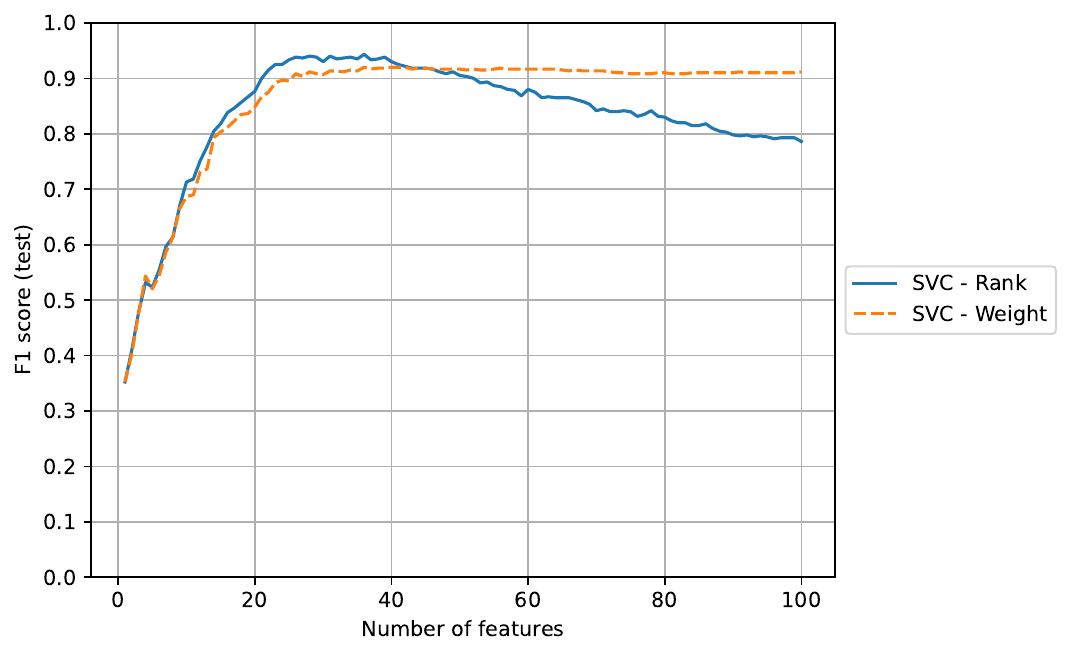}
         \caption{DeepSHAP}
         \label{subfig:predictor_comparison_for_synthA_specific_algorithms_deepshap}
     \end{subfigure}
     \hfill
     \begin{subfigure}[b]{0.45\textwidth}
         \centering
         \includegraphics[width=\textwidth]{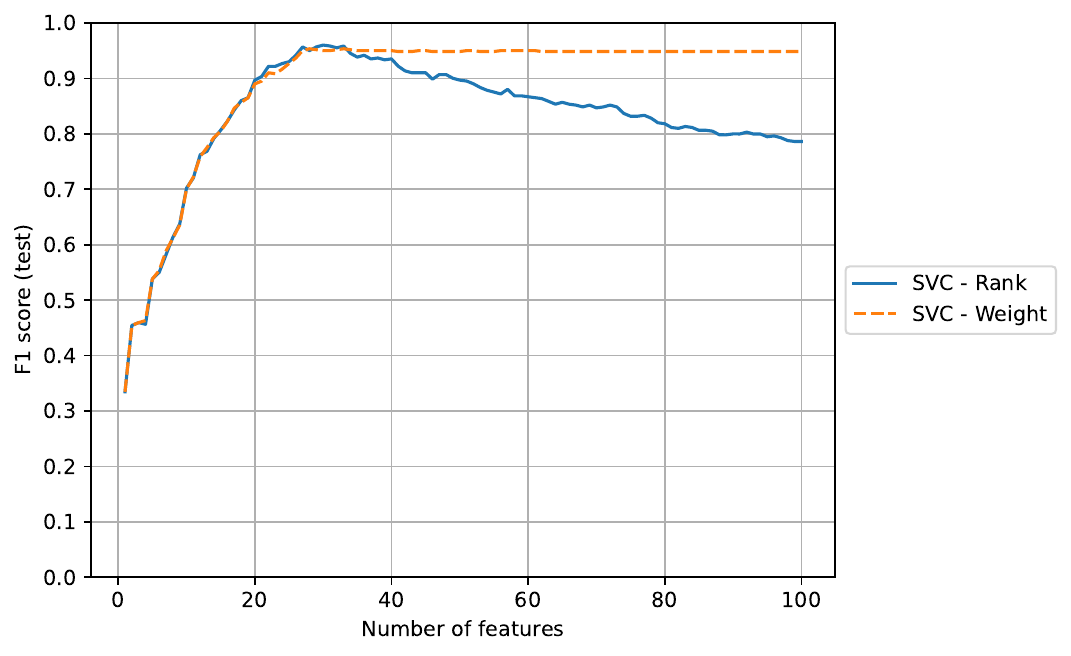}
         \caption{Multiple FSL}
        \label{subfig:predictor_comparison_for_synthA_specific_algorithms_multiplefsl}
     \end{subfigure}
     \caption[
        Comparison of SVC prediction performance trained with the top-n features ranked by weight for the SynthA dataset. 
        Multiple feature subsets of increasing size are generated based on the feature weights assigned by each method, which are then used to train an SVC model to report its prediction score. 
        Each algorithm's scores are presented separately for better individual analysis. 
        Visually, the figure consists of three individual line charts labeled (a) Random Forest, (b) DeepSHAP, and (c) Multiple FSL.
        Each chart's vertical axis represents the test F1-score (from 0.0 to 1.0), and the horizontal axis represents the number of features used (from 0 to 100). 
        Every chart plots two lines: a solid blue line representing standard ranked feature subsets (SVC - Rank) and a dashed orange line representing weighted features (SVC - Weight). 
        Across all three charts, the visual pattern is highly consistent: both lines rise steeply from the origin and reach a peak F1-score above 0.9 between 20 and 30 features. After this peak, the solid blue line (Rank) begins a steady downward slope, declining toward 0.8 as the number of features reaches 100. 
        In stark contrast, the dashed orange line (Weight) plateaus near its peak, forming a flat, horizontal line across the remainder of the chart. 
        This visually confirms that these models maintain high scores and demonstrate resilience even when additional noisy features are included.
    ]
    {\textbf{Comparing SVC prediction performance trained with top-$n$ features ranked by weight from FSL and proposed extensions for SynthA dataset.}
        Multiple feature subsets of increasing size are generated based on the feature weights assigned by each method. 
        These subsets are then used to train an SVC model, and its prediction score is reported.
        Each algorithm's scores are presented separately for a better individual analysis.
    }
    \label{fig:predictor_comparison_for_synthA_specific_algorithms}
\end{figure*}

\autoref{fig:predictor_comparison_for_synthA} illustrates the impact of different feature subsets on training a Support Vector Classifier (SVC) model to obtain a high accuracy. 
These subsets are defined based on the general feature weights assigned by each algorithm.
All algorithms demonstrated comparable performance, with Random Forest, Multiple FSL, and DeepSHAP achieving the highest F1 scores of 0.96, 0.96, and 0.94, respectively. 
This aligns with the observation that these models assigned higher weights to a larger number of relevant features.
Moreover, when utilizing weighted features (indicated by the top dashed lines), Multiple FSL, DeepSHAP, Random Forest, Lasso, and Decision Tree (in descending order of performance) exhibited greater robustness. 
\autoref{fig:predictor_comparison_for_synthA_specific_algorithms} shows the effect of the feature erasure individually for Random Forest, DeepSHAP and Multiple FSL. 
They maintained higher scores when considering weighted features even with the inclusion of additional noisy features in the dataset. 
This resilience can be attributed to the assignment of near-zero weights to these less relevant features.

\begin{table}[!b]
\centering
\caption[Stability comparison between algorithms feature weights on SynthA dataset.] {\textbf{Stability comparison between algorithms feature weights on SynthA dataset.}
Pearson and Spearman was measured using all features (100), while Jaccard was measured using the top-30 features given by the feature weighting algorithm, which is the number of informative features.
In bold is the highest average of each dataset and label.
Best results with statistical significance ($p < 0.01$) according to the Kruskal-Wallis and Dunn’s test are highlighted in orange.
}
\label{tab:comparative_stability_for_synthA}
\begin{tabular}{lrrr}
\hline
Algorithm & Pearson & Spearman & Jaccard \\ 
\hline
\multirow{1}{*}{Decision Tree} 
& 0.833$\pm$.044  & 0.607$\pm$.097 & 0.633$\pm$.079  \\ 
\multirow{1}{*}{Random Forest}  
& 0.987$\pm$.002  & 0.688$\pm$.042 & 0.935$\pm$.000  \\ 
\multirow{1}{*}{Kruskal Wallis}
& \cellcolor{orange!20} \textbf{0.999$\pm$.000}  & \cellcolor{orange!20} \textbf{0.999$\pm$.000} & \cellcolor{orange!20}\textbf{1.000$\pm$.000}  \\ 
\multirow{1}{*}{ReliefF}
& \cellcolor{orange!20} 0.997$\pm$.001  & \cellcolor{orange!20} 0.977$\pm$.007 & 0.748$\pm$.061  \\ 
\multirow{1}{*}{Lasso}
& \cellcolor{orange!20} 0.992$\pm$.002  & \cellcolor{orange!20} 0.964$\pm$.012 & 0.925$\pm$.047  \\ 
\multirow{1}{*}{LIME}
& 0.755$\pm$.043  & 0.475$\pm$.101 & 0.489$\pm$.067  \\ 
\multirow{1}{*}{DeepSHAP}
& 0.944$\pm$.011  & 0.744$\pm$.034 & \cellcolor{orange!20} 0.947$\pm$.026  \\ 
\multirow{1}{*}{Multiple FSL}
& 0.966$\pm$.005  & 0.854$\pm$.036 & \cellcolor{orange!20} 0.953$\pm$.030  \\ 
\hline
\end{tabular}
\end{table}

\autoref{tab:comparative_stability_for_synthA} presents stability metrics for Multiple FSL and comparison algorithms using the SynthA dataset. 
We employed the $m-1$ approach, where the algorithm is executed twice: first with the complete feature set, being $m$ equal to 100 features (\autoref{tab:synthetic_used_datasets}), and subsequently with the top-ranked feature removed, totaling 99 features.
Kruskal-Wallis proves to be the most stable algorithm across all metrics. 
ReliefF and Lasso also demonstrate stability regarding feature weights (Pearson) and rankings (Spearman). 
However, their stability decreases, particularly for ReliefF, when evaluating the Jaccard score for the top-30 features.
Finally, while neural network-based algorithms generally exhibit lower stability, DeepSHAP and Multiple FSL achieved strong results across all metrics, performing particularly well on the Jaccard index.
Complementary results using SynthB dataset are provided in \autoref{tab:comparative_stability_for_synthB} of \ref{sec:complementary_experiments}.

\begin{figure*}[!h]
     \centering
     \begin{subfigure}[b]{0.495\textwidth}
         \centering
         \includegraphics[width=\textwidth]{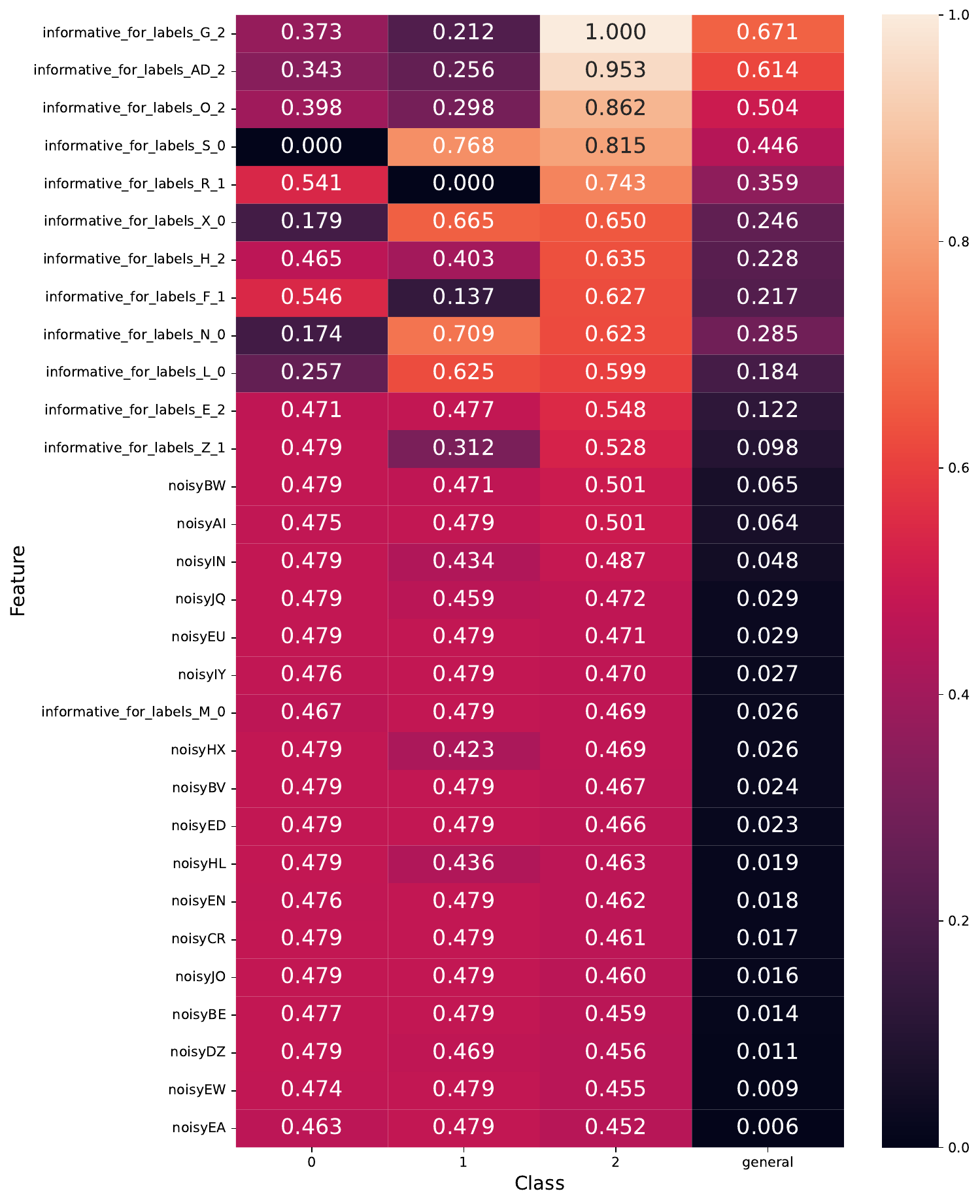}
         \caption{Lasso}
         \label{subfig:heatmap_on_comparison_for_synthB_multiple_lasso}
     \end{subfigure}
     \hfill
     \begin{subfigure}[b]{0.495\textwidth}
         \centering
         \includegraphics[width=\textwidth]{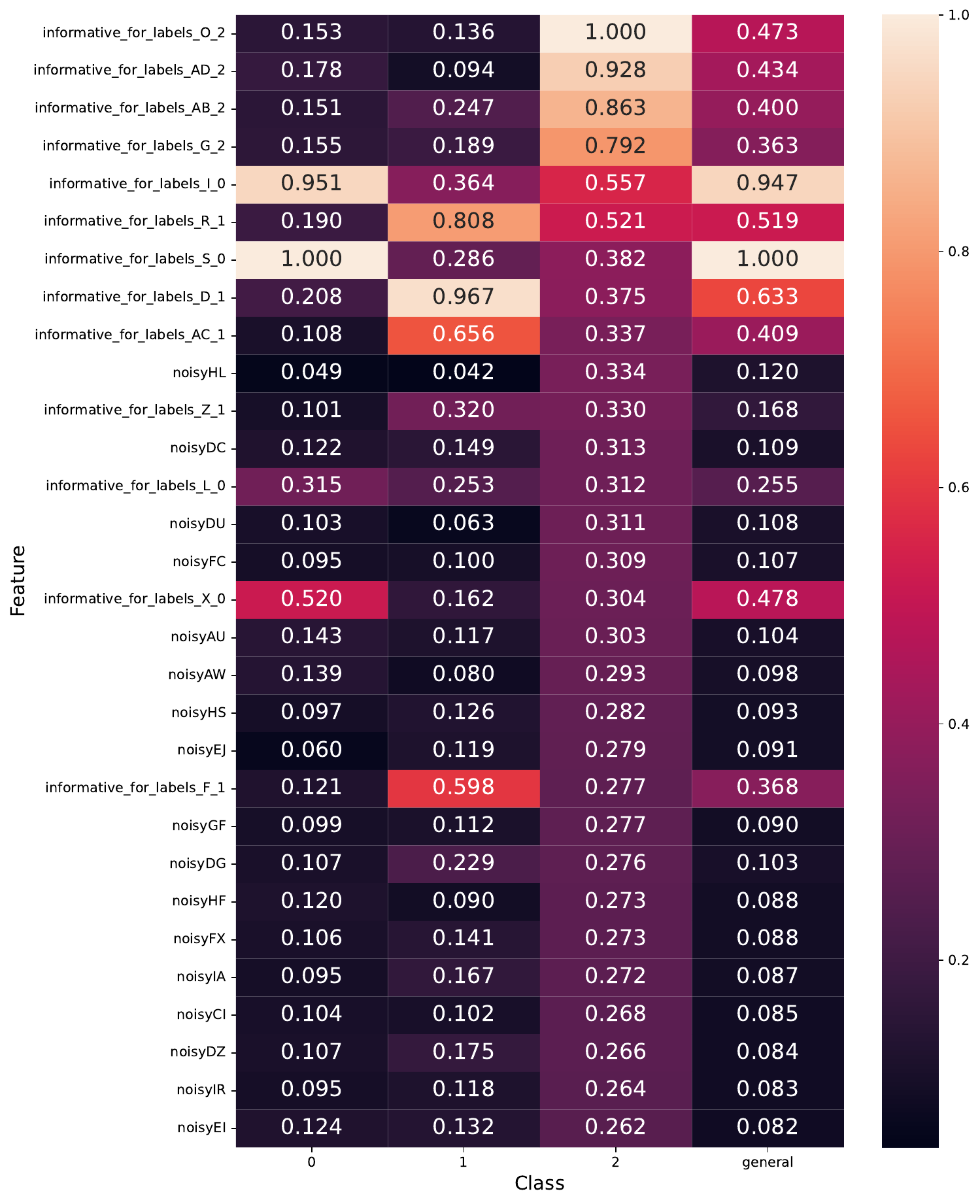}
         \caption{LIME}
         \label{subfig:heatmap_on_comparison_for_synthB_lime}
     \end{subfigure}
     \begin{subfigure}[b]{0.495\textwidth}
         \centering
         \includegraphics[width=\textwidth]{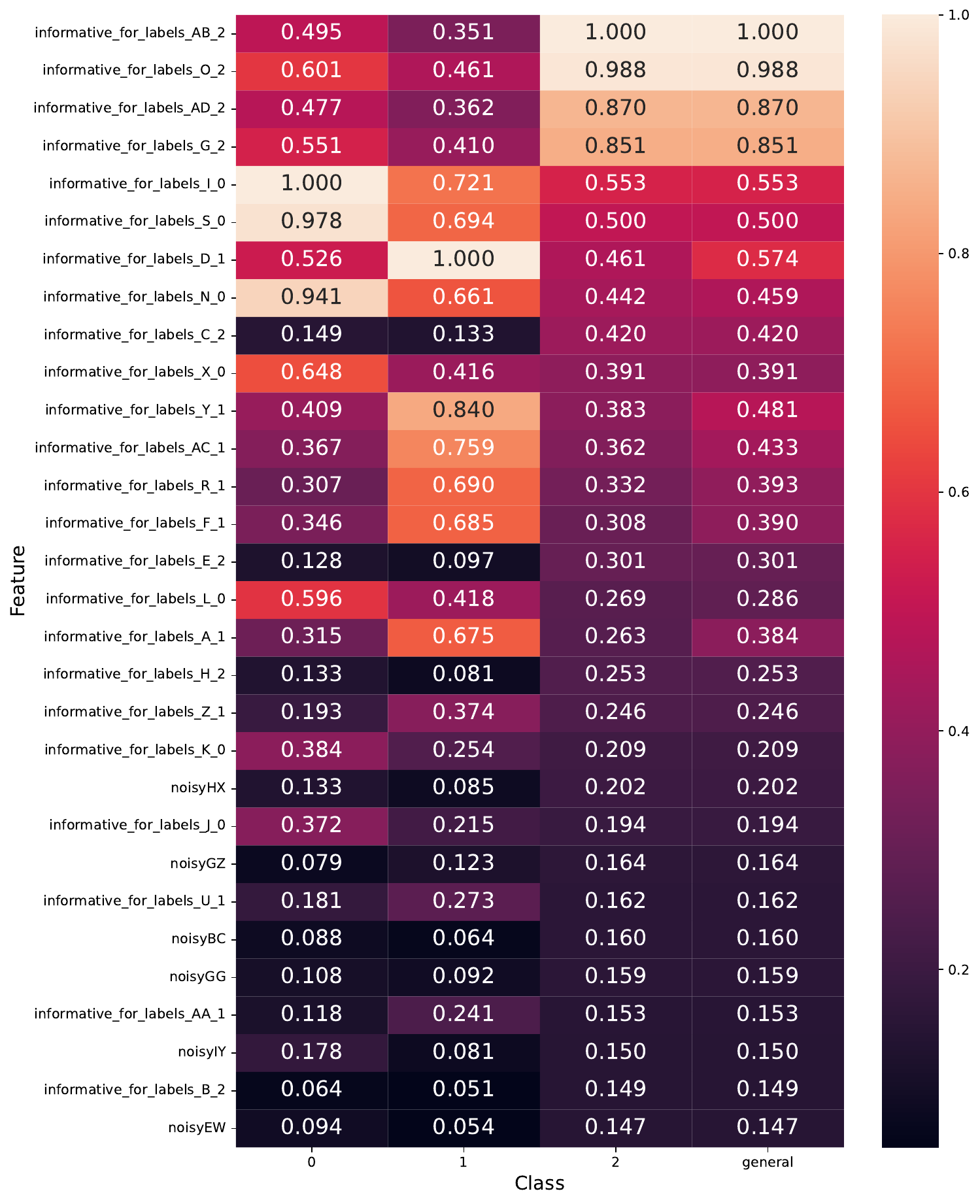}
         \caption{DeepSHAP}
         \label{subfig:heatmap_on_comparison_for_synthB_deepshap}
     \end{subfigure}
     \hfill
     \begin{subfigure}[b]{0.495\textwidth}
         \centering
         \includegraphics[width=\textwidth]{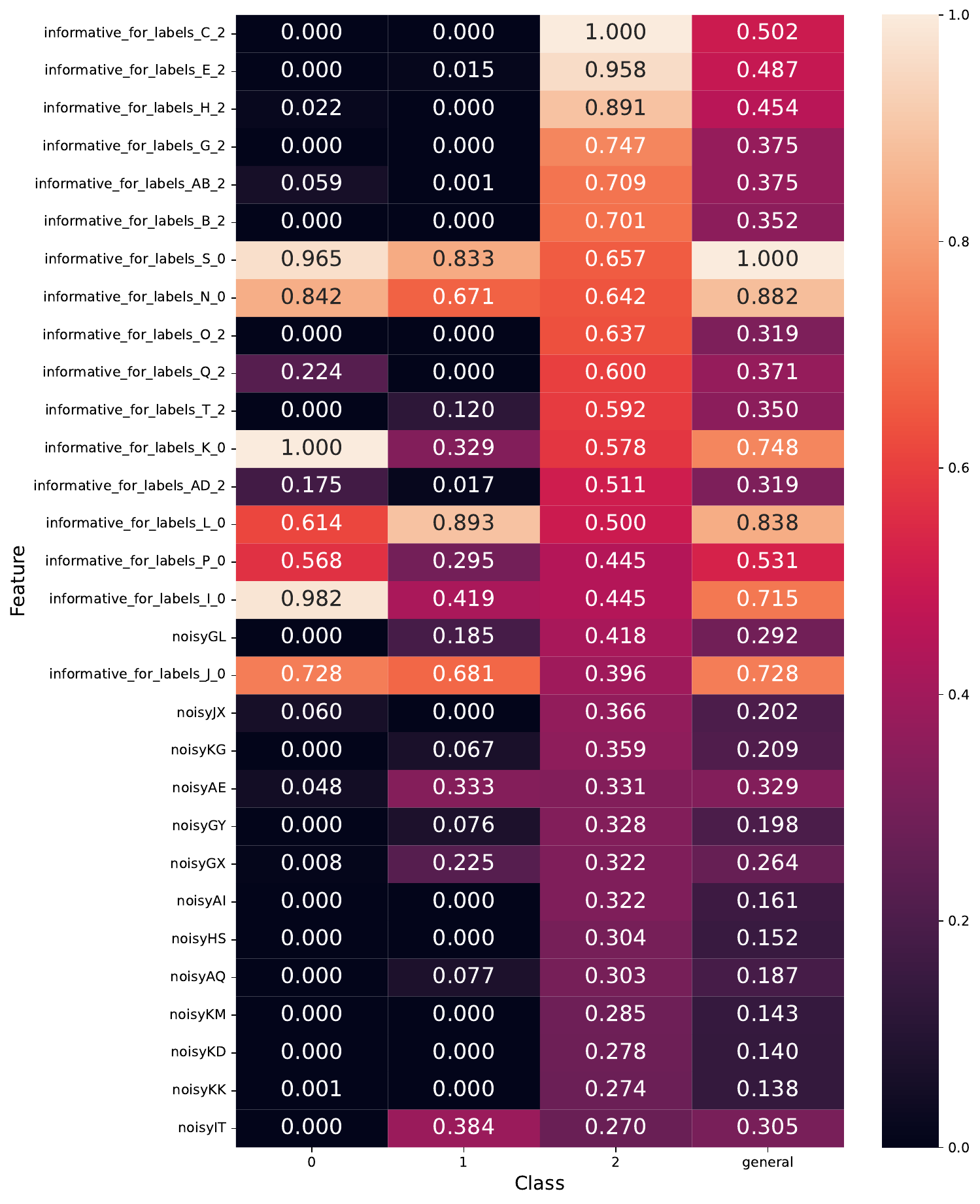}
         \caption{Multiple FSL}
         \label{subfig:heatmap_on_comparison_for_synthB_multiple_fsl}
     \end{subfigure}
    \caption[Heatmap visualization of top-30 features for Lasso, LIME, DeepSHAP, and Multiple FSL, based on their label-specific weights for label 2 of the SynthB dataset.
    Four images are displayed.
    The first shows the heatmap generated by Lasso. 
    In this heatmap, the top three features are features relevant to label 2 that received the highest weights for label 2 and minimal weights for labels 0 and 1.
    The next three features are not directly related to label 2 and exhibit an interesting inversion pattern: the lowest weights are assigned to the label for which the feature is actually relevant, while higher weights are given to unrelated labels. 
    For example, the feature ``informative\_for\_labels\_S\_0'' has a weight of 0.000 for label 0 (its relevant label), 0.768 for label 1 and 0.815 for label 2.
    Similarly, ``informative\_for\_labels\_R\_1'' received a weight of 0.000 for label 1, but 0.541 and 0.743 for labels 0 and 2.
    The second image corresponds to the heatmap generated by LIME. 
    Here, the top four features with the highest weights for label 2 are all features exclusively relevant to that label. 
    Unlike Lasso, LIME does not display the same inversion pattern. 
    The relevant labels consistently receive the highest weights.
    The third image presents the DeepSHAP heatmap. 
    As with LIME, the top four features most heavily weighted for label 2 are also those exclusively relevant to that label. 
    These four features are the same as in LIME, though their order differs. 
    DeepSHAP, like LIME, consistently assigns the highest weights to the labels that the features are actually relevant for.
    The fourth image shows the heatmap produced by Multiple FSL. 
    In this case, the top six features with the highest weights for label 2 are all exclusively relevant to label 2, and they receive near-zero or zero weights for the other labels. 
    The seventh and eighth ranked features are relevant to label 0, but they also received high weights for other labels. 
    The following three features are again exclusively relevant to label 2 and have weights close to zero for the others. 
    As with LIME and DeepSHAP, Multiple FSL avoids the inversion pattern observed in Lasso. 
    However, one exception is the feature ``informative\_for\_labels\_L\_0'', which, although intended to be relevant only for label 0, received a higher score for label 1.]
    {\textbf{Heatmap visualization of top-30 features for Lasso, LIME, DeepSHAP, and Multiple FSL, based on their label-specific weights for label 2 of the SynthB dataset.}
        Normalized feature weights for each algorithm, ranked by their label 2 (third column) feature weights. 
        Feature names prefix indicate if they are informative or noisy, followed by an identification letter sequence, which ensures that all names are unique, and the label which they are related. 
        For example, \textit{informative\_for\_labels\_A\_1} is the informative feature A that is related to label 1, while \textit{noisyAI} is a noisy feature identified by the letters ``AI''.
    }
    \label{fig:heatmap_on_comparison_for_synthB}
\end{figure*}

Lasso, LIME, DeepSHAP and Multiple FSL are able assign label-specific feature weights.
To assess their capacity, we used the SynthB dataset, which has distinct features per each label.
\autoref{fig:heatmap_on_comparison_for_synthB} presents a heatmap with all label-specific and general weights for the top-30 features following SynthB label 2 specific weights.
It confirms that all algorithms were able to assign different weights to features based on the label.
Additional experiments involving the synthetic datasets are provided in Figures \ref{fig:weighted_tsne_on_comparison_for_synthA}, \ref{fig:weighted_tsne_on_comparison_for_synthB} and \ref{fig:comparative_weighted_tsne_label_1_synth_B_new} of \ref{sec:complementary_experiments}.
To further evaluate the capabilities of feature weighting algorithms we compare them on three microarray datasets, we aim to assess their performance on real world high-dimensional datasets with limited samples.
One of these datasets, Liver, is binary. 
For Liver, we evaluate the algorithms ability to assign higher weights to the most relevant features without considering label-specific relations.
Due to the limited sample size of microarray datasets and the high amount of features, the background size for DeepSHAP was reduced to 50. 
Similarly, LIME was configured to evaluate 50 stratified samples to alleviate the necessary processing time associated with the high-dimensional feature space.

\begin{figure*}[h]
    \centering
    \begin{subfigure}[b]{0.32\textwidth}
         \centering
         \includegraphics[width=\textwidth]{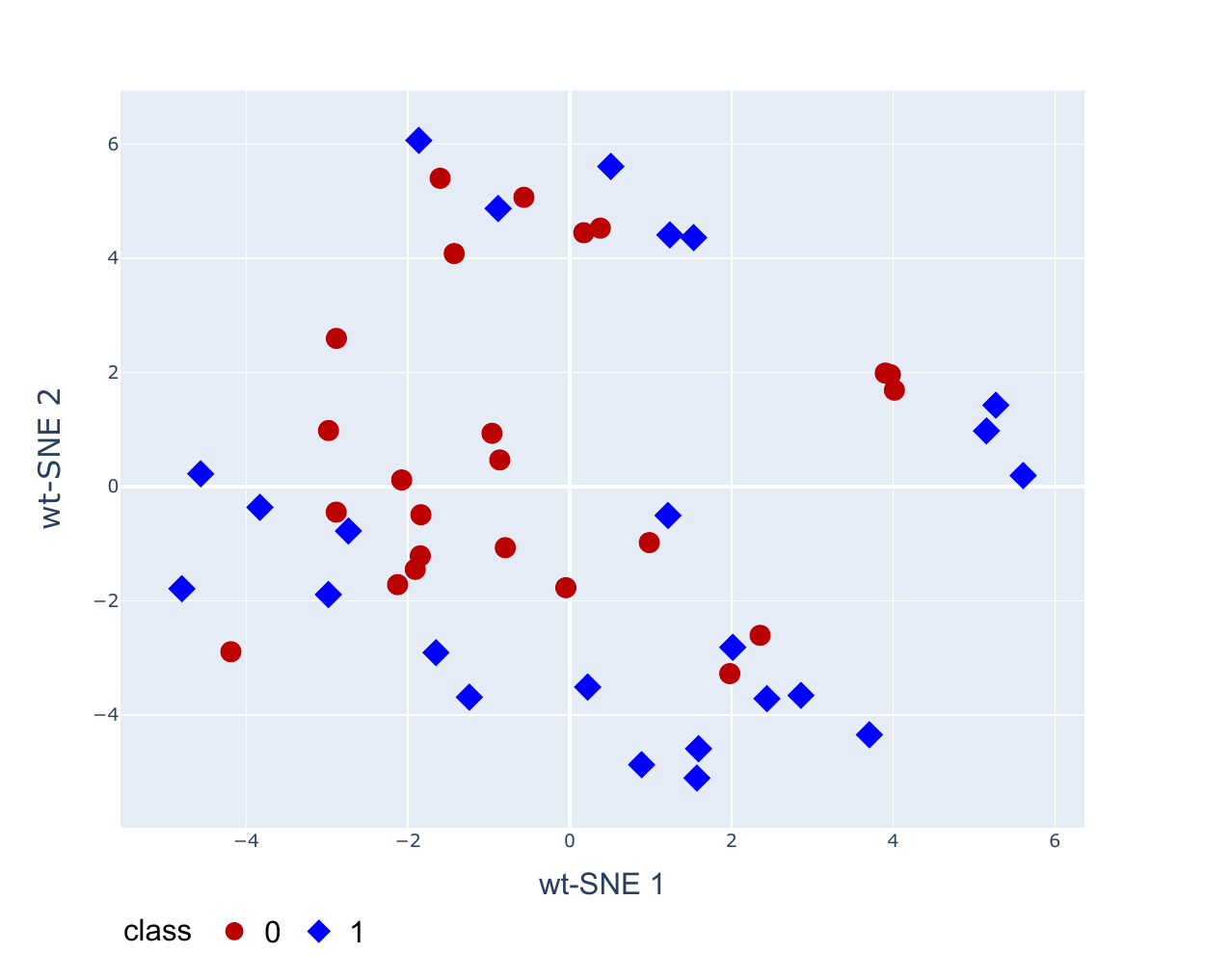}
         \caption{Without weights}
         \label{subfig:comparative_wtsne_model_liver_without_weights}
     \end{subfigure}
     \hfill
     \begin{subfigure}[b]{0.32\textwidth}
         \centering
         \includegraphics[width=\textwidth]{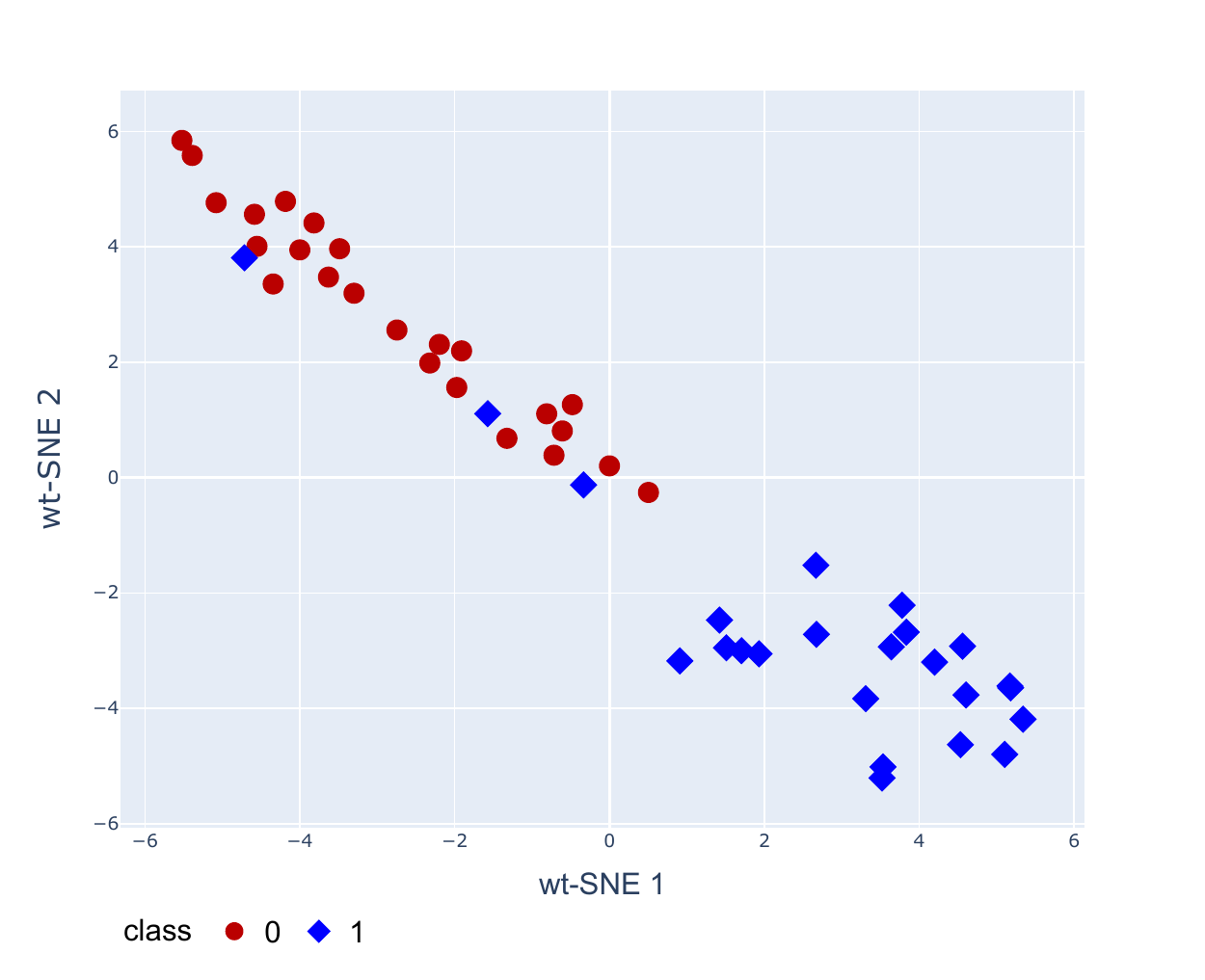}
         \caption{Decision Tree}
         \label{subfig:comparative_wtsne_model_liver_random_decision_tree}
     \end{subfigure}
     \hfill
     \begin{subfigure}[b]{0.32\textwidth}
         \centering
         \includegraphics[width=\textwidth]{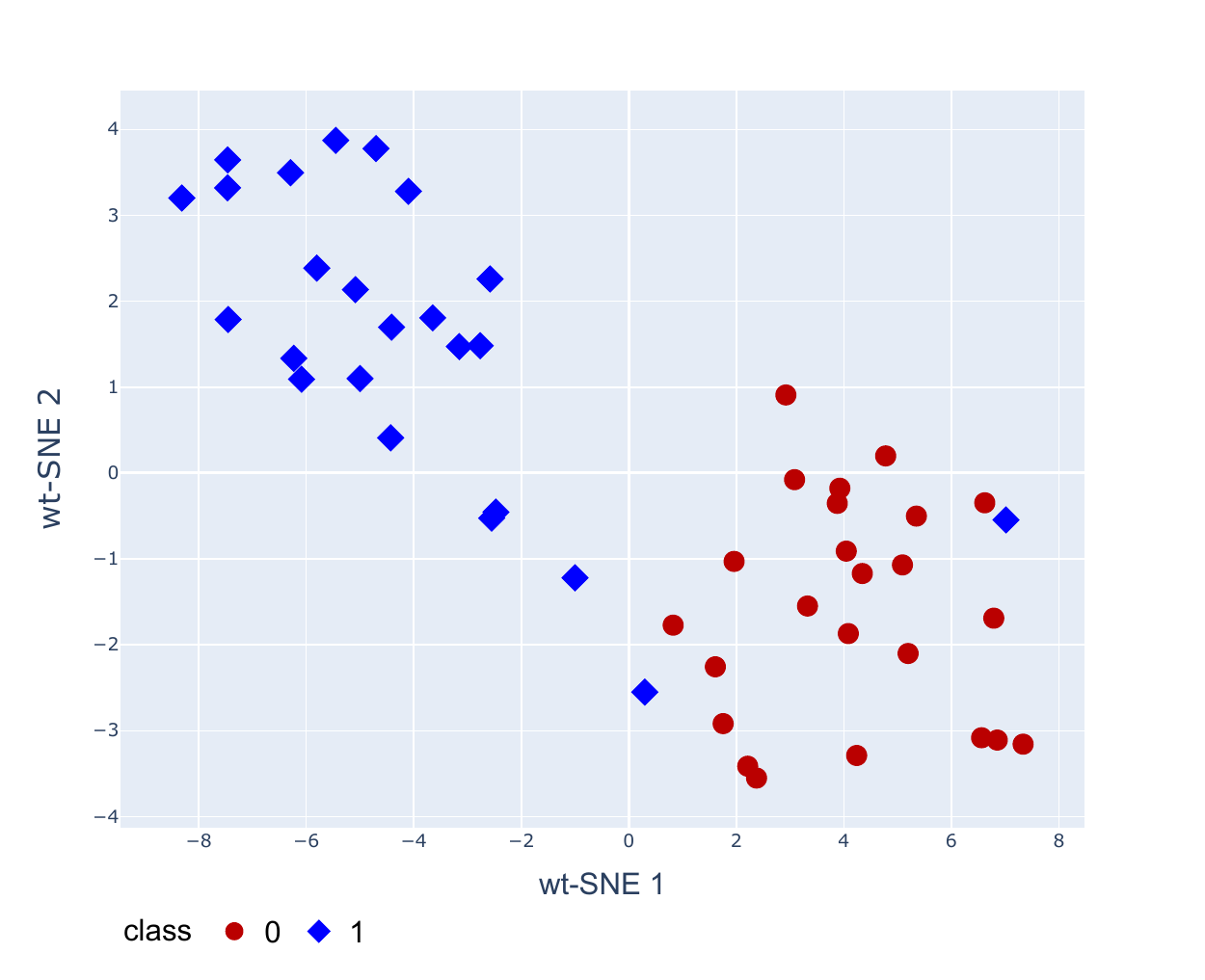}
         \caption{Random Forest}
         \label{subfig:comparative_wtsne_model_liver_random_forest}
     \end{subfigure}
     \begin{subfigure}[b]{0.32\textwidth}
         \centering
         \includegraphics[width=\textwidth]{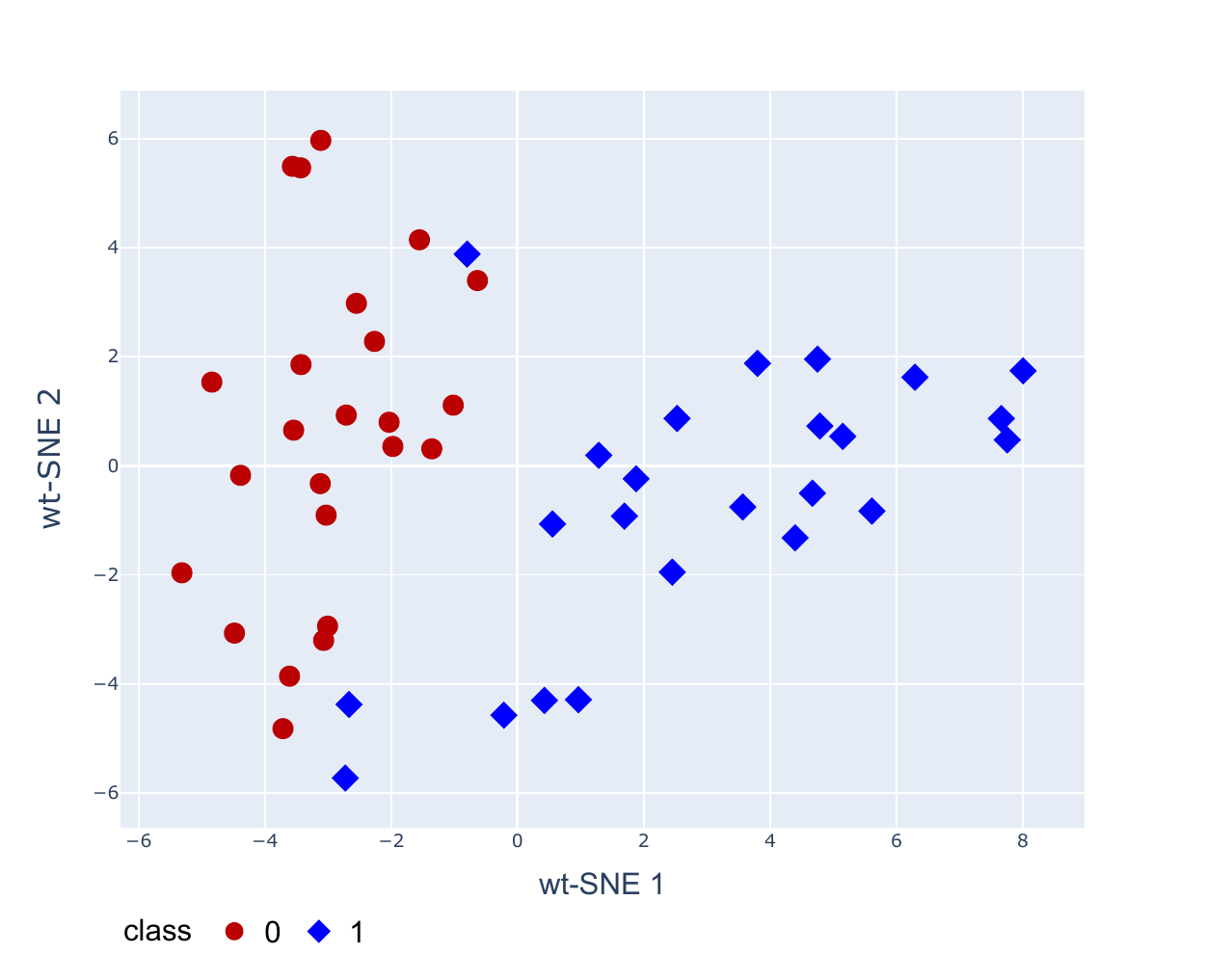}
         \caption{Kruskal Wallis}
         \label{subfig:comparative_wtsne_model_liver_kruskal_wallis}
     \end{subfigure}
     \hfill
     \begin{subfigure}[b]{0.32\textwidth}
         \centering
         \includegraphics[width=\textwidth]{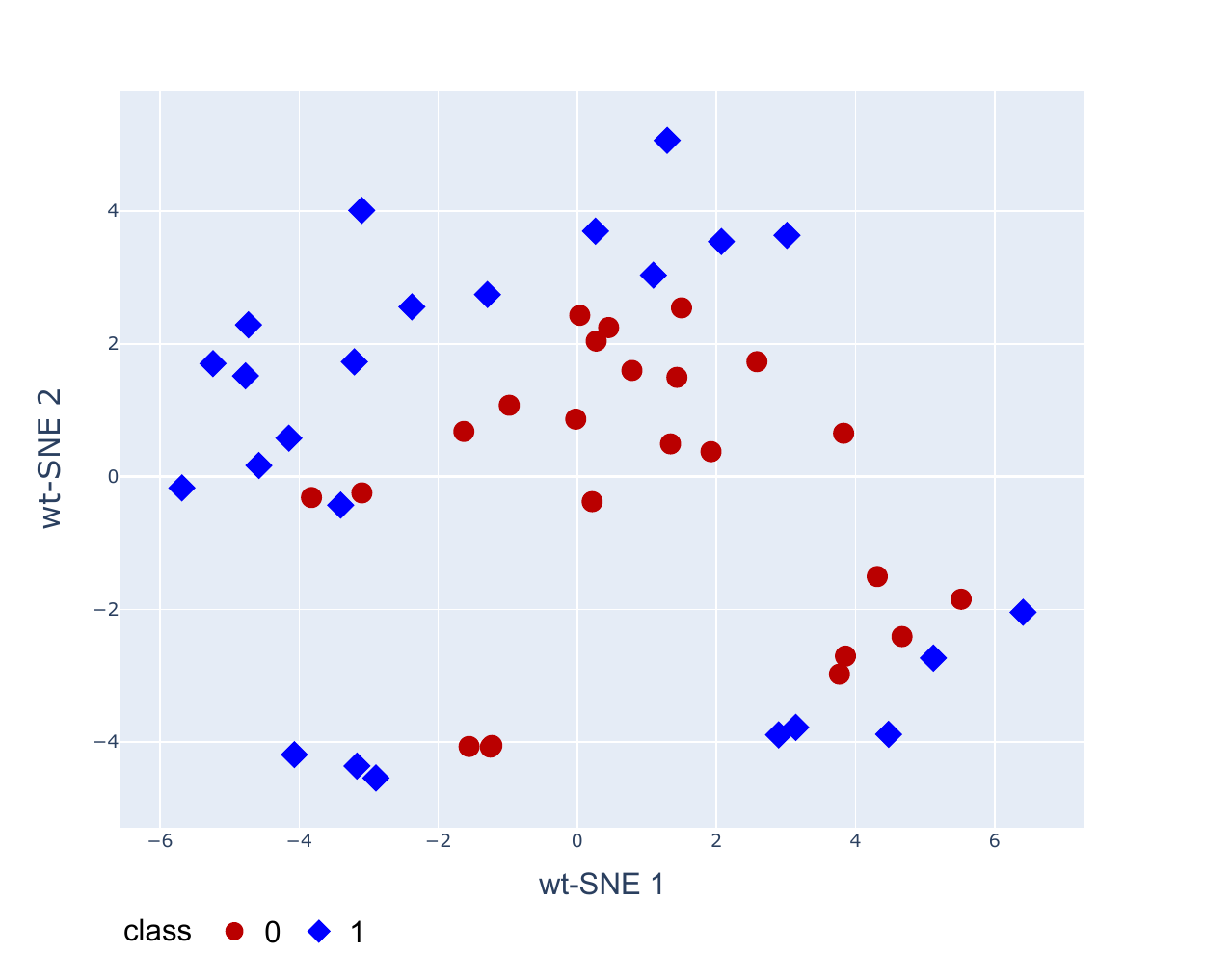}
         \caption{ReliefF}
         \label{subfig:comparative_wtsne_model_liver_relieff}
     \end{subfigure}
     \hfill
     \begin{subfigure}[b]{0.32\textwidth}
         \centering
         \includegraphics[width=\textwidth]{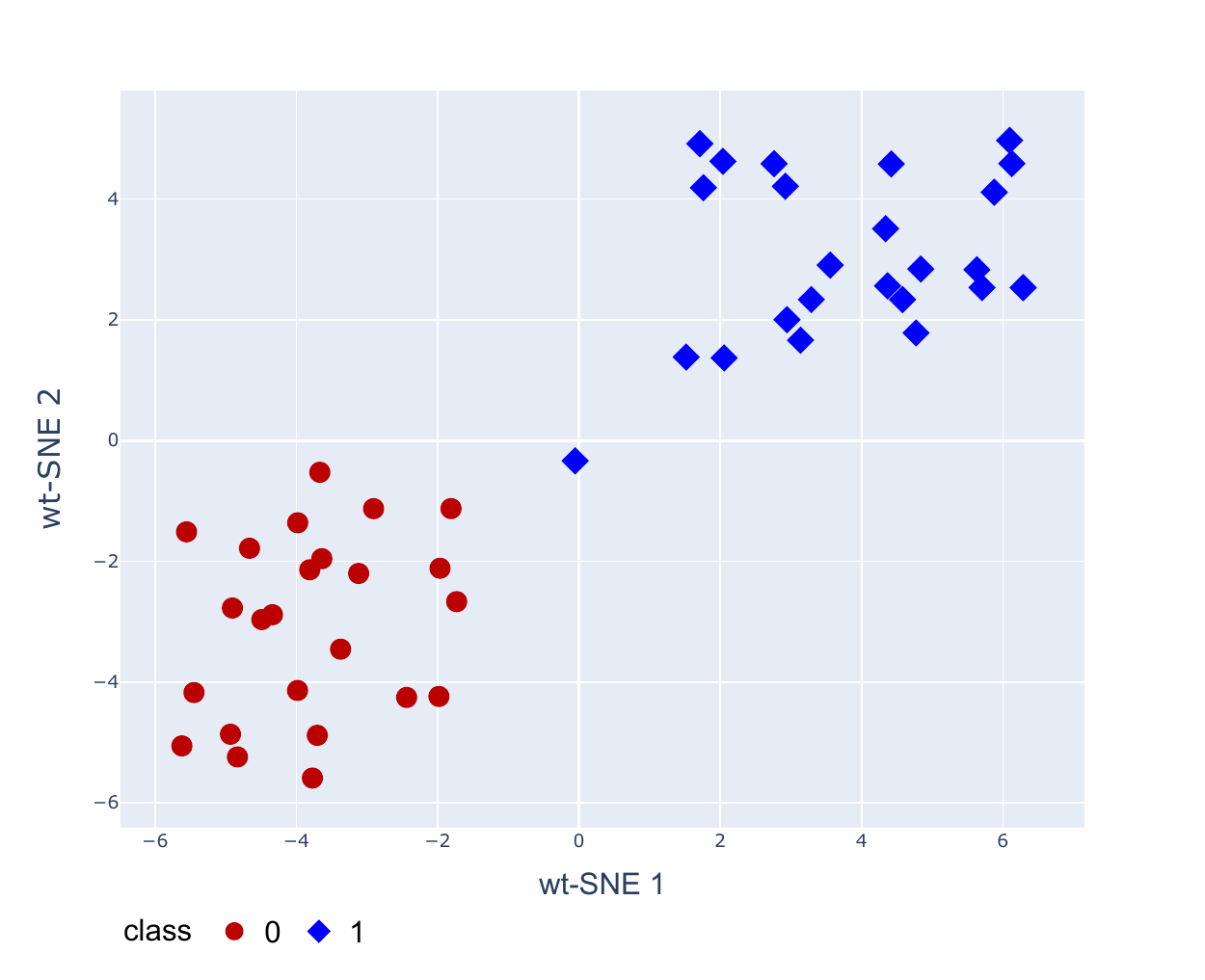}
         \caption{Lasso}
         \label{subfig:comparative_wtsne_model_liver_lasso}
     \end{subfigure}
     \begin{subfigure}[b]{0.32\textwidth}
         \centering
         \includegraphics[width=\textwidth]{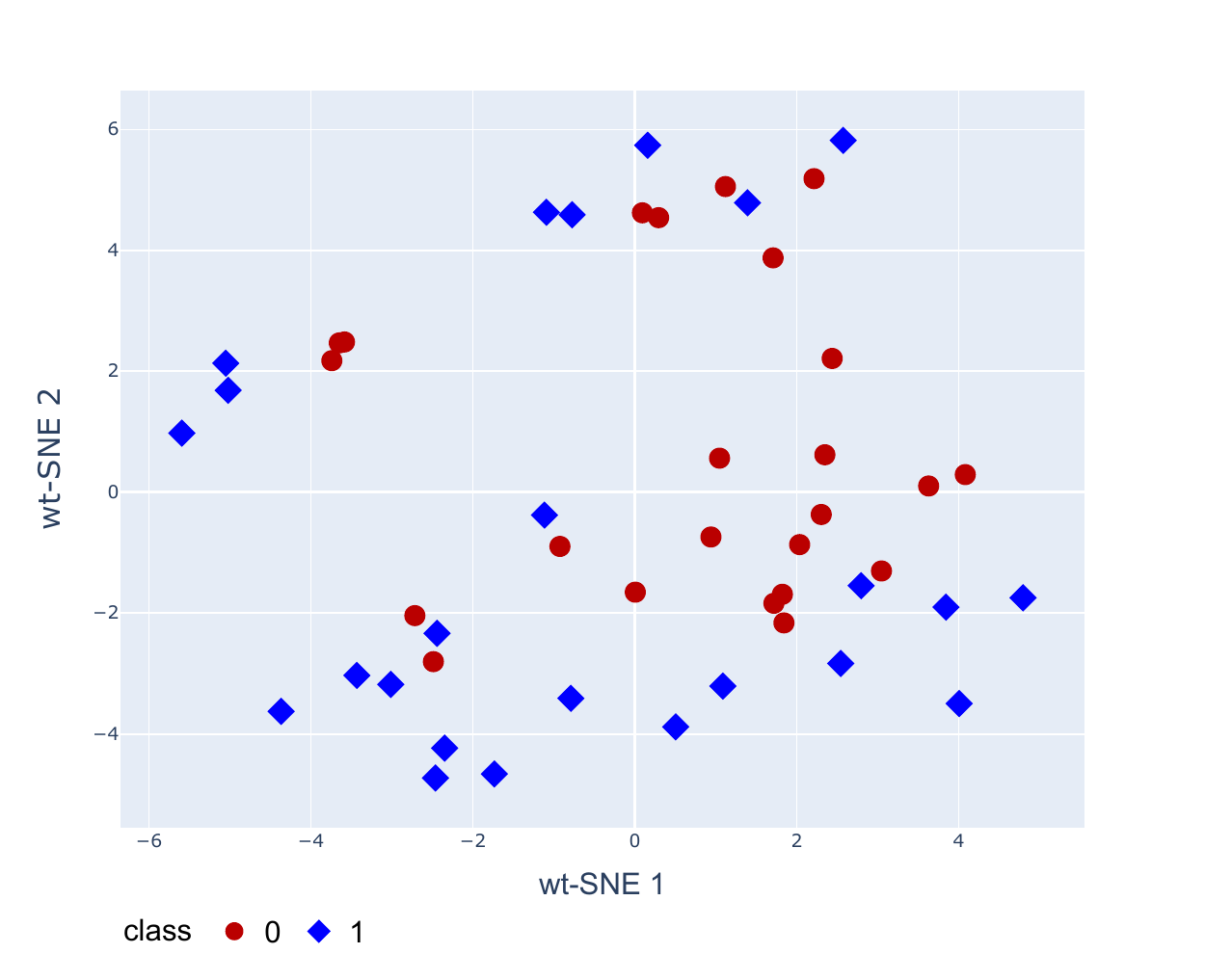}
         \caption{LIME}
         \label{subfig:comparative_wtsne_model_liver_lime}
     \end{subfigure}
     \hfill
     \begin{subfigure}[b]{0.32\textwidth}
         \centering
         \includegraphics[width=\textwidth]{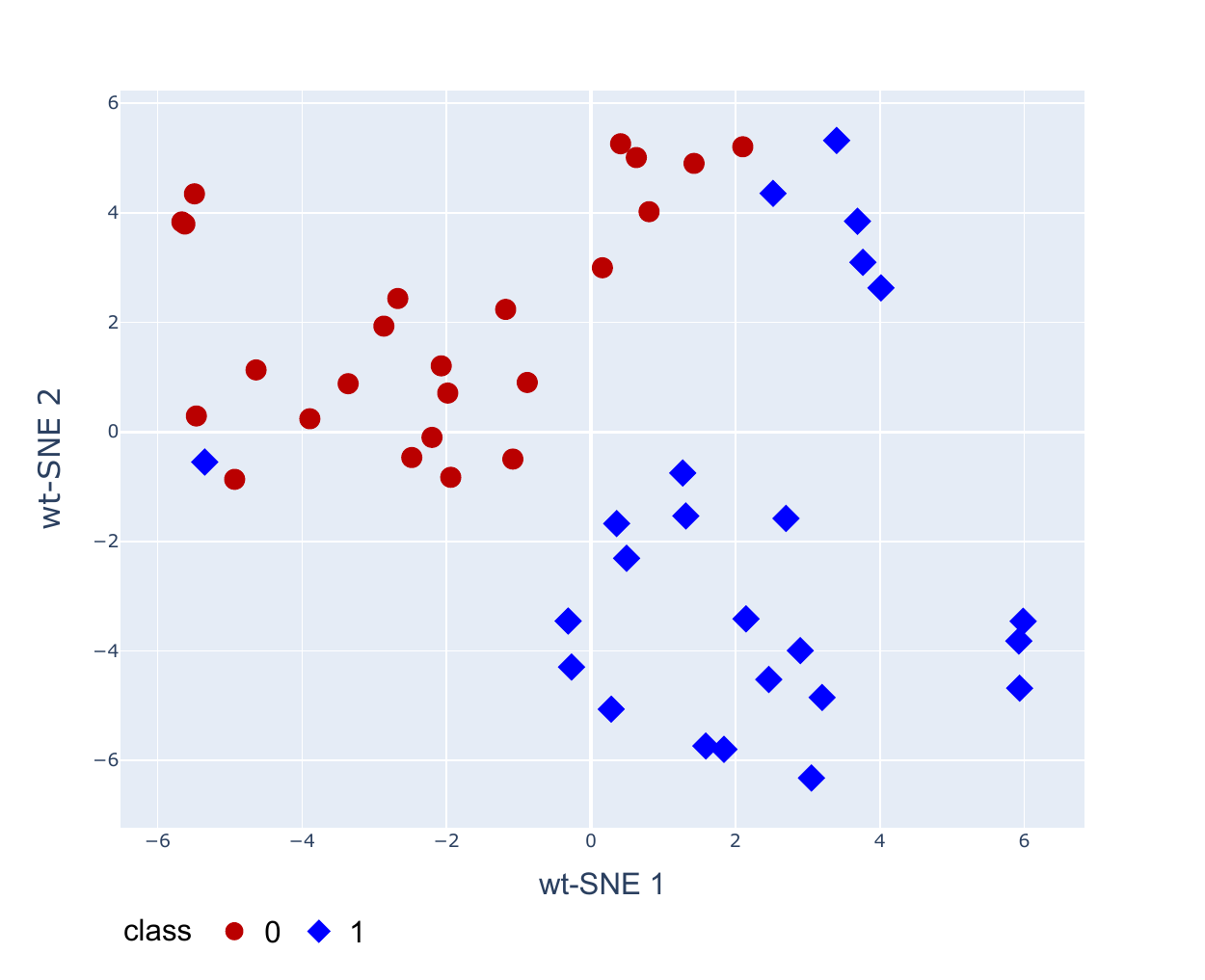}
         \caption{DeepSHAP}
         \label{subfig:comparative_wtsne_model_liver_deepshap}
     \end{subfigure}
     \hfill
     \begin{subfigure}[b]{0.32\textwidth}
         \centering
         \includegraphics[width=\textwidth]{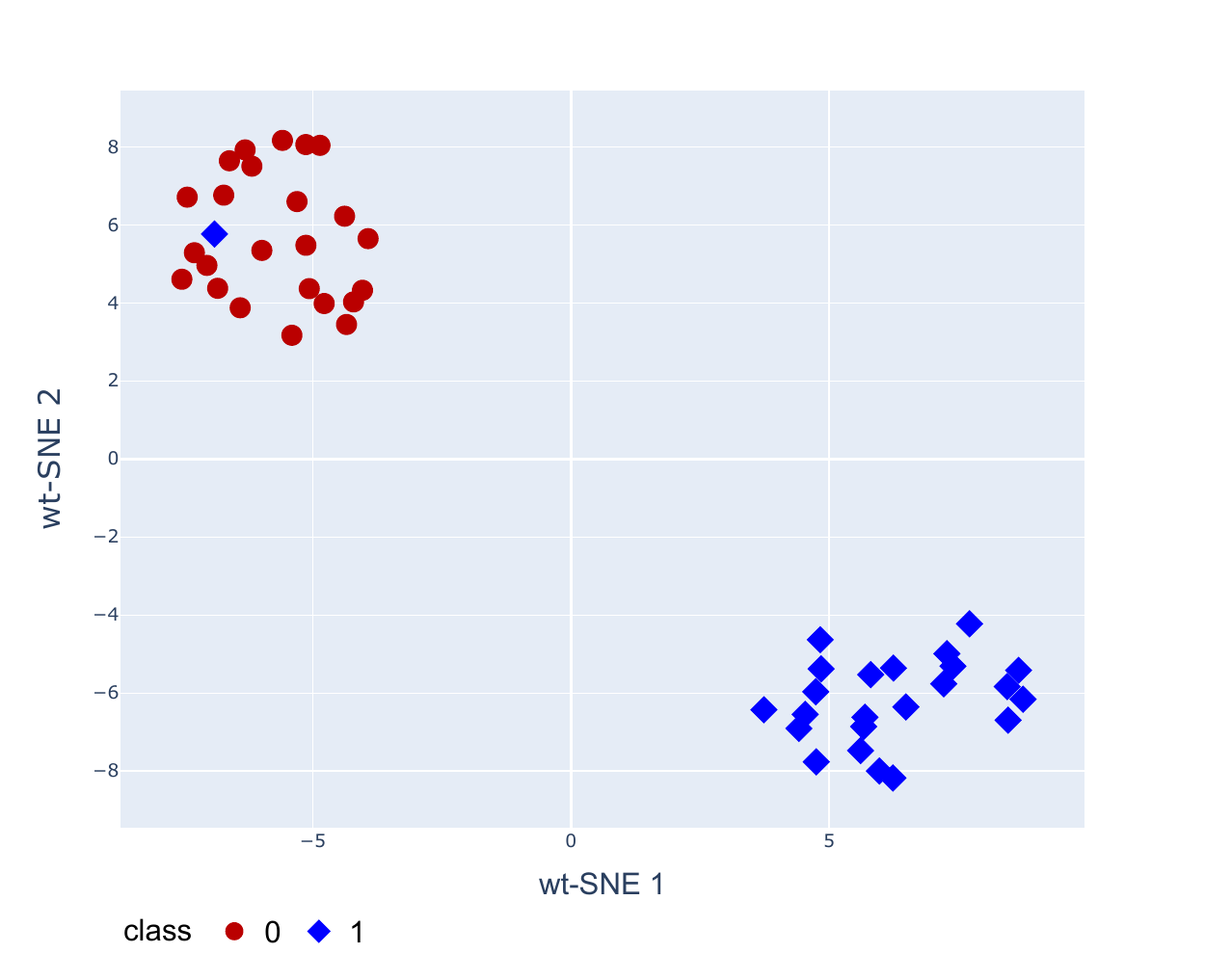}
         \caption{Multiple FSL}
         \label{subfig:comparative_wtsne_model_liver_multiple_fsl}
     \end{subfigure}
    \caption[Liver dataset weighted t-SNE visualization using general feature weights.
    The figure displays nine plots. 
    The first shows the result of applying plain t-SNE on the Liver dataset, while the remaining eight show the results of applying weighted t-SNE using the weights generated by Decision Tree, Random Forest, Kruskal Wallis, ReliefF, Lasso, LIME, DeepSHAP and Multiple FSL, respectively, being used the general feature weights.
    In the first plot both labels are mixed, without a clean separation between the data that belongs to label 0 and to label 1 in the 2D projection space.
    In the second plot, from Decision Tree, a better separation is created with only three points from label 1 being mixed with the label 0.
    In the third plot, from Random Forest, we can also see a huge improvements in the clustering. One data point from label 1 is mixed to label 0 cluster and a few others are closer to label 0 cluster than label 1.
    In the forth plot, from Kruskal Wallis, we can also see an improvement in the label clusters separation with a few mixed data points.
    In the fifth plot, from ReliefF we can see a slight better separation between the label data points, but worst than the other observed algorithms. 
    In the sixth plot, from Lasso, we can see a huge improvement where both clusters can be completely separated by drawing a simple vertical line, but there is a huger distance between each data point and each cluster centroid.
    In the seventh plot, from LIME, we can observe results similar to ReliefF, where the improvement is close to none.
    In the eighth plot, from DeepSHAP, an improvement can be noticed, similar to what was observed with Kruskal Wallis. Some data points stills mixed, not being able to easily define a simple border between each label's cluster.
    In the ninth plot, from Multiple FSL, two cluster were clearly created, one at the left top side of the image and another at the right bottom. As Lasso is easy to draw a simple line that separates both clusters, but a single data point from label 1 is present inside the label 0 cluster.
    This can indicate a mislabeled samples.
    Compared with the other algorithms that were able to clearly separate the data points from each label in the 2D space on Multiple FSL those data points are near of each other inside each cluster and far from each other between the different clusters.]
    {\textbf{Liver dataset weighted t-SNE visualization using general feature weights.} 
        \autoref{subfig:comparative_wtsne_model_liver_without_weights} presents the t-SNE visualization.
        The other figures presents the weighted t-SNE using the general feature weights extracted from each algorithm.
        The two classes are displayed in red circles (0) and blue diamonds (1). 
    }
    \label{fig:comparative_wtsne_on_liver}
\end{figure*}

\autoref{fig:comparative_wtsne_on_liver} shows the best weighted t-SNE result obtained on Liver dataset between 11 executions.
Multiple FSL and Lasso were able to achieve better separation between the samples by their label.
Decision Tree, Random Forest, Kruskal Wallis, and DeepSHAP also had good results.
\autoref{tab:weighted_tsne_on_comparsion_for_liver} presents the average HD and 2D silhouettes after applying the feature weights compared with the silhouettes without them (``None''). 
As observed in the visual representations, Multiple FSL and Lasso had superior performance in 2D silhouette.

\begin{table}[h]
\centering
\caption[Average HD and 2D silhouette coefficients for Liver dataset. 
For HD silhouette, Decision Tree had the highest average, and Decision Tree and Lasso had the highest result with statistical significance.
For 2D silhouette, Multiple FSL had the highest average, and Lasso and Multiple FSL had the highest result with statistical significance.] 
{\textbf{Average HD and 2D silhouette coefficients for Liver dataset.}
All feature weights were provided from the general feature weights, not being label-specific.
Averages were calculated with the weighted t-SNE results given the feature weights generated by each algorithm through 11 different executions.
In bold is the highest average for each column.
Best results with statistical significance ($p < 0.01$) according to the Kruskal-Wallis and Dunn’s test are highlighted in orange.
}
\label{tab:weighted_tsne_on_comparsion_for_liver}
\begin{tabular}{lrr}
\hline
    Weight source               & HD silhouette                    & 2D silhouette        \\ \hline
    None                        & .038                           & .049       \\
    Decision Tree               & \cellcolor{orange!20} \textbf{.409$\pm$.073}                & .498$\pm$.056     \\
    Random Forest               & .220$\pm$.019                & .517$\pm$.047     \\
    Kruskal Wallis                     & .148$\pm$.013                & .305$\pm$.061     \\
    ReliefF                     & .054$\pm$.003                & .066$\pm$.016     \\
    Lasso                       & \cellcolor{orange!20} .382$\pm$.038                & \cellcolor{orange!20} .661$\pm$.033     \\
    LIME                        & .039$\pm$.001                & .051$\pm$.010     \\
    DeepSHAP                    & .130$\pm$.007                & .232$\pm$.057     \\
    Multiple FSL                & .302$\pm$.046                & \cellcolor{orange!20} \textbf{.668$\pm$.061}     \\
\hline
\end{tabular}
\end{table}

\begin{figure*}[h]
    \centering
    \begin{subfigure}[b]{0.32\textwidth}
         \centering
         \includegraphics[width=\textwidth]{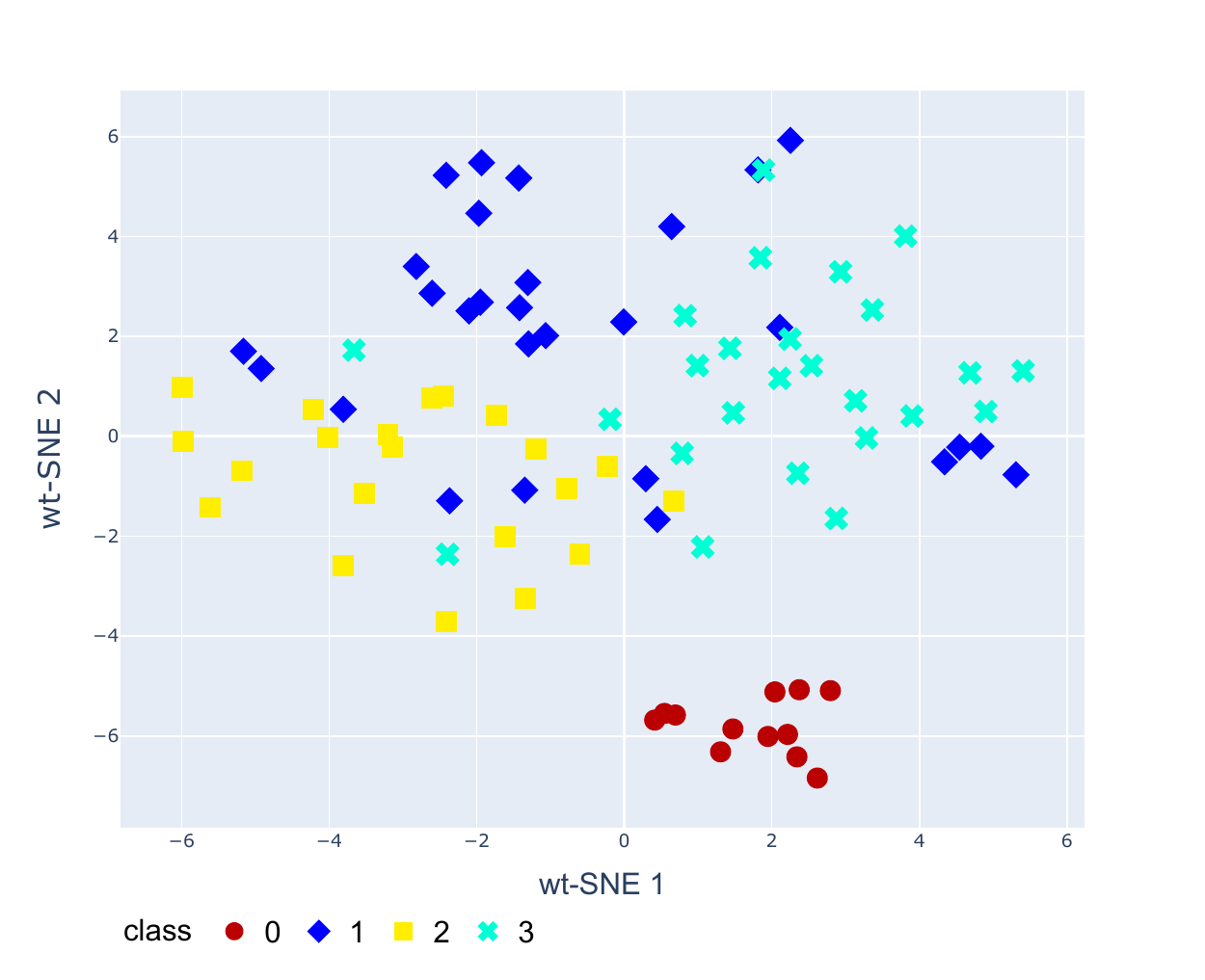}
         \caption{Without weights}
         \label{subfig:comparative_wtsne_model_colorectal_without_weights}
     \end{subfigure}
     \hfill
     \begin{subfigure}[b]{0.32\textwidth}
         \centering
         \includegraphics[width=\textwidth]{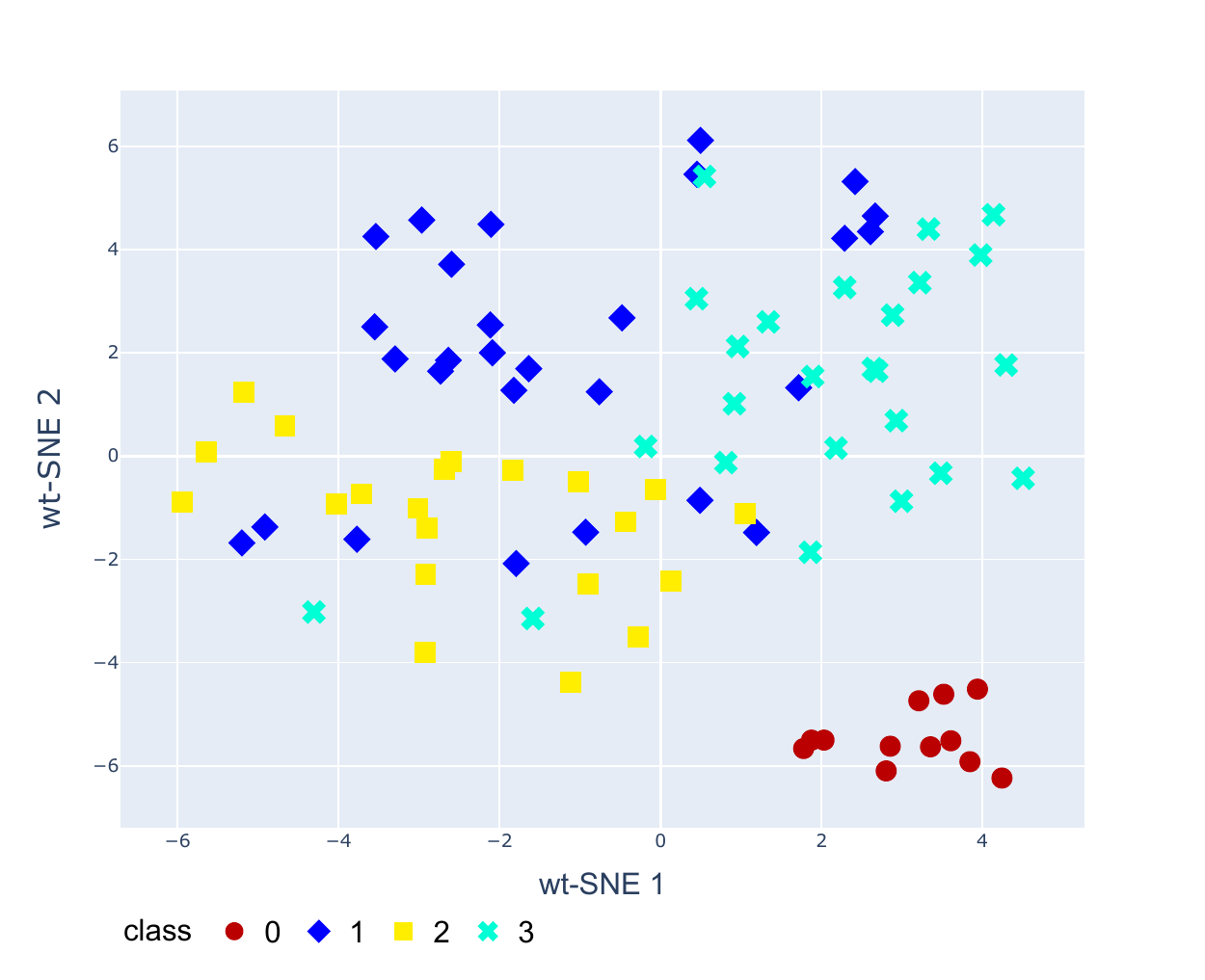}
         \caption{Lasso}
         \label{subfig:comparative_wtsne_model_colorectal_lasso_label_1}
     \end{subfigure}
     \hfill
     \begin{subfigure}[b]{0.32\textwidth}
         \centering
         \includegraphics[width=\textwidth]{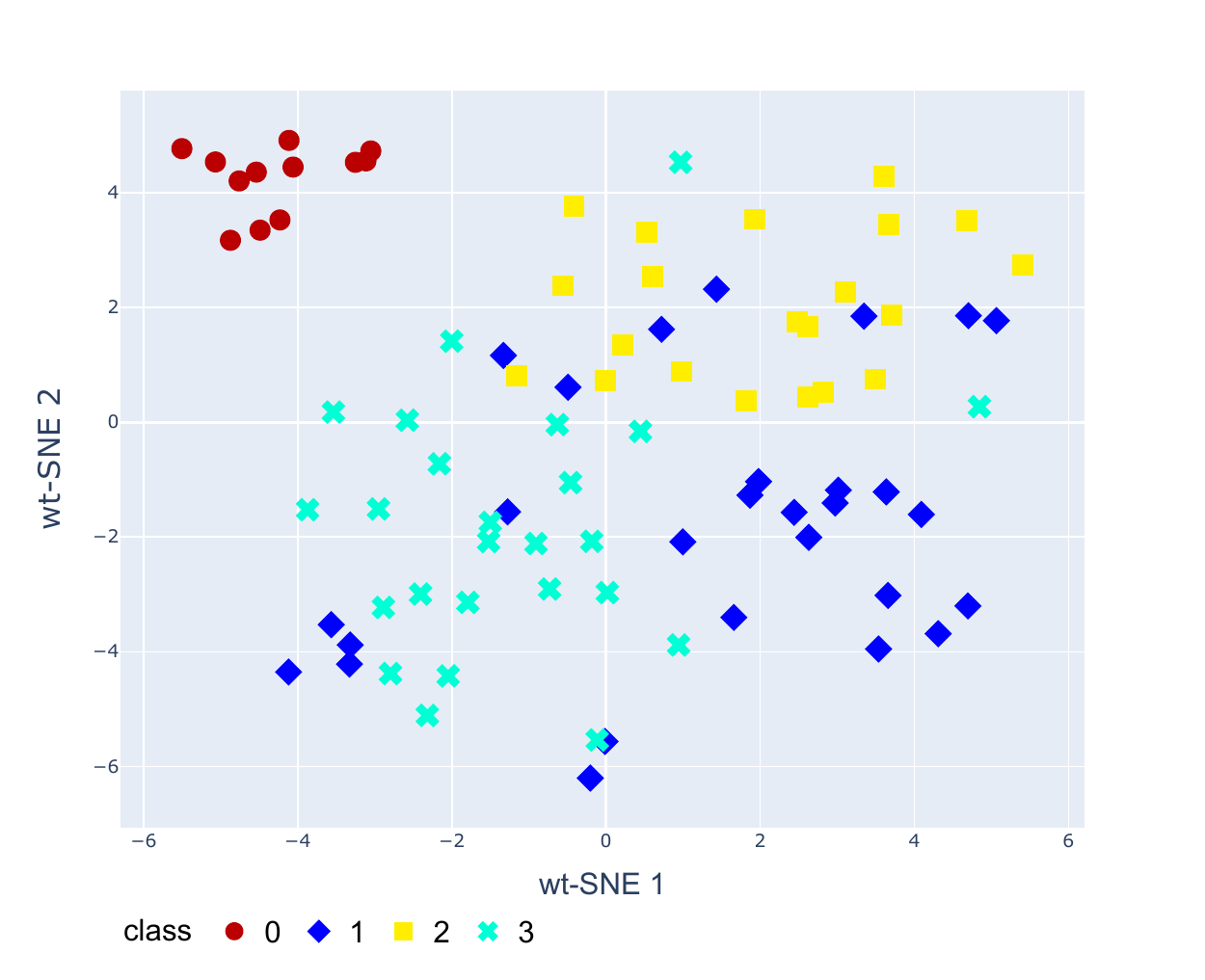}
         \caption{LIME}
         \label{subfig:comparative_wtsne_model_colorectal_lime_label_1}
     \end{subfigure}
     \begin{subfigure}[b]{0.32\textwidth}
         \centering
         \includegraphics[width=\textwidth]{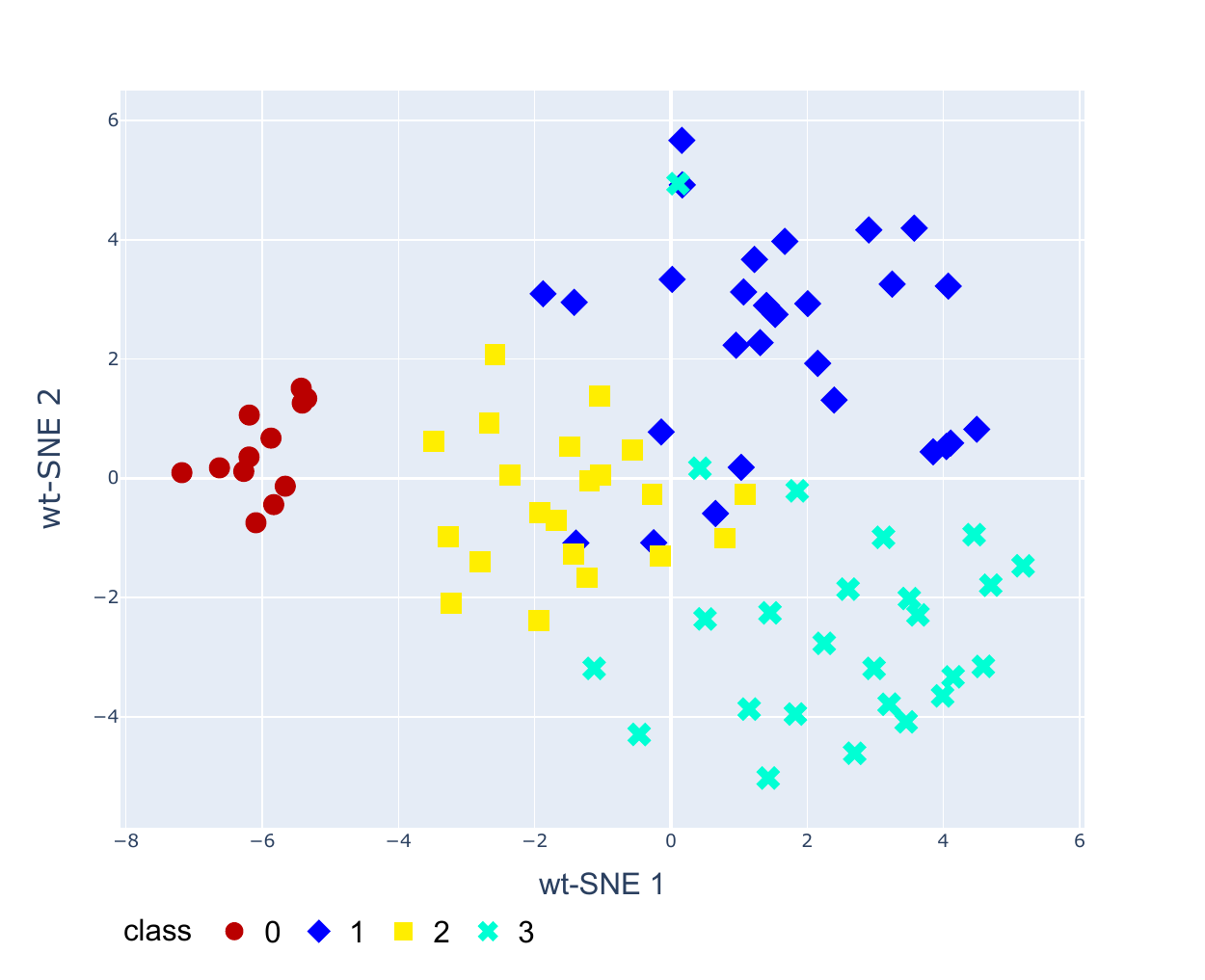}
         \caption{DeepSHAP}
         \label{subfig:comparative_wtsne_model_colorectal_random_deepshap_label_1}
     \end{subfigure}
     \begin{subfigure}[b]{0.32\textwidth}
         \centering
         \includegraphics[width=\textwidth]{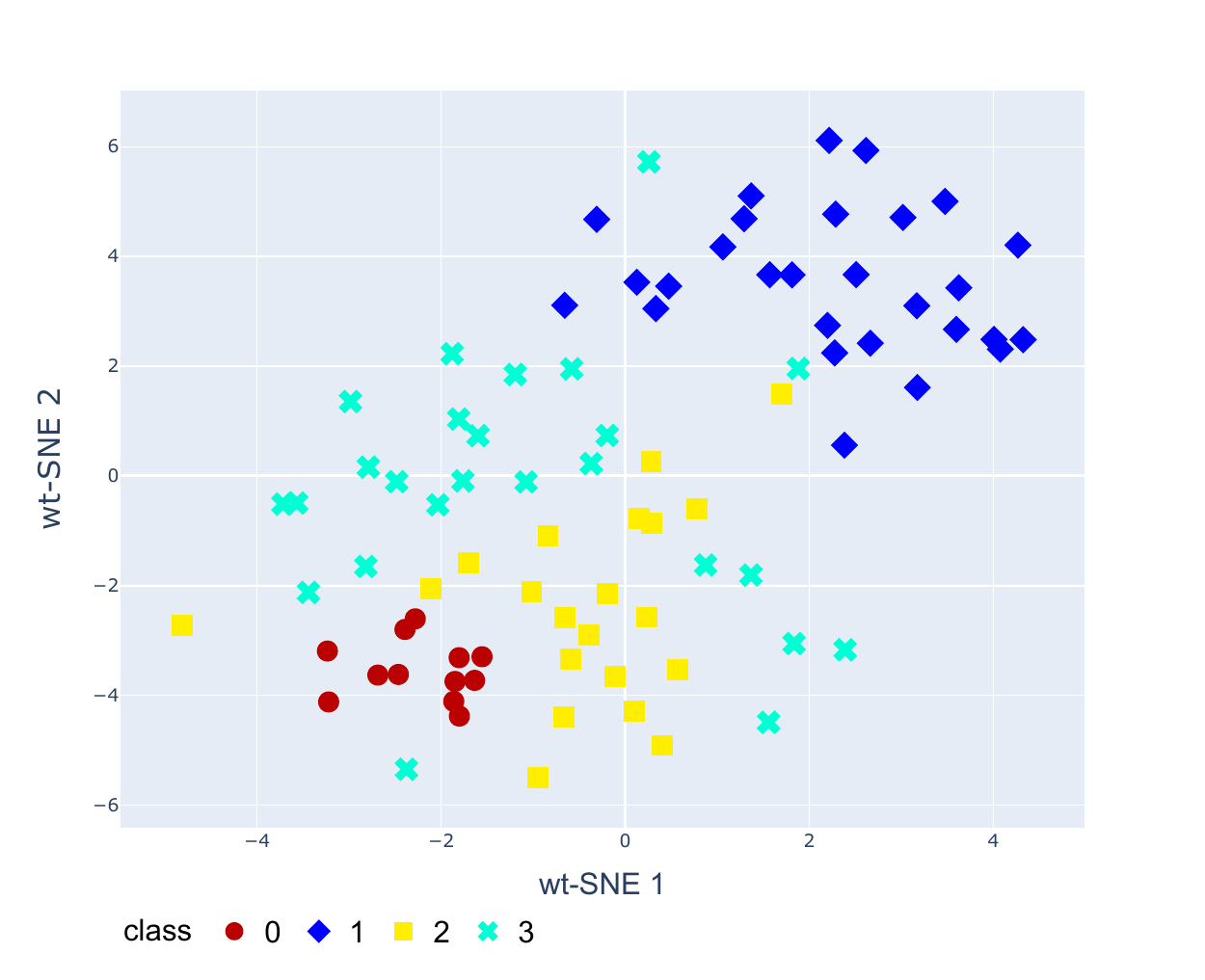}
         \caption{Multiple FSL}
         \label{subfig:comparative_wtsne_model_colorectal_multiple_fsl_label_1}
     \end{subfigure}
    \caption[Colorectal dataset weighted t-SNE visualization using feature weights related to label 1.
    The figure displays five plots. 
    The first shows the result of applying plain t-SNE on the Colorectal dataset, while the remaining four show the results of applying weighted t-SNE using the weights generated by Lasso, LIME, DeepSHAP and Multiple FSL, respectively, being used the label 1 specific feature weights.
    In the first plot all labels are mixed, without a clean separation between the data that belongs to label 1, label 2 and label 3 in the 2D projection space. 
    The only label that can be easily separated from all others is the label 0.
    In the second plot, from Lasso, there is no improvement. 
    The data points are in different positions, but the overall view was not changed when compared to the plain t-SNE.
    In the third plot, from LIME, we have similar results, no real improvement can be visually noticed.
    The main difference is that the points positions appear to be inverted along both axes.
    In the fourth plot, from DeepSHAP, an overall improvement can be noticed. 
    Label 0 data points stills well separated from the other and even though they remain close, there is a noticeable improvement in the separation of data points for labels 1, 2, and 3.
    Most of the label 3 data points were dislocated to the bottom, label 1 to the top and label 2 to the middle.
    In the fifth plot, from Multiple FSL, a huge improvement can be noticed to the label 1 data points, were all of them were dislocated to the top right of the plot with just a few label 3 and 2 data points near to the label 1 ones.
    Another change noticed on the fifth plot is on the label 0 data points.
    In contrast to the other results, in Multiple FSL the data points for label 0 showed worse separation, becoming mixed with those of labels 2 and 3.]
    {\textbf{Colorectal dataset weighted t-SNE visualization using feature weights related to label 1.} 
        \autoref{subfig:comparative_wtsne_model_colorectal_without_weights} presents a t-SNE visualization without feature weights.
        Label 1 is represented in blue diamonds.
    }
    \label{fig:comparative_wtsne_colorectal}
\end{figure*}

\autoref{fig:comparative_wtsne_colorectal} presents the weighted t-SNE results from DeepSHAP, LIME, Lasso and Multiple FSL on Colorectal dataset using label 1 (blue diamonds) specific weights.
Only Multiple FSL demonstrated a significant improvement specifically related to label 1 samples.
The findings presented in \autoref{tab:hd_and_2d_silhouette_comparison_colorectal} corroborate our visual observations with HD and 2D silhouette coefficients resulted from an one-versus-all (OVA) approach centered on label 1. 
Multiple FSL demonstrated the highest performance across all scenarios while DeepSHAP presented a slight improvement.

\begin{figure*}[h]
    \centering
     \begin{subfigure}[b]{0.45\textwidth}
         \centering
         \includegraphics[width=\textwidth]{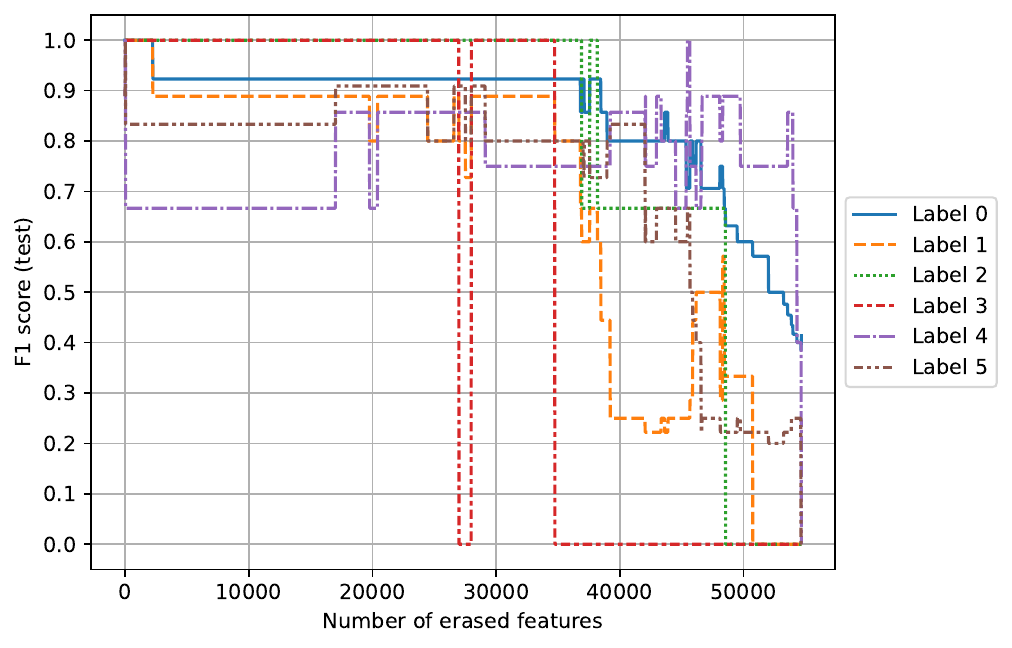}
         \caption{Lasso}
         \label{subfig:comparative_feature_erasure_breast_label_4_lasso}
     \end{subfigure}
     \hfill
     \begin{subfigure}[b]{0.45\textwidth}
         \centering
         \includegraphics[width=\textwidth]{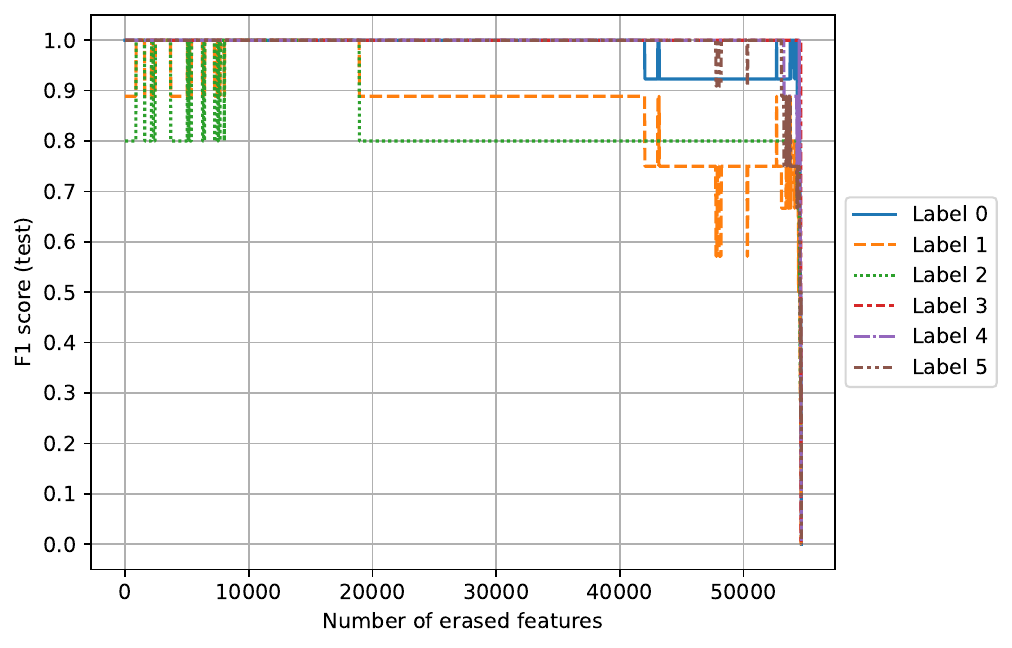}
         \caption{LIME}
         \label{subfig:comparative_feature_erasure_breast_label_4_lime}
     \end{subfigure}
     \hfill
     \begin{subfigure}[b]{0.45\textwidth}
         \centering
         \includegraphics[width=\textwidth]{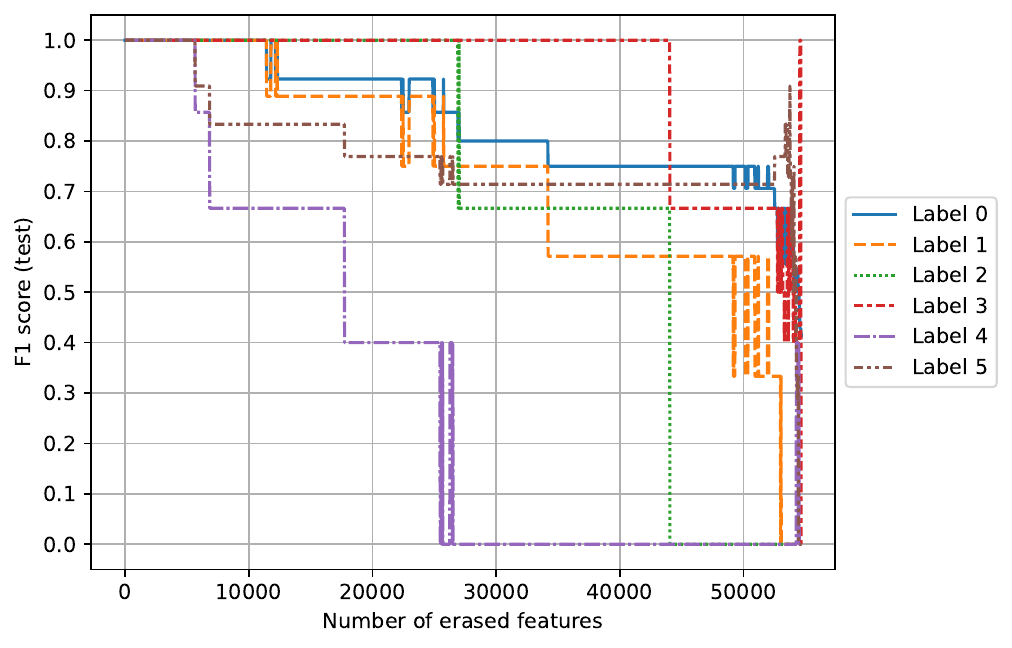}
         \caption{DeepSHAP}
        \label{subfig:comparative_feature_erasure_breast_label_4_deepshap}
     \end{subfigure}
     \hfill
     \begin{subfigure}[b]{0.45\textwidth}
         \centering
         \includegraphics[width=\textwidth]{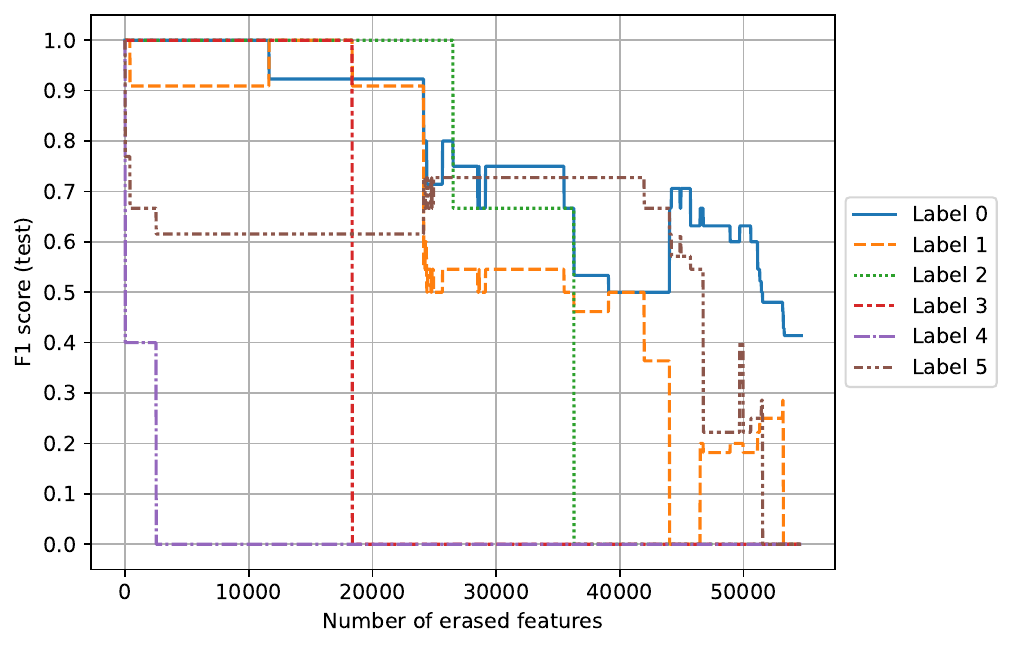}
         \caption{Multiple FSL}
        \label{subfig:comparative_feature_erasure_breast_label_4_multiple_fsl}
     \end{subfigure}
     \caption[
        The image contains four graphs.
        General Layout and Axes:
        Horizontal Axis (X-axis): Displays the "Number of erased features," ranging from 0 to approximately 60,000. Features were erased sequentially based on how highly each algorithm ranked them for Label 4.
        Vertical Axis (Y-axis): Displays the "F1 score (test)," ranging from 0.0 at the bottom to 1.0 at the top.
        Each graph contains six distinct lines representing the F1 scores for Labels 0 through 5. 
        The focus of the analysis is on Label 4, represented by a purple dash-dotted line.
        Individual Graph Summaries:
        (a) Lasso: The F1 score for Label 4 experiences a substantial and immediate decline early in the feature erasure process, indicating the algorithm correctly identified the most critical features for this label. The F1 scores drops below 0.7.
        (b) LIME: The F1 score for Label 4 remains high (near 1.0) even as tens of thousands of features are removed, only dropping at the very end. This visualizes that LIME, likely constrained by the dataset's limited sample size, failed to accurately identify and rank the features specifically relevant to Label 4.
        (c) DeepSHAP: The graph shows a significant drop in the F1 score for Label 4, demonstrating sensitivity to feature erasure similar to Lasso and Multiple FSL. However, the line drops steeply further along the x-axis (between 10,000 and 20,000 features), showing it required a larger amount of erased features to degrade performance.
        (d) Multiple FSL: Demonstrates the most dramatic early sensitivity. The F1 score for Label 4 plummets almost vertically to near 0.0 at the very beginning of the x-axis, showing a severe decline in performance when only a very small number of its top-ranked features are removed.
    ]
    {\textbf{Impact of feature erasure based on label 4 feature weights for Breast dataset.}
        The features were erased from the dataset following a ranking based on the feature weights given for label 4 by each algorithm.
        The horizontal axis shows the number of erased features, while the vertical axis displays the F1 score calculated from the test dataset.
    }
    \label{fig:comparative_feature_erasure_breast_label_4}
\end{figure*}

\begin{figure*}[h]
    \centering
    \begin{subfigure}[b]{0.32\textwidth}
         \centering
         \includegraphics[width=\textwidth]{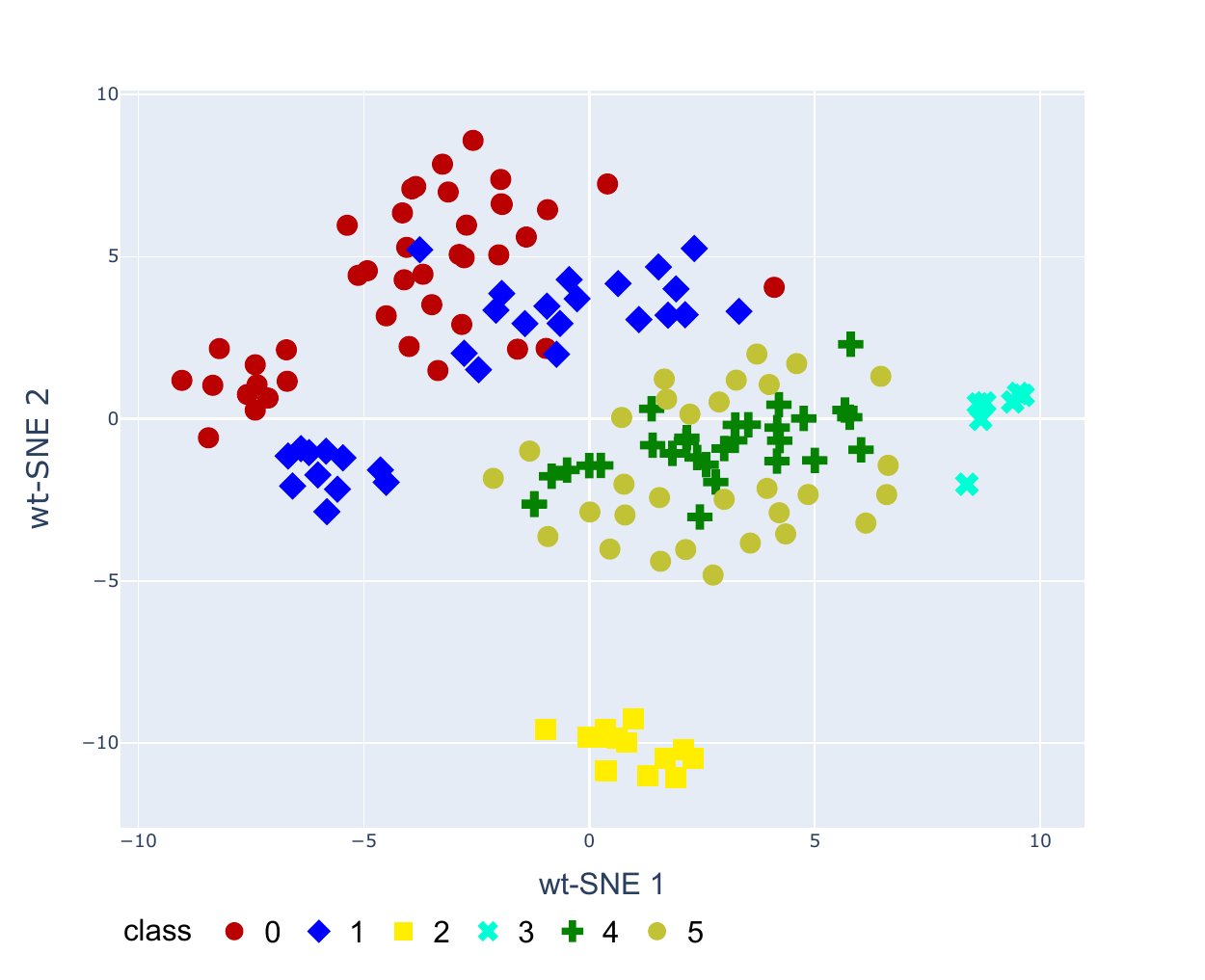}
         \caption{Without weights}
         
         \label{subfig:comparative_wtsne_model_breast_without_weights_mfs}
     \end{subfigure}
     \hfill
     \begin{subfigure}[b]{0.32\textwidth}
         \centering
         \includegraphics[width=\textwidth]{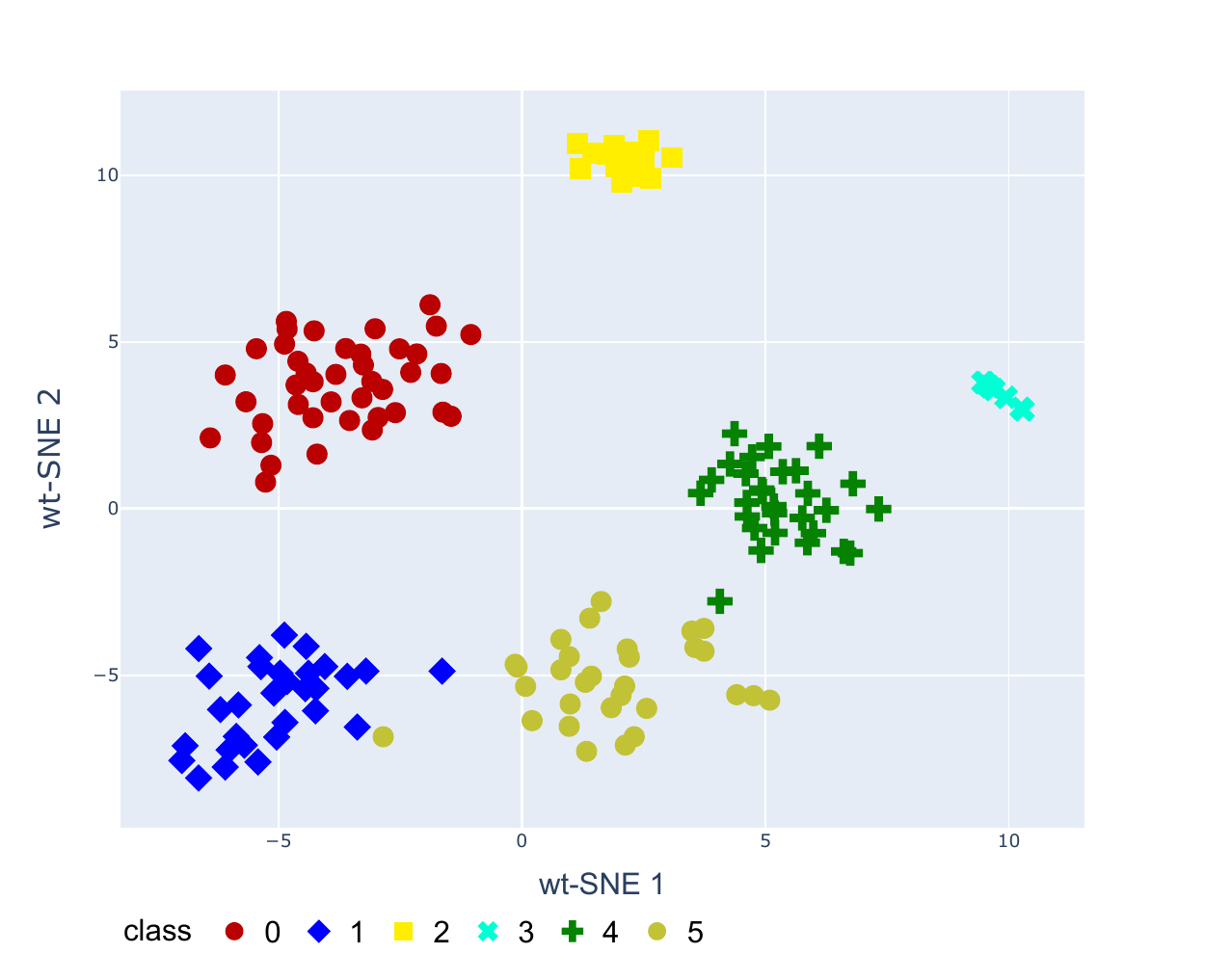}
         \caption{General weights}
         \label{subfig:comparative_wtsne_model_breast_general}
     \end{subfigure}
     \hfill
     \begin{subfigure}[b]{0.32\textwidth}
         \centering
         \includegraphics[width=\textwidth]{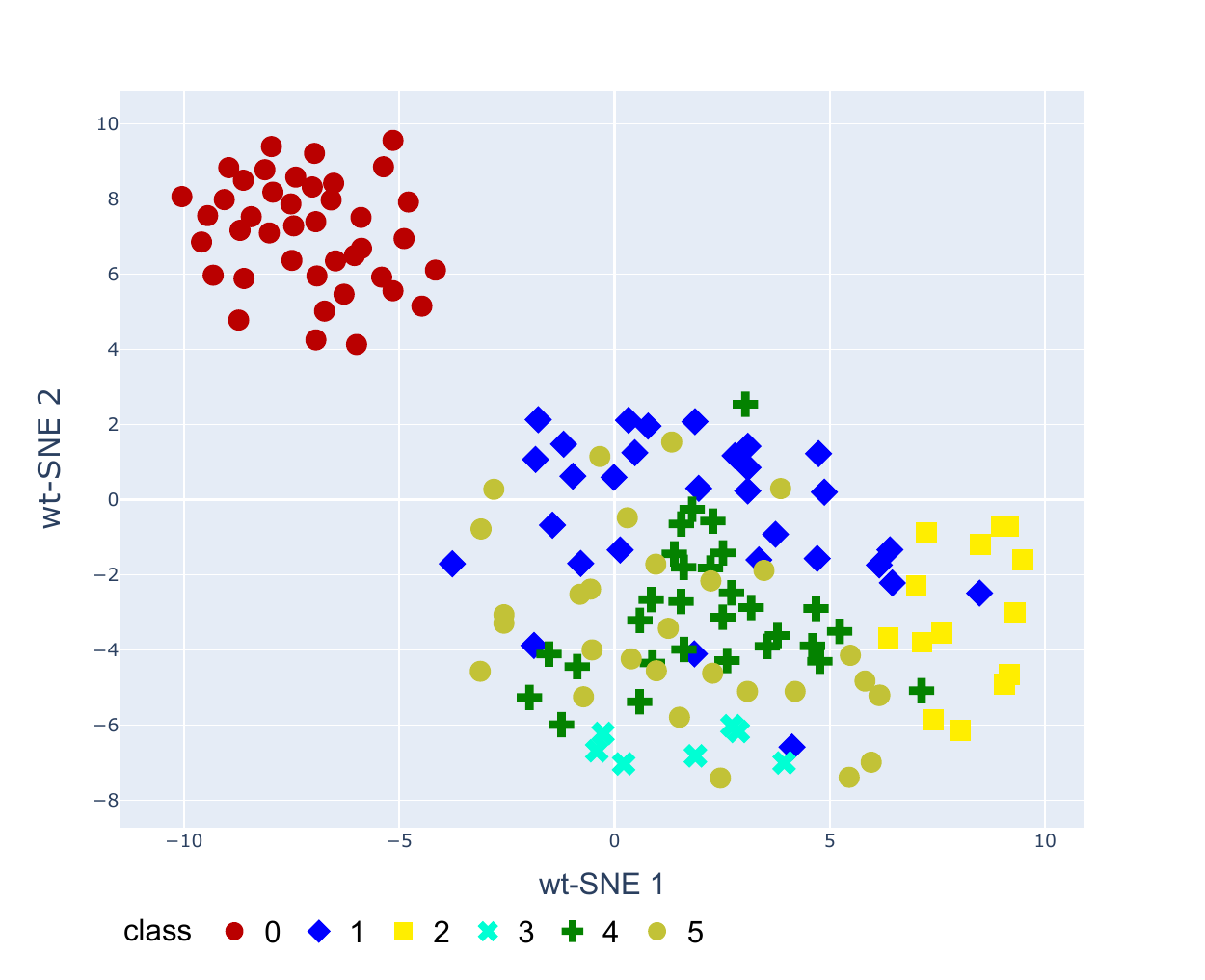}
         \caption{Label 0 (red)}
         \label{subfig:comparative_wtsne_model_breast_label_0}
     \end{subfigure}
     \begin{subfigure}[b]{0.32\textwidth}
         \centering
         \includegraphics[width=\textwidth]{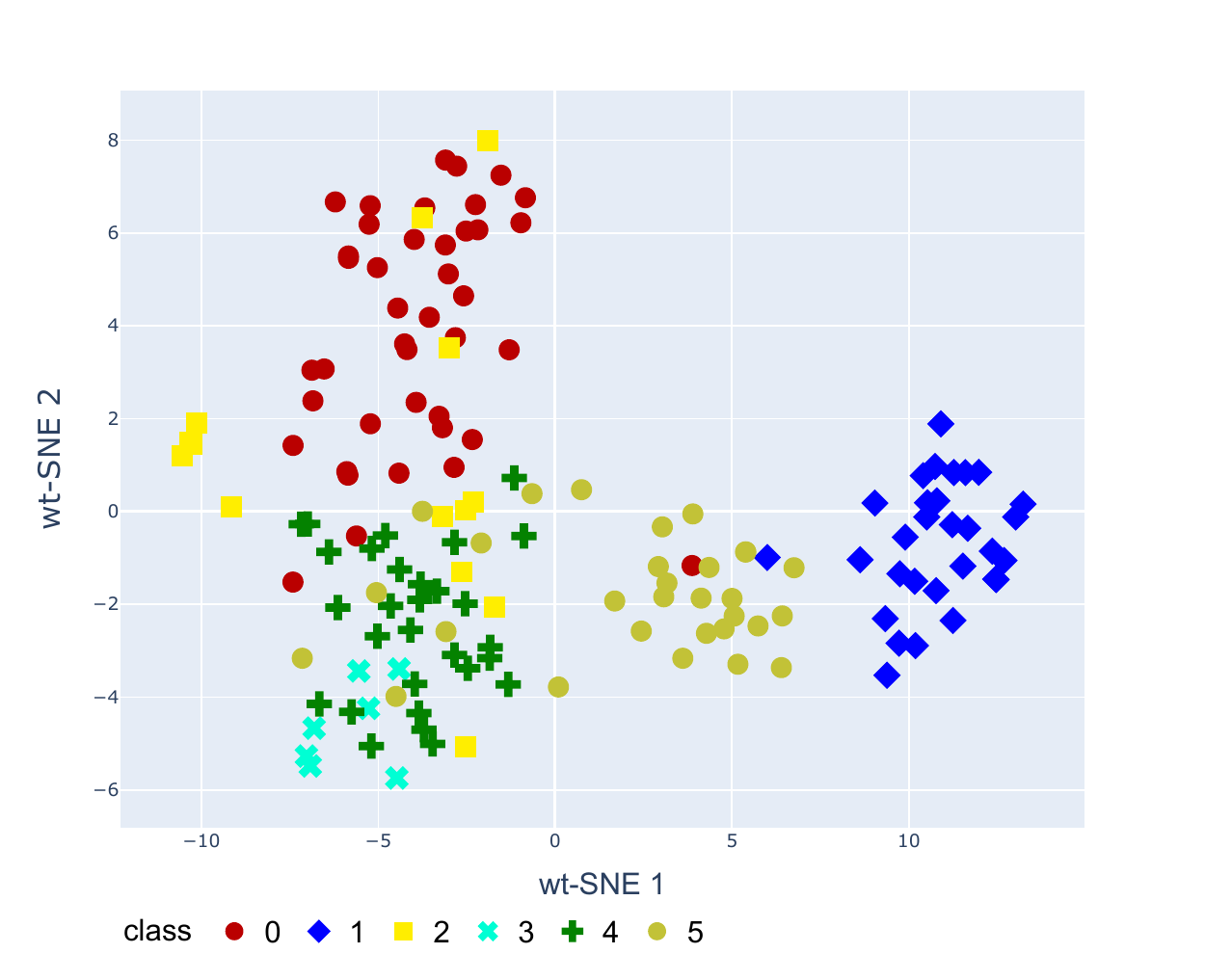}
         \caption{Label 1 (blue)}
         \label{subfig:comparative_wtsne_model_breast_label_1}
     \end{subfigure}
     \hfill
     \begin{subfigure}[b]{0.32\textwidth}
         \centering
         \includegraphics[width=\textwidth]{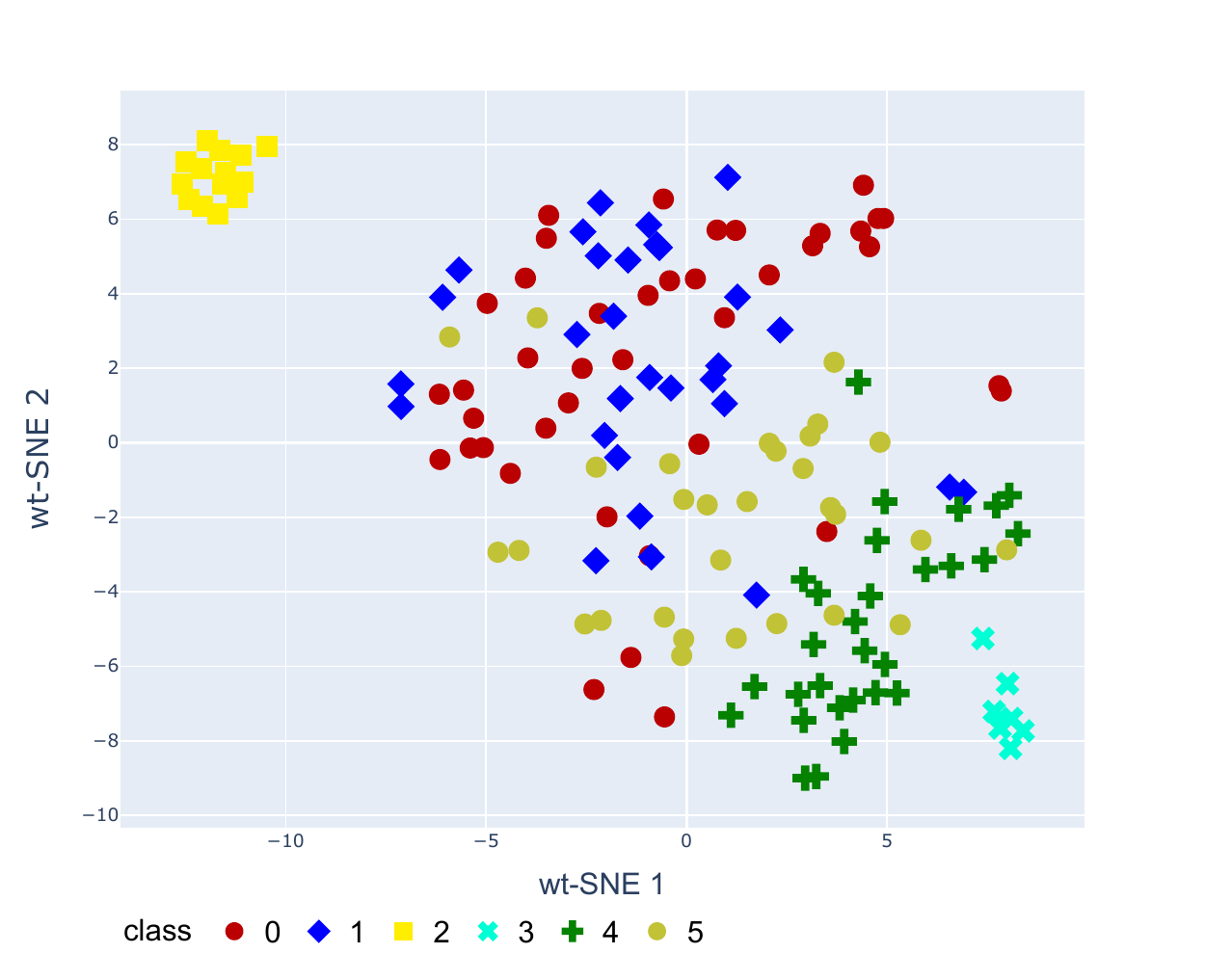}
         \caption{Label 2 (yellow)}
         \label{subfig:comparative_wtsne_model_breast_label_2}
     \end{subfigure}
     \hfill
     \begin{subfigure}[b]{0.32\textwidth}
         \centering
         \includegraphics[width=\textwidth]{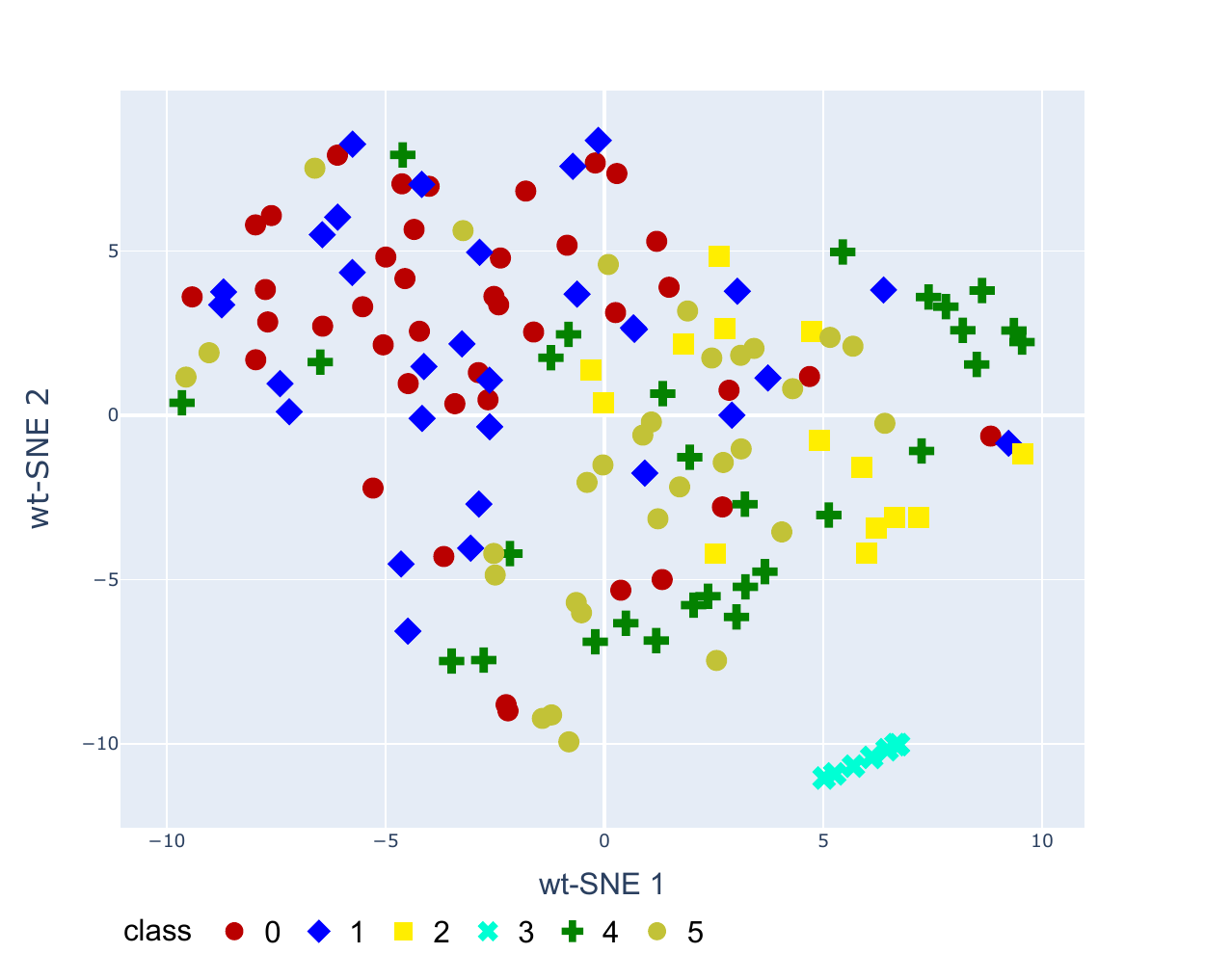}
         \caption{Label 3 (cyan)}
         \label{subfig:comparative_wtsne_model_breast_label_3}
     \end{subfigure}
     \begin{subfigure}[b]{0.32\textwidth}
         \centering
         \includegraphics[width=\textwidth]{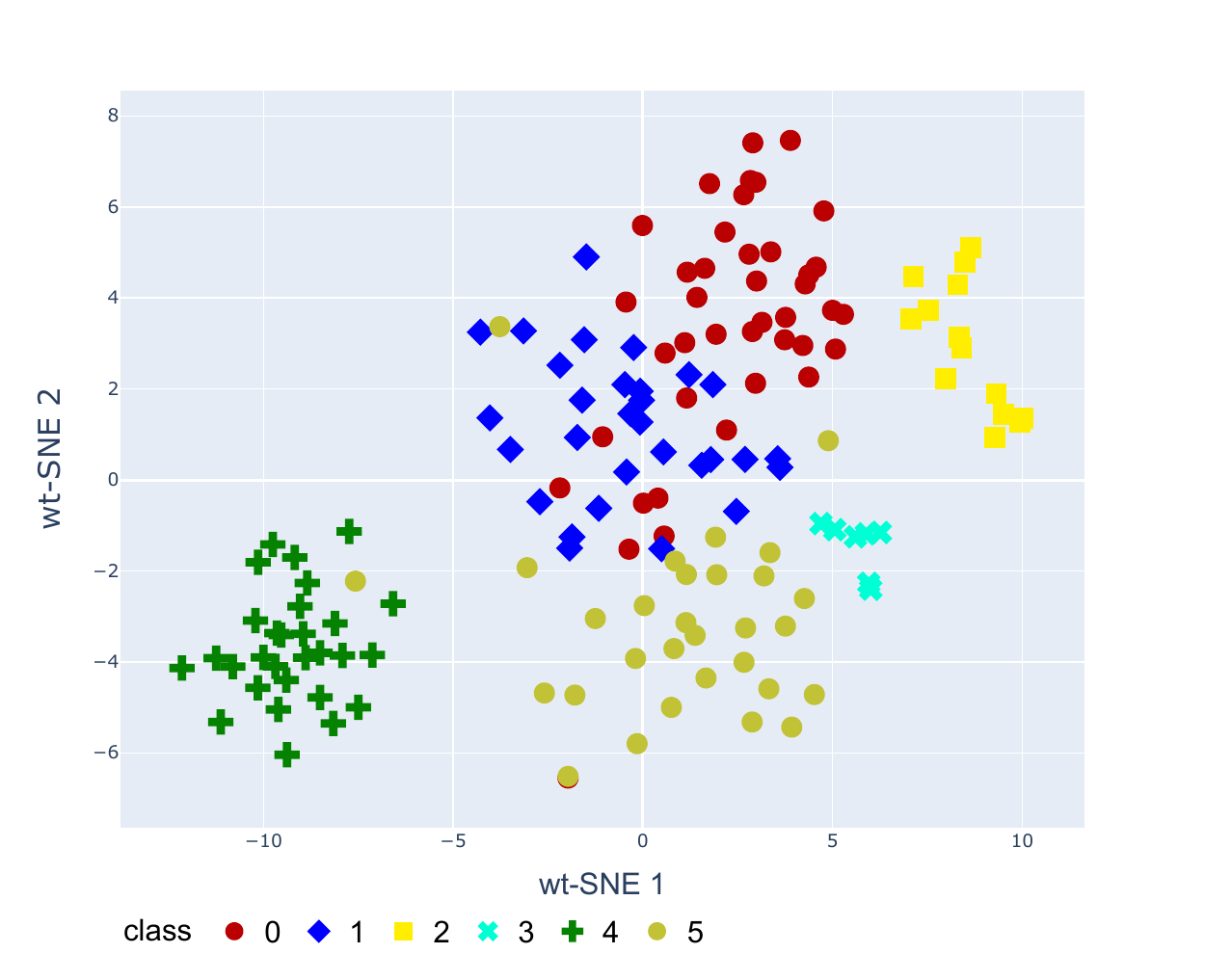}
         \caption{Label 4 (dark green)}
         \label{subfig:comparative_wtsne_model_breast_label_4}
     \end{subfigure}
     \begin{subfigure}[b]{0.32\textwidth}
         \centering
         \includegraphics[width=\textwidth]{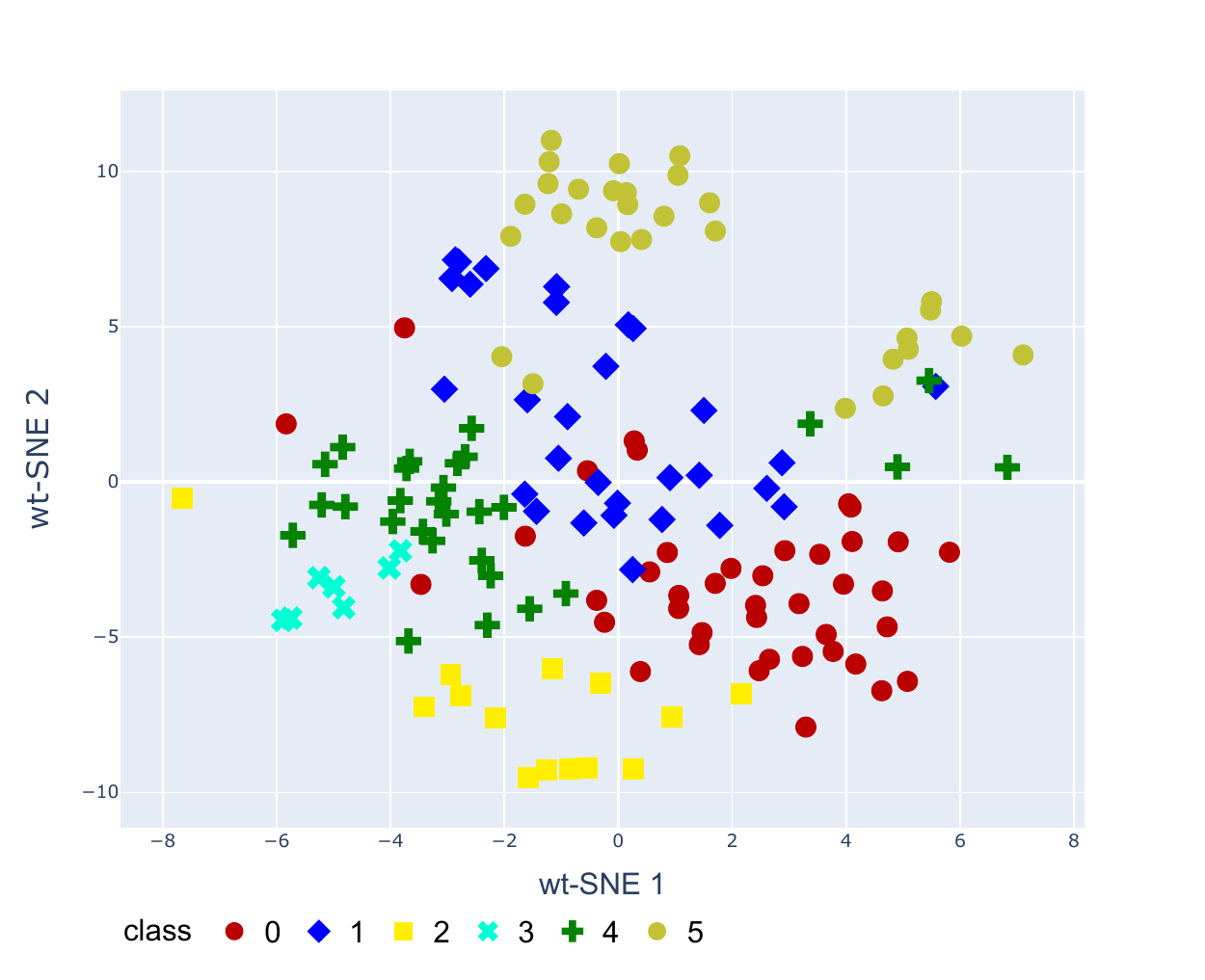}
         \caption{Label 5 (light green)}
         \label{subfig:comparative_wtsne_model_breast_label_5}
     \end{subfigure}
    \caption[Breast dataset weighted t-SNE visualization using label-specific feature weights from Multiple FSL.
    Eight charts are displayed.
    The first chart shows the result of applying plain t-SNE on the Breast dataset.
    While the data points belonging to labels 2 and 3 appear more clearly separated, those related to labels 0 and 1, as well as labels 4 and 5, appear overlapping in the 2D projection.
    The second chart, which uses the overall weights generated by Multiple FSL, reveals a clear separation between the clusters of all labels, with only small regions of proximity.
    In the third chart, which uses the weights generated specifically for label 0, the corresponding cluster stands out from the rest, while all other labels end up merging into a large secondary cluster.
    The fourth chart, using weights for label 1, shows a similar pattern: label 1 data points are separated in a single well defined cluster, while a second cluster is formed by data points from all other labels, with a slight separation observed for label 5 were one data point from label 1 got mixed with label 5 data points.
    In the fifth chart, using weights generated for label 2, a clear separation is observed between the label 2 points in the upper left corner and a second cluster that contains all other labels extending from the center to the bottom right corner of the chart.
    The sixth chart, using weights for label 3, shows the same pattern, with label 3 points separating toward the lower left corner of the chart. And all other labels spread on the rest of the chart.
    In the seventh chart, using weights for label 4, a similar pattern is observed. 
    However, one data point from label 5 is mixed into the cluster formed by label 4, which separates toward the lower right corner.
    In the eighth chart, using weights for label 5, a clear separation is observed between label 5 and the other labels. 
    However, this separation is less pronounced compared to the results for other labels using their respective weights, and label 5 points seem to split into two distinct clusters, being the first one better defined and the second one mixed with a few data points from other labels.]
    {\textbf{Breast dataset weighted t-SNE visualization using feature weights from Multiple FSL.} 
        \autoref{subfig:comparative_wtsne_model_breast_without_weights_mfs} presents the t-SNE visualization.
        \autoref{subfig:comparative_wtsne_model_breast_general} presents a weighted t-SNE visualization with general weights.
        \autoref{subfig:comparative_wtsne_model_breast_label_0} to \ref{subfig:comparative_wtsne_model_breast_label_5} presents the weighted t-SNE visualization with label-specific weights.
    }
    \label{fig:comparative_wtsne_model_breast}
\end{figure*}

\autoref{fig:comparative_feature_erasure_breast_label_4} presents the outcomes of feature erasure, conducted based on the feature weights assigned to label 4 from the Breast dataset by Lasso, LIME, DeepSHAP, and Multiple FSL. 
In this validation process, a rapid drop in predictive performance for a specific label indicates that an algorithm successfully identified the most critical label-specific features. 
As desired, Lasso and Multiple FSL exhibited a substantial decline in performance when only a small number of their top-ranked features for label 4 were removed, confirming their accuracy. 
DeepSHAP demonstrated a similar sensitivity, albeit requiring a larger number of removed features to achieve the same drop. 
LIME, likely constrained by the limited sample size of the Breast dataset, was unable to accurately identify label-specific features.
\autoref{fig:comparative_wtsne_model_breast} present the weighted t-SNE visualization using general and label-specific feature weights generated by Multiple FSL for each label available on Breast dataset.
Features 2 (yellow) and 3 (cyan) were already distinguishable in the original t-SNE visualization. 
Note that labels 2 (yellow) and 3 (cyan) are already well defined in the original t-SNE visualization, but on the weighted t-SNE we can notice a slight improvement for them and a higher decrease for the other labels, which means that features relevant to other labels were ``erased''.
However, a huge improvement can be noticed on all other labels, confirming Multiple FSL advantage on identifying label-specific feature weights.

\autoref{tab:comparative_pifs_for_synthB} compares Multiple FSL, LassoNet, TabNet, and SAGE using the SynthB dataset.
Regarding general feature weights, SAGE performed better on average, with values slightly better than Multiple FSL and LassoNet. 
However, Multiple FSL outperformed the other methods when considering label-specific feature weights.  
\autoref{fig:weighted_tsne_on_comparison_for_synthB} of \ref{sec:complementary_experiments} presents a weighted t-SNE visualization using the label 1 specific weights generated by the methods. 
This visualization demonstrates visually Multiple FSL's superior capability in defining label-specific feature weights.

\autoref{tab:all_execution_time} presents the execution times (in seconds) for the evaluated feature weighting methods across five distinct datasets. 
The computational efficiency of the algorithms can be broadly categorized into traditional baseline models and more advanced neural or interpretation-based frameworks.
As expected, traditional machine learning and statistical methods (Decision Tree, Random Forest, Kruskal-Wallis, ReliefF, and Lasso) achieve the lowest execution times, frequently running in under two seconds. 
These methods consistently achieve statistically significant speed advantages due to their lower algorithmic complexity and the absence of iterative deep learning training cycles. However, when focusing on deep learning-based feature selection and model explanation methods, the proposed Multiple FSL approach exhibits strong scalability and highly competitive computational efficiency. 
Specifically, it introduces a small computational overhead compared to the Base Neural Network and significantly outperforms complex interpretation and embedded deep learning methods like LassoNet, TabNet, LIME, and DeepSHAP. 
Furthermore, Multiple FSL demonstrates notable robustness on high-dimensional data. 
For instance, on the Breast dataset, the SAGE method resulted in an out of memory error. 
In our experiments, we utilized SAGE's default Kernel estimator, which is highly sample-efficient but tracks a full $(M \times M)$ covariance matrix for its confidence intervals, leading to an $\mathcal{O}(M^2)$ memory complexity (where $M$ is the number of features). 
For the Breast dataset (54,675 features), this accumulator requires tens of gigabytes of RAM, quickly exceeding the system's capacity. 
While SAGE offers an alternative Permutation estimator with an $\mathcal{O}(M)$ memory footprint, it computes marginal contributions by adding features sequentially, requiring roughly $M$ model calls per permutation. 
Without the Kernel's variance-reduction techniques, it needs more permutations to converge, rendering it computationally prohibitive in terms of execution time for this dataset. 
In contrast, Multiple FSL handles the data efficiently in $39.89 \pm 5.54$ seconds.
This execution time vastly outperforms other complex alternatives like LIME ($487.27\pm21.86$ s) and LassoNet ($386.92\pm48.85$ s). 
Crucially, the transition from the standard FSL framework to the proposed Multiple FSL achieves these advantages without incurring a notable performance penalty.

Peak GPU memory usage (VRAM) per feature weighting method across the five datasets is reported in \autoref{tab:all_memory_usage}.
Although the Base Neural Network is not itself a feature weighting method, it is included for comparison: DeepSHAP directly reuses this same trained network as its underlying classifier, while FSL, Multiple FSL, and LassoNet each match its hidden-layer architecture but train an independent instance as part of their own optimization.
LassoNet, in contrast, relies on a dedicated architecture unrelated to the base model.
On XOR, SynthB, Isolet, and Letter Recognition, memory consumption remains modest and comparable across methods, ranging from roughly 18 MB to 145 MB.
TabNet is the most memory-efficient method on three of these four datasets (17.7 MB on XOR, 21.1 MB on SynthB, 26.2 MB on Isolet), while the FSL achieves the best result on Letter Recognition (23.8 MB).
The Breast dataset, however, exposes an exception to this pattern.
While every other method's memory usage on Breast remains within the same order of magnitude (83.1–127.7 MB), TabNet's peak GPU allocation increases to over 12 GB.
This behavior is consistent with TabNet's internal feature-grouping mechanism, which allocates a dense matrix scaling quadratically with the number of input features when no explicit feature grouping is provided.
Breast, with 54,675 features, is by far the highest-dimensional dataset evaluated, and this quadratic growth accounts for the disproportionate memory demand observed.
This result highlights an important practical limitation of TabNet's default configuration when applied to high-dimensional, ungrouped feature spaces such as gene-expression data.
Multiple FSL, on the other hand, is the only method whose memory usage grows considerably with the number of labels, as can be seen on Isolet (114.3 MB) and Letter Recognition (144.8 MB).
This demonstrates that Multiple FSL's memory usage is primarily driven by the number of output labels rather than the number of input features, consistent with its architecture maintaining a dedicated per-label weighting structure. 
However, unlike TabNet's quadratic blowup, this growth is linear rather than quadratic.

\begin{table}[t]
\centering
\caption{
\textbf{Execution times (s) for each feature weighting method across the XOR, SynthB, Breast, Isolet, and Letter Recognition datasets.} 
Values represent the mean $\pm$ standard deviation over 11 executions. 
To ensure a fair comparison with embedded methods, the times reported for post-hoc methods represent the sum of the base model training time and the method's execution time.
Note that SAGE's computational cost is not reported for the Breast dataset because it resulted in an out of memory error. 
The method's default Kernel estimator requires tracking an $\mathcal{O}(M^2)$ covariance matrix, which demands tens of gigabytes of RAM for the 54,675 features of the Breast dataset, exceeding standard hardware limits.
In bold is the best average per dataset.
Best results with statistical significance ($p < 0.01$) according to the Kruskal-Wallis and Dunn’s test are highlighted in orange.
}
\label{tab:all_execution_time}
\resizebox{\textwidth}{!}{%
\begin{tabular}{llllll}
\toprule
Model & XOR & SynthB & Breast & Isolet & Letter Recognition \\
\midrule
Base Neural Network & 4.59 $\pm$ 0.60 & 39.40 $\pm$ 4.22 & 14.83 $\pm$ 0.31 & 13.88 $\pm$ 0.53 & 98.19 $\pm$ 1.65 \\
Decision Tree & \cellcolor{orange!20}\textbf{0.00 $\pm$ 0.00} & \cellcolor{orange!20}0.76 $\pm$ 0.21 & \cellcolor{orange!20}1.24 $\pm$ 0.11 & \cellcolor{orange!20}2.13 $\pm$ 0.05 & \cellcolor{orange!20}0.04 $\pm$ 0.00 \\
Random Forest & \cellcolor{orange!20}0.06 $\pm$ 0.00 & \cellcolor{orange!20}1.94 $\pm$ 0.07 & \cellcolor{orange!20}0.66 $\pm$ 0.01 & \cellcolor{orange!20}5.73 $\pm$ 0.05 & \cellcolor{orange!20}0.85 $\pm$ 0.01 \\
Kruskal Wallis & \cellcolor{orange!20}0.02 $\pm$ 0.00 & \cellcolor{orange!20}0.16 $\pm$ 0.03 & 18.09 $\pm$ 0.56 & \cellcolor{orange!20}\textbf{0.83 $\pm$ 0.00} & \cellcolor{orange!20}\textbf{0.02 $\pm$ 0.00} \\
ReliefF & \cellcolor{orange!20}0.02 $\pm$ 0.01 & \cellcolor{orange!20}0.23 $\pm$ 0.04 & \cellcolor{orange!20}\textbf{0.47 $\pm$ 0.01} & \cellcolor{orange!20}1.84 $\pm$ 0.02 & \cellcolor{orange!20}4.06 $\pm$ 0.06 \\
Lasso & \cellcolor{orange!20}\textbf{0.00 $\pm$ 0.00} & \cellcolor{orange!20}\textbf{0.06 $\pm$ 0.01} & \cellcolor{orange!20}7.00 $\pm$ 0.74 & \cellcolor{orange!20}2.57 $\pm$ 0.51 & \cellcolor{orange!20}0.15 $\pm$ 0.27 \\
LIME & 8.96 $\pm$ 0.62 & 81.10 $\pm$ 4.86 & 487.27 $\pm$ 21.86 & 22.20 $\pm$ 0.58 & 128.10 $\pm$ 1.83 \\
DeepSHAP & 6.75 $\pm$ 0.66 & 94.54 $\pm$ 4.76 & 17.63 $\pm$ 0.56 & 18.79 $\pm$ 0.56 & 888.77 $\pm$ 64.91 \\
LassoNet & 190.77 $\pm$ 3.57 & 299.53 $\pm$ 19.94 & 386.92 $\pm$ 48.85 & 446.22 $\pm$ 45.68 & 367.71 $\pm$ 2.80 \\
TabNet & 11.68 $\pm$ 1.24 & 65.12 $\pm$ 15.62 & 178.61 $\pm$ 28.27 & 256.43 $\pm$ 39.44 & 196.93 $\pm$ 7.47 \\
SAGE & 3.00 $\pm$ 0.15 & 36.37 $\pm$ 2.05 & -- & 228.40 $\pm$ 58.02 & 142.65 $\pm$ 21.44 \\
FSL & 9.36 $\pm$ 2.16 & 41.36 $\pm$ 2.37 & 19.21 $\pm$ 0.44 & 15.33 $\pm$ 0.65 & 114.09 $\pm$ 3.64 \\
Multiple FSL & 6.99 $\pm$ 0.94 & 44.32 $\pm$ 1.90 & 39.89 $\pm$ 5.54 & 16.55 $\pm$ 0.78 & 122.56 $\pm$ 2.49 \\
\bottomrule
\end{tabular}%
}
\end{table}

\begin{figure*}[h]
    \centering
    \begin{subfigure}[b]{0.49\textwidth}
         \centering
         \includegraphics[width=\textwidth]{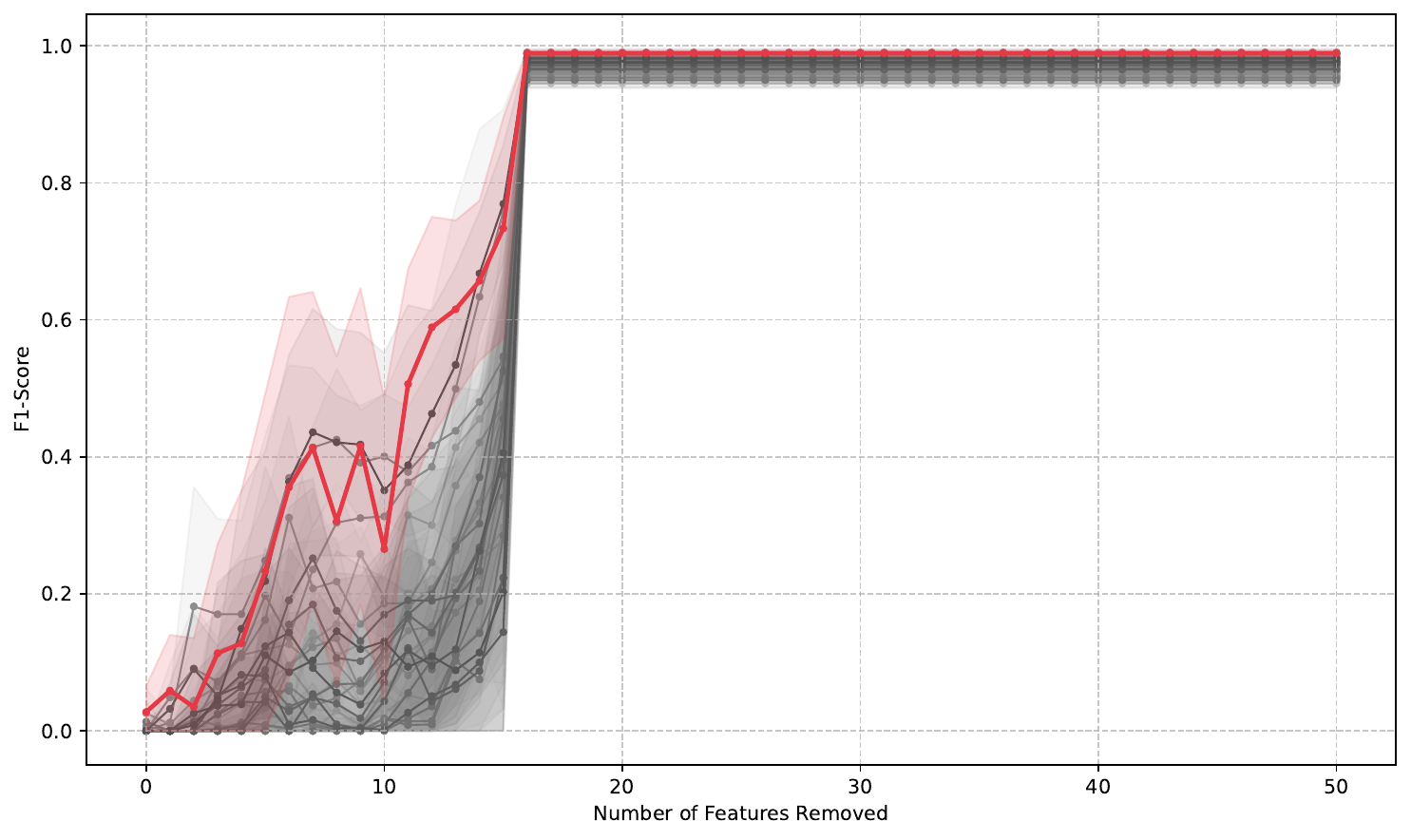}
         \caption{DeepSHAP}
         \label{subfig:comparative_wtsne_model_colorectal_lasso_label_1}
     \end{subfigure}
     \hfill
     \begin{subfigure}[b]{0.49\textwidth}
         \centering
         \includegraphics[width=\textwidth]{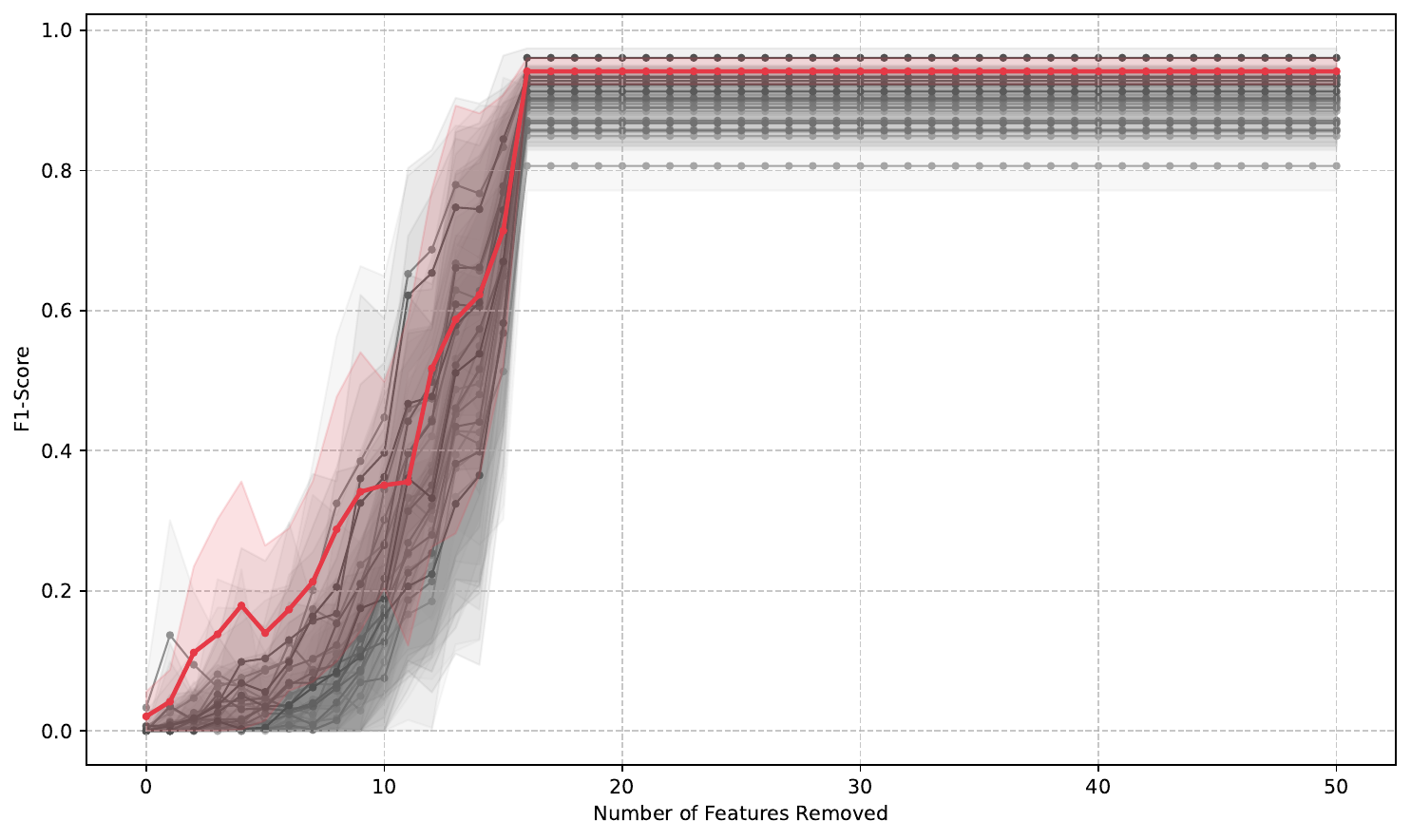}
         \caption{TabNet}
         \label{subfig:comparative_wtsne_model_colorectal_lime_label_1}
     \end{subfigure}
     \begin{subfigure}[b]{0.49\textwidth}
         \centering
         \includegraphics[width=\textwidth]{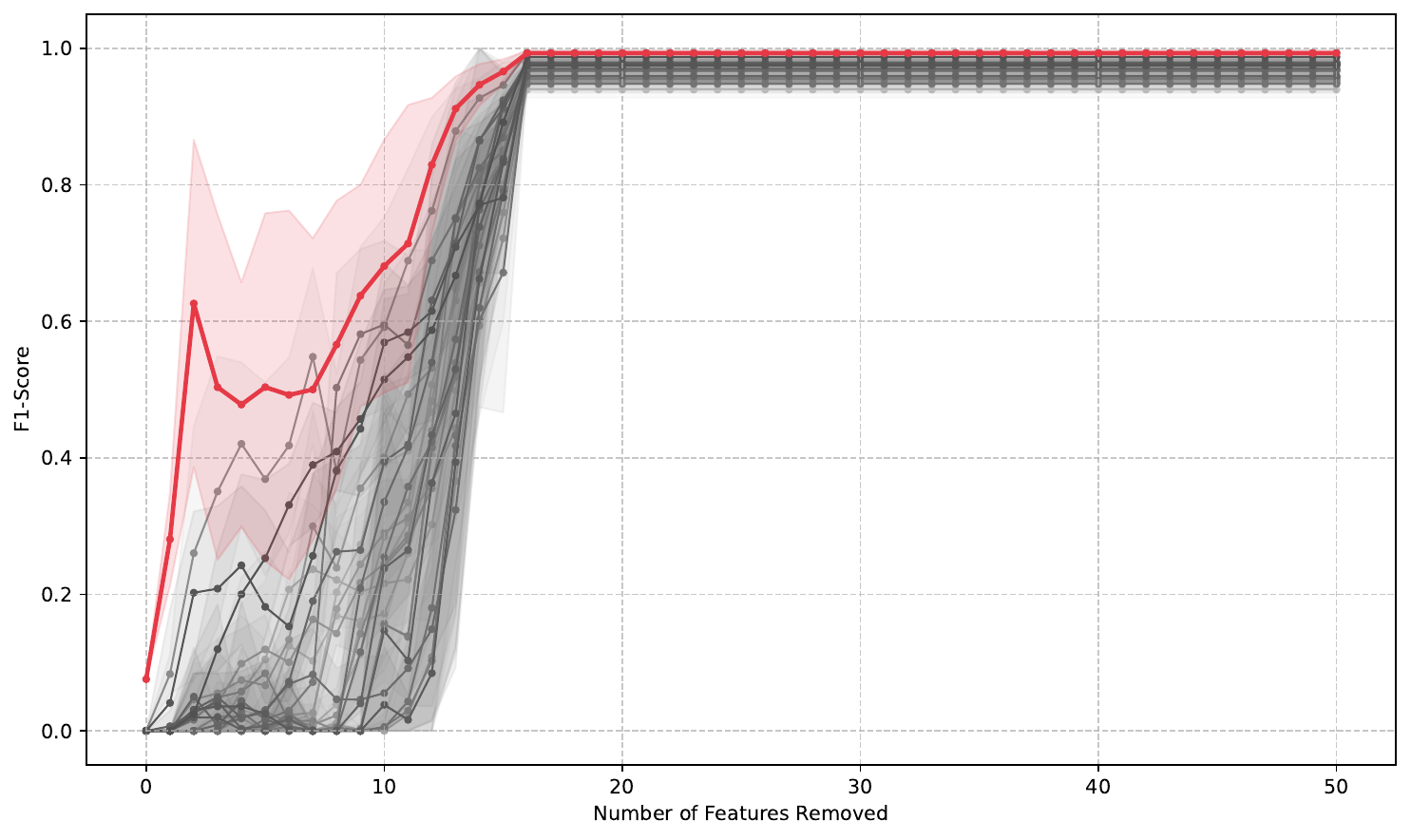}
         \caption{Multiple FSL}
         \label{subfig:comparative_wtsne_model_colorectal_random_deepshap_label_1}
     \end{subfigure}
    \caption[
    Impact of the most relevant label-specific features on predictive performance in the Letter Recognition dataset.
    The graph illustrates the effect of initially erasing all features and then sequentially re-enabling them, one by one, according to the label-specific feature weight ranking for label 0. 
    Visually, the figure comprises three individual line charts labeled (a) DeepSHAP, (b) TabNet, and (c) Multiple FSL. 
    Each chart's vertical axis measures the F1-score from 0.0 to 1.0, while the horizontal axis tracks the number of features re-enabled. 
    In every chart, a distinct red line (with a light red shaded variance band) represents the target label (label 0), accompanied by numerous overlapping gray lines (with gray shaded bands) representing all other labels.
    The charts visually reinforce the analytical findings: starting near zero, as the first few features are added, the red line generally rises faster and visibly separates itself above the dense cluster of gray lines, confirming that label 0 is the most heavily impacted by these initial top-ranked features. 
    However, by the time the horizontal axis reaches 15 features, a sharp, steep climb occurs for all lines simultaneously.
    At this point, the red and gray lines converge into a solid, flat, horizontal plateau just below 1.0. 
    This dense plateau continues for the remainder of the chart, visually demonstrating that beyond 15 features, the inclusion of additional features affects all classes uniformly and yields no further improvements in the F1-score.
    ]
    {
    \textbf{Impact of the most relevant label-specific features on predictive performance in the Letter Recognition dataset.}
    The graph illustrates the effect of initially erasing all features and then sequentially re-enabling them, one by one, according to the label-specific feature weight ranking for label 0. The red line denotes the F1-score of the target label, while the gray lines represent the F1-scores of all other labels.
    }
    \label{fig:feature_erasing_letter_recognition}
\end{figure*}

The Isolet and Letter Recognition datasets represent a distinct challenge for label-specific feature weighting methods, as both combine a small number of features with a large number of classes. 
All methods evaluated in the present work produced similar results, consistently identifying the similar features as relevant for multiple labels. 
Among the methods capable of producing label-specific feature weights, DeepSHAP, TabNet, and Multiple FSL achieved the best scores. 
\autoref{tab:isolet_silhouette} shows the 2D silhouette coefficients per label for the Isolet dataset, computed using the label-specific feature weights produced by these three methods. 
As can be observed, most coefficients are negative or near zero, indicating that the weighted t-SNE embedding was not able to separate the samples belonging to the target label from the remaining samples. 
Although some improvement over the baseline can be observed, it remains minimal.
\autoref{fig:feature_erasing_letter_recognition} shows the impact of initially erasing all features from the Letter Recognition dataset and sequentially re-enabling them, one by one, based on the label-specific feature rankings for label 0. 
The red line represents the F1-score for label 0, while the gray lines represent the performance of the remaining labels. 
For all three methods, label 0 is, on average, the most heavily impacted class during the inclusion of the first few features. By the time 15 features are enabled, the F1-scores for all labels have already plateaued. 
This demonstrates that while all methods successfully select appropriate features for each label, the majority of features affect multiple classes simultaneously, which complicates the task of identifying strictly label-specific features.

\begin{table}[h]
\centering
\caption{
\textbf{Peak GPU memory usage (MB) for each feature weighting method across the XOR, SynthB, Breast, Isolet, and Letter Recognition datasets.}
Standard deviation is omitted because the variation in peak GPU allocation is minimal, given the same architecture and dataset, the memory footprint remains nearly identical across all executions. 
All peak GPU memory allocations were measured using an NVIDIA GeForce RTX 5070 Ti (16 GB VRAM).
In bold is the best average per dataset and metric.
}
\label{tab:all_memory_usage}
\resizebox{\textwidth}{!}{%
\begin{tabular}{lrrrrr}
\toprule
Model & XOR & SynthB & Breast & Isolet & Letter Recognition \\
\midrule
Base Neural Network & 19.8 & 23.3 & \textbf{83.1} & 35.8 & \textbf{23.7} \\
DeepSHAP & 21.7 & 39.7 & 97.0 & 35.8 & 34.3 \\
LassoNet & 20.2 & 29.3 & 102.6 & 47.9 & 62.1 \\
TabNet & \textbf{17.7} & \textbf{21.1} & 12063.8 & \textbf{26.2} & 28.7 \\
FSL & 19.8 & 24.0 & 83.7 & 38.2 & 23.8 \\
Multiple FSL & 21.0 & 27.1 & 127.7 & 114.3 & 144.8 \\
\bottomrule
\end{tabular}%
}
\end{table}

\begin{table}[h]
\centering
\caption{
\textbf{Isolet dataset 2D silhouette coefficients per label.}
Averages were calculated over 11 executions.
A one-vs-all (OVA) strategy was adopted, where all labels other than the target label were collapsed into a single category.
In bold is the highest average.
Best results with statistical significance (p < 0.01) according to the Kruskal-Wallis and Dunn’s test are highlighted in orange.
}
\label{tab:isolet_silhouette}
\begin{tabular}{lllll}
\toprule
Label & Baseline & DeepSHAP & TabNet & Multiple FSL \\
\midrule
A & -0.182 $\pm$ 0.000 & \cellcolor{orange!20}-0.173 $\pm$ 0.027 & \cellcolor{orange!20}\textbf{-0.128 $\pm$ 0.039} & -0.181 $\pm$ 0.031 \\
B & -0.103 $\pm$ 0.000 & \cellcolor{orange!20}-0.039 $\pm$ 0.069 & \cellcolor{orange!20}\textbf{-0.002 $\pm$ 0.025} & \cellcolor{orange!20}-0.041 $\pm$ 0.063 \\
C & 0.144 $\pm$ 0.000 & \cellcolor{orange!20}0.135 $\pm$ 0.052 & \cellcolor{orange!20}0.146 $\pm$ 0.100 & \cellcolor{orange!20}\textbf{0.179 $\pm$ 0.034} \\
D & -0.052 $\pm$ 0.000 & \cellcolor{orange!20}-0.041 $\pm$ 0.038 & \cellcolor{orange!20}-0.017 $\pm$ 0.033 & \cellcolor{orange!20}\textbf{0.003 $\pm$ 0.093} \\
E & -0.048 $\pm$ 0.000 & \cellcolor{orange!20}-0.034 $\pm$ 0.043 & \cellcolor{orange!20}\textbf{0.038 $\pm$ 0.036} & -0.131 $\pm$ 0.050 \\
F & 0.020 $\pm$ 0.000 & \cellcolor{orange!20}0.097 $\pm$ 0.038 & \cellcolor{orange!20}\textbf{0.116 $\pm$ 0.054} & \cellcolor{orange!20}0.104 $\pm$ 0.050 \\
G & -0.029 $\pm$ 0.000 & -0.061 $\pm$ 0.043 & \cellcolor{orange!20}0.055 $\pm$ 0.026 & \cellcolor{orange!20}\textbf{0.062 $\pm$ 0.049} \\
H & 0.102 $\pm$ 0.000 & \cellcolor{orange!20}\textbf{0.172 $\pm$ 0.021} & \cellcolor{orange!20}0.078 $\pm$ 0.125 & 0.089 $\pm$ 0.037 \\
I & -0.038 $\pm$ 0.000 & \cellcolor{orange!20}0.083 $\pm$ 0.069 & \cellcolor{orange!20}0.091 $\pm$ 0.107 & \cellcolor{orange!20}\textbf{0.157 $\pm$ 0.043} \\
J & -0.159 $\pm$ 0.000 & \cellcolor{orange!20}-0.065 $\pm$ 0.055 & \cellcolor{orange!20}\textbf{0.024 $\pm$ 0.071} & \cellcolor{orange!20}-0.060 $\pm$ 0.135 \\
K & -0.161 $\pm$ 0.000 & -0.027 $\pm$ 0.074 & -0.039 $\pm$ 0.068 & \cellcolor{orange!20}\textbf{0.046 $\pm$ 0.040} \\
L & 0.016 $\pm$ 0.000 & \cellcolor{orange!20}-0.020 $\pm$ 0.069 & \cellcolor{orange!20}0.052 $\pm$ 0.080 & \cellcolor{orange!20}\textbf{0.070 $\pm$ 0.051} \\
M & -0.139 $\pm$ 0.000 & -0.121 $\pm$ 0.066 & \cellcolor{orange!20}-0.021 $\pm$ 0.084 & \cellcolor{orange!20}\textbf{0.052 $\pm$ 0.057} \\
N & -0.173 $\pm$ 0.000 & -0.150 $\pm$ 0.033 & \cellcolor{orange!20}-0.097 $\pm$ 0.071 & \cellcolor{orange!20}\textbf{0.008 $\pm$ 0.079} \\
O & 0.155 $\pm$ 0.000 & 0.007 $\pm$ 0.046 & \cellcolor{orange!20}0.124 $\pm$ 0.044 & \cellcolor{orange!20}\textbf{0.154 $\pm$ 0.046} \\
P & -0.019 $\pm$ 0.000 & -0.122 $\pm$ 0.025 & \cellcolor{orange!20}\textbf{-0.010 $\pm$ 0.041} & \cellcolor{orange!20}-0.073 $\pm$ 0.081 \\
Q & 0.105 $\pm$ 0.000 & \cellcolor{orange!20}\textbf{0.162 $\pm$ 0.024} & 0.065 $\pm$ 0.082 & 0.087 $\pm$ 0.077 \\
R & 0.100 $\pm$ 0.000 & \cellcolor{orange!20}0.141 $\pm$ 0.049 & \cellcolor{orange!20}\textbf{0.158 $\pm$ 0.051} & \cellcolor{orange!20}0.155 $\pm$ 0.027 \\
S & 0.067 $\pm$ 0.000 & 0.159 $\pm$ 0.028 & \cellcolor{orange!20}0.193 $\pm$ 0.073 & \cellcolor{orange!20}\textbf{0.202 $\pm$ 0.015} \\
T & 0.014 $\pm$ 0.000 & -0.029 $\pm$ 0.048 & \cellcolor{orange!20}0.077 $\pm$ 0.037 & \cellcolor{orange!20}\textbf{0.100 $\pm$ 0.028} \\
U & 0.162 $\pm$ 0.000 & \cellcolor{orange!20}0.094 $\pm$ 0.120 & \cellcolor{orange!20}0.075 $\pm$ 0.073 & \cellcolor{orange!20}\textbf{0.136 $\pm$ 0.035} \\
V & -0.063 $\pm$ 0.000 & \cellcolor{orange!20}-0.063 $\pm$ 0.061 & \cellcolor{orange!20}\textbf{-0.018 $\pm$ 0.022} & -0.073 $\pm$ 0.055 \\
W & 0.021 $\pm$ 0.000 & \cellcolor{orange!20}0.097 $\pm$ 0.076 & -0.037 $\pm$ 0.081 & \cellcolor{orange!20}\textbf{0.127 $\pm$ 0.073} \\
X & 0.110 $\pm$ 0.000 & \cellcolor{orange!20}0.109 $\pm$ 0.054 & \cellcolor{orange!20}\textbf{0.146 $\pm$ 0.055} & \cellcolor{orange!20}0.140 $\pm$ 0.025 \\
Y & 0.156 $\pm$ 0.000 & \cellcolor{orange!20}\textbf{0.192 $\pm$ 0.028} & 0.128 $\pm$ 0.055 & \cellcolor{orange!20}0.177 $\pm$ 0.023 \\
Z & 0.090 $\pm$ 0.000 & \cellcolor{orange!20}0.080 $\pm$ 0.034 & \cellcolor{orange!20}\textbf{0.104 $\pm$ 0.040} & -0.004 $\pm$ 0.055 \\
\bottomrule
\end{tabular}
\end{table}

In deep learning architectures that rely on raw predictions to infer feature importance, it is crucial to ensure that the label-wise logits are well-calibrated. 
A well-calibrated model outputs confidence scores that accurately reflect its true probability of correctness, mitigating the overconfidence issue commonly found in modern neural networks. 
Unlike the original FSL implementation, our proposed Multiple FSL generates each label's output logit based on a distinct feature weighting vector. 
To empirically assess whether these label-wise logits remain well-calibrated despite originating from independent propagations (\autoref{eq:multiple_fslp_propagation}), we measured the top-1 Expected Calibration Error (ECE) \cite{calibration:2017} of the predicted-class confidence against actual accuracy for five models: the baseline neural network, LassoNet, TabNet, FSL, and Multiple FSL.
The baseline neural network is identical to the one used internally by FSL and Multiple FSL, but without any feature gating. 
The original FSL, which relies on a single shared feature weighting vector to generate predictions for all classes, is theoretically less prone to producing miscalibrated logits. 
All five models were evaluated across the same nine benchmark datasets used throughout this work (\autoref{tab:expected_calibration_error}). 
The baseline neural network yielded an average ECE of 0.084 across the nine datasets, noticeably higher than every gated model we evaluated: FSL (0.054), LassoNet (0.066), TabNet (0.073), and Multiple FSL (0.042). 
Notably, Multiple FSL achieved the lowest overall mean ECE and secured the best individual calibration on four of the nine datasets (SynthB, Liver, Breast, and Isolet), while tying for the lowest score on XOR.
Even in datasets where it did not achieve the absolute lowest ECE, the performance gap was minimal. 
On Letter Recognition, it trailed the baseline neural network by a negligible margin of 0.001. 
On Colorectal, the smallest dataset in this comparison with only 86 samples and 4 classes, it trailed TabNet by 0.027, a difference well within the variance observed among the other models on that same dataset (0.129–0.190).
\autoref{tab:expected_calibration_error_complete} of \ref{sec:complementary_experiments} presents the complete results, including comparisons with other embedded feature weighting methods. 
While Multiple FSL does not always achieve the absolute lowest average across all datasets, statistical analysis over the 11 executions reveals no significant difference between its performance and that of the top-performing models.
If assembling label-wise logits from independent feature-weighting layers introduced a systematic scale mismatch, Multiple FSL would be expected to exhibit poorer calibration than both the non-gated baseline and models that derive all logits from a single shared layer, as FSL.
Instead, it matches or outperforms them on average.
Its rare instances of underperformance are minor and more plausibly attributed to the affected dataset's sample size rather than the multi-propagation mechanism itself. 
Consequently, we did not apply temperature scaling or a shared cross-label normalization to Multiple FSL, as the empirical evidence strongly indicates that its per-label logits are natively well-calibrated relative to comparable single-propagation alternatives.

\begin{table}[!b]
\centering
\caption{
\textbf{Expected Calibration Error (ECE) comparison across datasets.}
Lower values indicate better calibration. 
The proposed Multiple FSL consistently achieves highly calibrated probabilities without the need for post-hoc adjustments. 
For a complete comparison with additional baselines, please refer to \autoref{tab:expected_calibration_error_complete}.
In bold is the lowest average.
Best results with statistical significance (p < 0.01) according to the Kruskal-Wallis and Dunn’s test are highlighted in orange.
}
\label{tab:expected_calibration_error}
\resizebox{\textwidth}{!}{%
\begin{tabular}{llllll}
\toprule
Dataset & Baseline & LassoNet & TabNet & FSL & Multiple FSL \\
\midrule
XOR & 0.078 $\pm$ 0.051 & \cellcolor{orange!20}\textbf{0.003 $\pm$ 0.001} & 0.067 $\pm$ 0.028 & \cellcolor{orange!20}\textbf{0.003 $\pm$ 0.007} & \cellcolor{orange!20}\textbf{0.003 $\pm$ 0.006} \\
SynthA & 0.086 $\pm$ 0.027 & \cellcolor{orange!20}\textbf{0.044 $\pm$ 0.017} & 0.101 $\pm$ 0.025 & 0.073 $\pm$ 0.010 & \cellcolor{orange!20}0.062 $\pm$ 0.010 \\
SynthB & 0.090 $\pm$ 0.038 & 0.087 $\pm$ 0.067 & 0.067 $\pm$ 0.018 & \cellcolor{orange!20}0.037 $\pm$ 0.011 & \cellcolor{orange!20}\textbf{0.029 $\pm$ 0.005} \\
SynthC & 0.076 $\pm$ 0.032 & \cellcolor{orange!20}\textbf{0.041 $\pm$ 0.013} & \cellcolor{orange!20}0.054 $\pm$ 0.020 & 0.086 $\pm$ 0.007 & 0.066 $\pm$ 0.016 \\
Liver & \cellcolor{orange!20}0.094 $\pm$ 0.142 & \cellcolor{orange!20}0.050 $\pm$ 0.074 & \cellcolor{orange!20}0.111 $\pm$ 0.074 & \cellcolor{orange!20}0.033 $\pm$ 0.080 & \cellcolor{orange!20}\textbf{0.023 $\pm$ 0.053} \\
Colorectal & \cellcolor{orange!20}0.172 $\pm$ 0.080 & \cellcolor{orange!20}0.190 $\pm$ 0.046 & \cellcolor{orange!20}\textbf{0.109 $\pm$ 0.091} & \cellcolor{orange!20}0.129 $\pm$ 0.064 & \cellcolor{orange!20}0.136 $\pm$ 0.079 \\
Breast & 0.071 $\pm$ 0.032 & 0.138 $\pm$ 0.044 & 0.117 $\pm$ 0.024 & \cellcolor{orange!20}0.034 $\pm$ 0.022 & \cellcolor{orange!20}\textbf{0.033 $\pm$ 0.030} \\
Isolet & 0.079 $\pm$ 0.019 & \cellcolor{orange!20}0.027 $\pm$ 0.007 & \cellcolor{orange!20}0.023 $\pm$ 0.004 & 0.079 $\pm$ 0.014 & \cellcolor{orange!20}\textbf{0.016 $\pm$ 0.005} \\
Letter R. & \cellcolor{orange!20}\textbf{0.010 $\pm$ 0.002} & \cellcolor{orange!20}0.011 $\pm$ 0.002 & \cellcolor{orange!20}0.012 $\pm$ 0.003 & \cellcolor{orange!20}0.011 $\pm$ 0.002 & \cellcolor{orange!20}0.011 $\pm$ 0.002 \\
\bottomrule
\end{tabular}
}%
\end{table}

\autoref{fig:comparative_wtsne_model_breast} presents the weighted t-SNE visualization using general and label-specific feature weights generated by Multiple FSL for each label available on Breast dataset.
A substantial improvement is observed in the visualization of each target label, accompanied by a deterioration in the others, indicating that not only were the label-specific features effectively highlighted, but features relevant to other labels were also suppressed, confirming the ability of Multiple FSL to identify label-specific feature weights.
More experiments are in Figure \ref{fig:comparative_wtsne_model_colorectal_general} and Tables \ref{tab:hd_and_2d_silhouette_comparison_colorectal_general_weights},\ref{tab:label_f1_score_letter_recognition} and \ref{tab:label_f1_score_isolet} of \ref{sec:complementary_experiments}.

\begin{table}[!b]
\centering
\caption[Colorectal dataset HD and 2D silhouette coefficients with feature weights related to label 1.
For HD silhouette and 2D silhouette, Multiple FSL had the highest average and also the highest result with statistical significance.] {\textbf{Colorectal dataset HD and 2D silhouette coefficients with feature weights related to label 1.}
Averages were calculated over 11 executions.
``None'' represents the t-SNE without using weighted features, being a fixed value.
Averages were calculated with the weighted t-SNE results given the feature weights generated by each algorithm through 11 different executions.
A one-vs-all (OVA) strategy was adopted, where all labels other than label 1 were converted into a single label.
In bold is the highest average.
Best results with statistical significance ($p < 0.01$) according to the Kruskal-Wallis and Dunn’s test are highlighted in orange.
}
\label{tab:hd_and_2d_silhouette_comparison_colorectal}
\begin{tabular}{lrr}
\hline
    Weight source               & HD silhouette                                         & 2D silhouette        \\ 
\hline
    None                        & .024                                                & .090       \\
    Lasso                       & .024$\pm$.000                                     & .060$\pm$.032     \\
    LIME                        & .024$\pm$.000                                     & .036$\pm$.030     \\
    DeepSHAP                    & .048$\pm$.008                                     & .141$\pm$.042     \\
    Multiple FSL                & \cellcolor{orange!20} \textbf{.143$\pm$.025}      & \cellcolor{orange!20} \textbf{.446$\pm$.027}     \\
\hline
\end{tabular}
\end{table}

\begin{table}[h]
\centering
\caption[PIFS comparison between Multiple FSL, LassoNet, TabNet and SAGE with SynthB dataset.] {\textbf{PIFS comparison between Multiple FSL, LassoNet, TabNet and SAGE with SynthB dataset.}
The measure considers a subset of 30 features, which is the amount of informative features on SynthB dataset.
SAGE is the only method unable to define label-specific weights. 
In bold is the highest average of each dataset and label.
Best results with statistical significance ($p < 0.01$) according to the Kruskal-Wallis and Dunn’s test are highlighted in orange.
}
\label{tab:comparative_pifs_for_synthB}
\begin{tabular}{lrrrr}
\hline
Algorithm & General & Label 0 & Label 1 & Label 2 \\ 
\hline
\multirow{1}{*}{LassoNet}  
& \cellcolor{orange!20} 0.948$\pm$.031  & 0.936$\pm$.067 & \cellcolor{orange!20} \textbf{1.0$\pm$.000} & 0.882$\pm$.098 \\ 
\multirow{1}{*}{TabNet}
& 0.530$\pm$.071  & 0.518$\pm$.147 &  0.591$\pm$.130 & 0.509$\pm$.138 \\ 
\multirow{1}{*}{SAGE}
& \cellcolor{orange!20} \textbf{0.951$\pm$.035}  & - & - & - \\ 
\multirow{1}{*}{Multiple FSL} 
& \cellcolor{orange!20} 0.921$\pm$.040  & \cellcolor{orange!20} \textbf{0.982$\pm$.040} & \cellcolor{orange!20} \textbf{1.0$\pm$.000} & \cellcolor{orange!20} \textbf{1.0$\pm$.000} \\ 
\hline
\end{tabular}
\end{table}

\subsection{Results}
\label{subsection:results}
To improve both the interpretability and performance of neural networks, we propose an extension to FSL that introduces label-specific feature weighting. 
The extension was evaluated on synthetic and real-world datasets to assess its ability to: (1) assign general feature weights, as in the original FSL, to identify the most relevant features; (2) assign label-specific weights to prioritize features for individual labels; and (3) examine the relationship between the feature relevance assigned by the algorithm and the resulting model prediction performance.
Initial experiments on the XOR and SynthA datasets validated Multiple FSL ability to access general feature relevance. 
Subsequently, the SynthB and SynthC datasets were employed to assess the algorithm's capacity to discriminate between features relevant to specific labels.
Through the application of feature erasure, weighted t-SNE, and a SVC model trained with subsets of the highest weighted features to three microarray datasets, we validate the effectiveness of Multiple FSL in handling high-dimensional real-world datasets with limited sample sizes.
Compared to existing algorithms, Multiple FSL performs similarly for general relevant features while demonstrating a superior ability to capture label-specific feature relationships, mainly on high-dimensional datasets with a small sample size.

%%%%%%%%%%%%%%%%%%%%%%%%%%%%%%%%%%%%%%%%%%

\section{Conclusions}
\label{sec:conclusios}
The present work has the objective of improving Deep Neural Network interpretability for multi-class data by introducing a new embedded feature weighting algorithm. 
While post-hoc methods can assign label-specific weights, embedded approaches use the internal mechanisms of neural networks during the training phase.
However, most existing embedded methods lack label-specific relevance, limiting interpretability in multi-class scenarios.
To address this, we posed the research question: ``Can an embedded FSL be capable of learning label-specific feature relevance?''. 
To answer it, we proposed four FSL extensions and identified Multiple FSL as the most promising. 
This novel embedded feature weighting algorithm builds upon FSL adds label-specific selection layers, enabling identification of relevant features per label.
To address the complementary question, ``How to evaluate a label-specific feature weighting algorithm?'', we applied different methodologies to evaluate the identification of relevant features and the assignment of higher weights to them. 
Synthetic datasets demonstrated that the algorithm assigns high weights to relevant and low weights to noisy features, distinguishing between general and label-specific relevance.
In real-world data, weighted t-SNE and SVC models highlighted the algorithm ability to define label-specific feature weighting and feature erasure confirmed the predictor relies most on top-weighted features.
Our results demonstrated that Multiple FSL is capable of improving the model interpretability by indicating which features impacts most each label.

Finally, to address the question, ``How does this approach compare to other algorithms?'', we conducted comparative analyses with multiple established methods.
Our findings demonstrate that Multiple FSL exhibits competitive performance defining general relevant features and a superior performance to identify label-specific feature relevance. 
In contrast to post-hoc explainability techniques such as LIME, SHAP and SAGE, the proposed method bypasses significant computational overhead by directly aligning feature weights with the model's predictions at the cost of a linear increase on memory usage. 
This intrinsic efficiency makes it highly suitable for demanding real-world applications.

It is also important to identify the limitations of the proposed work.
The proposed Multiple FSL algorithm was specifically developed for, and is currently limited to, tabular datasets due to its reliance on fixed feature semantics.
The method operates by learning global feature weights that are applied uniformly across the entire dataset. 
This mechanism inherently requires that the meaning of each individual feature remains constant across all samples. 
In unstructured data such as images, however, individual features (pixels) lack a fixed semantic meaning, a specific pixel's relevance changes entirely from one image to another and must be interpreted within the local context of that specific sample. 
Furthermore, because Multiple FSL optimizes these global weights to reduce or explicitly disable the impact of features it considers irrelevant, applying it to image data would permanently blind the neural network to specific spatial locations. 
This global suppression would severely restrict the network's flexibility to detect varying visual patterns.

Like most feature weighting methods, Multiple FSL depends on the predictive capacity of the underlying predictive model. 
When the model performs poorly, the identified features may not reflect true relevance to the real-world problem, such as true biomarkers in microarray datasets. 
Instead, the model might be relying on illegitimate shortcuts that lead to overfitting. 
However, isolating the features the model primarily relies on in these situations remains valuable, as it allows researchers to detect and mitigate the specific patterns causing the overfitting.
Multiple FSL is designed for easy integration with any neural network model. 
For future work, this approach could be explored in two distinct scenarios: (1) applying a trained Multiple FSL to an untrained model to enhance performance and reinforce relevant labels; and (2) attaching an untrained Multiple FSL to a pre-trained model to learn relevant features. 
This second technique has already been successfully implemented with the original FSL \cite{posthoc_fsl:2026}.
Our experiments demonstrated the effectiveness of Multiple FSL in identifying label-specific features across diverse dataset types, including gene expression data. 
As a next step, these identified features could be analyzed from a biological perspective to validate them against known biomarkers and investigate potential novel ones. 
Additionally, future work could explore the integration of Multiple FSL with other neural network architectures, such as CNNs and Transformers.

\section*{Data and code availability}
\textbf{Paper source code} is available at \href{https://github.com/JPedroSilveira/label-specific-feature-selection-layer} {https:\/\/github.com/JPedroSilveira/label-specific-fsl}.
\textbf{Weighted t-SNE \cite{WTSNE:2025}} at   \href{https://github.com/sbcblab/weighted_tSNE}{https:\/\/github.com/sbcblab/weighted\_tSNE}.
And \textbf{Feature selection analysis framework \cite{barbieri_analysis_feature_selection:2024}} at \href{https://github.com/sbcblab/GenExpFS}{https:\/\/github.com/sbcblab/GenExpFS}.
\textbf{CuMiDa datasets \cite{feltes_cumida:2019}} can be found at \href{https://sbcb.inf.ufrgs.br/cumida}{https:\/\/sbcb.inf.ufrgs.br/cumida}. \textbf{Synthetic datasets} are available within the source code.

\section*{Funding}

This research  was funded by Fundaç\~ao de Amparo à Pesquisa do Estado do Rio Grande do Sul (\texttt{FAPERGS}) [24/2551-0001392-0; 23/2551-0001894-2; 19/2551-000190\newline6-8; 24/2551-0000725-3; 25/2551-0002116-2].
This work was supported by the Serrapilheira Institute (grant number \texttt{Serra -  R-2501-51351}).
This study was financed in part by the Coordination for the Improvement of Higher Education Personnel – Brazil (CAPES) – Finance Code 001.

\section*{CRediT authorship contribution statement}

\textbf{JPSS:} Software, Investigation, Writing - Review and Editing, Visualization.
\textbf{MD:} Conceptualization of this study, Resources, Writing - Review and Editing, Supervision, Project Administration, Funding Acquisition.
\textbf{BIG:} Conceptualization of this study, Resources, Methodology, Writing - Review and Editing, Project Administration, Funding Acquisition.

\section*{Declaration of competing interest}

The authors declare that they have no known competing financial
interests or personal relationships that could have appeared to influence this work.

%\bibliographystyle{cas-model2-names}
%\bibliographystyle{unsrtnat}
%\bibliographystyle{apa}
%\bibliographystyle{plainnat}
\bibliographystyle{unsrtnat}
\bibliography{cas-refs}

% \section*{Appendix}

% \appendix

\renewcommand\thefigure{\thesection.\arabic{figure}}  
\setcounter{figure}{0}    

\renewcommand\thetable{\thesection.\arabic{table}}  
\setcounter{table}{0}    

% \section{Section}

\newpage

\appendix

%%% SECTION

\section{Feature weighting algorithms} 
\label{sec:feature_weighting_algorithms}
In this appendix, we introduce five consolidated feature weighting algorithms used on our experiments, the Kruskal–Wallis filter (\ref{subsec:kruskal_wallis}), ReliefF (\ref{subsec:relieff_alg}), Decision Tree (\ref{subsec:decision_tree}), Random Forest (\ref{subsec:random_forest}) and LASSO (\ref{subsec:lasso}).

\subsection{Kruskal–Wallis filter} 
\label{subsec:kruskal_wallis}
The Kruskal-Wallis test \cite{kruskal_test:1952} is a non-parametric statistical test that evaluates variation within and between groups to determine if a group of random variables originates from the same distribution \cite{barbieri_analysis_feature_selection:2024}. Kruskal-Wallis test generates feature weights that are particularly useful as a filter which identifies features that significantly differentiate between labels within a dataset \cite{relevance_aggregation:2021}.
Given a labeled dataset with $L$ labels, to test whether a feature $f$ is relevant to the label, first $f$'s values are partitioned into $L$ groups according to their label membership. 
Then, the Kruskal-Wallis statistic is calculated to determine if these $L$ groups are significantly different \cite{vora_feature_selection_algorithms:2017}.
The Kruskal-Wallis test is computed by calculating the $H$ value using \autoref{eq:kruskal_wallis_test_1} and \ref{eq:kruskal_wallis_test_2}.
In these equations, $L$ is the number of compared groups, $n_i$ is the number of samples in the $i$-th group, $N$ is the total number of samples across all groups, and $R_i$ is the sum of the values in the $i$-th group \cite{barbieri_analysis_feature_selection:2024}.

\begin{equation}
    \label{eq:kruskal_wallis_test_1}
    N = \sum_{i=1}^{L} n_i
\end{equation}

\begin{equation}
    \label{eq:kruskal_wallis_test_2}
    H = \frac{12}{N(N+1)} \sum_{i=1}^{L} \frac{R_i^2}{n_i} - 3(N+1) 
\end{equation}

\subsection{ReliefF} 
\label{subsec:relieff_alg}
ReliefF \cite{kononenko_relieff:1994} is an extension of Relief \cite{kira_relief_practical:1992} that can handle multi-class problems. 
It is also more robust and capable of dealing with incomplete and noisy data \cite{robnik_theoretical_relief_analysis:2003}, being a popular choice of filter method through feature weighting \cite{barbieri_analysis_feature_selection:2024}.
The pseudo-code for ReliefF is described in Algorithm \ref{alg:relieff}. 
It starts with all feature weights, $W$, set to zero. For each given instance $R_i$, it searches for $k$ of its nearest neighbors from the same label, $\text{label}(R_i)$, called nearest hits $H$, and also $k$ nearest neighbors from each of the different labels, called nearest misses $M_l$. 
Then it updates the weights estimation $W$ for all attributes $A$ based on their values for $R_i$, hits $H_j$, and misses $M_l$. The function $\textit{diff}$ calculates the probability that two given instances have different values for a given feature, conditioned on a label value. The contribution for each label of the misses is weighted with the prior probability of that label, $P(L)$, which is estimated from the training set, divided by $1 - P(\text{label}(R_i))$ (the sum of probabilities for the misses labels) to keep the values between 0 and 1 \cite{robnik_theoretical_relief_analysis:2003,relevance_aggregation:2021}. 

\begin{algorithm}
    \caption{ReliefF}
    \label{alg:relieff}
    \setcounter{AlgoLine}{0}
    \SetKwInOut{Input}{Inputs}\SetKwInOut{Output}{Output}
    \Input{
    \\
        \textbf{R}: Set of all instances\\
        \textbf{A}: Set of all features \\
        \textbf{L}: Set of all labels\\
        \textbf{k}: Number of neighbors to lookup\\
    }
    \Output{
    \\
            \textbf{W}: Vector of feature weights
    }
    $W_{0\dots a} \gets 0.0$\\
    $m \gets |R|$\\
    \For{$i \gets 1$ \KwTo $m$}{
        $H \gets k\_nearest\_hits(k, R_i, R)$\\
        \For{$l$ \textbf{in} $L \neq label\left(R_i\right)$}{
            $M_l \gets k\_nearest\_misses(k, l, R_i, R)$\\
        }
        \For{$a \gets 1$ \KwTo $|A|$}{
            \For{$hit$ \textbf{in} $H$}{
                $W_a \gets W_a - \frac{diff(a,R_{i}, \:hit)}{m * k}$\\
            }

            \For{$l$ \textbf{in} $L \neq label\left(r\right)$}{
                \For{$miss$ \textbf{in} $M_{l}$}{
                    $W_a \gets W_a +
                    {\frac{P(l)}{1-P\left(label\left(R_{i}\right)\right)}} * {\frac{diff(a,R_{i},\:miss)}{m * k}}$\\
                }
            }
        }
    }
    \KwRet{$W$}
\end{algorithm}

The hyperparameter $k$ is defined by the user and controls the locality of the estimates \cite{robnik_theoretical_relief_analysis:2003}. For most purposes it can be set to 10 \cite{kononenko_relieff:1994}.

\subsection{Decision Tree}
\label{subsec:decision_tree}
Decision Tree is a hierarchical, rule-based model that recursively splits the feature space into subdomains until only single-label regions remain \cite{suthaharanmachine:2016}. 
It consists of branch nodes (multi-label) and leaf nodes (single-label). 
Predictions follow a path through branch nodes based on input features until reaching a leaf. 
Common splitting criteria include Information Gain and Gini Index \cite{myles_introduction_decision_tree:2004}.
In Decision Tree models, feature relevance can be computed as the (normalized) total reduction of the criterion brought by that feature, also known as Gini importance.

\subsection{Random Forest}
\label{subsec:random_forest}
Random Forest is an ensemble of Decision Trees designed to reduce overfitting by combining multiple models. 
Each tree votes on the label for a given input, improving accuracy but reducing interpretability \cite{breimanrandomforest:2001}.
Feature relevance is computed by averaging the importance scores from all trees, yielding a single feature weight vector \cite{barbieri_analysis_feature_selection:2024}.

\subsection{Least absolute shrinkage and selection operator (LASSO)}
\label{subsec:lasso}
LASSO \cite{tibshirani_selection_lasso:1996} is a powerful embedded method used for regularization and feature weighting. 
It penalizes the coefficients of regression variables, shrinking some of them to zero through $l_1$ regularization.
For feature selection, the features with non-zero coefficients are selected to be part of the model.
Consider the linear regression model in \autoref{eq:linear_regression}, where $x_1, x_2, \ldots, x_p$ are the values of the $p$ features for an input, $y$ is the output, $\beta_1, \beta_2, \ldots, \beta_p$ are the coefficients, and $\epsilon$ is the error.
Given an input matrix $X_{n \times p}$, in which each row denotes a vector of $p$ features for an input $n$, and a vector $Y_n$, in which each value represents the correct output for each input $n$. 
Then, for a given input with features $x_1, x_2, \ldots, x_p$, the estimated response $\hat{y}$ is predicted by \autoref{eq:linear_regression_prediction}.

\begin{equation}
    \label{eq:linear_regression}
    y = \beta_1 x_1 + \beta_2 x_2 + \cdots + \beta_p x_p + \epsilon
\end{equation}

\begin{equation}
    \label{eq:linear_regression_prediction}
    \hat{y} = \hat{\beta}_1 x_1 + \hat{\beta}_2 x_2 + \cdots + \hat{\beta}_p x_p.
\end{equation}

The ordinary least squares estimate, $\hat{\beta} = \{ \hat{\beta}_1, \hat{\beta}_2, \ldots, \hat{\beta}_p \}$, is obtained by minimizing the residual sum of squares, as formulated in \autoref{eq:linear_regression_optimization}.

\begin{equation}
    \label{eq:linear_regression_optimization}
    \hat{\beta} = \underset{\beta}{\operatorname{argmin}} \sum_{i=1}^n \left( y_i - \sum_{j=1}^p \beta_j x_{ij} \right)^2.
\end{equation}

The LASSO algorithm introduces an extra penalty into \autoref{eq:linear_regression_optimization}, generating \autoref{eq:linear_regression_optimization_with_lasso}, where $\lambda \sum_{j=1}^{p} |\beta_j|$ is called the LASSO penalty, also known as $l_1$ regularization, and $\lambda$ is a non-negative hyperparameter.
The regularization causes the $\hat{\beta}$ values to continuously shrink towards zero as the parameter $\lambda$ increases. The coefficients $\hat{\beta} = \{ \hat{\beta}_1, \hat{\beta}_2, \ldots, \hat{\beta}_p \}$ shrink exactly to zero if $\lambda$ is sufficiently large, which implies that all variables are eliminated \cite{tibshirani_selection_lasso:1996, sun_design_lasso:2016}.

\begin{equation}
    \label{eq:linear_regression_optimization_with_lasso}
    \hat{\beta} = \underset{\beta}{\operatorname{argmin}} \sum_{i=1}^n \left( y_i - \sum_{j=1}^p \beta_j x_{ij} \right)^2 + \lambda \sum_{j=1}^p |\beta_j|
\end{equation}

There are many advantages in using LASSO. 
It provides good prediction accuracy by shrinking and removing coefficients, which can reduce variance without a substantial increase in bias, being especially useful in ``large p, small n'' problems. 
LASSO also enhances model interpretability by eliminating irrelevant features that are not associated with the response variable, thereby reducing overfitting.

%%%%%%%%%%%%%%%%%%%%%%%% SECTION %%%%%%%%%%%%%%%%%%%%%%%%%%%%%%
\section{Feature weighting evaluation} 
\label{sec:feature_selection_metrics}
Given an arbitrary dataset and a feature weighting algorithm, evaluating the quality of the generated feature relevance is a complex task.
To address this challenge, we propose different evaluation methods. 
First, we utilize synthetic datasets to assess the ability of feature weighting algorithms to identify truly relevant features (\ref{subsection:informative_features}).
Second, we employ feature erasure (\ref{subsection:occlusion}) in conjunction with classification metrics to evaluate the impact of most relevant features, ordering by weight, on the model performance. 
Third, we assess the performance of new models trained on subsets of selected features with and without using the feature weights (\ref{subsection:selection_performance_new_models}). 
Fourth, we utilize weighted t-SNE (\ref{subsection:t_sne}) to visualize how the feature weights can differentiate between different labels compared with the standard t-SNE.
Finally, we measure and compare the stability of different feature weighting algorithms (\ref{subsection:stability}).

%%% SUBSECTION
\subsection{Informative features} \label{subsection:informative_features}
Feature weighting algorithms are particularly valuable for problems with a high number of features and a limited number of samples which makes difficult to obtain good classification performance and interpretability on most models.
As we lack complete control over the complex interactions and correlations between features and labels in high-dimensional real-world datasets, we use synthetic datasets where we can explicitly define the relevant features for each label.
Given this precise knowledge of feature relevance, we can compute two key metrics:
\begin{itemize}
    \label{list:synthetic_dataset_metrics}
    \item \textbf{Percentage of informative features selected (PIFS)}: Measures the proportion of informative features included in a feature subset. It is calculated by dividing the number of informative features present in the subset by the total number of informative features available in the dataset. A score of 1 indicates that all informative features are present in the subset, while a score of 0 indicates that none of the informative features are present.
    \item \textbf{Percentage of selected features that are informative (PSFI)}: Evaluates the proportion of informative features within a given feature subset. It is calculated by dividing the number of informative features present in the subset by the total number of features in the subset. A score of 1 indicates that all features in the subset are informative, while a score of 0 signifies that none of the features in the subset are informative.
\end{itemize}

\autoref{eq:percentage_of_informative_features_selected} and \ref{eq:percentage_of_selected_features_that_are_informative} outline the calculation of these metrics, where $S_{\textit{selected}}$ is a set of selected features and $S_{\textit{informative}}$
is the set of informative, or relevant, features for a given dataset \cite{barbieri_analysis_feature_selection:2024}. 
For feature weighting algorithms, these metrics can be calculated multiple times by considering different top-$n$ feature subsets defined by a threshold.

\begin{equation}
    \label{eq:percentage_of_informative_features_selected}
    \begin{aligned}
        PIFS = \frac{|S_{\textit{selected}} \cap S_{\textit{informative}}|}{|S_{\textit{informative}}|}
    \end{aligned}
\end{equation}

\begin{equation}
    \label{eq:percentage_of_selected_features_that_are_informative}
    \begin{aligned}
        PSFI = \frac{|S_{\textit{selected}} \cap S_{\textit{informative}}|}{|S_{\textit{selected}}|}
    \end{aligned}
\end{equation}

%%% SUBSECTION
\subsection{Feature erasure} \label{subsection:occlusion}
Feature erasure is a method for assessing the relevance of features subsets in a trained model, adapted by \citet{relevance_aggregation:2021} for tabular data from the perturbation analysis presented by \citet{samek_evaluating_w_heatmap:2016}.
It involves systematically masking specific features and observing the impact on the model's predictive performance. 
This is achieved by setting the values of the target features to zero.
The value zero is chosen because, given the initial data normalization, all input features have a mean equal to zero.
If a feature is crucial for the model's predictions, masking it should lead to a significant decrease in performance. 
Conversely, masking irrelevant features should have a minimal impact or even increase the predictive power.

This technique is primarily applicable to feature weighting algorithms that are integrated into the model training process or applied post-hoc. 
For algorithms that access feature weights independently of a model, the evaluation approach described in \ref{subsection:selection_performance_new_models} is more appropriate.

%%% SUBSECTION
\subsection{Selection performance on new models} \label{subsection:selection_performance_new_models}
Different machine learning predictors, as Decision Trees, SVMs, and Deep Neural Networks can be employed to assess the predictive power of different feature subsets. 
For each feature subset generated by selecting the most relevant features given by a feature weighting algorithm, we train a predictor and evaluate it's predictive performance using the metrics as F1 score, prediction, recall, and accuracy.

%%% SUBSECTION
\subsection{Weighted t-SNE} \label{subsection:t_sne}
t-SNE (t-Distributed Stochastic Neighbor Embedding) is a popular technique for visualizing high-dimensional data in two dimensions, originally proposed by \citet{van_t_sne:2008} as a variation of SNE (Stochastic Neighbor Embedding), introduced by \citet{hinton_sne:2002}, using the Student t-distribution with one degree of freedom.
It calculates the distances between high-dimensional data points, typically using the Euclidean distance (\autoref{eq:euclidean_distance}).

\begin{equation}
    \label{eq:euclidean_distance}
    \begin{aligned}
        d(p, q) = \sqrt{\sum_{i=1}^{n} (q_i - p_i)^2}
    \end{aligned}
\end{equation}

Weighted t-SNE, proposed by \citet{relevance_aggregation:2021}, introduces a weighted Euclidean distance (\autoref{eq:weighted_euclidean_distance}), adapted from \citet{carroll_weighted_euclidean_distance:1970}, to enhance the visualization based on features weights given by a feature weighting algorithm. 
Those weights range from 0 (less relevant) to 1 (more relevant). 
When scaling each dimension by its weight, dimensions with higher weights account for a greater portion of the distance between points and thus have more influence on their final positions in the visualization.
Another significant advantage of weighted t-SNE is its ability to create visualizations based on label-specific feature weights, allowing us to use the weights from a specific label to determine if they can better distinguish it from the others.

\begin{equation}
    \label{eq:weighted_euclidean_distance}
    \begin{aligned}
        d_w(p, q) = \sqrt{\sum_{i=1}^{n} (w_i(q_i - p_i))^2}
    \end{aligned}
\end{equation}

%%% SUBSECTION
\subsection{Stability} \label{subsection:stability}
Feature weighting is not only crucial for improving model performance but also plays a significant role in research and interpretability. 
For instance, as an example, in Bioinformatics, it can help identify biomarkers associated with specific diseases. 
In such domains, the stability of a feature weighting algorithm is an important metric, as unstable algorithms can lead to non-replicable results. 
The issue of non-replicable gene selections has been a major concern for many researchers \cite{awada_review_feature_selection_stability:2012}.
And several factors can contribute to unstable feature weighting, some of them are \cite{he_stable_feature_selection:2010}:
\begin{itemize}
    \label{list:unstable_feature_selection_reasons}
    \item \textbf{Multiple informative features}: The presence of multiple highly correlated features can make it difficult for algorithms to consistently select the most relevant ones.
    \item \textbf{High-dimensional, low-sample data}: In such scenarios, feature weighting algorithms may be prone to overfitting and unstable results.
\end{itemize}

Stability metrics assess the consistency of feature selection and weighting results across different data samples. 
Data perturbation, such as adding, removing, or resampling instances, can be used to simulate variations in the training data \cite{barbieri_analysis_feature_selection:2024}.
K-Fold Cross-Validation is a widely used technique to evaluate model performance that can be used to measure feature weighting stability \cite{awada_review_feature_selection_stability:2012}. 
In this context it involves:
\begin{itemize}
    \label{list:cross_validation_to_measure_stability}
    \item \textbf{Splitting the dataset}: Divide the dataset into $k$ multiple stratified folds.
    \item \textbf{Execute feature selection}: Execute the feature weighting method using $k-1$ folds and persist the result which can be a simple feature subset, a rank or a weighted features set.
    \item \textbf{Repeating}: Repeat the process $k$ times using all available folds $k-1$ times each.
\end{itemize}

By comparing different subsets of features based on their weights from each execution, we can assess the stability of the method using different metrics, considering different perspectives over the output. 
For this work, we employ Jaccard, Spearman, and Pearson similarity measures:
\begin{itemize}
    \label{list:stability_metrics}
    \item \textbf{Jaccard index}: Measures the similarity between two sets, in our case the top $n$ weighted features. It ranges from 0 (no overlap) to 1 (identical sets), being defined by:
    \begin{equation}
        \label{eq:jaccard}
        \begin{aligned}
            J(A, B) = \frac{|A \cap B|}{|A \cup B|},
        \end{aligned}
    \end{equation}
    where $A$ and $B$ are the two subsets being analyzed.
    \item \textbf{Spearman’s rank correlation coefficient}: Measures the relationship between two ranked lists. It ranges from -1 (linearly inverse correlated) to 1 (linearly correlated), where 0 indicates no correlation. It is defined by:
    \begin{equation}
        \label{eq:spearman}
        \begin{aligned}
            r_s(p, q) = 1 - \frac{6\sum_{i=1}^{n}(p_i - q_i)^2}{n(n^2 - 1)},
        \end{aligned}
    \end{equation}
    where $p$ and $q$ are two rankings presented as permutation lists and $n$ is the number of element in the rankings.
    \item \textbf{Pearson correlation coefficient}: Measures the linear correlation between two sets of numerical values, being used to measure the correlation between feature weights. 
    It's boundaries are the same of Spearman’s rank correlation coefficient, being defined by:
    \begin{equation}
        \label{eq:pearson}
        \begin{aligned}
            r_{xy} = \frac{\sum_{i=1}^{n}(x_i - \bar{x})(y_i - \bar{y})}{\sqrt{\sum_{i=1}^{n}(x_i - \bar{x})^2}\sqrt{\sum_{i=1}^{n}(y_i - \bar{y})^2}}
        \end{aligned}
    \end{equation}
    where $x$ and $y$ represent each subset, $\bar{x}$ and $\bar{y}$ represent the mean value of each subset and $n$ represents the subset size.
\end{itemize}

%%%% SECTION %%%%
\section{Complementary Experiments} 
\label{sec:complementary_experiments}

In this appendix, we introduce complementary experiments that further support our findings.
All neural networks shared a common architecture of four hidden layers (100, 200, 100, 100), with $10\%$ Dropout applied between layers. 
The models were optimized using AdamW with a learning rate of $0.001$ and evaluated using Stratified K-Fold cross-validation ($K=11$). 
The test split was set to $20\%$ for synthetic datasets and $15\%$ for microarray datasets. 
Regarding regularization, FSL and its extensions used $\lambda=0.001$ for synthetic and $\lambda=0.0025$ for microarray datasets, while Lasso used $\lambda=0.01$.
Batch sizes and training epochs varied by dataset: synthetic datasets used a batch size of 32, while Prostate and Breast used 16, and Colorectal and Liver used 8. 
Training duration was 2000 epochs for XOR, 800 for Prostate, 400 for Breast, and 300 for the remaining datasets. 
Finally, for interpretability baselines, DeepSHAP and LIME utilized 1000 background samples for synthetic data (100 for SAGE). 
However, due to data dimensionality constraints, all methods were limited to 50 background samples for the microarray datasets.
More details are available in the provided source code.

\begin{figure}[!h]
    \caption[
    Figure composed of eight images, forming 4 pairs.
    Each pair is composed by the pure t-SNE of a synthetic dataset and the ``perfect'' t-SNE.
    The first pair presents the application of t-SNE to the XOR dataset.
    Samples from two classes are projected into a 2D space, each represented by distinct shapes and colors. 
    In the original projection, the two classes heavily overlap in the center of the plot, forming a circular pattern with no clear separation. 
    When incorporating the perfect weights, the projection changes to four small circular clusters located at the plot’s borders: the top-left and bottom-right clusters correspond to class 0, while the top-right and bottom-left clusters correspond to class 1.
    The second pair shows the application of t-SNE to the SynthA dataset.
    In the original projection, the three classes largely overlap in the center of the plot, forming a circular pattern with no evident separation. 
    When incorporating the perfect weights, the circular structure remains, but it becomes divided into  five distinct regions: the bottom-center corresponds to class 1, the top-left and top-right to class 0, and the top-center together with the bottom-left to class 2.
    The third pair illustrates the application of t-SNE to the SynthB dataset.
    In the original projection, the three classes largely overlap at the center of the plot, forming a circular pattern with no clear separation. 
    Nevertheless, class 1 samples are mostly located on the right, class 0 in the top-left and bottom-right, and class 2 spans from the bottom-left to the top-right, passing through the center. 
    When incorporating the perfect weights, the circular structure is distorted, forming three distinct regions: the top-left corresponds to class 0, the top-right to class 1, and the bottom region to class 2.
    The fourth pair illustrates the application of t-SNE to the SynthC dataset.
    In the original projection, the three classes largely overlap at the center of the plot, forming a circular pattern with no clear separation. 
    Nonetheless, class 1 samples are mostly concentrated at the top, class 0 in the bottom-left, and class 2 extends from the center to the right. 
    When incorporating the perfect weights, the circular structure is distorted, forming three distinct regions: the left corresponds to class 1, the center to class 2, and the right to class 0.
    ]
    {\textbf{t-SNE and ``perfect'' weighted t-SNE of each synthetic datasets.} 
    The ``perfect'' weighted t-SNE is achieved by assigning a weight of 1 for all informative features, while irrelevant features were assigned a weight of 0.
    }
     \centering
     \label{fig:synthetic_tsne}
     \begin{subfigure}{0.35\textwidth}
         \centering
         \includegraphics[width=\textwidth]{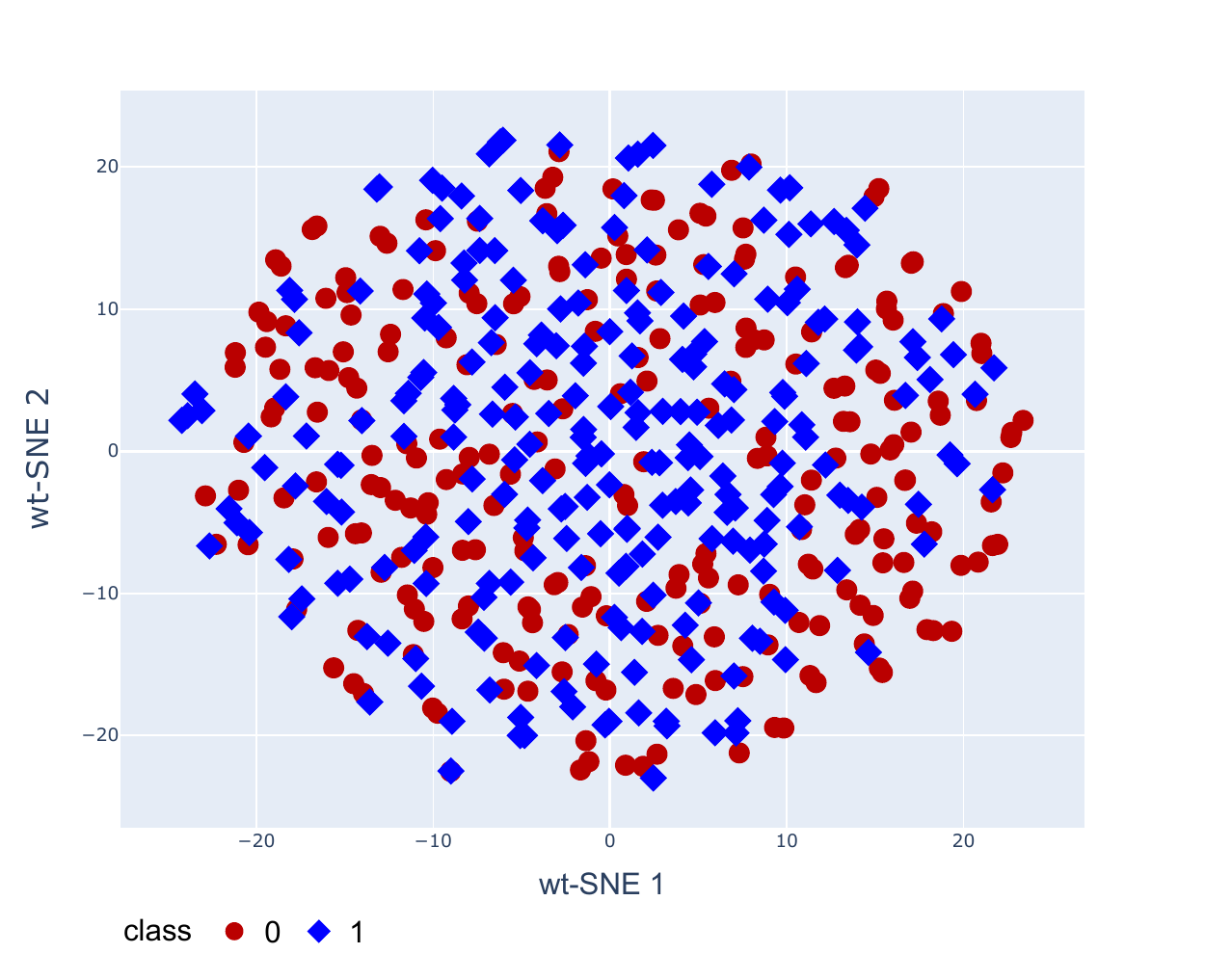}
         \caption{XOR t-SNE}
         \label{fig:breast_no_weight}
     \end{subfigure}
     \hfill
     \begin{subfigure}{0.35\textwidth}
         \centering
         \includegraphics[width=\textwidth]{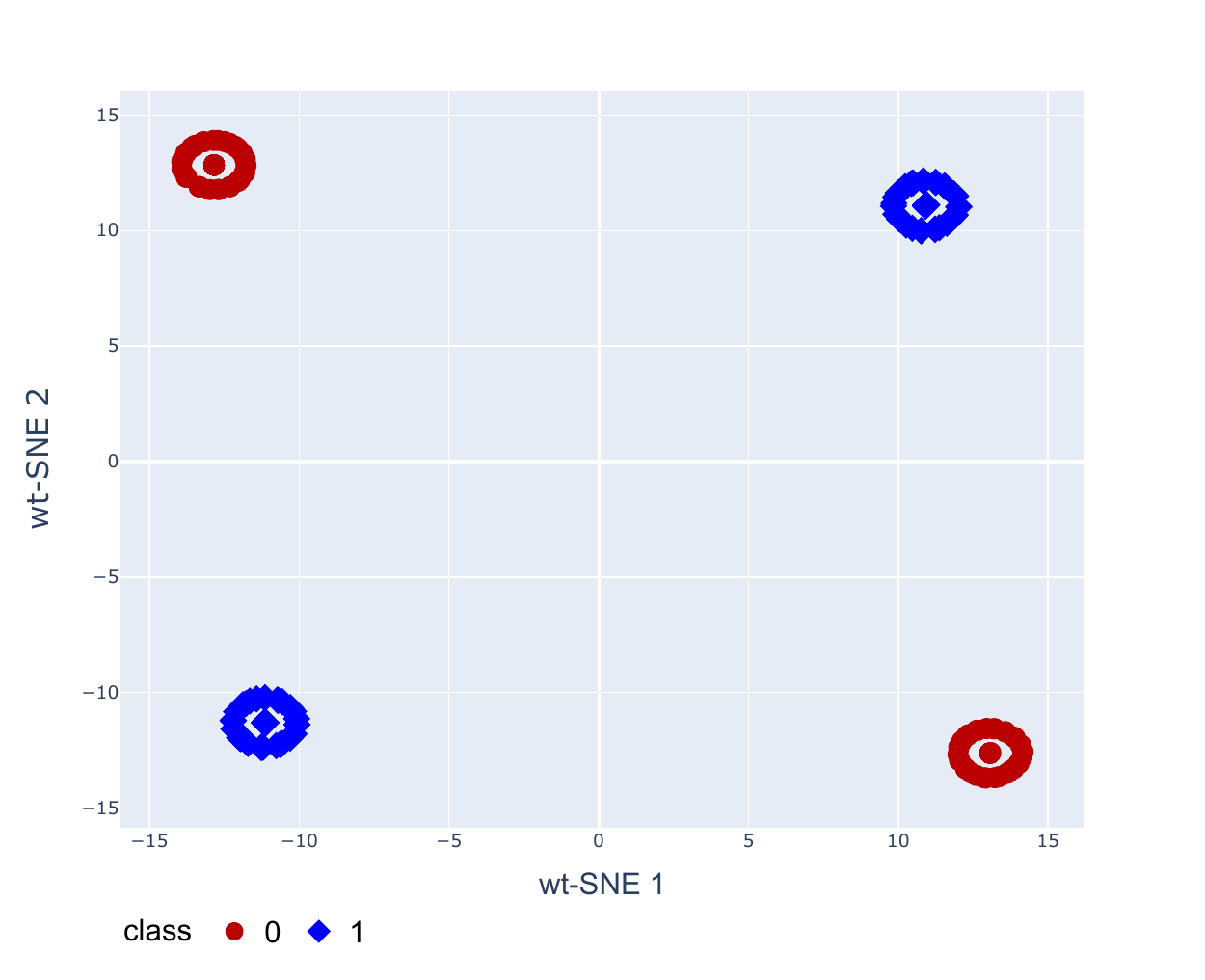}
         \caption{XOR ``perfect'' weighted t-SNE}
         \label{fig:liver_no_weight}
     \end{subfigure}
     \begin{subfigure}{0.35\textwidth}
         \centering
         \includegraphics[width=\textwidth]{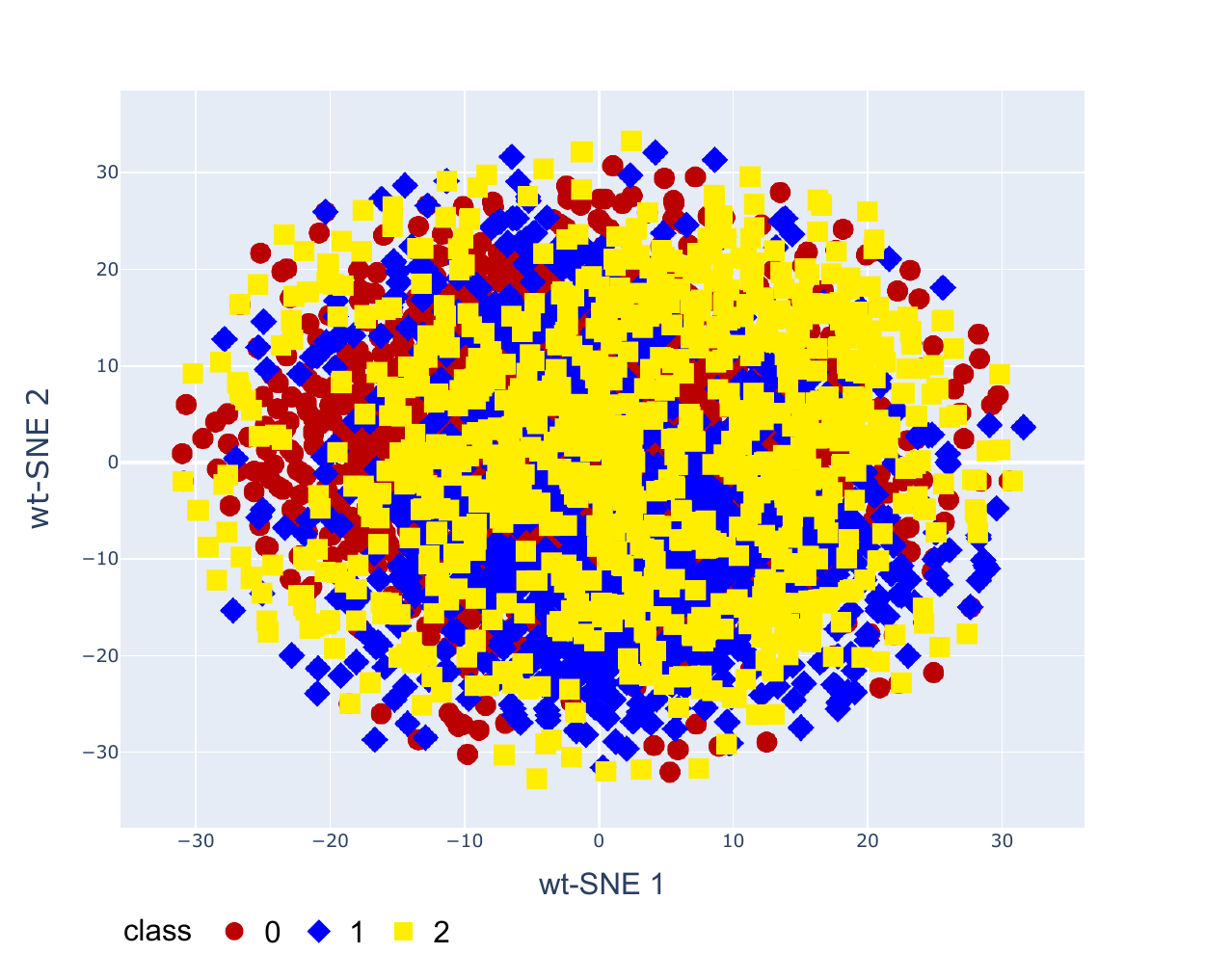}
         \caption{SynthA t-SNE visualization}
         \label{fig:prostate_no_weight}
     \end{subfigure}
     \hfill
     \begin{subfigure}{0.35\textwidth}
         \centering
         \includegraphics[width=\textwidth]{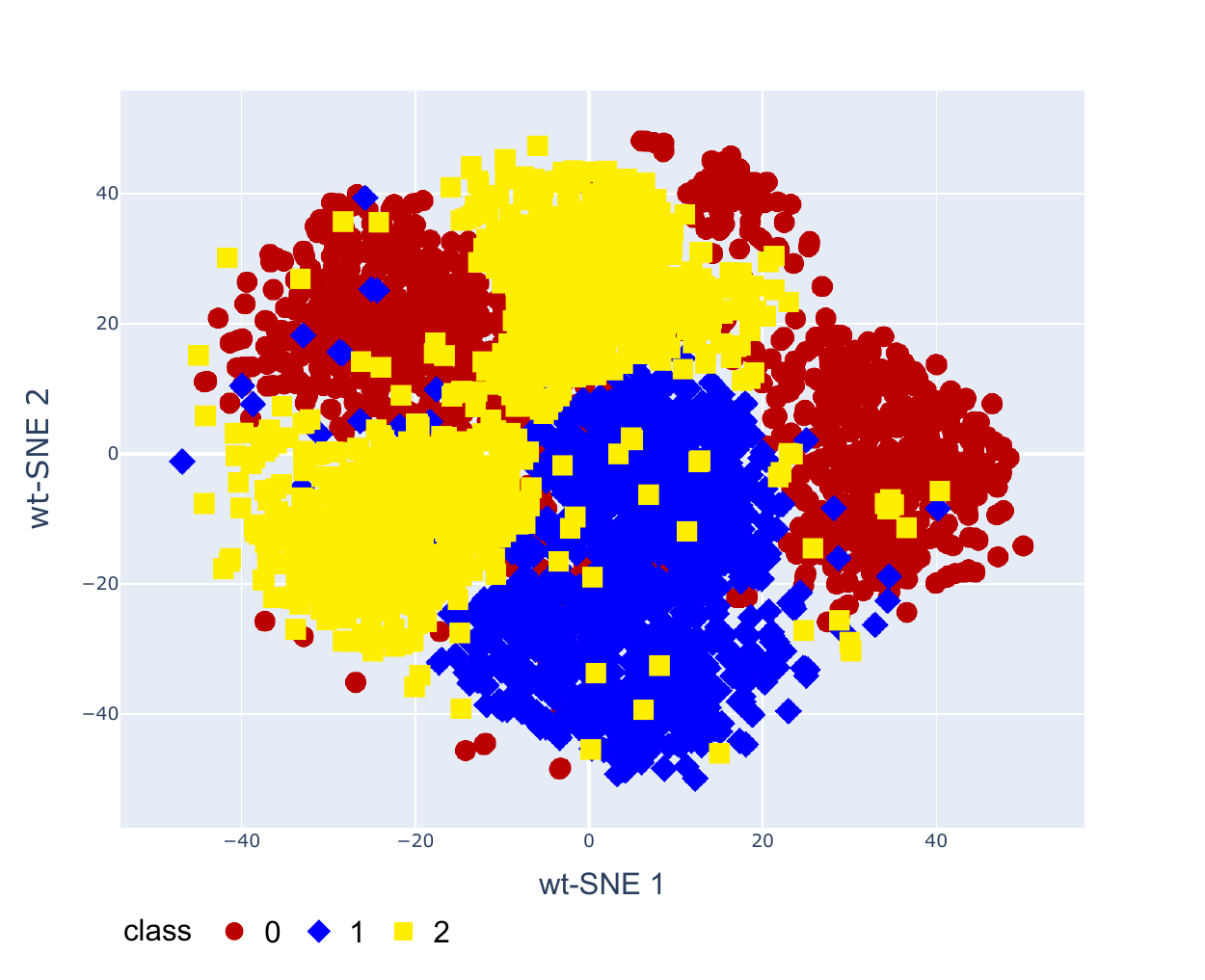}
         \caption{SynthA ``perfect'' weighted t-SNE visualization}
         \label{fig:colorectal_no_weight}
     \end{subfigure}
     \begin{subfigure}{0.35\textwidth}
         \centering
         \includegraphics[width=\textwidth]{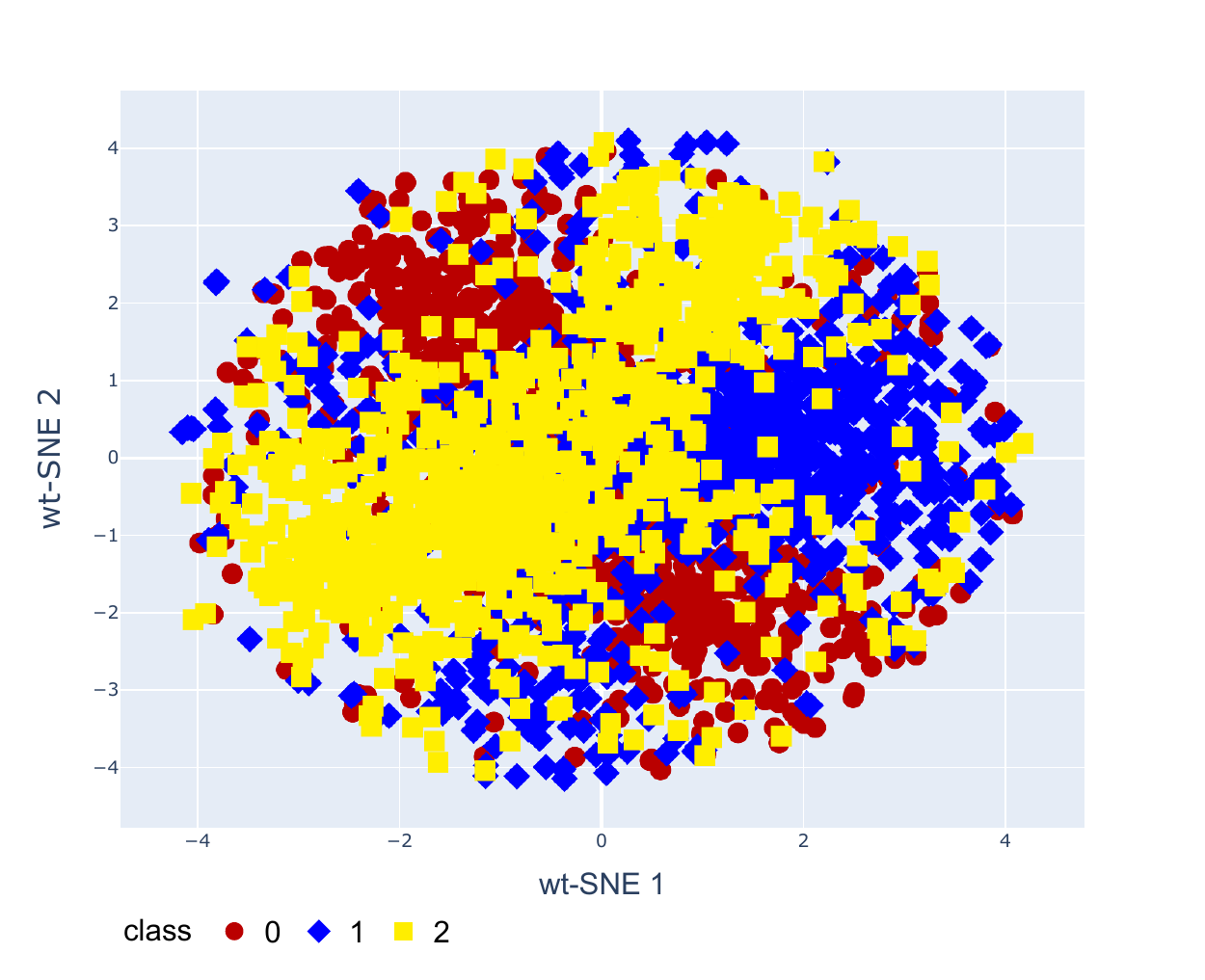}
         \caption{SynthB t-SNE visualization}
         \label{fig:prostate_no_weight}
     \end{subfigure}
     \hfill
     \begin{subfigure}{0.35\textwidth}
         \centering
         \includegraphics[width=\textwidth]{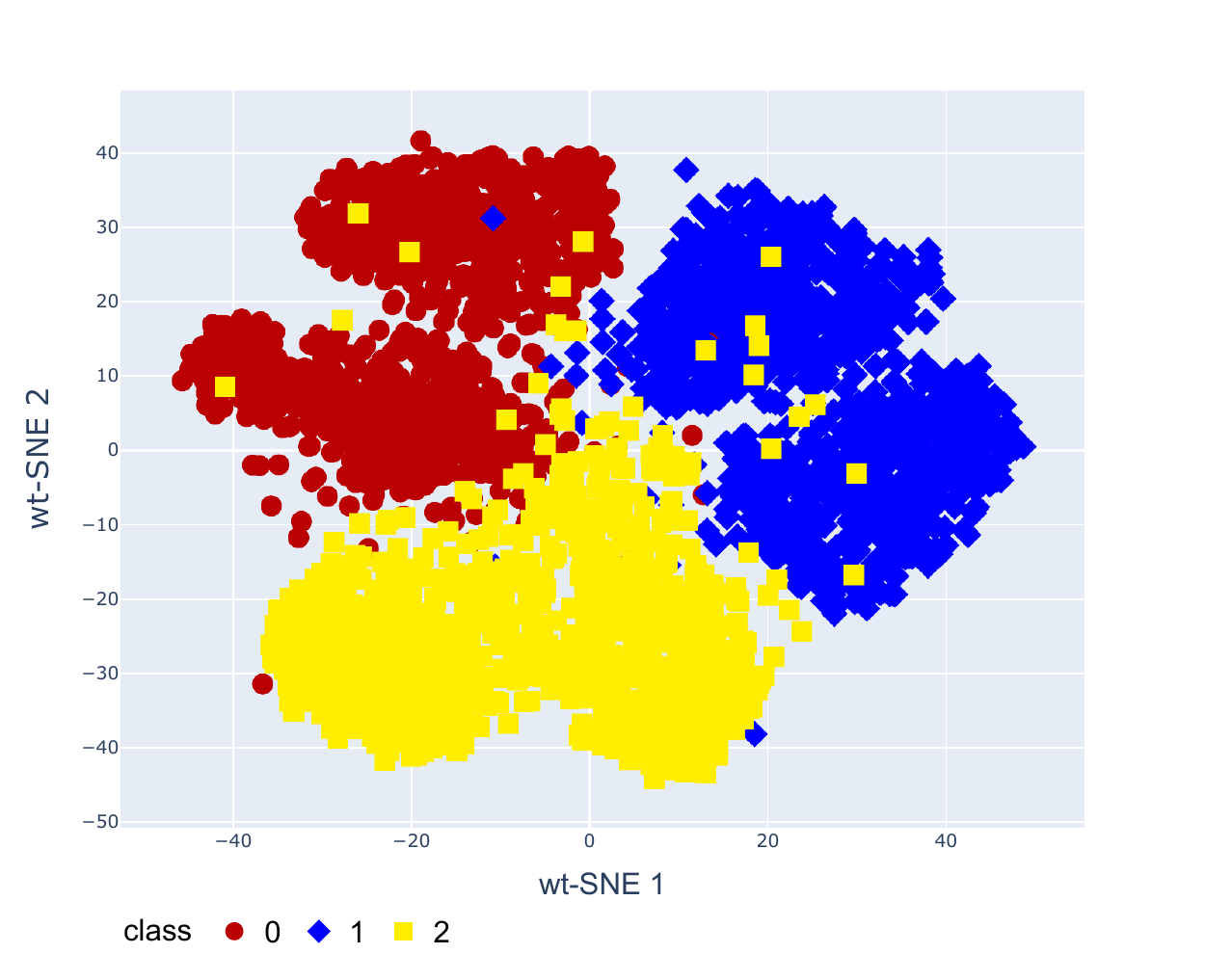}
         \caption{SynthB ``perfect'' weighted t-SNE visualization}
         \label{fig:colorectal_no_weight}
     \end{subfigure}
     \begin{subfigure}{0.35\textwidth}
         \centering
         \includegraphics[width=\textwidth]{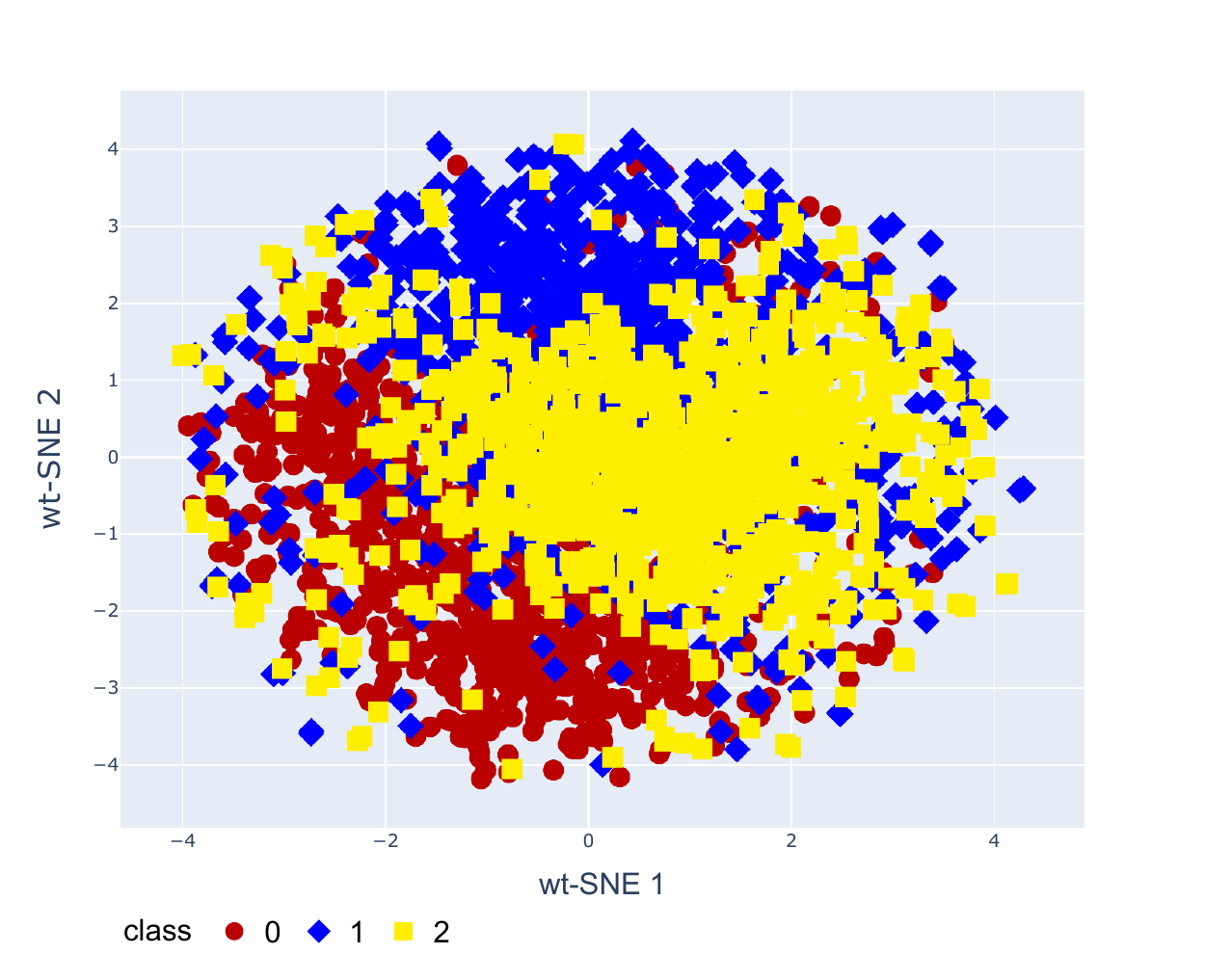}
         \caption{SynthC t-SNE visualization}
         \label{fig:prostate_no_weight}
     \end{subfigure}
     \hfill
     \begin{subfigure}{0.35\textwidth}
         \centering
         \includegraphics[width=\textwidth]{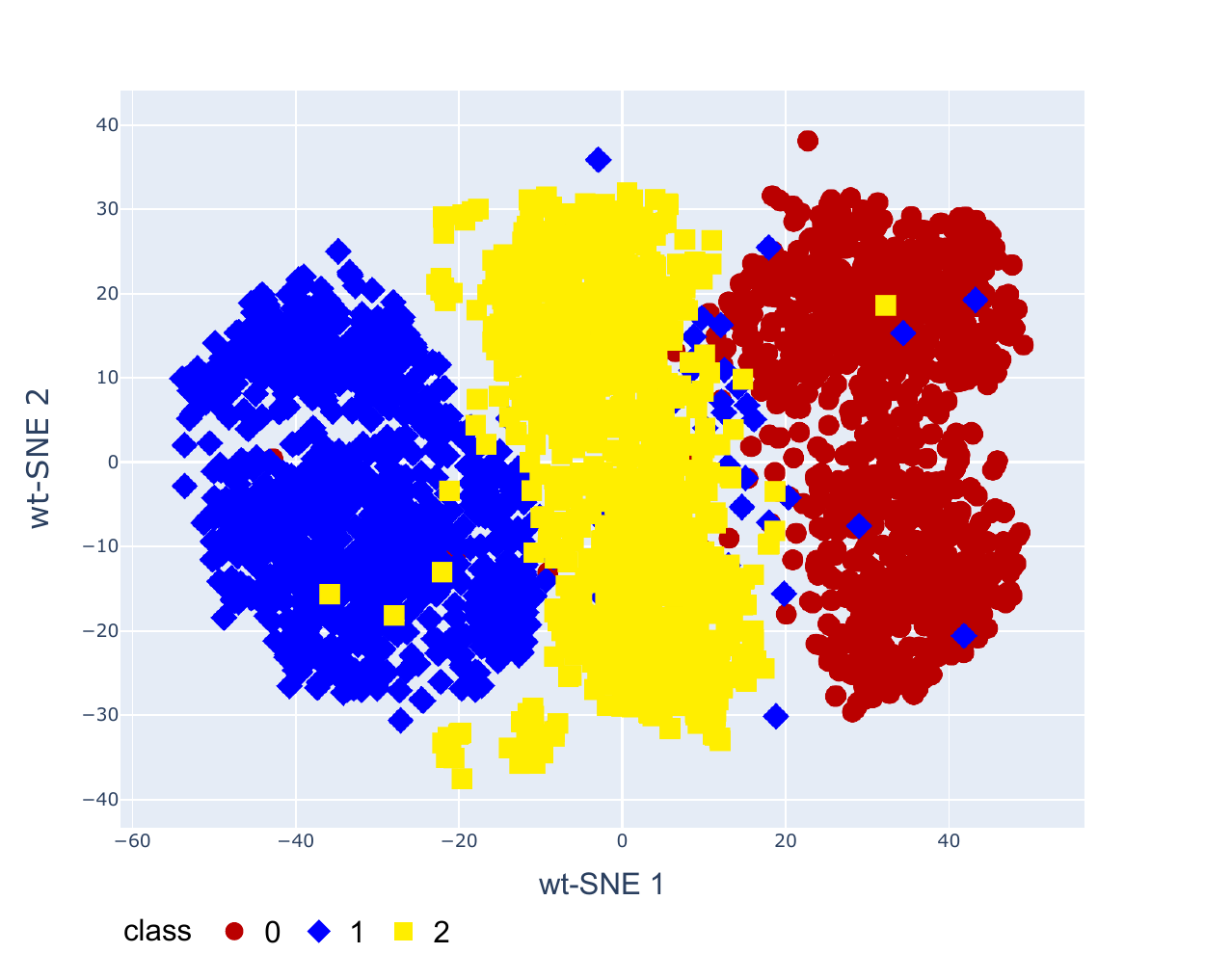}
         \caption{SynthC ``perfect'' weighted t-SNE visualization}
         \label{fig:colorectal_no_weight}
     \end{subfigure}
\end{figure}

\begin{figure}[!h]
    \caption[
    Figure composed of four images.
    The first one present the t-SNE applied to the Liver dataset.
    Samples from two classes are projected into a 2D space, each represented by distinct shapes and colors. 
    However, the two classes extensively overlap across the entire plot, with no clear separation observable in the 2D projection.
    The second one present T-SNE applied to the Prostate dataset.
    Samples from two classes are projected into a 2D space, each represented by distinct shapes and colors. 
    However, the two classes extensively overlap across the entire plot, with no clear separation observable in the 2D projection.
    The third one present the t-SNE applied to the Colorectal dataset.
    Samples from four classes are projected into a 2D space, each represented by distinct shapes and colors. 
    Class 0 forms a well-defined cluster, clearly separated from the others. 
    In contrast, classes 1, 2, and 3 show considerable overlap, extending from the top to the central region of the plot.
    The last one present the t-SNE applied to the Breast dataset.
    Samples from six classes are projected into a 2D space, represented with distinct shapes and colors.
    Classes 2 and 3 are clearly separated from the others, forming well-defined clusters. 
    In contrast, classes 0 and 1 overlap in the top-left region, while classes 4 and 5 exhibit overlap in the central area.
    ]
    {\textbf{t-SNE of CuMiDa datasets.}
    Analysis of each CuMiDa datasets used on this work. 
    A weighted t-SNE approach was not implemented due to the lack of prior knowledge about the relevance of each feature.
    }.
     \centering
     \label{fig:cumida_tsne}
     \begin{subfigure}[b]{0.45\textwidth}
         \centering
         \includegraphics[width=\textwidth]{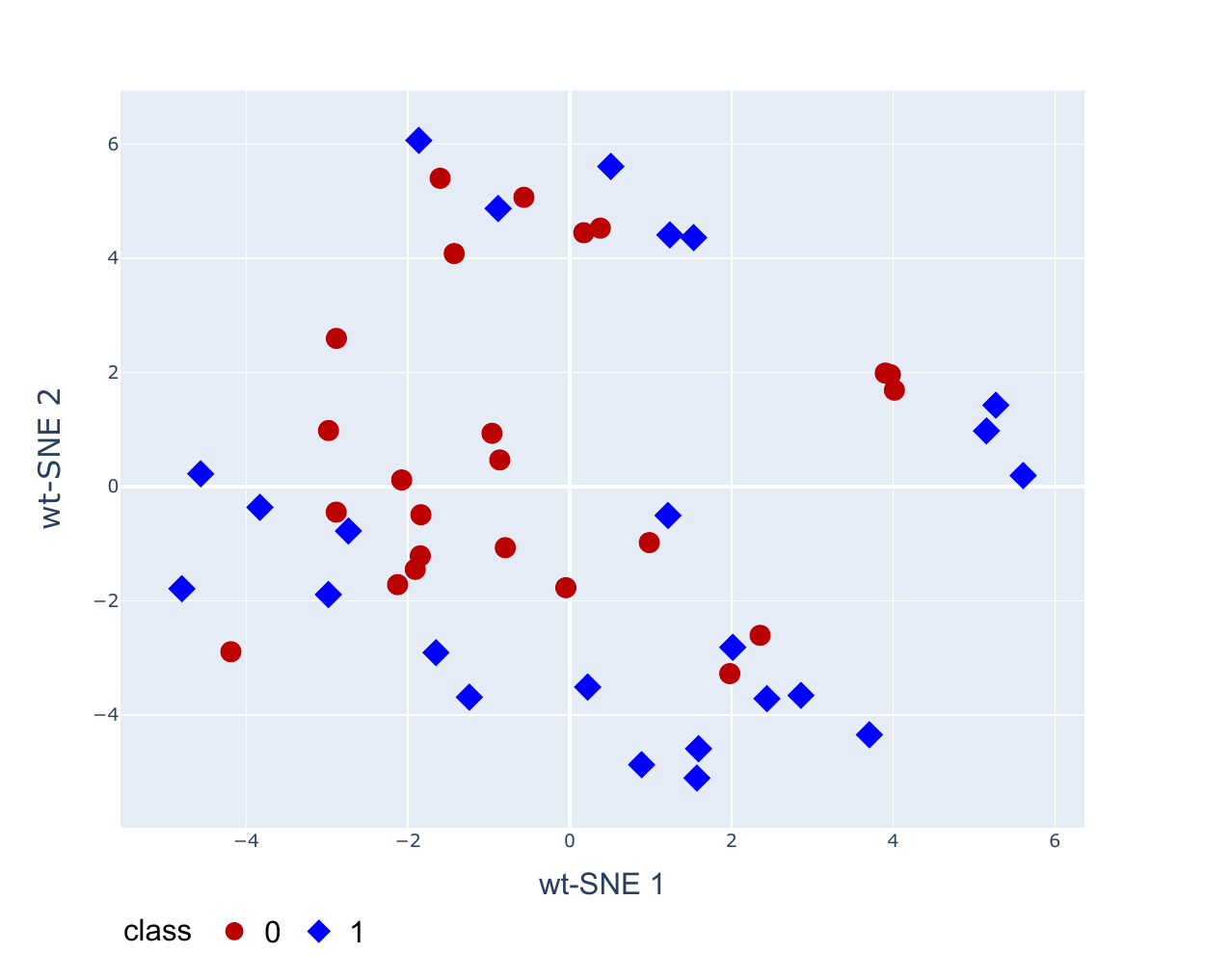}
         \caption{Liver t-SNE visualization}
         \label{fig:liver_no_weight}
     \end{subfigure}
     \hfill
     \begin{subfigure}[b]{0.45\textwidth}
         \centering
         \includegraphics[width=\textwidth]{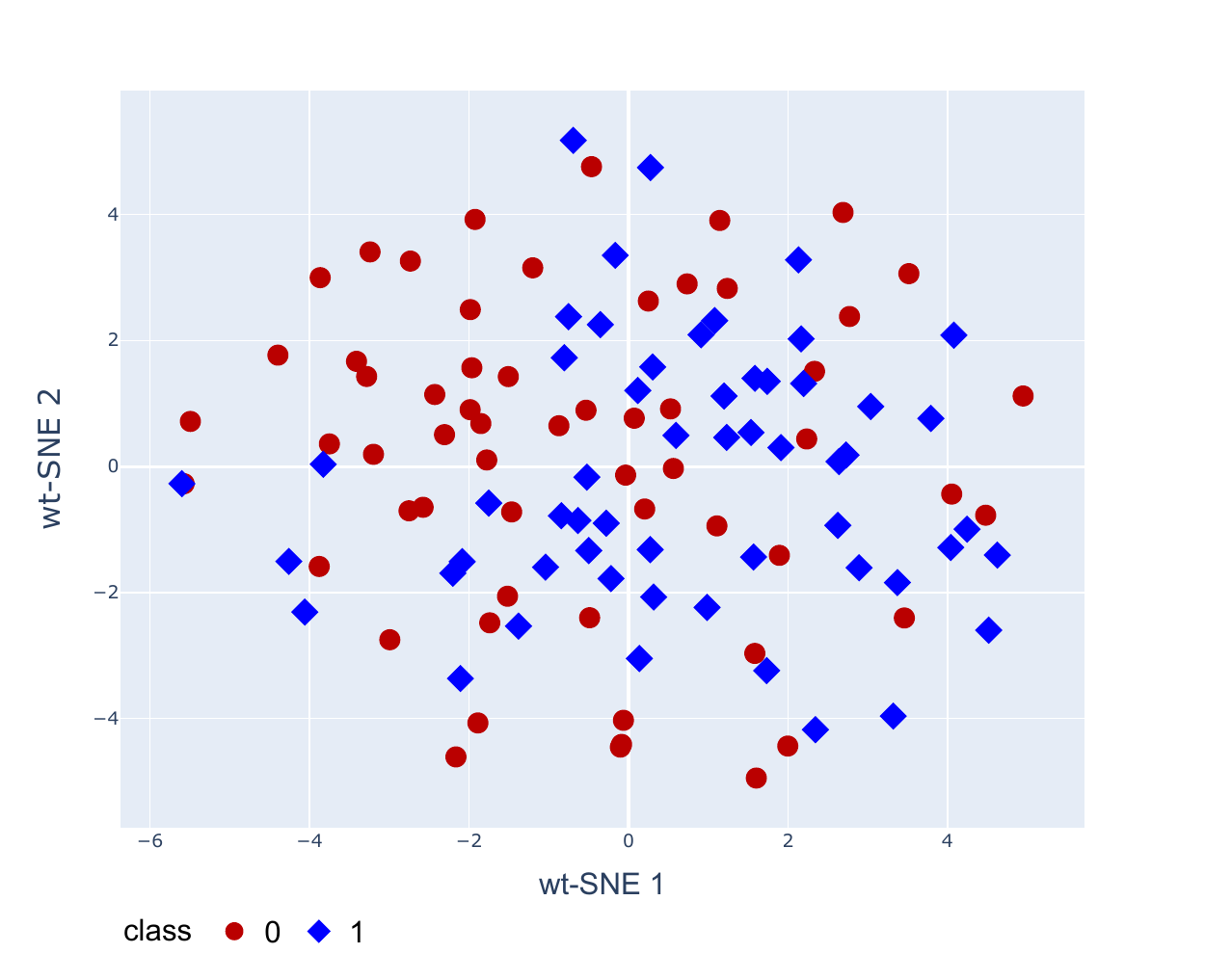}
         \caption{Prostate t-SNE visualization}
         \label{fig:prostate_no_weight}
     \end{subfigure}
     \begin{subfigure}[b]{0.45\textwidth}
         \centering
         \includegraphics[width=\textwidth]{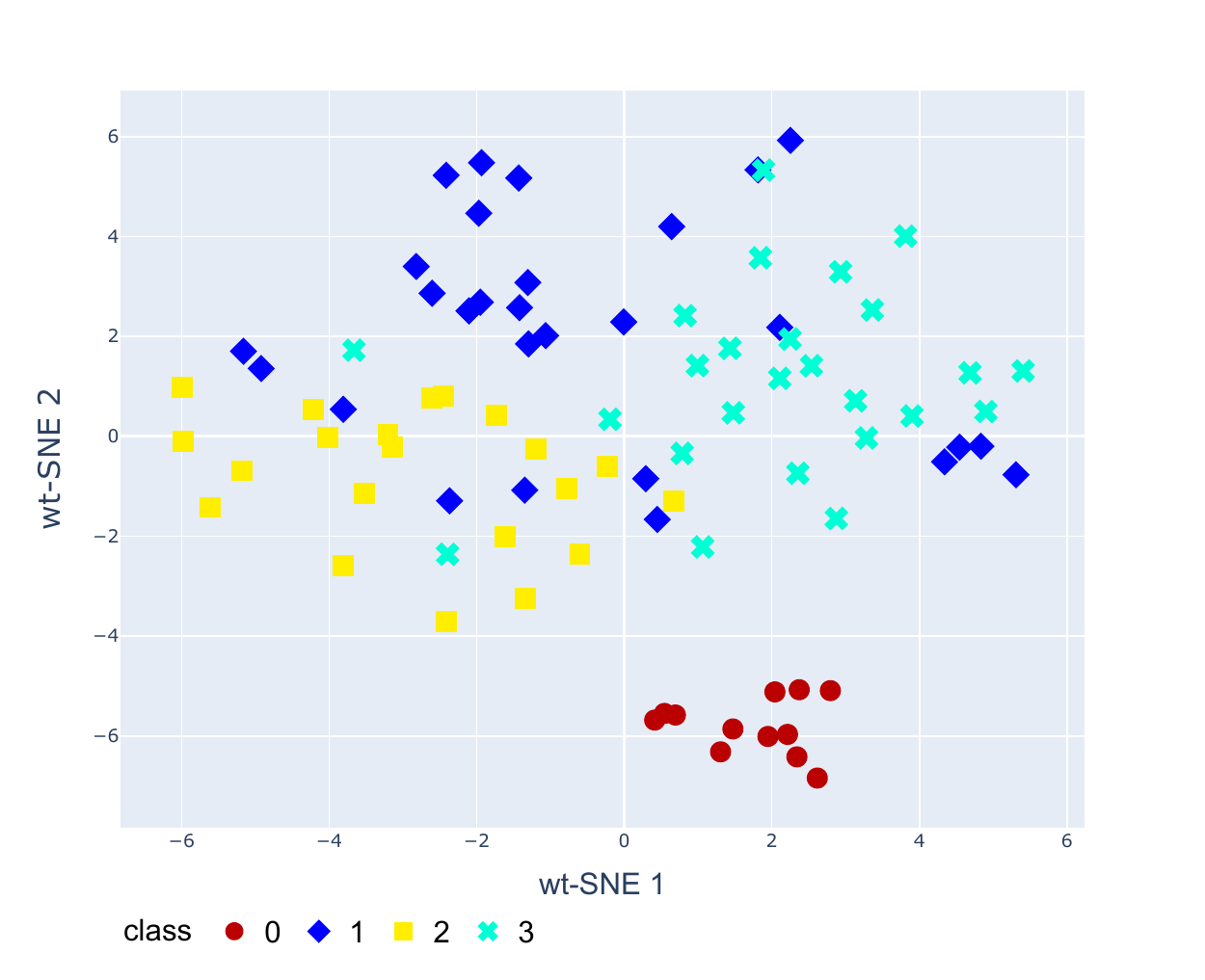}
         \caption{Colorectal t-SNE visualization}
         \label{fig:colorectal_no_weight}
     \end{subfigure}
     \hfill
     \begin{subfigure}[b]{0.45\textwidth}
         \centering
         \includegraphics[width=\textwidth]{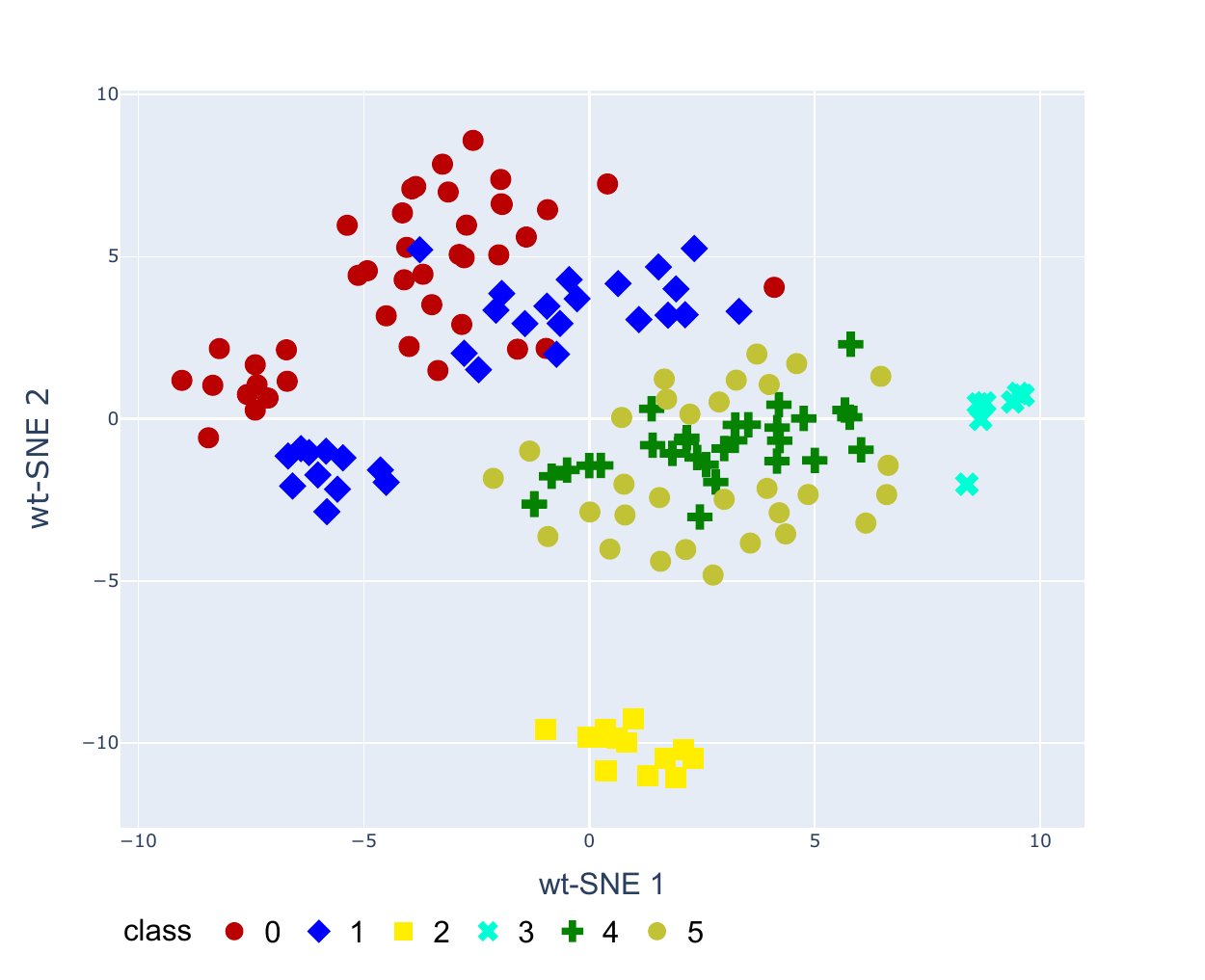}
         \caption{Breast t-SNE visualization}
         \label{fig:breast_no_weight}
     \end{subfigure}
\end{figure}

\autoref{fig:synthetic_tsne} presents the t-SNE and a “perfect’’ weighted t-SNE for each synthetic dataset. The “perfect’’ weighted version assigns a weight of 1.0 to all informative features and 0.0 to all non-informative ones.
\autoref{fig:cumida_tsne} shows the t-SNE visualizations for each microarray dataset. Since the informative features are unknown for these datasets, a “perfect’’ weighted t-SNE cannot be constructed.

\begin{figure}[!h]
    \centering
    \caption[
        The graph presents the evaluation of PIFS and PSFI on the XOR dataset, comparing the original FSL with the proposed extensions.
        Each metric for FSL and its extensions is represented by a dashed bar with distinct colors and spacing. 
        All metrics reach a score of 1.0 for PIFS using only two features, while PSFI remains at 1.0 up to the top three features before gradually decreasing.
    ]
    {\textbf{Evaluation of PIFS and PSFI on the XOR dataset between original FSL and proposed extensions.} 
        All scores are the average over rankings generated by 11 independent execution.
        Vertical axis represents the score constrained from $0$ to $1$ as both measures are percentages.
        Horizontal axis represents the subset size being evaluated.
        All subsets are defined by the top-$n$ features from a ranking based on the feature weights generated by each algorithm.
    }.
    \includegraphics[width=0.7\textwidth]{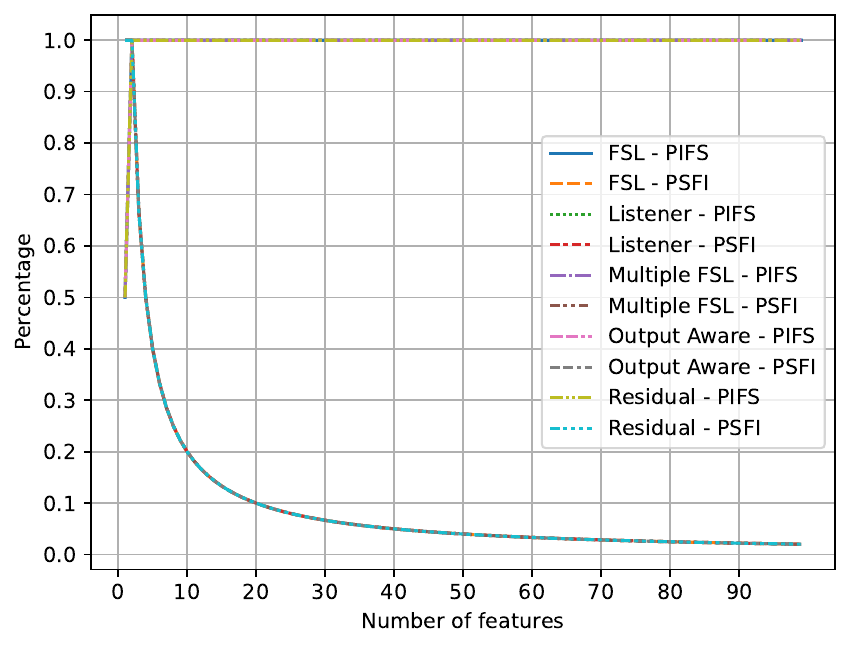}
    \label{fig:comparative_informative_features_graph_on_extensions_for_xor}
\end{figure}

\autoref{fig:comparative_informative_features_graph_on_extensions_for_xor} reports the PIFS and PSFI evaluations on the XOR dataset, comparing the original FSL with the proposed extensions.
%and \autoref{tab:comparative_execution_time_onextensions_for_xor} summarizes the average execution time of the FSL algorithm and its extensions on the XOR dataset.

%\begin{table}[!h]
%\centering
%\caption[Average execution time of FSL algorithm and proposed extensions for XOR dataset.] {\textbf{Average execution time of FSL algorithm and proposed extensions for XOR dataset.} 
%Values are defined in seconds being the average of 11 executions for each method using the same hardware, data partitions and hyper parameters over the same neural network architecture where the only difference was the implementation of the feature selection layer.
%In bold is the highest average.
%Best results with statistical significance ($p < 0.01$) according to the Kruskal-Wallis and Dunn’s test are highlighted in orange.
%}
%\label{tab:comparative_execution_time_onextensions_for_xor}
%\begin{tabular}{lr}
%\hline
% Algorithm    &   Execution time (s) \\
%\hline
% FSL          &       \cellcolor{orange!20} \textbf{19.650$\pm$0.588} \\
% Listener     &        24.199$\pm$1.308 \\
% Residual     &        22.017$\pm$0.045 \\
% Output Aware &        22.415$\pm$1.356 \\
% Multiple FSL &     31.797$\pm$2.075 \\
%\hline
%\end{tabular}
%\end{table}

\begin{figure}[!h]
    \caption[
        Heatmap visualization with the 15 highest weighted features for XOR dataset.
        FSL results are presented in a single columns as it only has the general feature weights.
        Both informative features are present in the top of the chart with score of 1.0 and 0.84. 
        The next noisy feature has the score 0.441 and it continues to reduce, being the last noisy feature with a score of 0.271.
        Listener results are presented in the second figure.
        It contains three columns, being the first one for the class 0 weights, followed by the class 1 weights and by the general feature weights.
        In all cases the two informative features are the first ones with scores above to 0.9 and the noisy feature have weights near to zero being the highest one 0.012.
        Residual results are presented in the third figure.
        It also contains the same three columns.
        In all cases the two informative features are the first ones with scores above to 0.75 and the noisy feature have weights near to zero being the highest score 0.294.     
        Output Aware results are presented in the fourth figure.
        In all tree columns the two informative features are the first ones with scores above to 0.85 and the noisy feature have weights near to zero being all of them lower than 0.001.
        Multiple FSL results are presented in the fifth figure.
        In all tree columns the two informative features are the first ones with scores above to 0.85 and the noisy feature have weights near to zero being the highest one 0.002.
    ]
    {\textbf{Heatmap visualization with the 15 highest weighted features for XOR dataset.}
        Normalized weights for each feature ranked from the highest general weight to the lowest.
        ``General'' represents the average label-specific weights for the proposed extensions Residual, Output Aware and Multiple FSL.
    }
     \centering
     \label{fig:heatmap_on_extensions_for_xor}
     \begin{subfigure}[b]{0.23\textwidth}
         \centering
         \includegraphics[width=\textwidth]{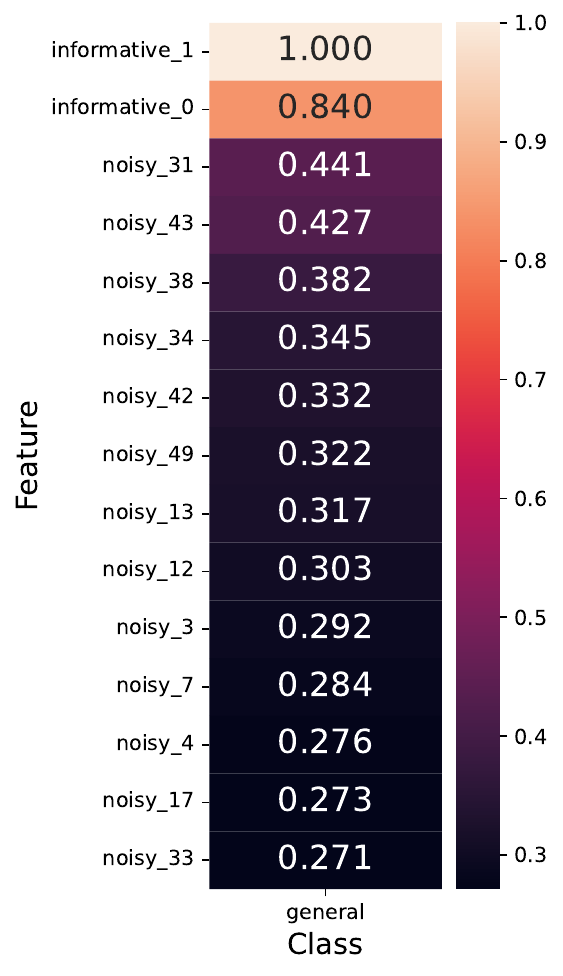}
         \caption{FSL}
     \end{subfigure}
     \hfill
     \begin{subfigure}[b]{0.44\textwidth}
         \centering
         \includegraphics[width=\textwidth]{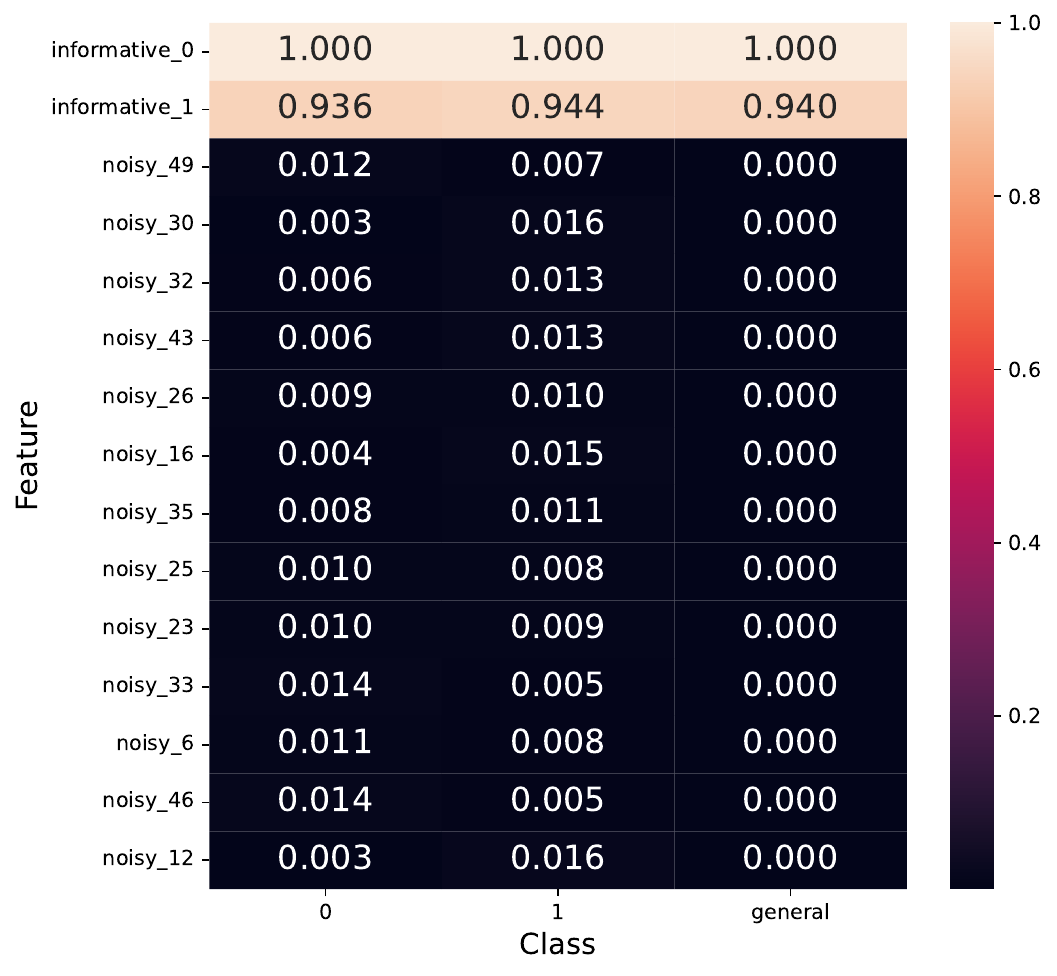}
         \caption{Listener}
     \end{subfigure}
     \begin{subfigure}[b]{0.44\textwidth}
         \centering
         \includegraphics[width=\textwidth]{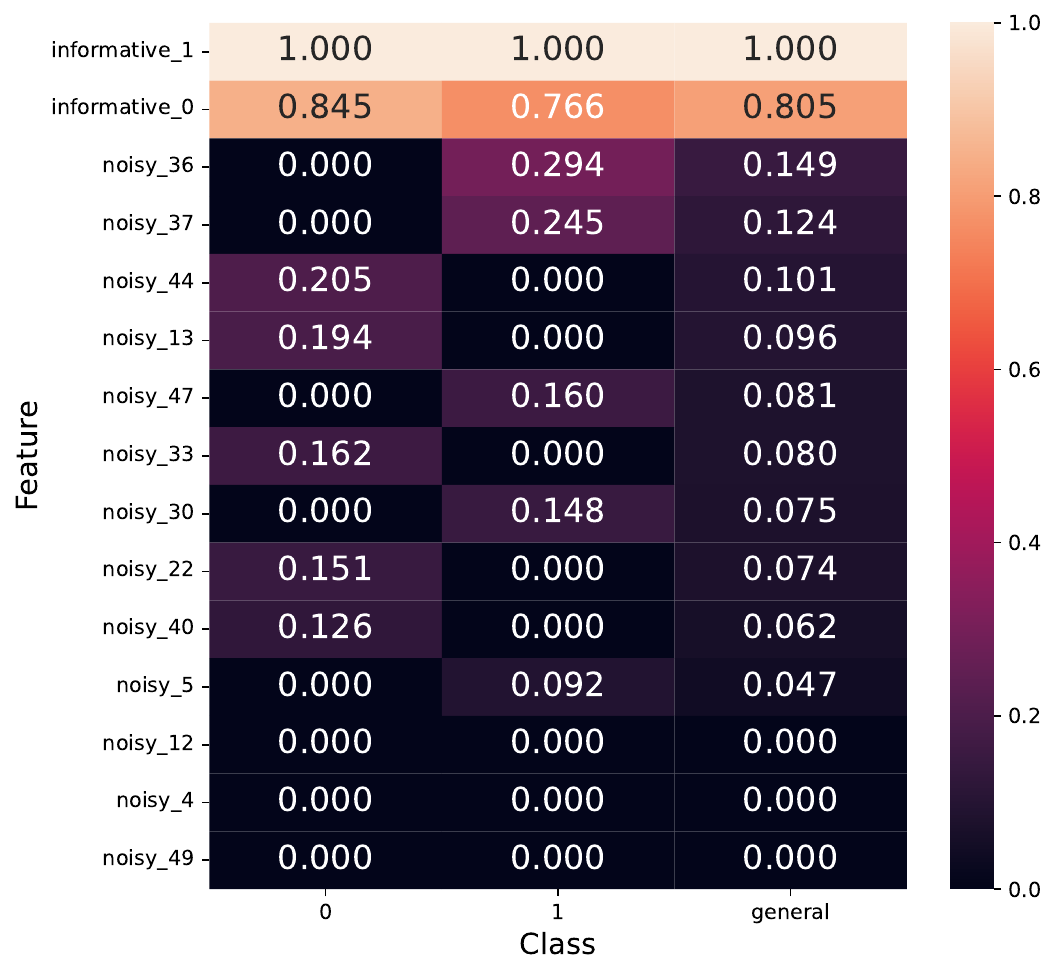}
         \caption{Residual}
     \end{subfigure}
     \hfill
     \begin{subfigure}[b]{0.44\textwidth}
         \centering
         \includegraphics[width=\textwidth]{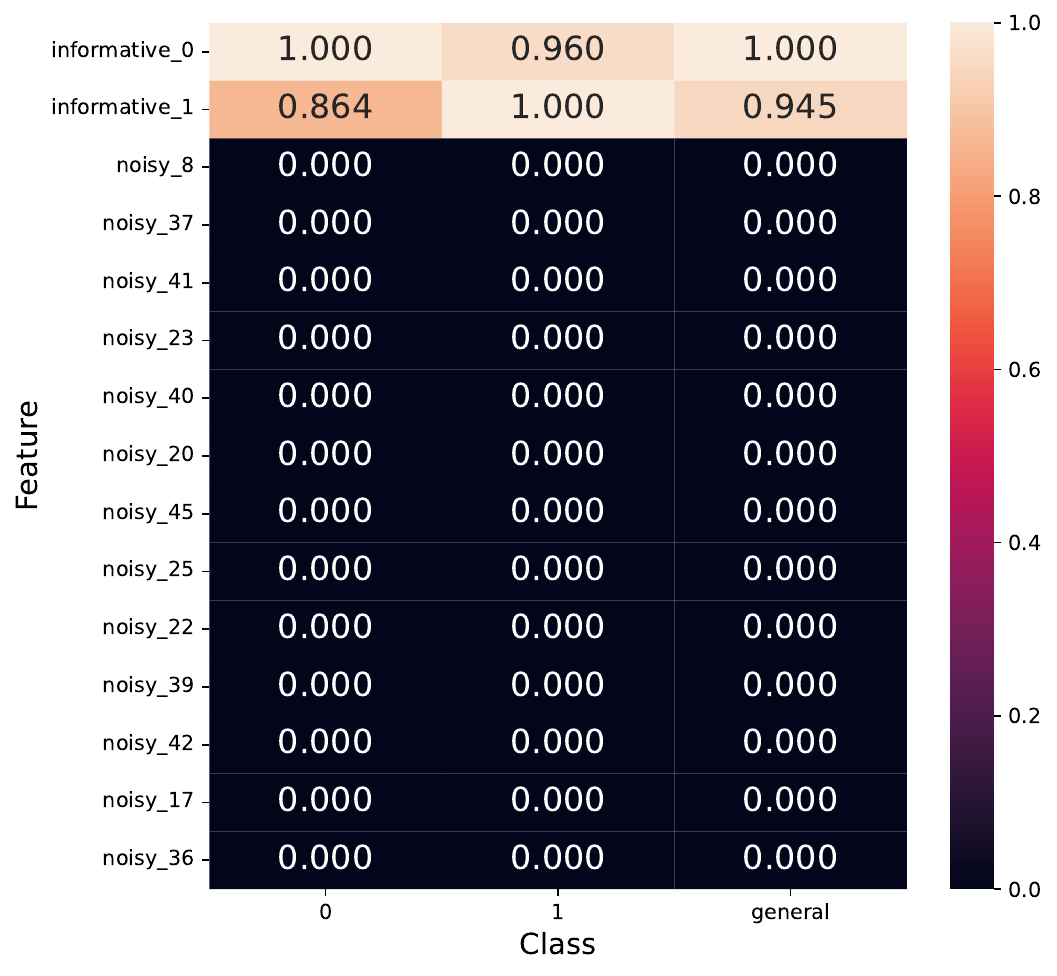}
         \caption{Output Aware}
     \end{subfigure}
     \hfill
     \begin{subfigure}[b]{0.44\textwidth}
         \centering
         \includegraphics[width=\textwidth]{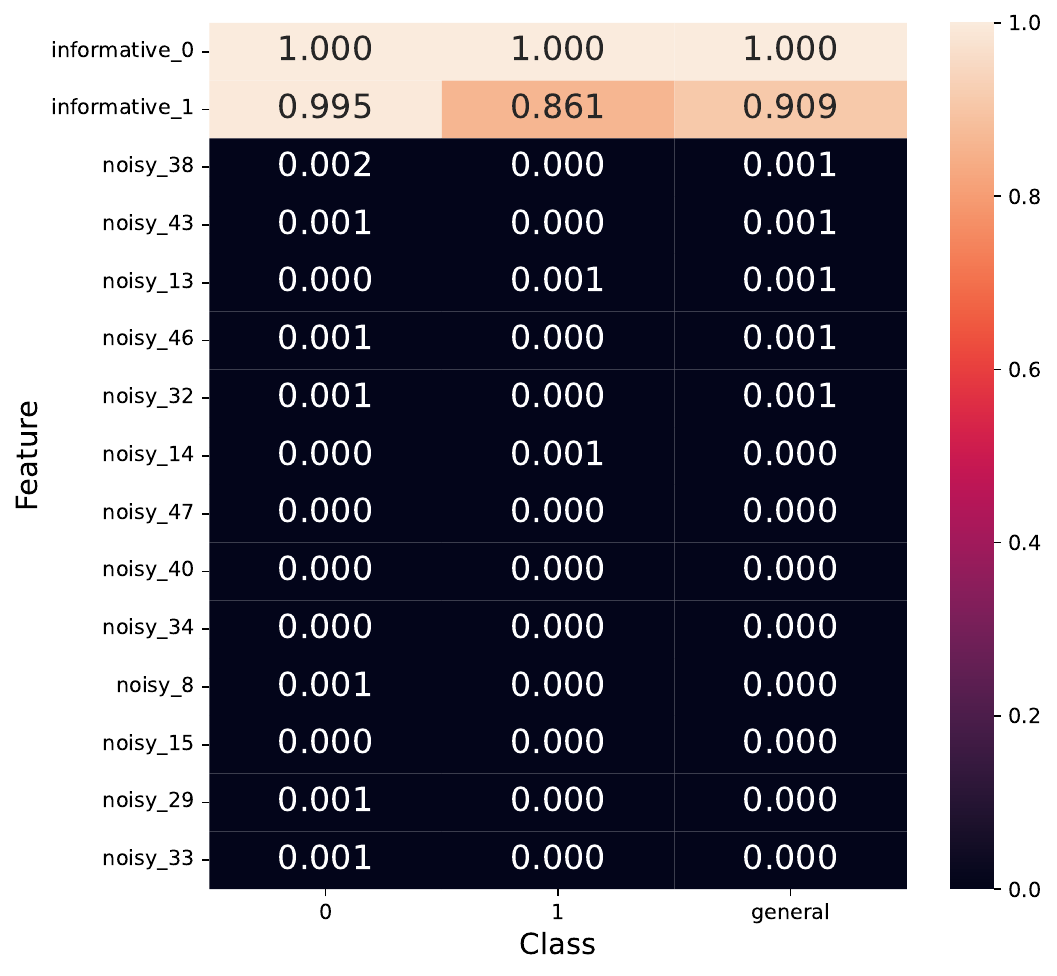}
         \caption{Multiple FSL}
     \end{subfigure}
\end{figure}

\begin{figure}[!h]
    \caption[
        Feature ranking positions across FSL and proposed extensions executions on XOR dataset.
        Each of the five features shows the positions of each feature in the feature weighting ranking across all eleven executions.
        For all algorithms is possible to notice that the two informative features are always on the first two positions and the other features are constantly in different positions on the bottom of the ranking.
        Its also possible to notice that the two informative features are not always constantly in first or second place.
    ]
    {\textbf{Feature ranking positions across FSL and proposed extensions executions on XOR dataset.}
        All rankings were created by sorting the features in descending order of their weights.
        Each algorithm was executed 11 times (horizontal axis). 
        The horizontal axis represents each individual execution, while the vertical axis shows the rank position of each feature, from 0 (top) to 49 (bottom), within that execution. 
        Each color represent a different feature, with blue and orange lines at the top indicating the two relevant features.
    }
     \centering
     \label{fig:feature_position_per_execution_on_extensions_for_xor}
     \begin{subfigure}[b]{0.49\textwidth}
         \centering
         \includegraphics[width=\textwidth]{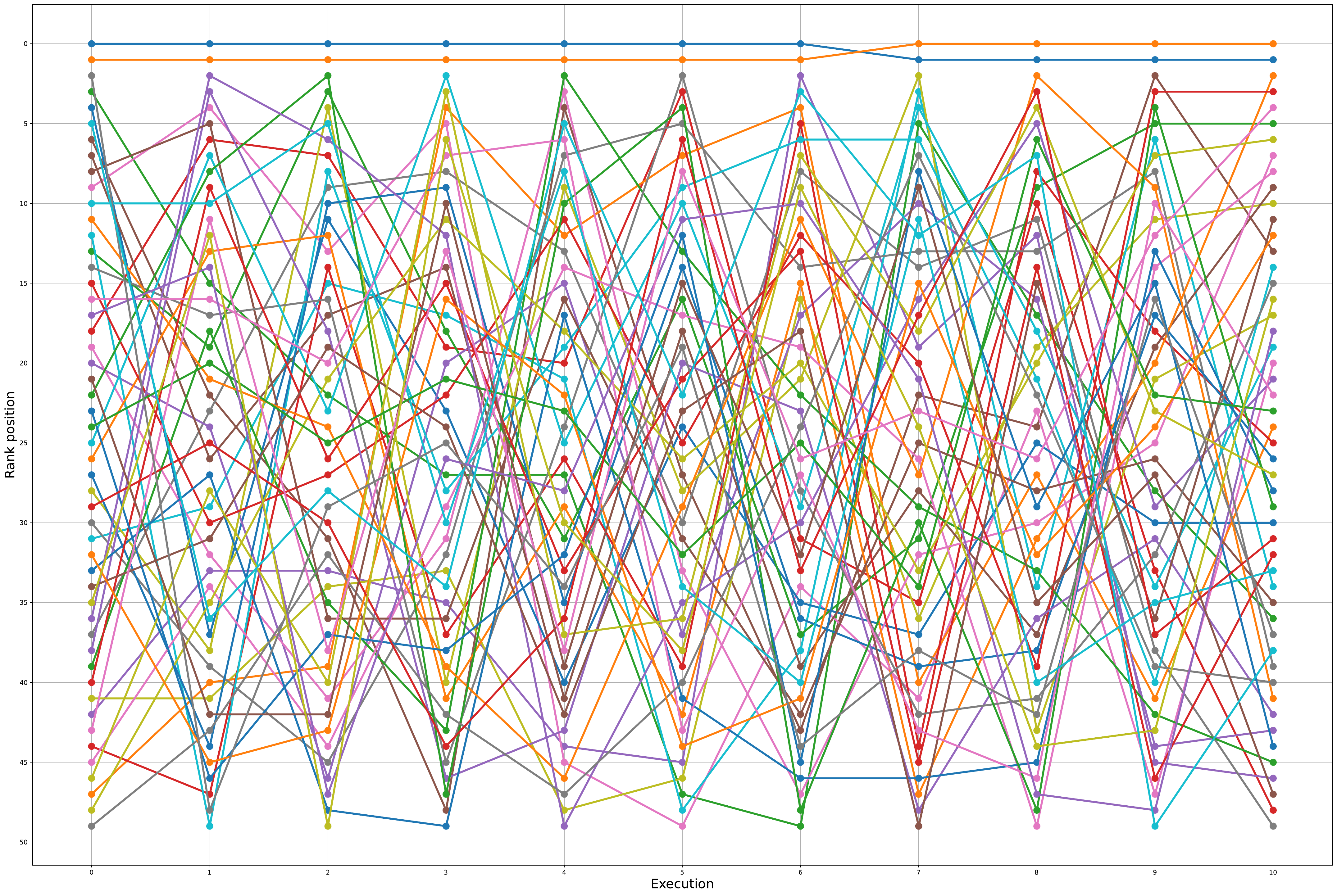}
         \caption{FSL}
         \label{subfig:fsl_feature_position_per_execution_on_extensions_for_xor}
     \end{subfigure}
     \hfill
     \begin{subfigure}[b]{0.49\textwidth}
         \centering
         \includegraphics[width=\textwidth]{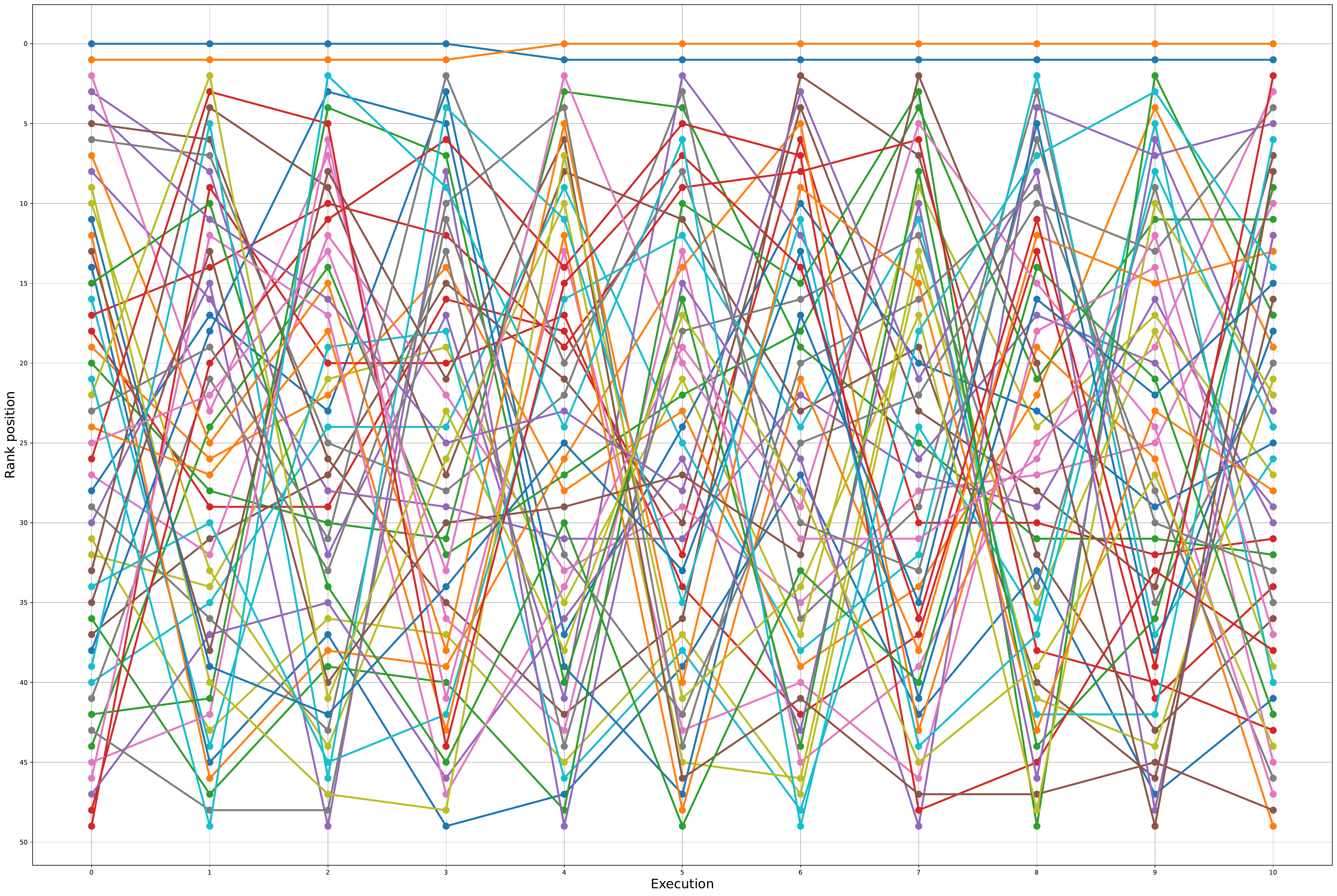}
         \caption{Listener}
         \label{subfig:listener_feature_position_per_execution_on_extensions_for_xor}
     \end{subfigure}
     \begin{subfigure}[b]{0.49\textwidth}
         \centering
         \includegraphics[width=\textwidth]{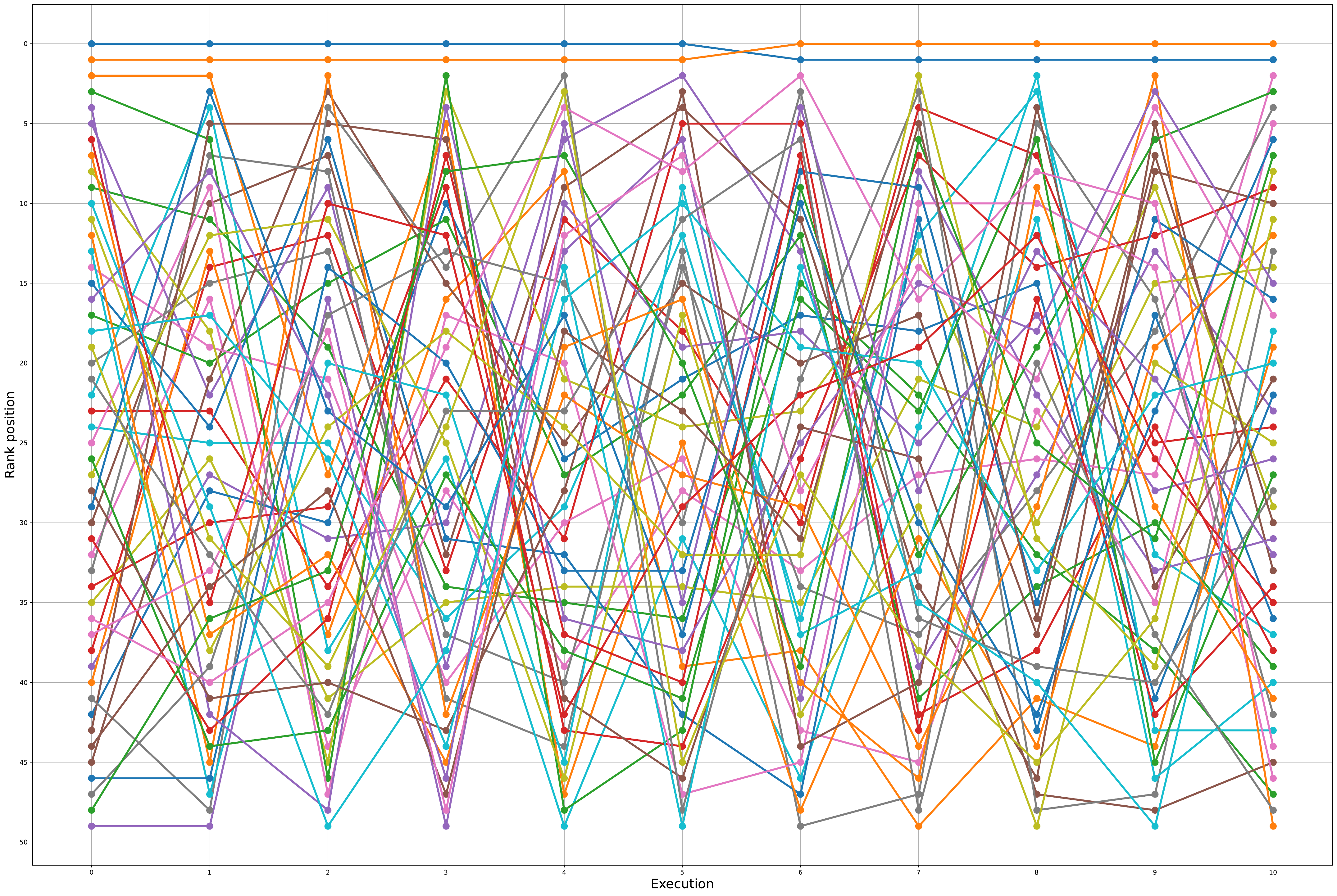}
         \caption{Residual}
         \label{subfig:residual_feature_position_per_execution_on_extensions_for_xor}
     \end{subfigure}
     \hfill
     \begin{subfigure}[b]{0.49\textwidth}
         \centering
         \includegraphics[width=\textwidth]{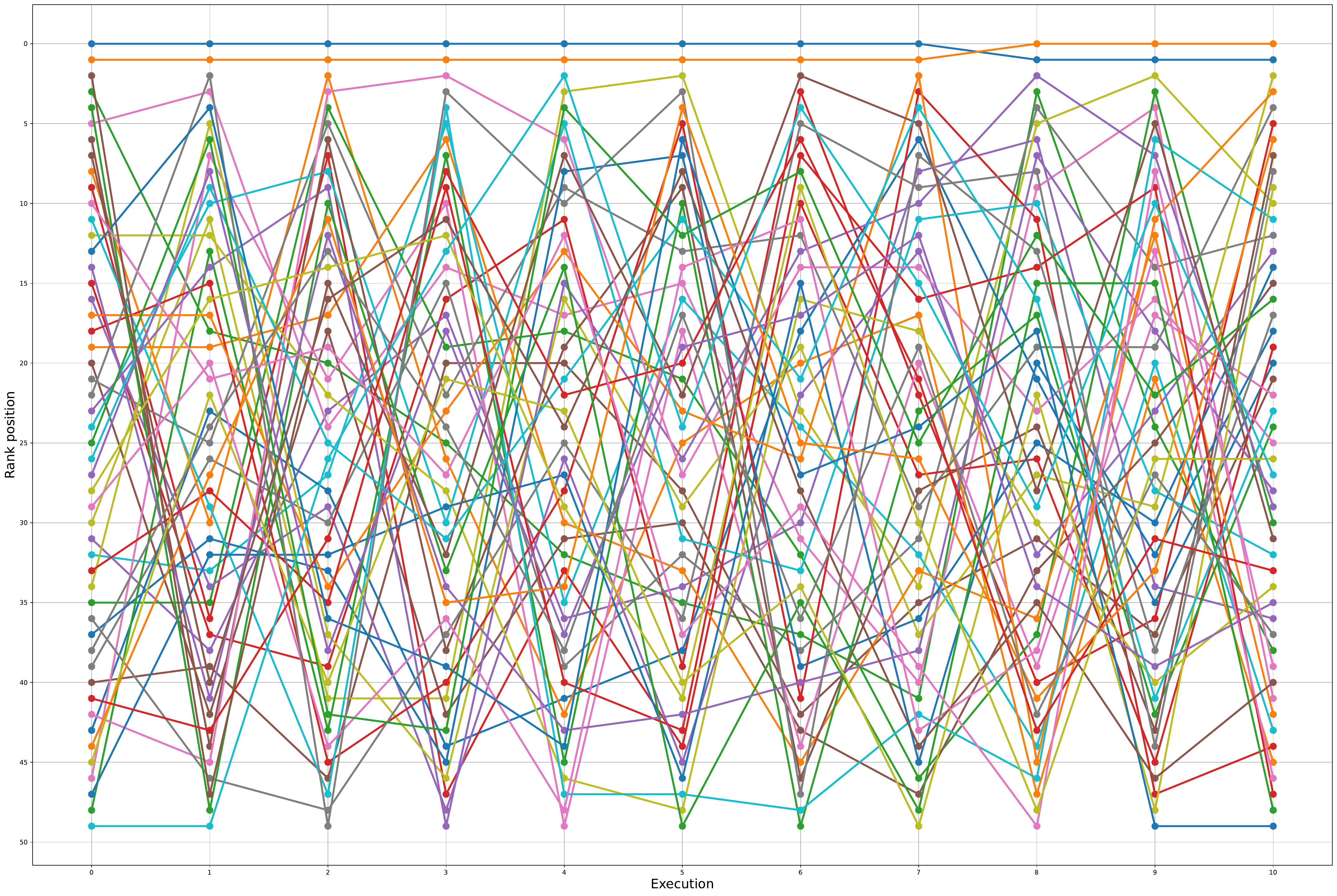}
         \caption{Output Aware}
         \label{subfig:output_aware_feature_position_per_execution_on_extensions_for_xor}
     \end{subfigure}
     \hfill
     \begin{subfigure}[b]{0.49\textwidth}
         \centering
         \includegraphics[width=\textwidth]{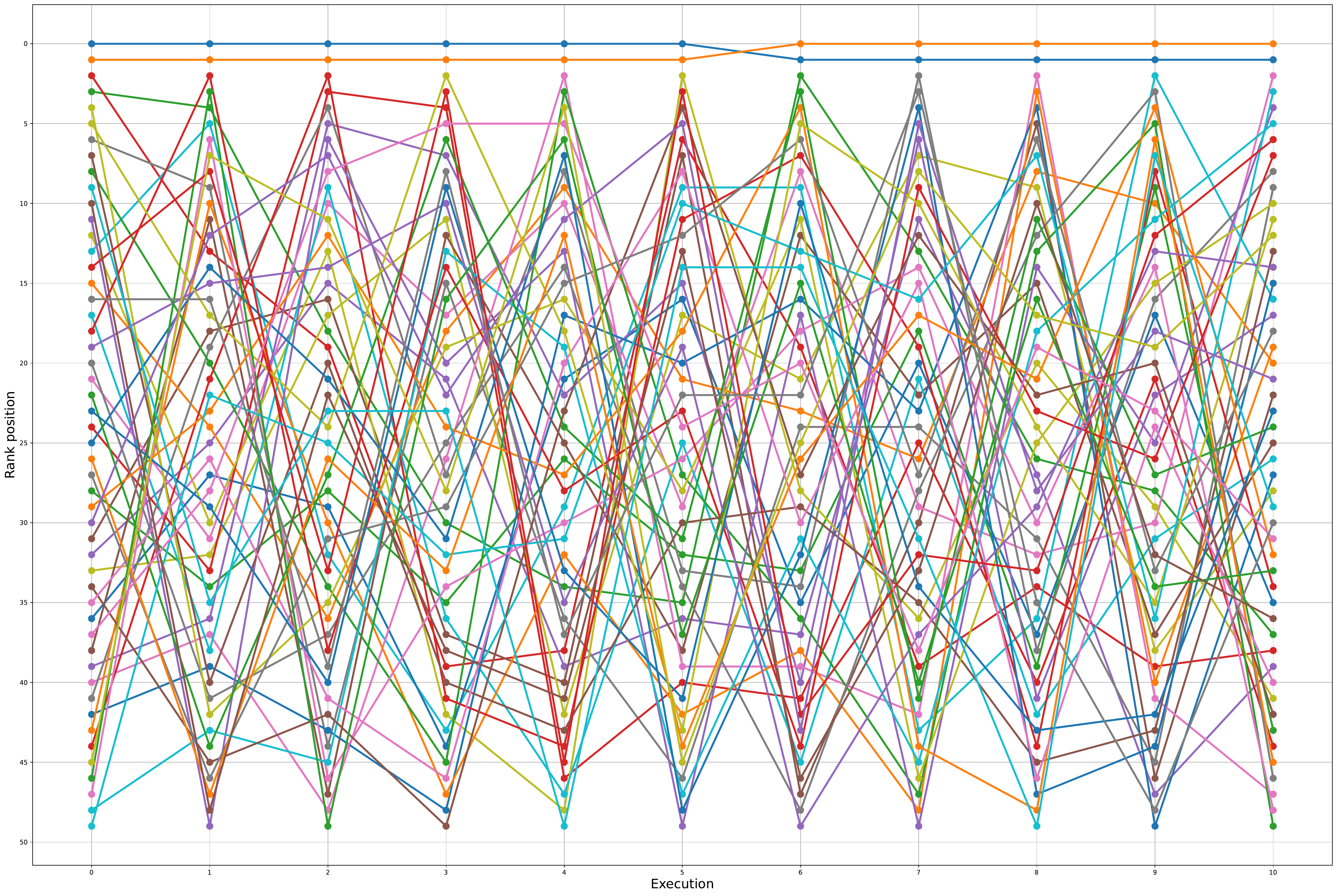}
         \caption{Multiple FSL}
         \label{subfig:mfs_feature_position_per_execution_on_extensions_for_xor}
     \end{subfigure}
\end{figure}

\begin{figure}[!h]
    \caption[
    Figure composed of six images, containing the t-SNE and weighted t-SNE results for XOR dataset based on FSL and the proposed extensions.
    The first figure presents the application of t-SNE to the XOR dataset without any weight.
    Samples from two classes are projected into a 2D space, each represented by distinct shapes and colors. 
    The two classes heavily overlap in the center of the plot, forming a circular pattern with no clear separation. 
    The second figure presents the application of t-SNE to the XOR dataset using FSL feature weights, which changes the projection to four smaller circular clusters located at the plot’s borders: the top-left and bottom-right clusters correspond to class 0, while the top-right and bottom-left clusters correspond to class 1.
    The third figure presents the application of t-SNE to the XOR dataset using Listener extension feature weights, which also changes the projection to four smaller circular clusters located at the plot’s borders: the top-left and bottom-right clusters correspond to class 0, while the top-right and bottom-left clusters correspond to class 1.
    However each cluster is composed of other four cluster with the same class.
    The forth figure presents the application of t-SNE to the XOR dataset using Residual extension feature weights, which changes the projection to four smaller circular clusters located at the plot’s borders: the top-left and bottom-right clusters correspond to class 0, while the top-right and bottom-left clusters correspond to class 1.
    Being very similar with the FSL result.
    The fifth figure presents the application of t-SNE to the XOR dataset using Output Aware extension feature weights, which also changes the projection to the same four smaller circular clusters.
    However, as Listener, the cluster are separated internally into different smaller cluster, being them near to each other and with the same class.
    The sixth figure presents the application of t-SNE to the XOR dataset using Multiple FSL extension feature weights, which also changes the projection to the same four smaller circular clusters.
    However, as Output Aware, the cluster are separated internally into different smaller cluster, being even more near to each other and with the same class.
    ]
    {\textbf{Visualization of XOR dataset with weighted t-SNE using the feature weights attributed by original FSL and the proposed extensions.}
        \autoref{subfig:no_weight_weighted_tsne_on_extensions_for_xor} presents a t-SNE visualization of the data without feature weights. 
        In contrast, \ref{subfig:fsl_weighted_tsne_on_extensions_for_xor}, \ref{subfig:listener_weighted_tsne_on_extensions_for_xor}, \ref{subfig:residual_weighted_tsne_on_extensions_for_xor},
        \ref{subfig:output_aware_weighted_tsne_on_extensions_for_xor} and \ref{subfig:multiple_fsl_weighted_tsne_on_extensions_for_xor} employ weighted t-SNE, utilizing the generated weights from original FSL and each proposed extension to enhance the separation between clusters.
        False labels (0) are represented in red, while true labels (1) are represented in blue.
     }
     \centering
     \label{fig:weighted_tsne_on_extensions_for_xor}
     \begin{subfigure}[b]{0.48\textwidth}
         \centering
         \includegraphics[width=\textwidth]{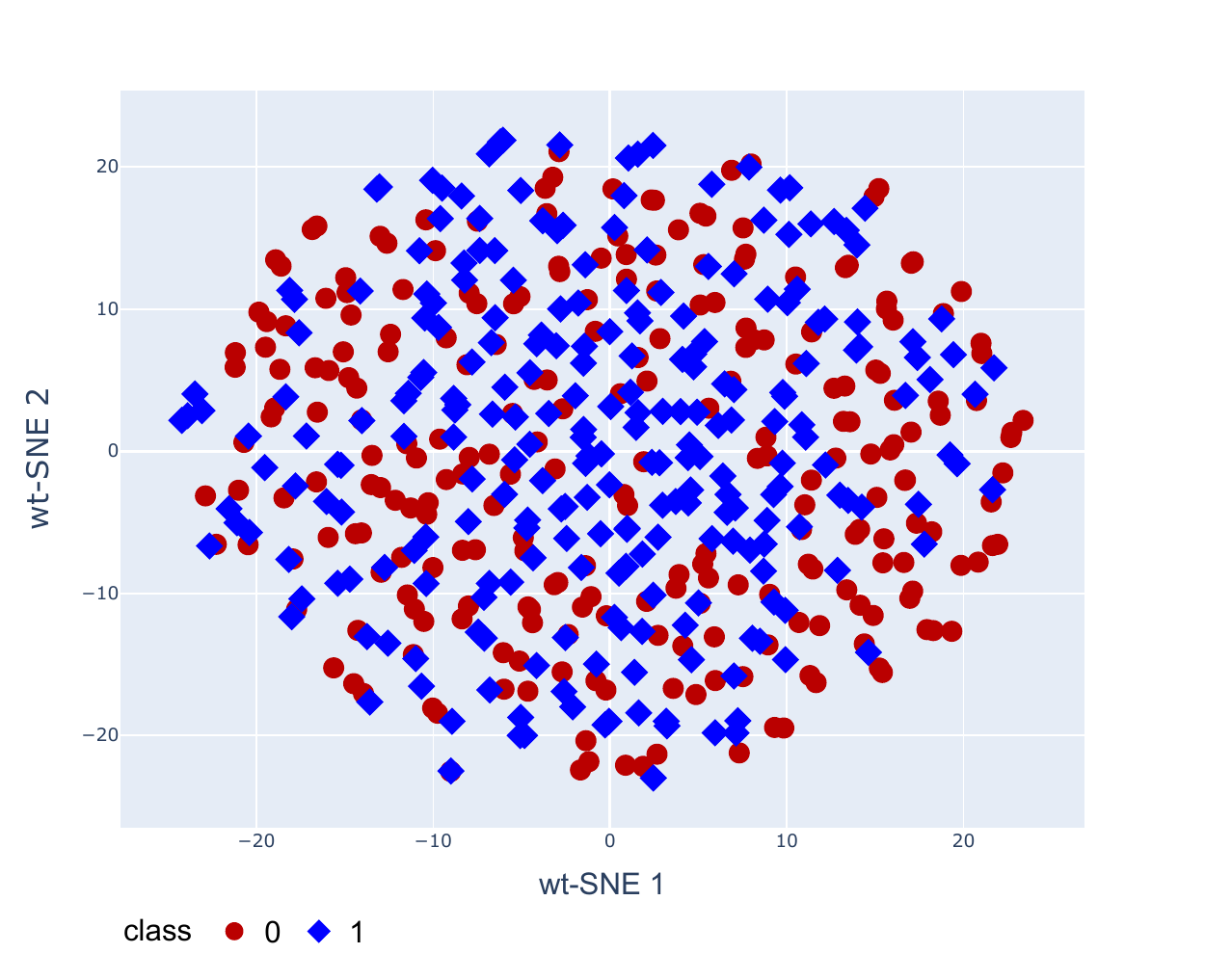}
         \caption{Without weights}
         \label{subfig:no_weight_weighted_tsne_on_extensions_for_xor}
     \end{subfigure}
     \hfill
     \begin{subfigure}[b]{0.48\textwidth}
         \centering
         \includegraphics[width=\textwidth]{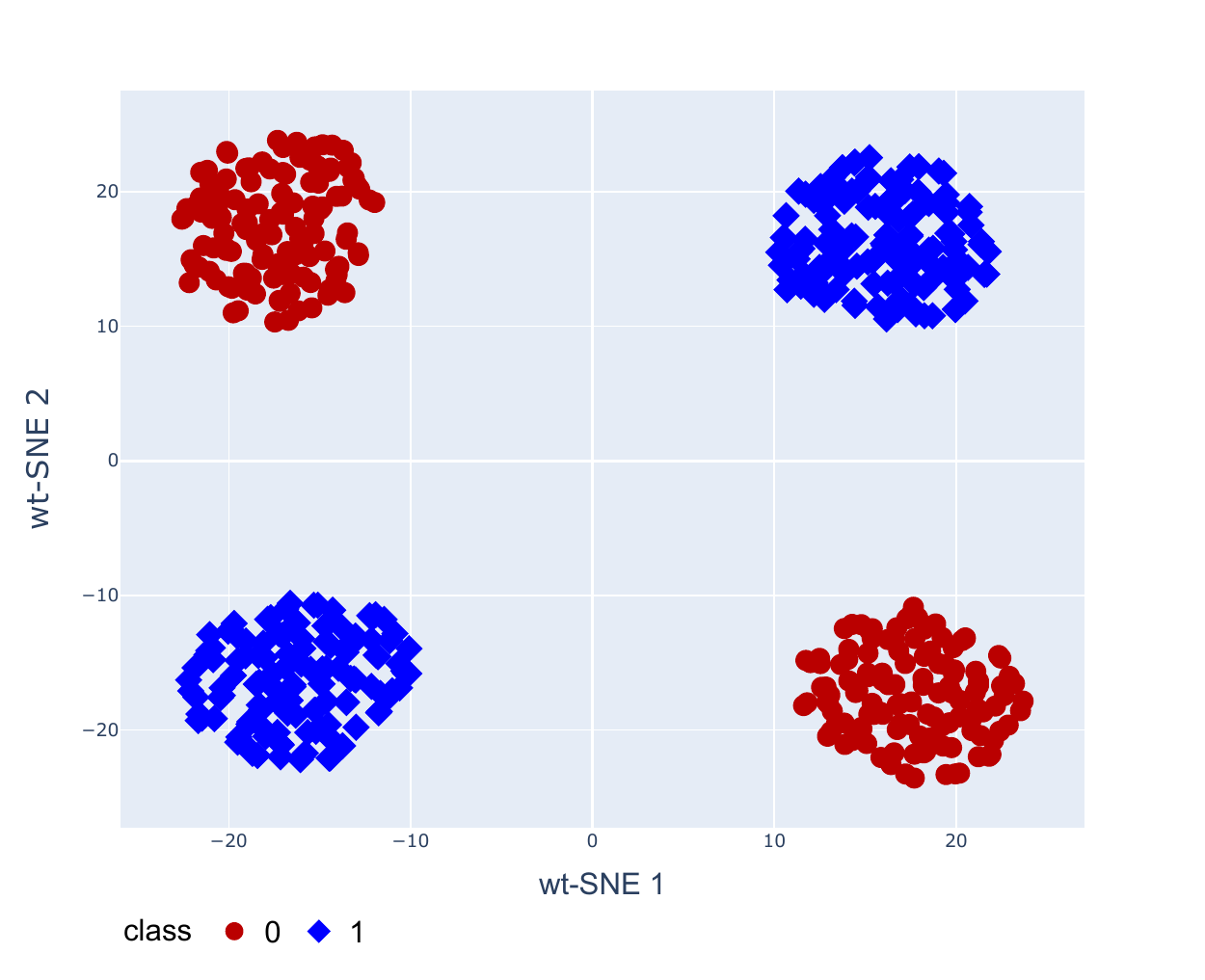}
         \caption{FSL}
        \label{subfig:fsl_weighted_tsne_on_extensions_for_xor}
     \end{subfigure}
     \begin{subfigure}[b]{0.48\textwidth}
         \centering
         \includegraphics[width=\textwidth]{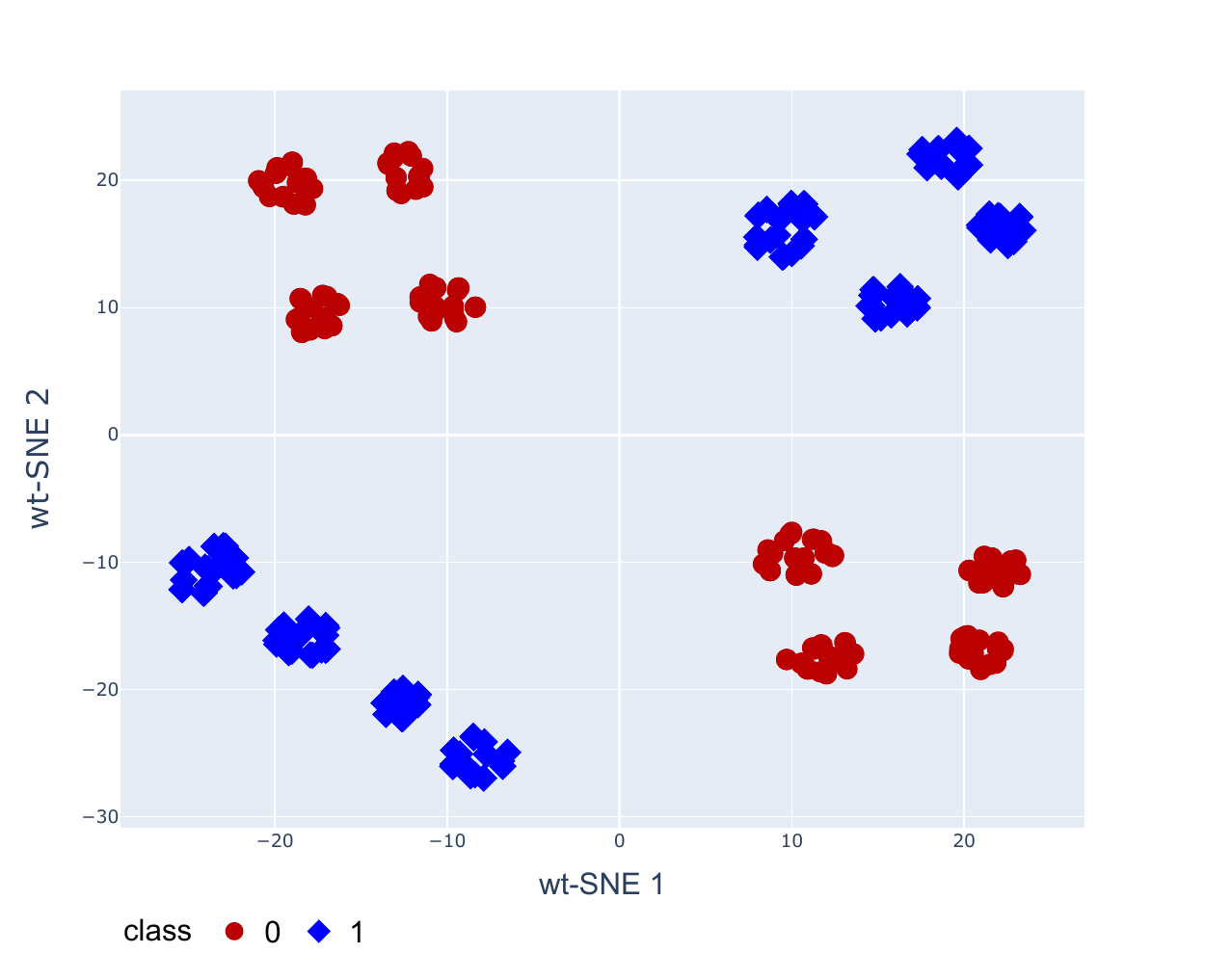}
         \caption{Listener}
         \label{subfig:listener_weighted_tsne_on_extensions_for_xor}
     \end{subfigure}
     \hfill
     \begin{subfigure}[b]{0.48\textwidth}
         \centering
         \includegraphics[width=\textwidth]{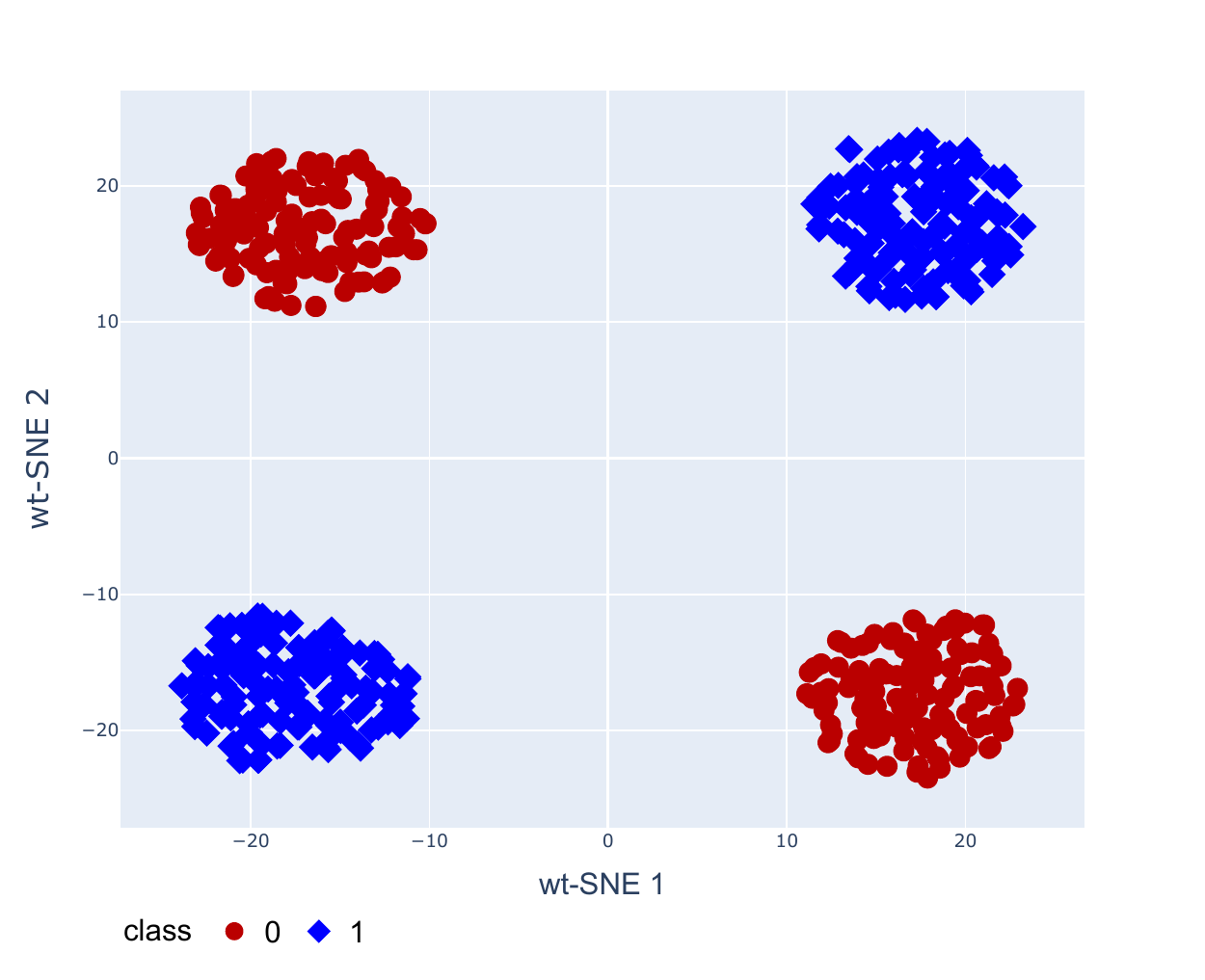}
         \caption{Residual}
         \label{subfig:residual_weighted_tsne_on_extensions_for_xor}
     \end{subfigure}
     \begin{subfigure}[b]{0.48\textwidth}
         \centering
         \includegraphics[width=\textwidth]{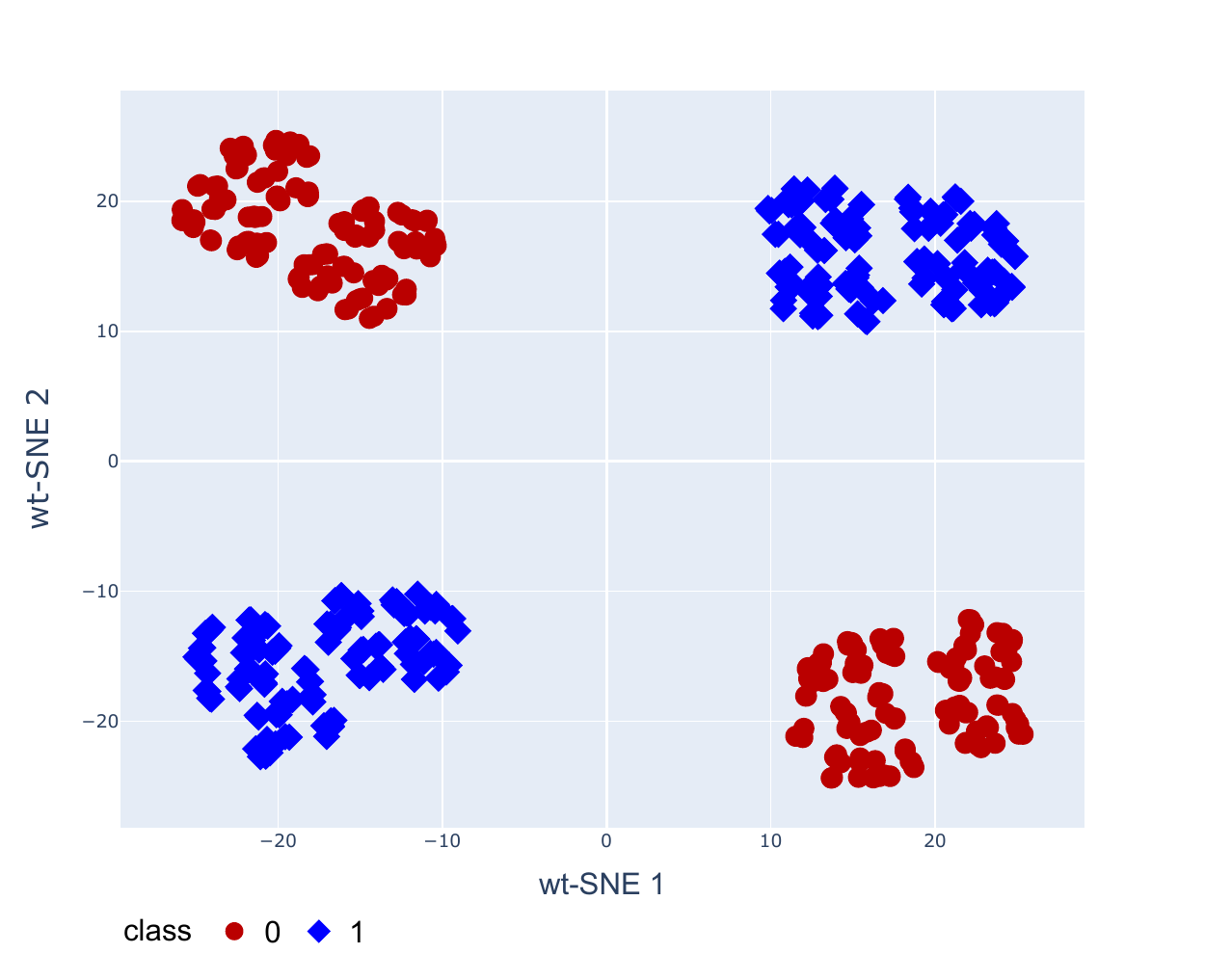}
         \caption{Output Aware}
         \label{subfig:output_aware_weighted_tsne_on_extensions_for_xor}
     \end{subfigure}
     \hfill
     \begin{subfigure}[b]{0.48\textwidth}
         \centering
         \includegraphics[width=\textwidth]{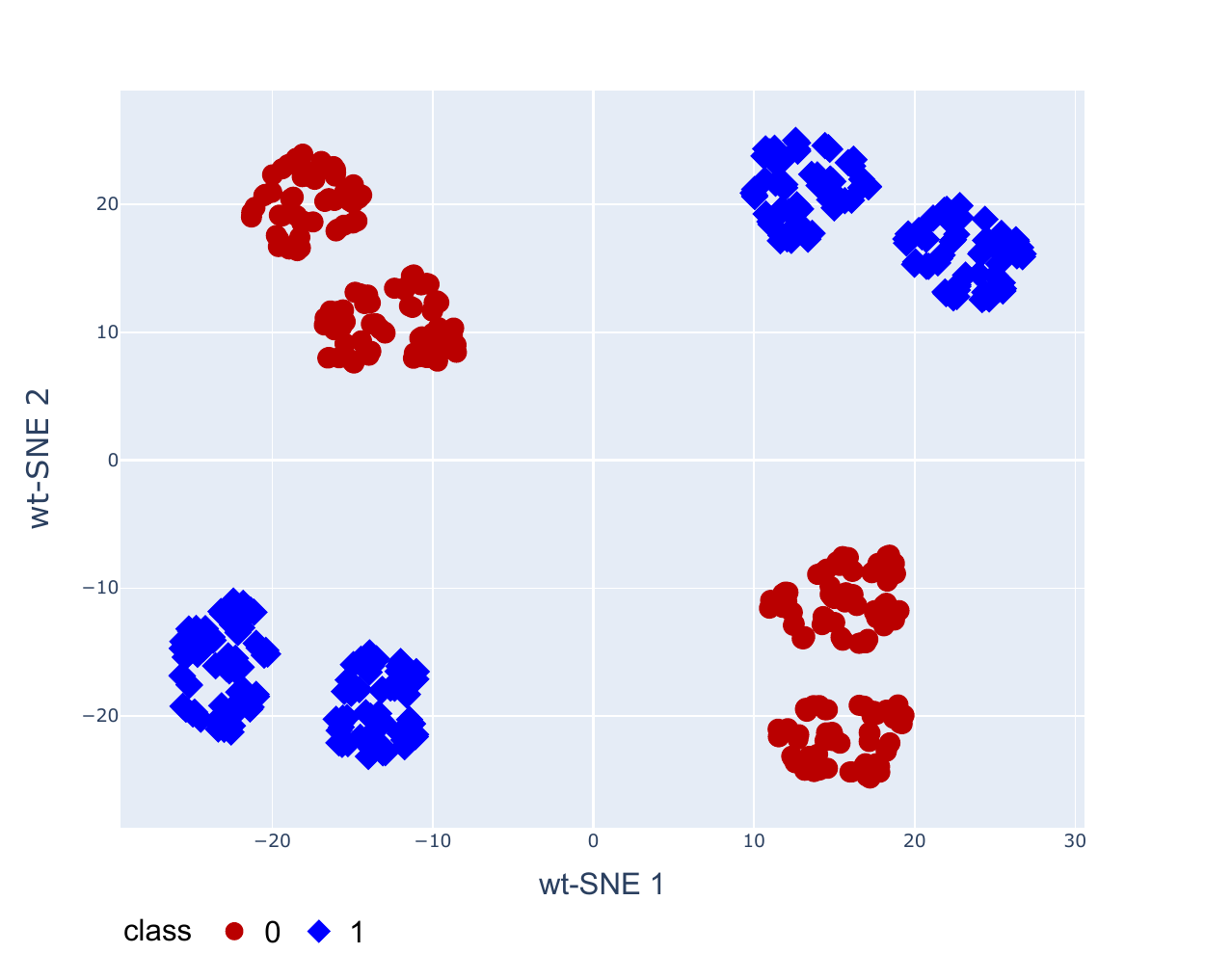}
         \caption{Multiple FSL}
         \label{subfig:multiple_fsl_weighted_tsne_on_extensions_for_xor}
     \end{subfigure}
\end{figure}

\autoref{fig:heatmap_on_extensions_for_xor} shows a heatmap of the 15 highest-weighted features for the XOR dataset for each extension and for the original FSL.
\autoref{fig:feature_position_per_execution_on_extensions_for_xor} presents the feature-ranking positions across multiple executions of FSL and the extensions. 
The goal is to determine whether the informative and non-informative features form clearly separated ranking groups. Since the models produce non-stable rankings, the results suggest a persistent separation between informative and non-informative features.
Finally \autoref{fig:weighted_tsne_on_extensions_for_xor} presents the weighted t-SNE results.

\begin{figure}[!h]
    \caption[
    Analysis of PIFS and PSFI on SynthA, SynthB and SynthC datasets between original FSL and proposed extensions for increasing selection subsets.
    The first figure shows the results for SynthA dataset, where FSL was able to create the bigger ranking with only informative features, followed by Listener and Multiple FSL.
    Output Aware and Residual had the worst results, selecting only a fraction of the informative features as the most important ones.
    The second figure shows the results for SynthB dataset, where all algorithms were able to select most of the informative features as the most important ones before starting including noisy features into the ranking.
    The third figure shows the results for SynthC dataset, where all algorithms had a result near to the perfect one where the raking with the top-30 features is composed most of the times with only informative features.
    ]
    {\textbf{Analysis of PIFS and PSFI on SynthA, SynthB and SynthC datasets between original FSL and proposed extensions for increasing selection subsets.} 
        All values scores are the average over rankings generated by 11 independent models execution.
        Vertical axis represents the scores constrained from 0 to 1 as both measures are percentages.
        Horizontal axis represents the subsets size selected from the top-n features with the highest weights.
        \autoref{subfig:pifs_psfi_analysis_synthA_on_extensions} represents the evaluation over SynthA dataset, \autoref{subfig:pifs_psfi_analysis_synthB_on_extensions} over SynthB and \autoref{subfig:pifs_psfi_analysis_synthC_on_extensions} over SynthC.
    }
     \centering
    \label{fig:pifs_psfi_analysis_synthA_synthB_synthC_on_extensions}
     \begin{subfigure}[b]{0.49\textwidth}
         \centering
         \includegraphics[width=\textwidth]{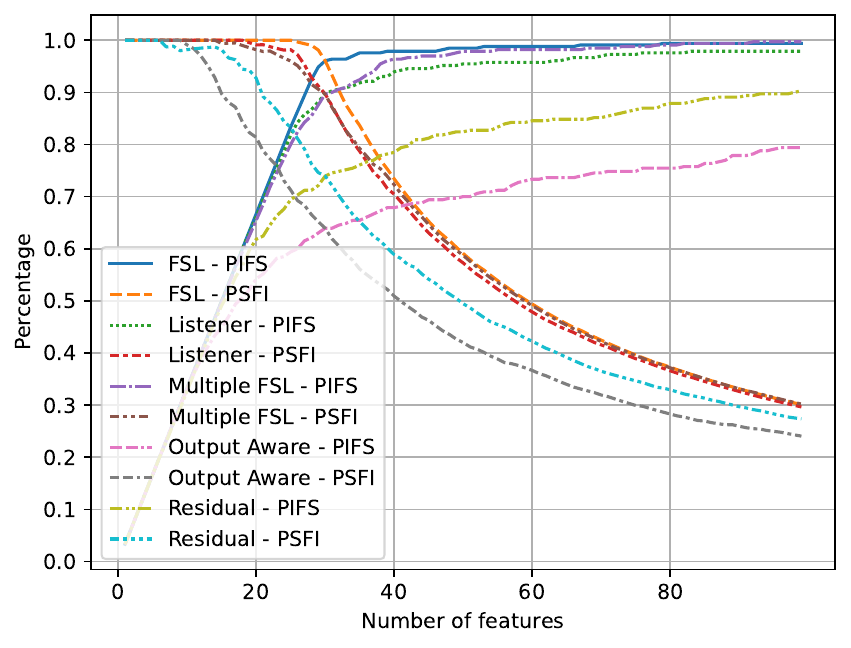}
         \caption{SynthA}
         \label{subfig:pifs_psfi_analysis_synthA_on_extensions}
     \end{subfigure}
     \hfill
     \begin{subfigure}[b]{0.49\textwidth}
         \centering
         \includegraphics[width=\textwidth]{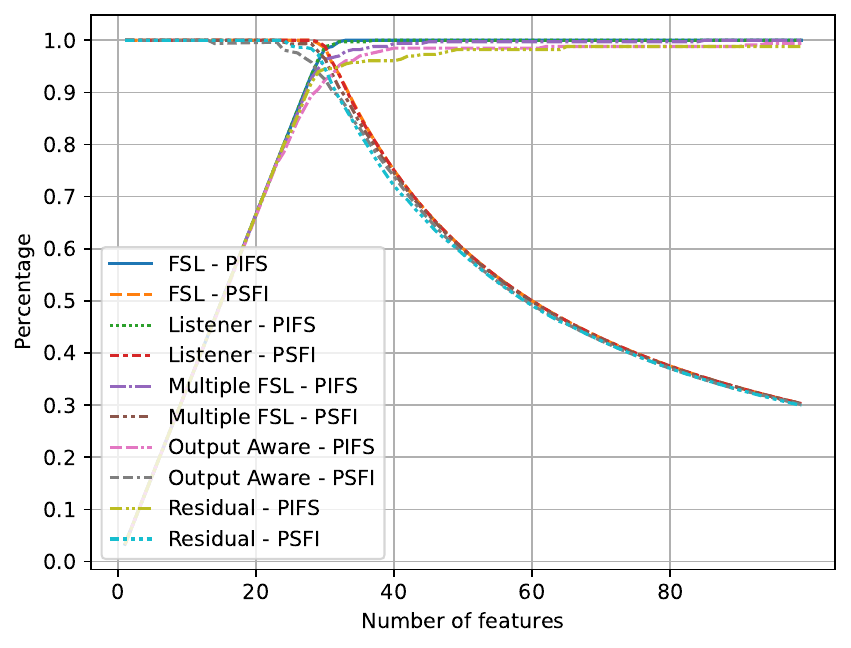}
         \caption{SynthB}
        \label{subfig:pifs_psfi_analysis_synthB_on_extensions}
     \end{subfigure}
     \hfill
     \begin{subfigure}[b]{0.49\textwidth}
         \centering
         \includegraphics[width=\textwidth]{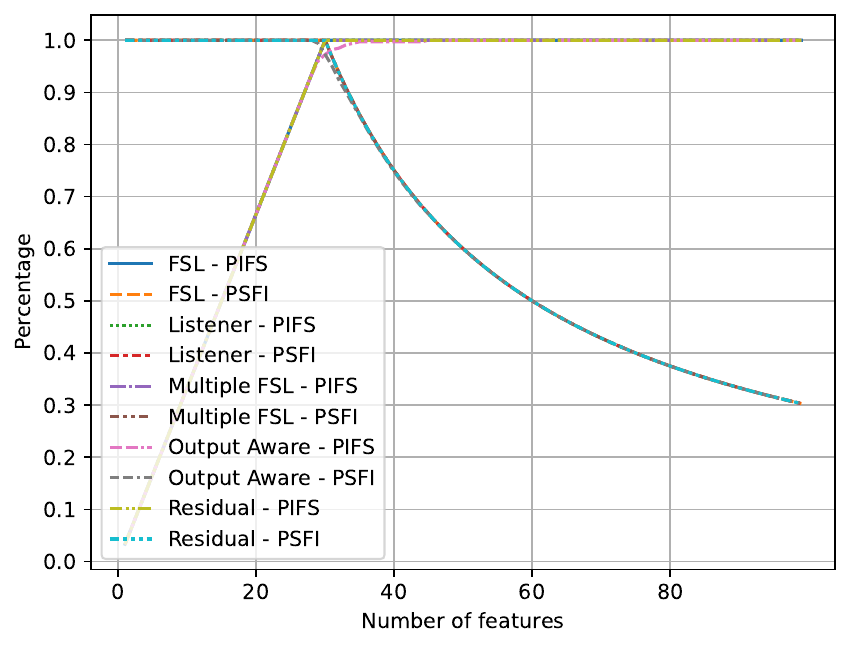}
         \caption{SynthC}
         \label{subfig:pifs_psfi_analysis_synthC_on_extensions}
     \end{subfigure}
\end{figure}

\autoref{fig:pifs_psfi_analysis_synthA_synthB_synthC_on_extensions} presents the PIFS (percentage of informative features selected) and PSFI (percentage of selected features that are informative) analyses for SynthA, SynthB, and SynthC across increasing selection subsets, comparing the original FSL with the proposed extensions.
This analysis highlights how consistently informative features receive higher weights across multiple runs.

\begin{figure}[!h]
    \caption[
    SynthA weighted t-SNE visualization using the general feature weights attributed by original FSL and the proposed extensions.
    The first figure shows the application of t-SNE to the SynthA dataset without feature weights.
    All three classes largely overlap in the center of the plot, forming a circular pattern with no evident separation. 
    The second figure uses the FSL weights, the circular structure remains, but it becomes divided into  tree distinct regions: the bottom-left corresponds to class 2, the top-left to class 0, and the top-right to class 1.
    The third figure uses the Listener weights, being divided into five distinct regions: the bottom-left and center-right corresponds to class 2, the top-left to class 0, and the top-right and bottom-right to class 1.
    The forth figure uses the Residual weights, being divided into four biggest distinct regions: the center and bottom-left corresponds to class 0, the top-left and bottom to class 2, and the right to class 1. However it also has smaller regions of each class between the biggest clusters.
    The fifth figure uses the Output Aware weights, being divided into four distinct regions: the left and bottom-right corresponds to class 0, the top to class 2, and the center and right to class 1.
    The sixth figure uses the Multiple FSL weights, being divided into four distinct regions: the top corresponds to class 0, the center and left to class 1, and the right and bottom to the class 2 being these two last regions near to each other.
    On all figures we can notice individual samples from each class overlapping other classes clusters.
    ]
    {\textbf{SynthA weighted t-SNE visualization using the general feature weights attributed by original FSL and the proposed extensions.}
        \autoref{subfig:no_weight_weighted_tsne_on_extensions_for_synthA} presents a t-SNE visualization of the data without feature weights. 
        In contrast, \ref{subfig:fsl_weighted_tsne_on_extensions_for_synthA}, \ref{subfig:listener_weighted_tsne_on_extensions_for_synthA}, \ref{subfig:residual_weighted_tsne_on_extensions_for_synthA}, \ref{subfig:output_aware_weighted_tsne_on_extensions_for_synthA} and  \ref{subfig:multiple_fsl_weighted_tsne_on_extensions_for_synthA} employ weighted t-SNE, utilizing the generated weights from original FSL and all proposed extensions to enhance the separation between clusters.
        Label 0 is represented in red, label 1 in blue and label 3 in yellow.
    }
     \centering
     \label{fig:weighted_tsne_on_extensions_for_synthA}
     \begin{subfigure}[b]{0.48\textwidth}
         \centering
         \includegraphics[width=\textwidth]{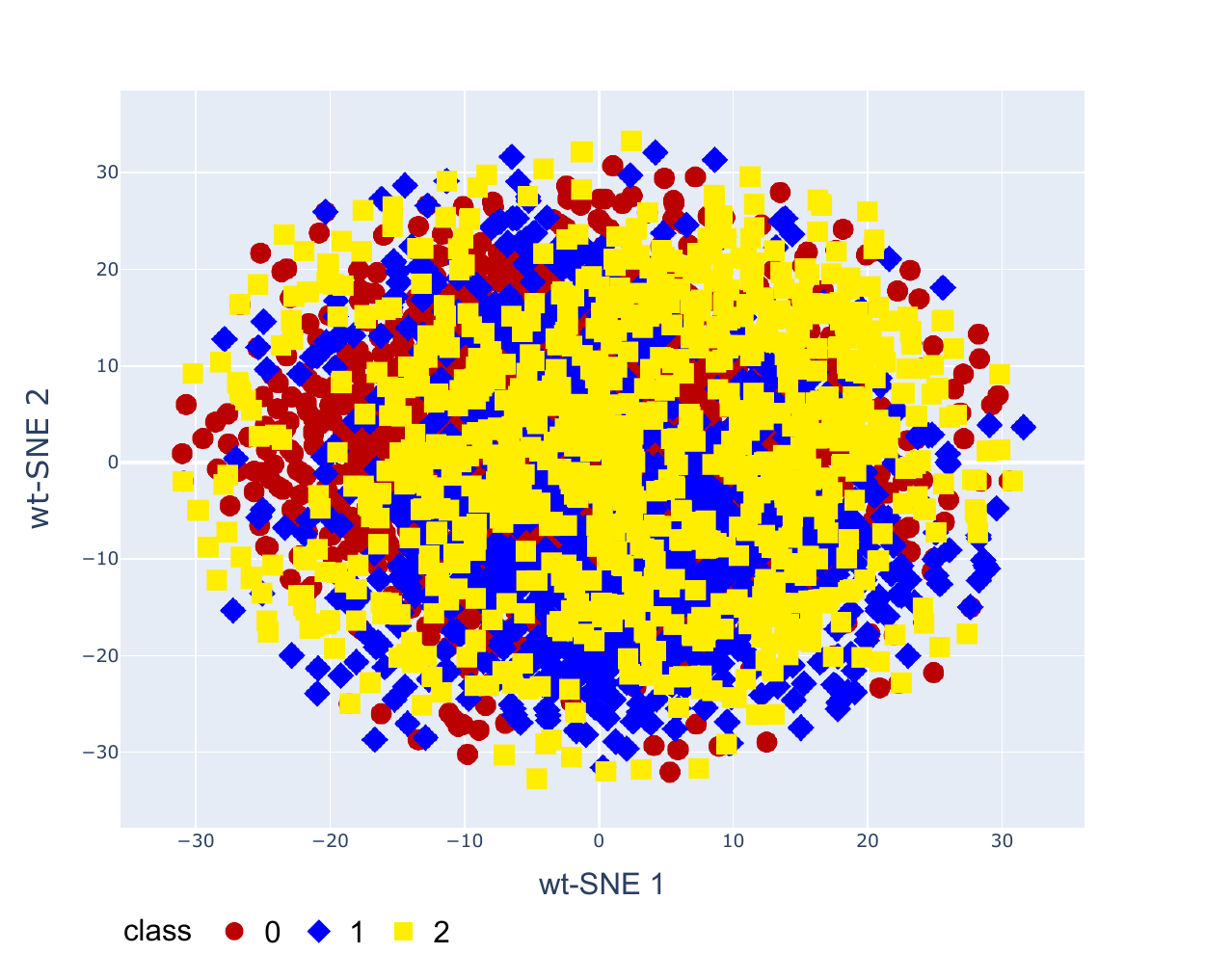}
         \caption{Without weights}
         \label{subfig:no_weight_weighted_tsne_on_extensions_for_synthA}
     \end{subfigure}
     \hfill
     \begin{subfigure}[b]{0.48\textwidth}
         \centering
         \includegraphics[width=\textwidth]{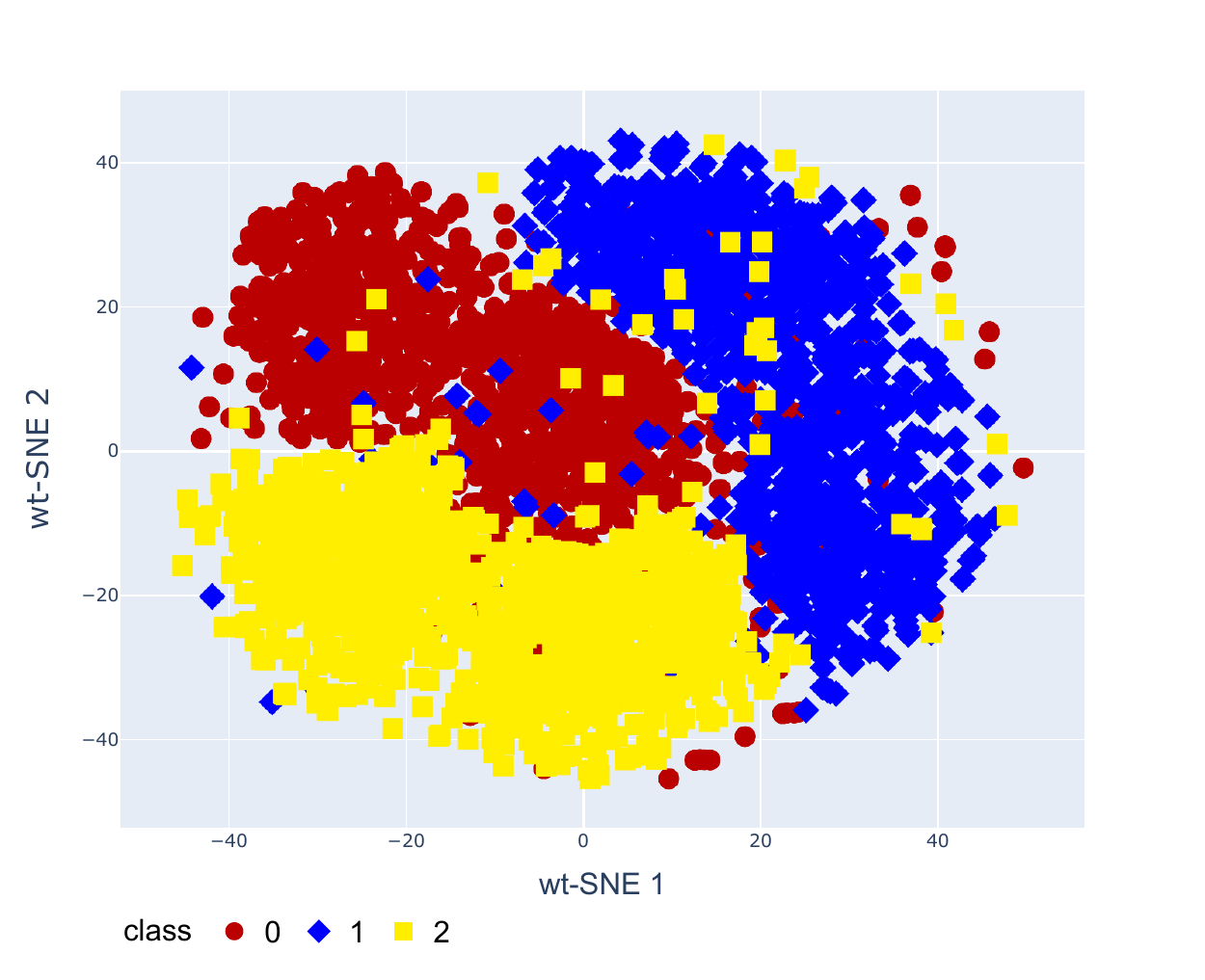}
         \caption{FSL}
        \label{subfig:fsl_weighted_tsne_on_extensions_for_synthA}
     \end{subfigure}
     \begin{subfigure}[b]{0.48\textwidth}
         \centering
         \includegraphics[width=\textwidth]{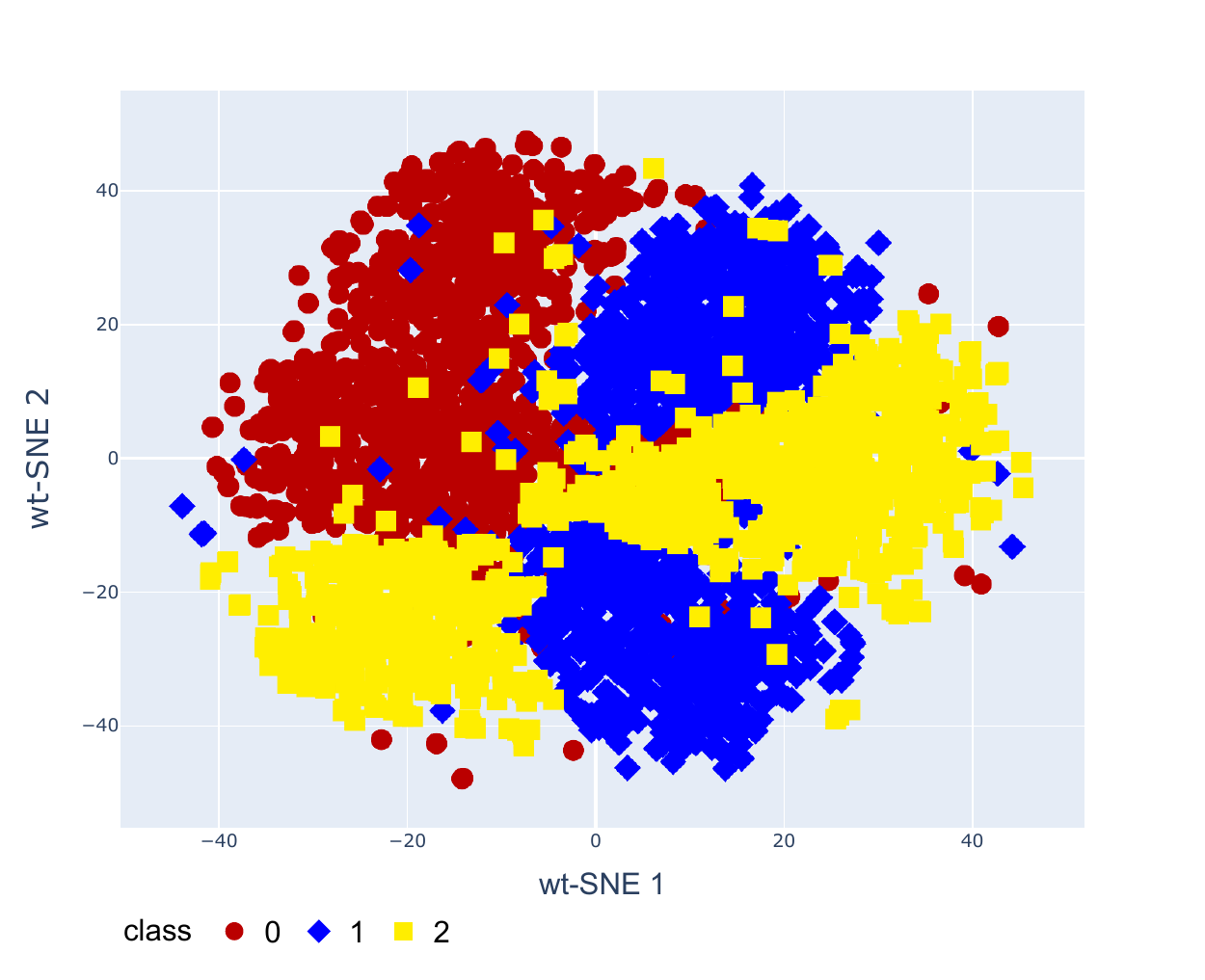}
         \caption{Listener}
         \label{subfig:listener_weighted_tsne_on_extensions_for_synthA}
     \end{subfigure}
     \hfill
     \begin{subfigure}[b]{0.48\textwidth}
         \centering
         \includegraphics[width=\textwidth]{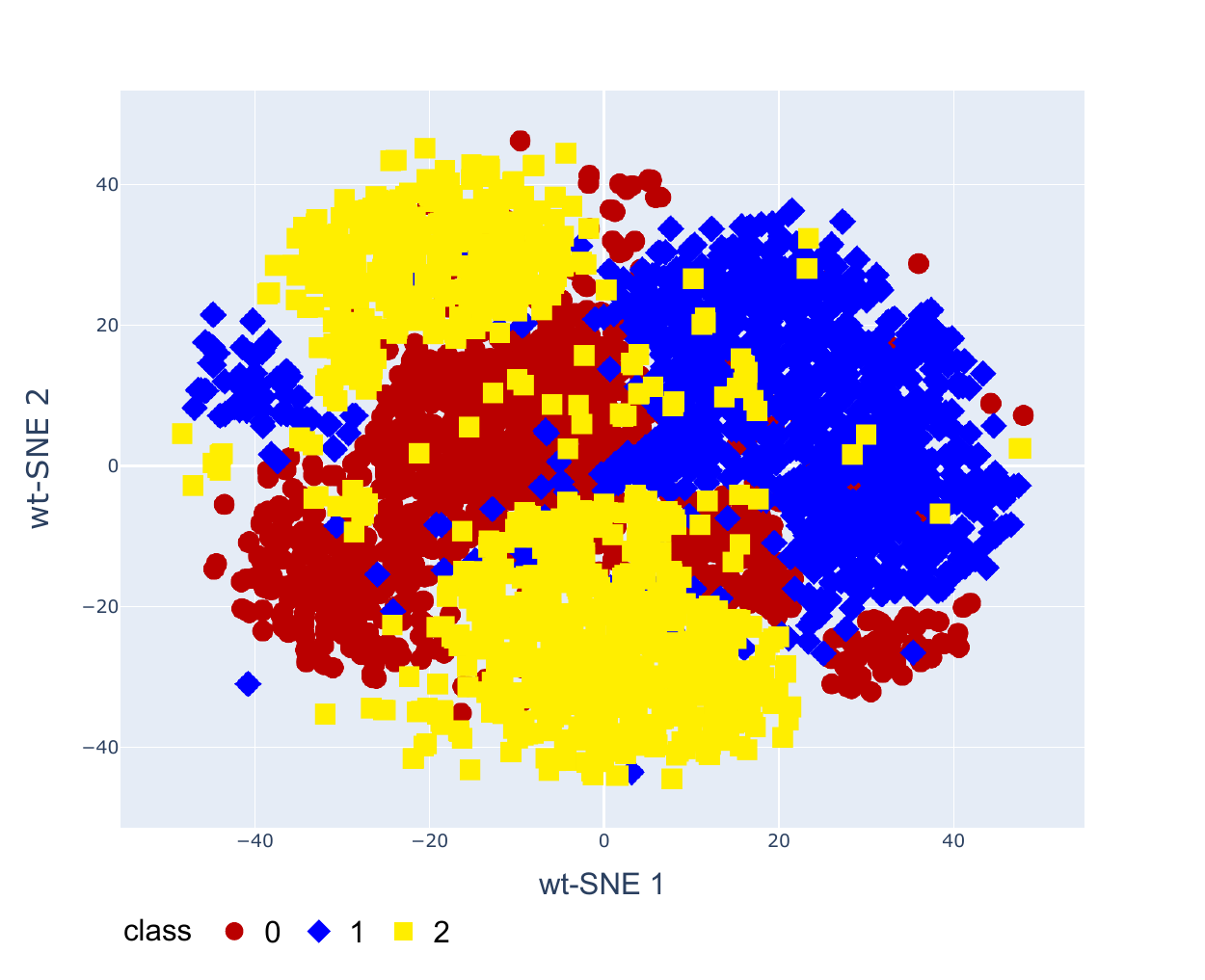}
         \caption{Residual}
         \label{subfig:residual_weighted_tsne_on_extensions_for_synthA}
     \end{subfigure}
     \begin{subfigure}[b]{0.48\textwidth}
         \centering
         \includegraphics[width=\textwidth]{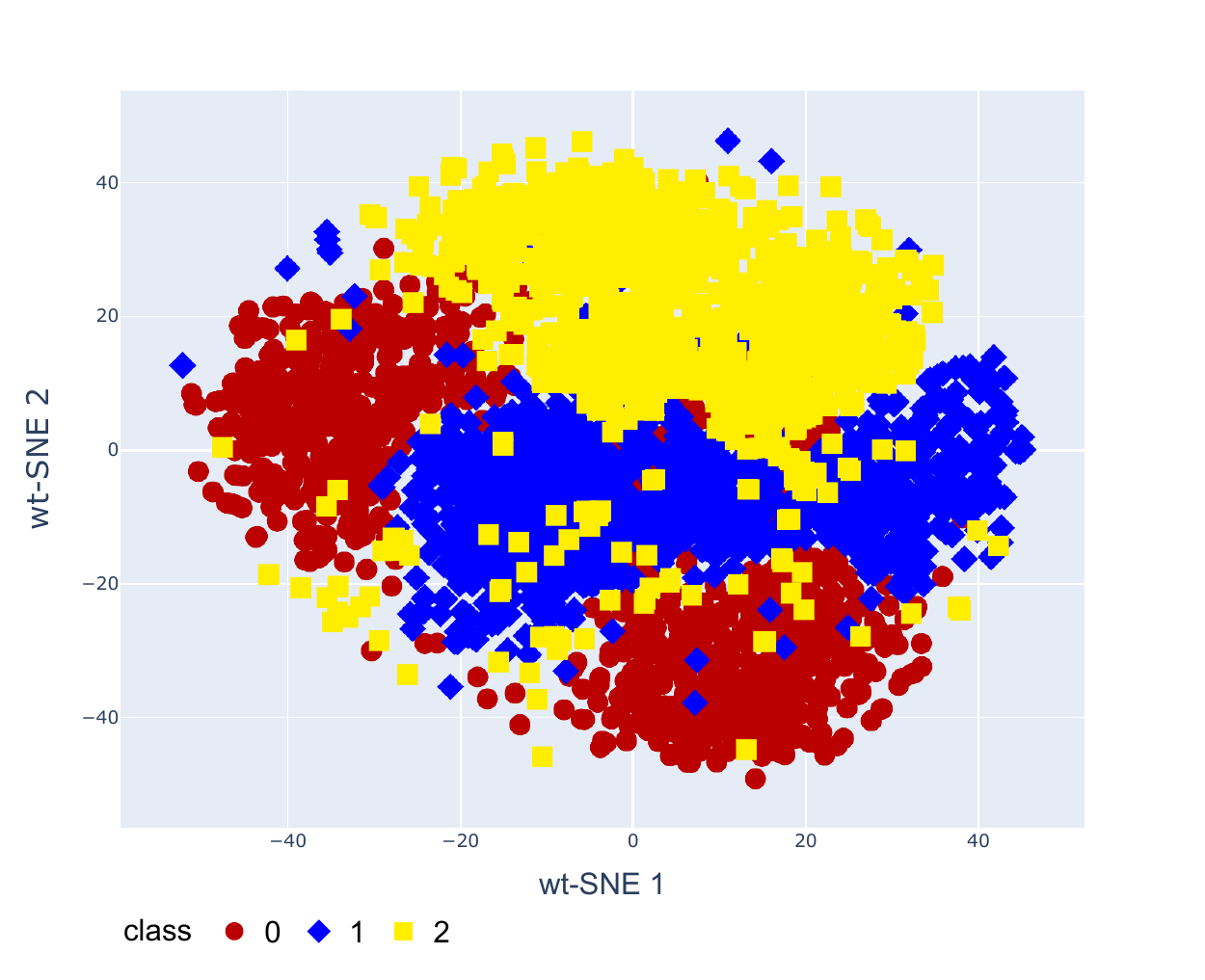}
         \caption{Output Aware}
         \label{subfig:output_aware_weighted_tsne_on_extensions_for_synthA}
     \end{subfigure}
     \hfill
     \begin{subfigure}[b]{0.48\textwidth}
         \centering
         \includegraphics[width=\textwidth]{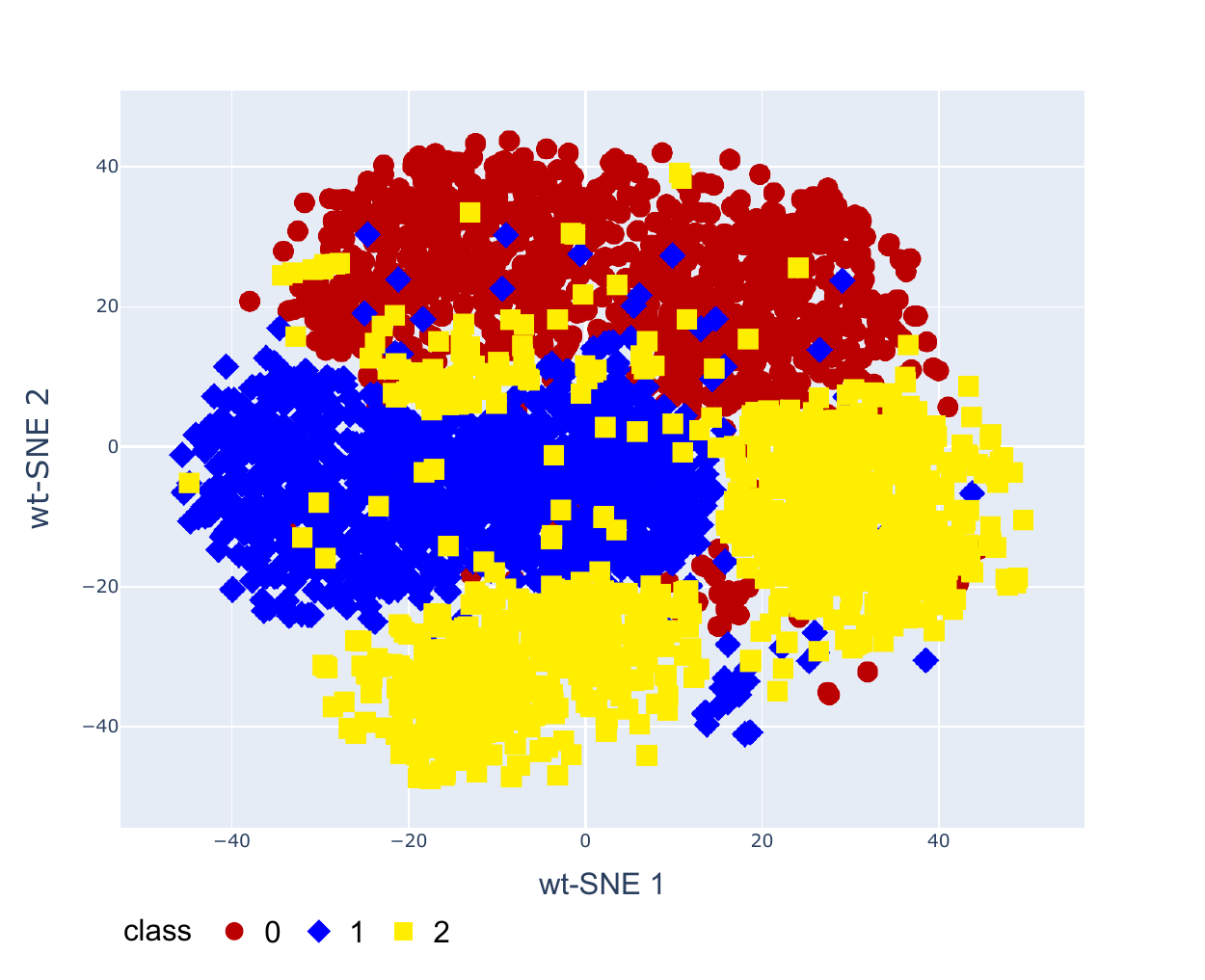}
         \caption{Multiple FSL}
         \label{subfig:multiple_fsl_weighted_tsne_on_extensions_for_synthA}
     \end{subfigure}
\end{figure}

\autoref{tab:weighted_tsne_on_extensions_for_synthA} reports the HD and 2D silhouette coefficients for the SynthA dataset based on the feature weights produced by the original FSL and the extensions, with corresponding visual results shown in \autoref{fig:weighted_tsne_on_extensions_for_synthA}.

\begin{table}[!h]
\centering
\caption[SynthA dataset HD and 2D silhouette coefficients based on the feature weights attributed by original FSL and the proposed extensions.] {\textbf{SynthA dataset HD and 2D silhouette coefficients based on the feature weights attributed by original FSL and the proposed extensions.}
All feature weights were provided from the general weights, not being label-specific.
High-dimensional (HD) silhouette coefficients are calculated before the dimension reduction performed by weighted t-SNE, while 2D silhouette coefficients are calculated after.
``None'' represents the silhouette coefficients without adding feature weights.
In bold is the highest average.
Best results with statistical significance ($p < 0.01$) according to the Kruskal-Wallis and Dunn’s test are highlighted in orange.
}
\label{tab:weighted_tsne_on_extensions_for_synthA}
\begin{tabular}{lrr}
    \hline
    Weight source       & HD silhouette & 2D silhouette        \\ \hline
    None     & .0030 &  -.013       \\
    FSL     & .0105$\pm$.0005 & \cellcolor{orange!20} .1755$\pm$.0560      \\
    Listener      & .0109$\pm$.0003 & \cellcolor{orange!20} .1291$\pm$.0433     \\
    Residual   & .0109$\pm$.0004 & \cellcolor{orange!20} .1755$\pm$.0560    \\ 
    Output Aware & \cellcolor{orange!20} \textbf{.0128$\pm$.0005}  & \cellcolor{orange!20} .1696$\pm$.0693    \\ 
    Multiple FSL & \cellcolor{orange!20} .0123$\pm$.0004 & \cellcolor{orange!20} \textbf{.1773$\pm$.0922}     \\
    \hline
\end{tabular}
\end{table}

\begin{figure*}[!h]
    \centering
     \begin{subfigure}[b]{0.495\textwidth}
         \centering
         \includegraphics[width=\textwidth]{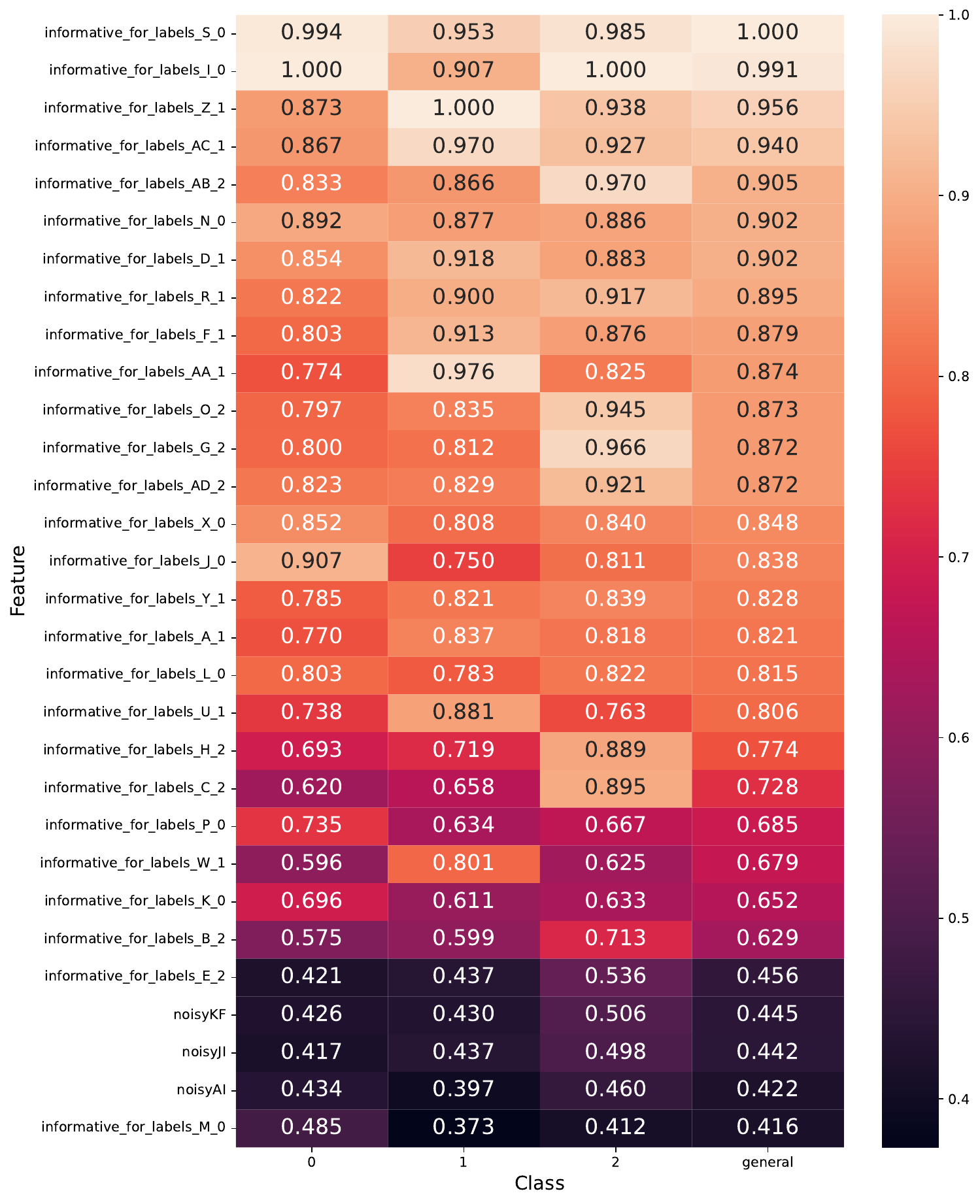}
         \caption{Listener}
         \label{subfig:heatmap_on_extensions_for_synthB_listener}
     \end{subfigure}
     \hfill
     \begin{subfigure}[b]{0.495\textwidth}
         \centering
         \includegraphics[width=\textwidth]{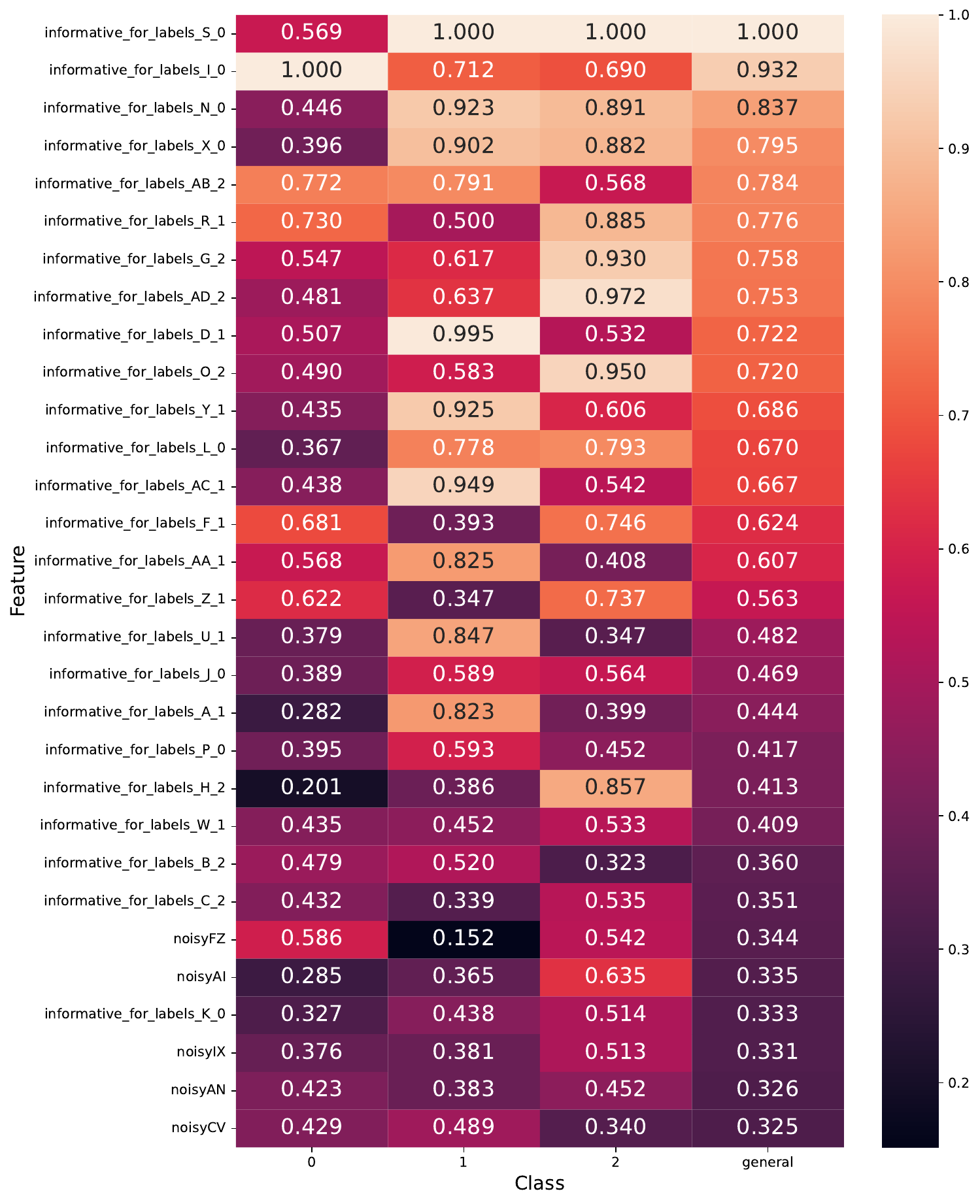}
         \caption{Residual}
         \label{subfig:heatmap_on_extensions_for_synthB_residual}
     \end{subfigure}
     \begin{subfigure}[b]{0.495\textwidth}
         \centering
         \includegraphics[width=\textwidth]{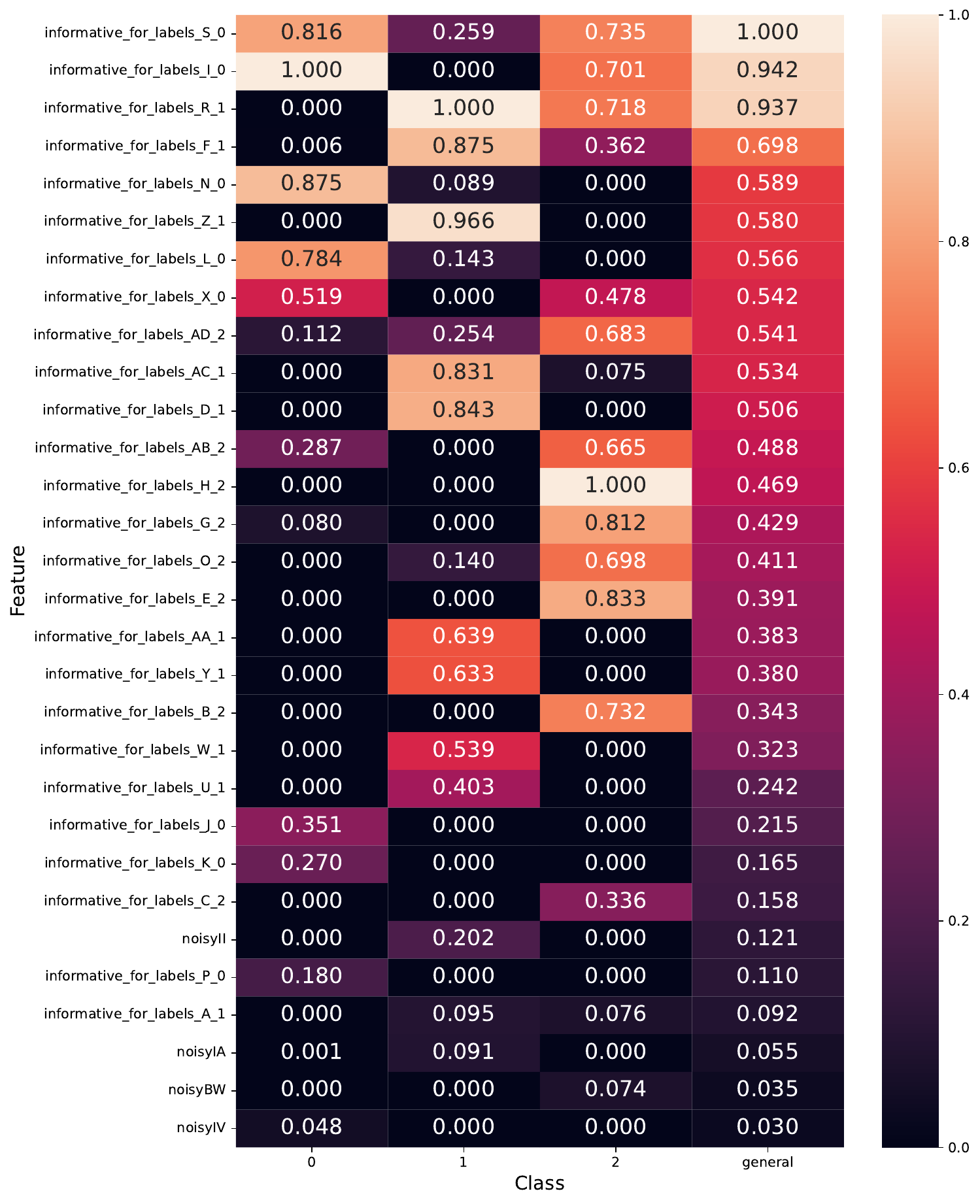}
         \caption{Output Aware}
         \label{subfig:heatmap_on_extensions_for_synthB_output_aware}
     \end{subfigure}
     \hfill
     \begin{subfigure}[b]{0.495\textwidth}
         \centering
         \includegraphics[width=\textwidth]{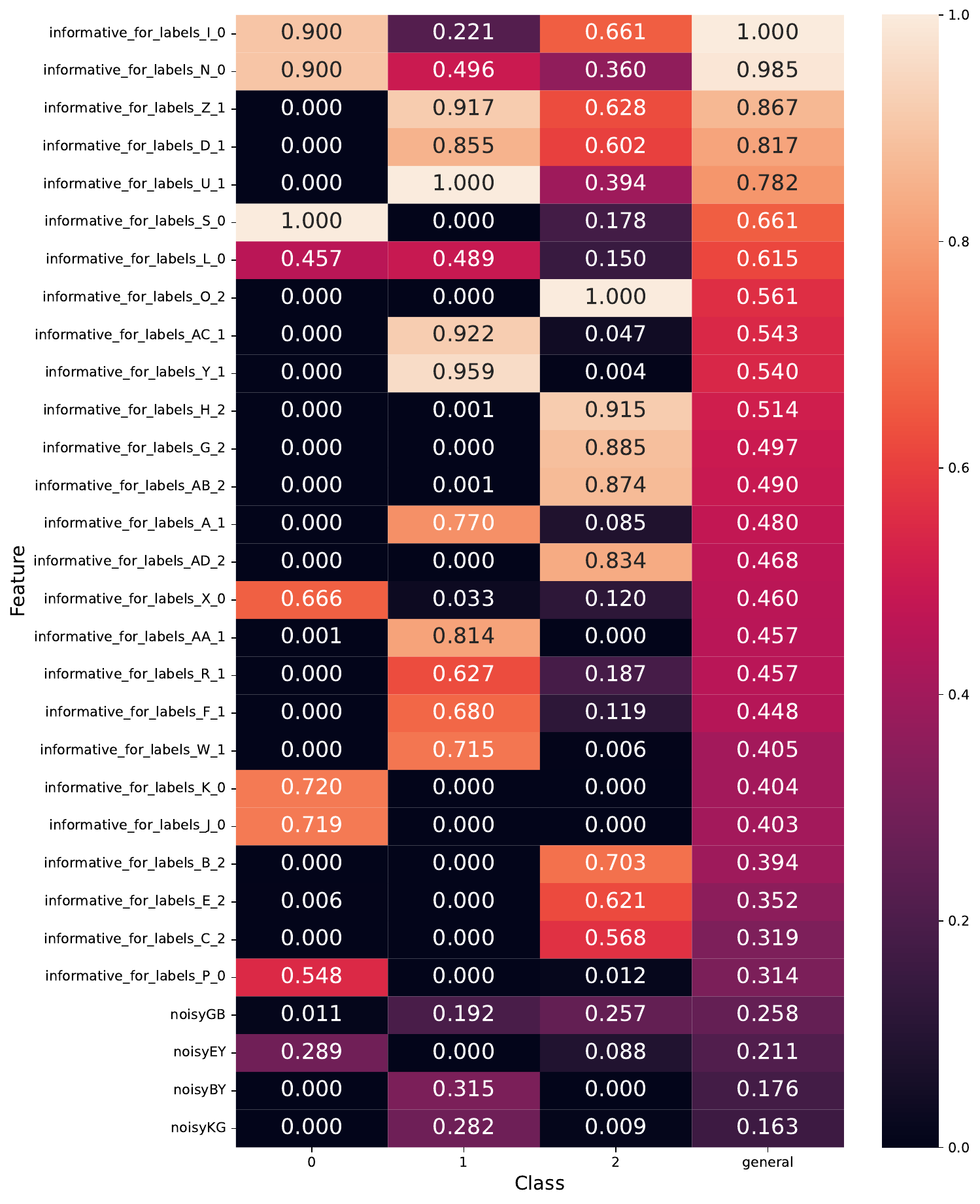}
         \caption{Multiple FSL}
         \label{subfig:heatmap_on_extensions_for_synthB_multiple_fsl}
     \end{subfigure}
    \caption[Heatmap visualization of top-30 general features for each extension from SynthB dataset.
    Four images are displayed.
    The first corresponds to the heatmap generated by the Listener, where all features have similar weights across each label.
    For example, the feature ``informative\_for\_labels\_S\_0'' has a weight of 0.944 for label 0, 0.953 for label 1, 0.985 for label 2, and 1.0 as the general weight.
    Despite this, the heatmap shows that the Listener was able to assign higher weights to relevant features and lower weights to noisy ones.
    The second figure corresponds to the heatmap generated by the Residual, where the values vary.
    Although the weights of each feature differ across labels, they are not necessarily higher for the labels to which the feature is actually relevant.
    The third and fourth figures show the Output Aware and Multiple FSL, respectively.
    In these, it is clearly visible that higher weights are assigned to the label for which the feature is relevant in most cases.
    There are multiple examples where the weight of a feature for one label is close to one, while for the other labels it is close to zero.]
    {\textbf{Heatmap visualization of top-30 general features for each extension from SynthB dataset.}
    Normalized feature weights for each algorithm, ranked by their overall general weight. 
        Feature names prefix indicate if they are informative or noisy, followed by an identification letter sequence, which ensures that all names are unique, and the label which they are related. 
        For example, \textit{informative\_for\_labels\_A\_1} is the informative feature identified by the letter 'A' that is related to label 1, while \textit{noisyAI} is a noisy feature identified by the letter sequence ``AI''.
    }
     \label{fig:heatmap_on_extensions_for_synthB}
\end{figure*}

\autoref{fig:heatmap_on_extensions_for_synthB} exhibits a heatmap with the normalized weights of the top-$30$ features ordered by the general features weights generated by each proposed extension over SynthB dataset.
Relevant features are prefixed with \textit{informative\_for\_labels\_} followed by an identification character sequence and the directly related label.
As example, \textit{informative\_for\_labels\_S\_0} is a relevant feature uniquely identified by the character sequence ``S'' and relevant only for label 0 (suffix ``\textit{\_0}'').
As shown in the heatmap, Multiple FSL and Output Aware assigned the highest weights to label-relevant features.
Listener assigned highest weight values to all relevant features despite their relation with only a specific label, struggling to accurately prioritize features by their relative label.
The \textit{informative\_for\_labels\_R\_1} is an example of feature relevant only for label 1 assigned with similar normalized weights for all labels by Listener, being $0.822$ for label 0, $0.9$ for label 1 and $0.917$ for label 2.
Residual exhibited a more inconsistent behavior. 
In some cases, such as \textit{informative\_for\_labels\_I\_0}, it correctly assigned the highest weight to the relevant label. 
However, in other cases, like \textit{informative\_for\_labels\_S\_0} and \textit{informative\_for\_labels\_N\_0}, it assigned the lowest weights to the correct label.

When comparing feature weights on SynthC, we observed instances where features related to two labels are sometimes considered more relevant to classify a third label, as exemplified in \autoref{tab:informative_per_label_L_0_1_extensions} for the feature \textit{informative\_per\_label\_L\_0\_1} (relevant for the labels 0 and 1) where the highest weight was associated with label 2 instead of 0 and 1.
One possible hypothesis is that the model prioritizes those attributes for differentiating the third label instead of defining the first and second labels that have informative values.

\begin{table}[!h]
\centering
\caption[Analysis of feature weights for \textit{informative\_per\_label\_L\_0\_1. Label 2 had the highest weight for all algorithms.} from SynthC dataset.] {\textbf{Analysis of feature weights for \textit{informative\_per\_label\_L\_0\_1} from SynthC dataset.}
    Each column represents the feature weight given by each algorithm to the feature \textit{informative\_per\_label\_L\_0\_1}.
    Although the feature is directly related to label 0 and label 1, all extensions gave a higher weight, highlighted in bold, related to label 2.
}
\label{tab:informative_per_label_L_0_1_extensions}
\begin{tabular}{lrlll}
    \hline
    Algorithm  & Label 0 & Label 1   & Label 2         \\
    \hline
    Listener  &   .834 & .727 & \textbf{.886}  \\
    Residual  &   .398 & .400 & \textbf{.862}  \\
    Output Aware &   .463 & .513 & \textbf{.599}  \\
    Multiple FSL &   .341 & .626 & \textbf{.811}  \\
    \hline
\end{tabular}
\end{table}

Because FSL and its extensions affect model training, \autoref{tab:comparative_selector_prediction_on_extensions_for_breast} reports their predictive performance on the Breast dataset.

\begin{table}[!h]
\centering
\caption[Individual predictive performance of FSL and proposed extensions on Breast dataset per label.] 
{\textbf{Individual predictive performance of FSL and proposed extensions on Breast dataset per label.}
     The average scores were calculated across 11 independent executions using test dataset.
     For each reported metric we present the average and standard.
     In bold is the highest average of each metric and label.
     Best results with statistical significance ($p < 0.01$) according to the Kruskal-Wallis and Dunn’s test are highlighted in orange.
}
\label{tab:comparative_selector_prediction_on_extensions_for_breast}
\begin{tabular}{lrlll}
    \hline
     Algorithm    &   Label & Precision                                         & Recall                                    & F1 score        \\
    \hline
    \multirow{6}{*}{FSL}            &       0 & \cellcolor{orange!20} \textbf{.990$\pm$.032}      & \cellcolor{orange!20} .977$\pm$.048               & \cellcolor{orange!20} .983$\pm$.029 \\
                                    &       1 & \cellcolor{orange!20} .951$\pm$.084               & \cellcolor{orange!20} \textbf{.985$\pm$.048}      & \cellcolor{orange!20} .965$\pm$.049 \\
                                    &       2 & \cellcolor{orange!20} \textbf{1.0$\pm$.0}         & \cellcolor{orange!20} \textbf{1.0$\pm$.0}         & \cellcolor{orange!20} \textbf{1.0$\pm$.0}         \\
                                    &       3 & \cellcolor{orange!20} \textbf{1.0$\pm$.0}         & \cellcolor{orange!20} \textbf{1.0$\pm$.0}         & \cellcolor{orange!20} \textbf{1.0$\pm$.0}         \\
                                    &       4 & \cellcolor{orange!20} \textbf{.974$\pm$.055}      & \cellcolor{orange!20} \textbf{1.0$\pm$.0}         & \cellcolor{orange!20} \textbf{.986$\pm$.030} \\
                                    &       5 & \cellcolor{orange!20} \textbf{1.0$\pm$.0}         & \cellcolor{orange!20} \textbf{.939$\pm$.080}      & \cellcolor{orange!20} \textbf{.967$\pm$.044} \\
    \hline
    \multirow{6}{*}{Listener}       &       0 & \cellcolor{orange!20} .980$\pm$.043               & \cellcolor{orange!20} \textbf{1.0$\pm$.0}         & \cellcolor{orange!20} .989$\pm$.023  \\
                                    &       1 & \cellcolor{orange!20} .974$\pm$.055               & \cellcolor{orange!20} .970$\pm$.064               & \cellcolor{orange!20} .969$\pm$.041 \\
                                    &       2 & \cellcolor{orange!20} \textbf{1.0$\pm$.0}         & \cellcolor{orange!20} \textbf{1.0$\pm$.0}         & \cellcolor{orange!20} \textbf{1.0$\pm$.0}         \\
                                    &       3 & \cellcolor{orange!20} \textbf{1.0$\pm$.0}         & \cellcolor{orange!20} \textbf{1.0$\pm$.0}         & \cellcolor{orange!20} \textbf{1.0$\pm$.0}         \\
                                    &       4 & \cellcolor{orange!20} .964$\pm$.079               & \cellcolor{orange!20} \textbf{1.0$\pm$.0}         & \cellcolor{orange!20} .980$\pm$.045 \\
                                    &       5 & \cellcolor{orange!20} \textbf{1.0$\pm$.0}         & \cellcolor{orange!20} .924$\pm$.109               & \cellcolor{orange!20} .957$\pm$.064 \\
    \hline
    \multirow{6}{*}{Residual}       &       0 & \cellcolor{orange!20} .954$\pm$.079               & .784$\pm$.193                                     & .840$\pm$.104 \\
                                    &       1 & .581$\pm$.397                                     & .403$\pm$.316                                     & .403$\pm$.283 \\
                                    &       2 & \cellcolor{orange!20} \textbf{1.0$\pm$.0}         & .727$\pm$.278                                     & .809$\pm$.207 \\
                                    &       3 & \cellcolor{orange!20} .727$\pm$.445               & \cellcolor{orange!20} .682$\pm$.440               & \cellcolor{orange!20} .697$\pm$.437   \\
                                    &       4 & .639$\pm$.320                                     & \cellcolor{orange!20} .818$\pm$.305               & .818$\pm$.256 \\
                                    &       5 & .626$\pm$.06859                                   & \cellcolor{orange!20} .682$\pm$.305               & .536$\pm$.110  \\
    \hline
    \multirow{6}{*}{Output Aware}   &       0 & \cellcolor{orange!20} .941$\pm$.094               & \cellcolor{orange!20} .898$\pm$.231               & \cellcolor{orange!20} .906$\pm$.185  \\
                                    &       1 & \cellcolor{orange!20} .802$\pm$.282               & .682$\pm$.297                                     & .705$\pm$.246  \\
                                    &       2 & \cellcolor{orange!20} .868$\pm$.236               & \cellcolor{orange!20} .970$\pm$.096               & \cellcolor{orange!20} .890$\pm$.173 \\
                                    &       3 & \cellcolor{orange!20} .920$\pm$.254               & \cellcolor{orange!20} \textbf{1.0$\pm$.0}         & \cellcolor{orange!20} .928$\pm$.227 \\
                                    &       4 & \cellcolor{orange!20} .847$\pm$.280               & \cellcolor{orange!20} .894$\pm$.287               & \cellcolor{orange!20} .867$\pm$.278  \\
                                    &       5 & \cellcolor{orange!20} .864$\pm$.02307             & \cellcolor{orange!20} .712$\pm$.285               & .715$\pm$.155  \\
    \hline
    \multirow{6}{*}{Multiple FSL}   &       0 & \cellcolor{orange!20} \textbf{.990$\pm$.032}      & \cellcolor{orange!20} \textbf{1.0$\pm$.0}         & \cellcolor{orange!20} \textbf{.995$\pm$.017} \\
                                    &       1 & \cellcolor{orange!20} \textbf{.987$\pm$.041}      & \cellcolor{orange!20} \textbf{.985$\pm$.048}      & \cellcolor{orange!20} \textbf{.985$\pm$.033} \\
                                    &       2 & \cellcolor{orange!20} \textbf{1.0$\pm$.0}         & \cellcolor{orange!20} \textbf{1.0$\pm$.0}         & \cellcolor{orange!20} \textbf{1.0$\pm$.0}        \\
                                    &       3 & \cellcolor{orange!20} \textbf{1.0$\pm$.0}         & \cellcolor{orange!20} \textbf{1.0$\pm$.0}         & \cellcolor{orange!20} \textbf{1.0$\pm$.0}         \\
                                    &       4 & \cellcolor{orange!20} .961$\pm$.064               & \cellcolor{orange!20} \textbf{1.0$\pm$.0}         & \cellcolor{orange!20} .979$\pm$.034 \\
                                    &       5 & \cellcolor{orange!20} \textbf{1.0$\pm$.0}         & \cellcolor{orange!20} \textbf{.939$\pm$.107}      & \cellcolor{orange!20} .965$\pm$.063 \\
    \hline
\end{tabular}
\end{table}

\begin{figure}[!h]
    \caption[
    SynthA weighted t-SNE visualization using the feature weights attributed by Multiple FSL and compared algorithms.
    The first figure shows the application of t-SNE to the SynthA dataset without feature weights.
    All three classes largely overlap in the center of the plot, forming a circular pattern with no evident separation. 
    The second figure uses the Decision Tree weights, the circular structure remains, but it becomes divided into  tree mostly distinct regions: the bottom-left corresponds to class 1, the top-left to class 0, and the right to class 2.
    However there is a high amount of samples from different classes that overlap other classes regions.
    The third figure uses the DeepSHAP weights, being divided into tree distinct regions: the bottom-left corresponds to class 2, the top-left to class 0, and the right and bottom-right to class 1.
    As Decision Tree, DeepSHAP also contains a high amount of samples from different classes that overlap other classes regions.
    The forth figure uses the Kruskal Wallis weights, showing only a slight improvement when compared with the t-SNE without feature weights.
    The samples from all classes stills overlapping each other with a high frequency, however it is possible to notice that samples from class 0 are mostly concentrated on the top, class 1 on the bottom and class 2 on the middle of the graph.
    The fifth figure uses the LIME weights, and as Kruskal Wallis, it shows only a slight improvement when compared with the t-SNE without feature weights. Its also more difficult to understand where the samples from each class are mostly concentrated.
    The sixth figure uses the LASSO weights, being divided into five slightly distinct regions: a smaller top corresponds to class 1 together with a second region that goes from the right to the bottom-center of the graph, class 2 is present on the top of the graph around the class 1 samples and on the center, and class 0 samples are mostly on the bottom-right and bottom-left of the graph.
    The seventh figure uses the Multiple FSL weights, having the better definition of regions between the algorithms, being five:
    class 2 samples are grouped on a big region at the top of the graph, class 1 are grouped on a big region that goes from the left to the right of the graph's center, and class 0 are separated into tree regions, being the bigger one at the bottom of the graph, followed by a region at the top-right and a smaller one at the top-left.
    Compared with the other samples, Multiple FSL contains less samples from a class overlapping other classes dominant regions.
    The eighth figure uses the Random Forest weights, having the six regions: class 0 is more present on the top to the center of the graph, class 1 is more present on the bottom-right, and class 2 is present on the left, top-right and bottom overlapping with the class 1 samples.
    The ninth figure uses the ReliefF weights, having five regions: class 0 are concentrated mostly on the right, class 1 on the left and class 2 on the middle, having also two smaller clusters overlapping with class 1 samples on the right of the graph.
    On all experiments the samples are occupying a circular regions in the middle of the graph.
    ]
    {\textbf{SynthA weighted t-SNE visualization using the feature weights attributed by Multiple FSL and compared algorithms.}
        \autoref{subfig:weighted_tsne_on_comparison_for_synthA_no_weight} presents a t-SNE visualization of the data without feature weights. 
        In contrast other figures employ weighted t-SNE, utilizing the generated weights from each feature weighting algorithm.
    }
     \centering
     \label{fig:weighted_tsne_on_comparison_for_synthA}
     \begin{subfigure}[b]{0.3285\textwidth}
         \centering
         \includegraphics[width=\textwidth]{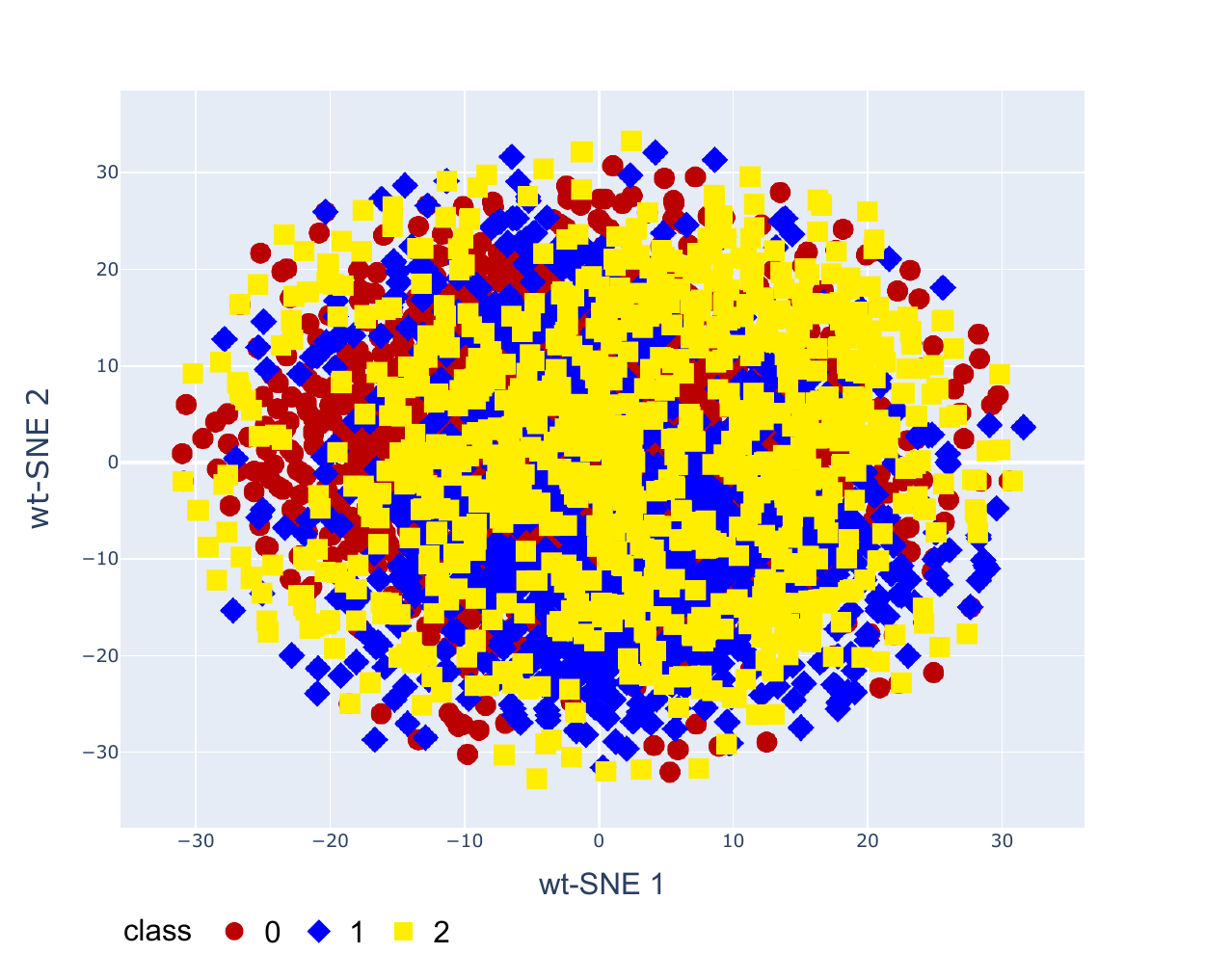}
         \caption{Without weights}
         \label{subfig:weighted_tsne_on_comparison_for_synthA_no_weight}
     \end{subfigure}
     \begin{subfigure}[b]{0.3285\textwidth}
         \centering
         \includegraphics[width=\textwidth]{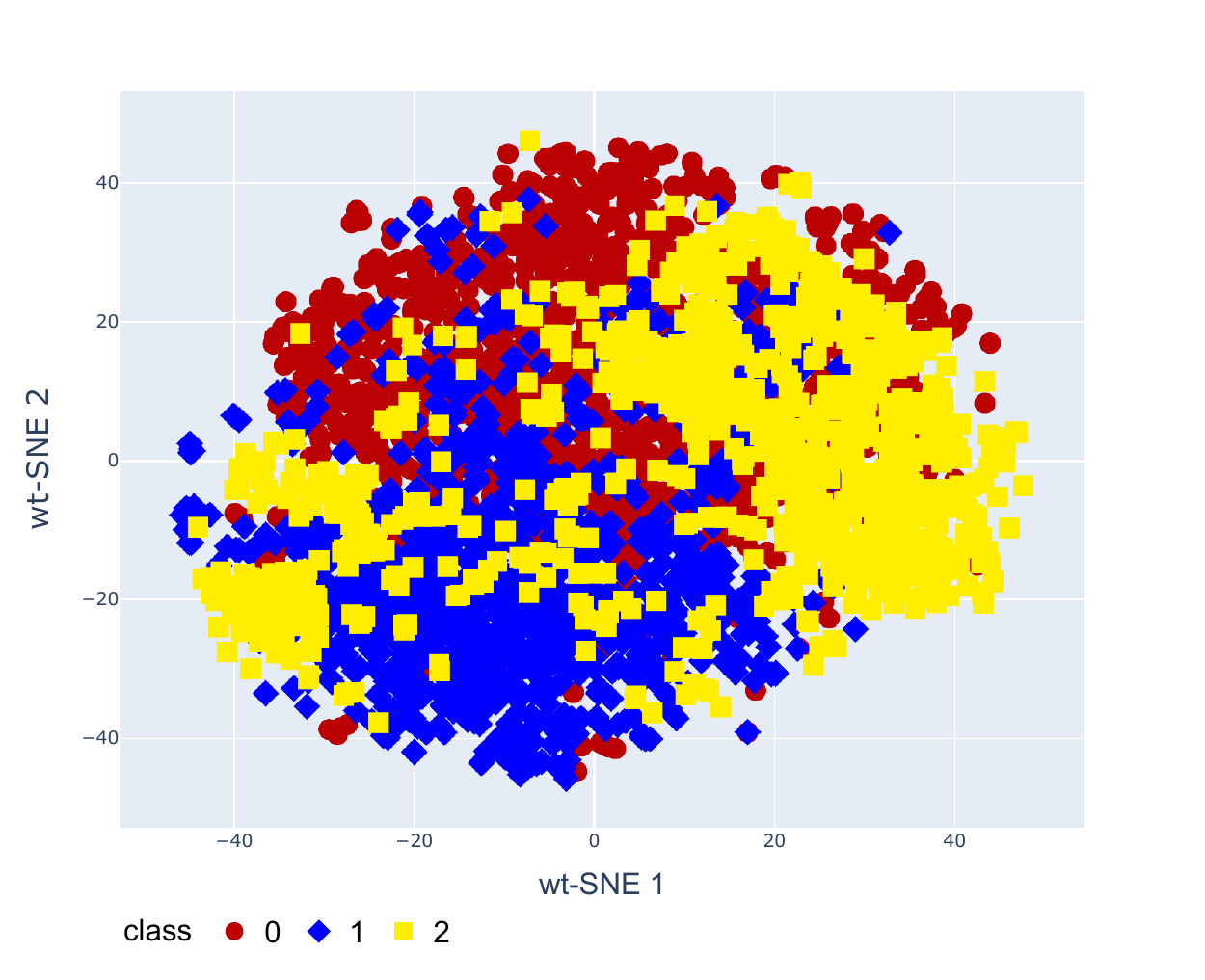}
         \caption{Decision Tree}
        \label{subfig:weighted_tsne_on_comparison_for_synthA_decision_tree}
     \end{subfigure}
     \begin{subfigure}[b]{0.3285\textwidth}
         \centering
         \includegraphics[width=\textwidth]{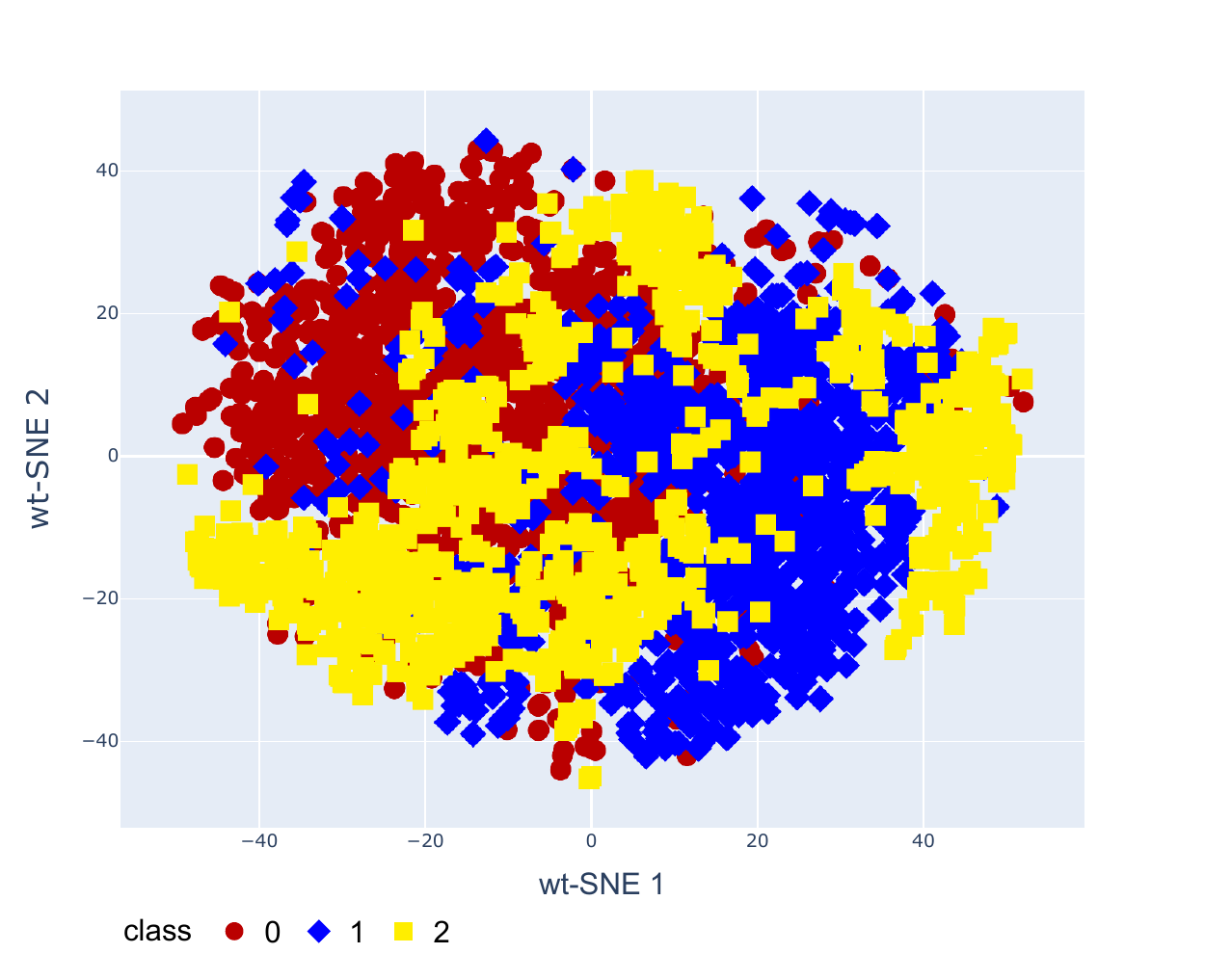}
         \caption{DeepSHAP}
         \label{subfig:weighted_tsne_on_comparison_for_synthA_deep_shap}
     \end{subfigure}
     \begin{subfigure}[b]{0.3285\textwidth}
         \centering
         \includegraphics[width=\textwidth]{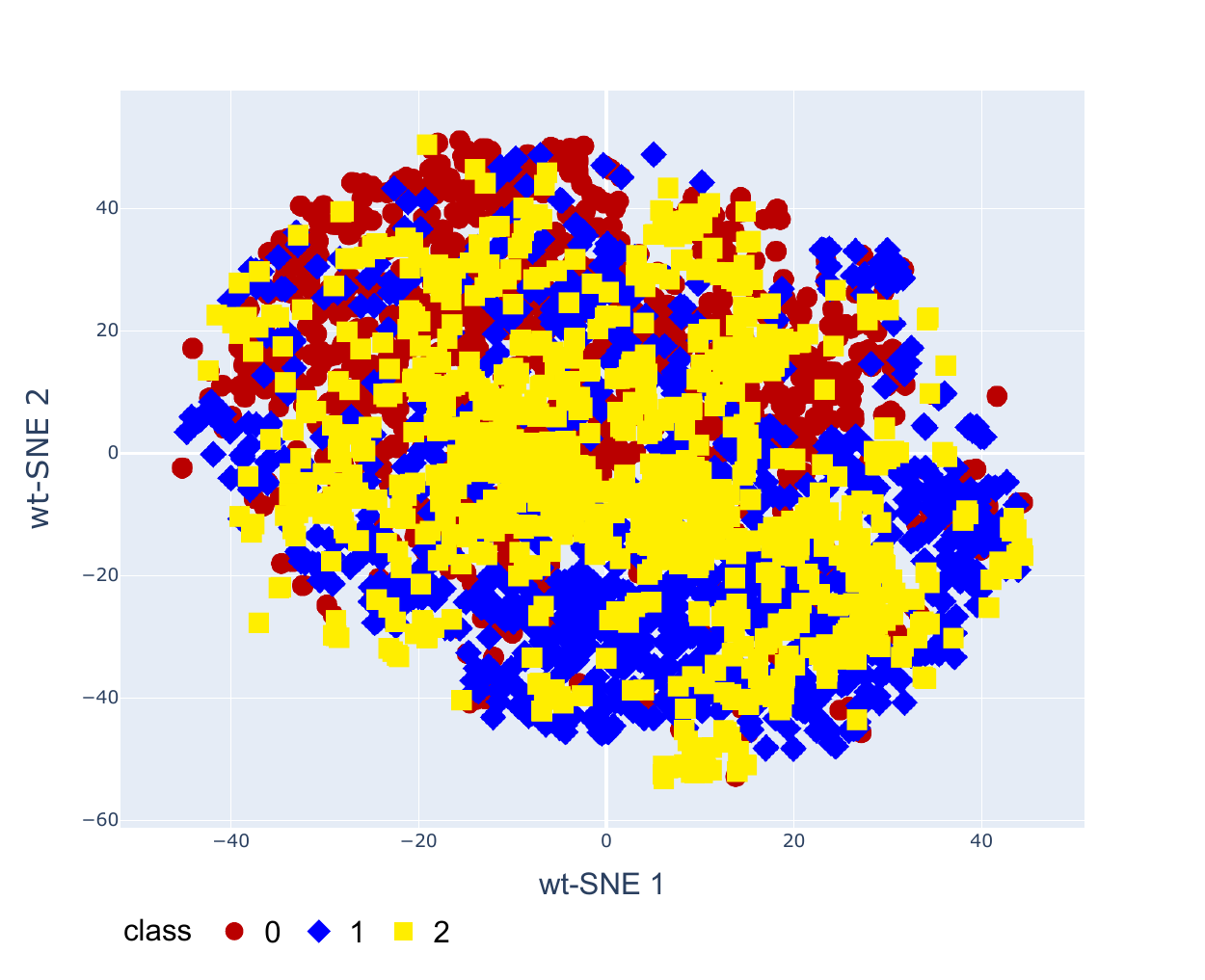}
         \caption{Kruskal Wallis}
         \label{subfig:weighted_tsne_on_comparison_for_synthA_kruskal}
     \end{subfigure}
     \begin{subfigure}[b]{0.3285\textwidth}
         \centering
         \includegraphics[width=\textwidth]{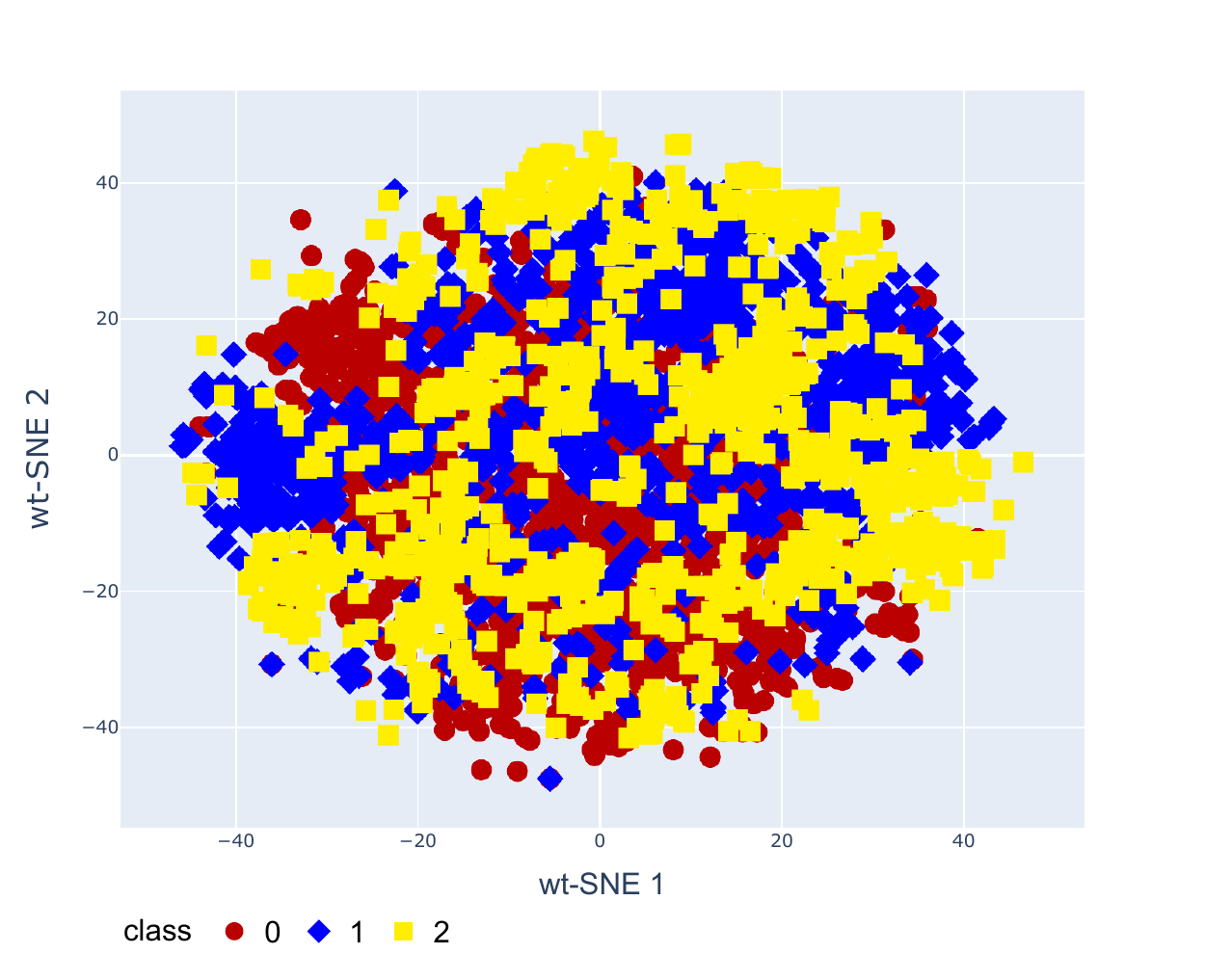}
         \caption{LIME}
         \label{subfig:weighted_tsne_on_comparison_for_synthA_lime}
     \end{subfigure}
     \begin{subfigure}[b]{0.3285\textwidth}
         \centering
         \includegraphics[width=\textwidth]{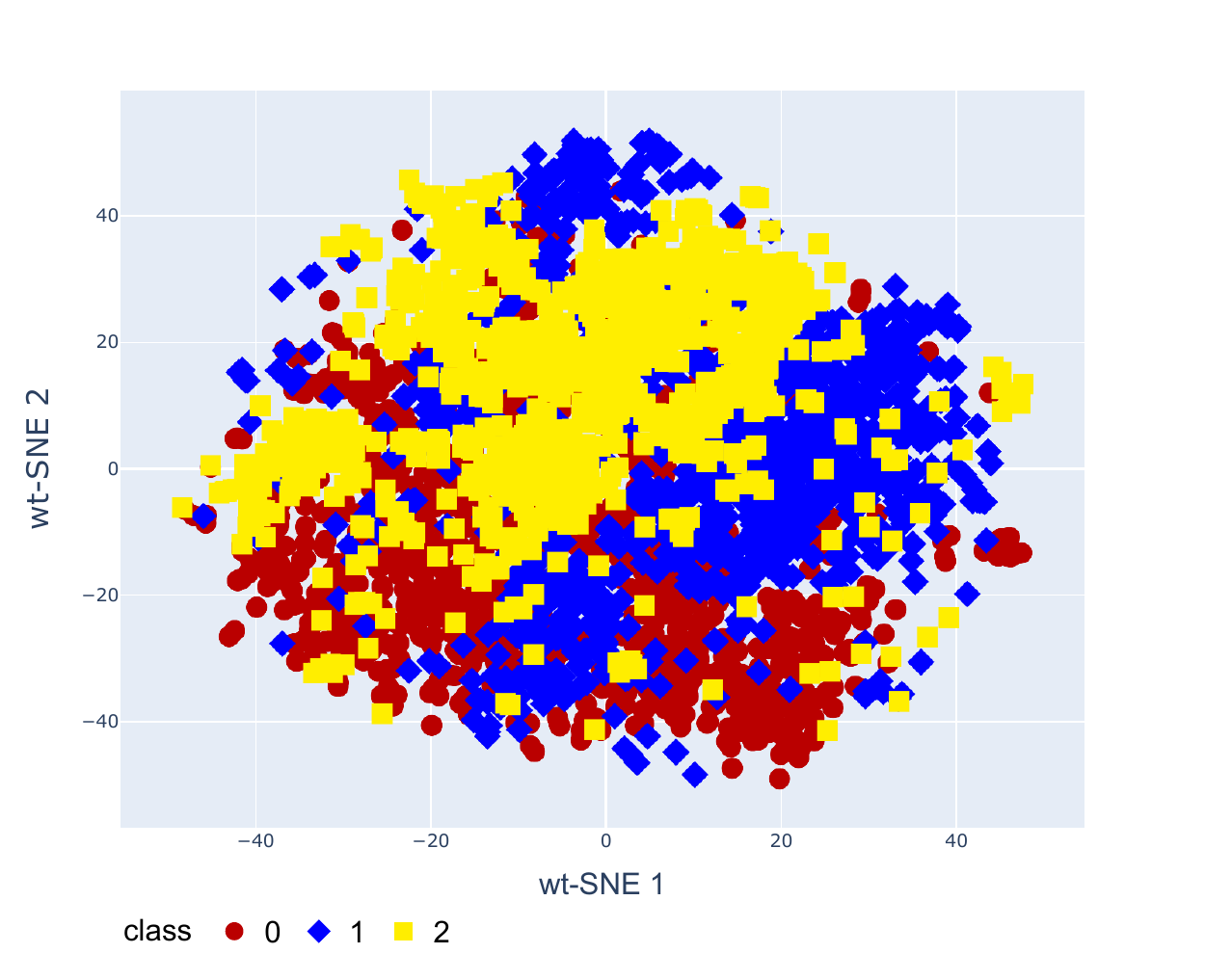}
         \caption{Lasso}
         \label{subfig:weighted_tsne_on_comparison_for_synthA_lasso}
     \end{subfigure}
     \begin{subfigure}[b]{0.3285\textwidth}
         \centering
         \includegraphics[width=\textwidth]{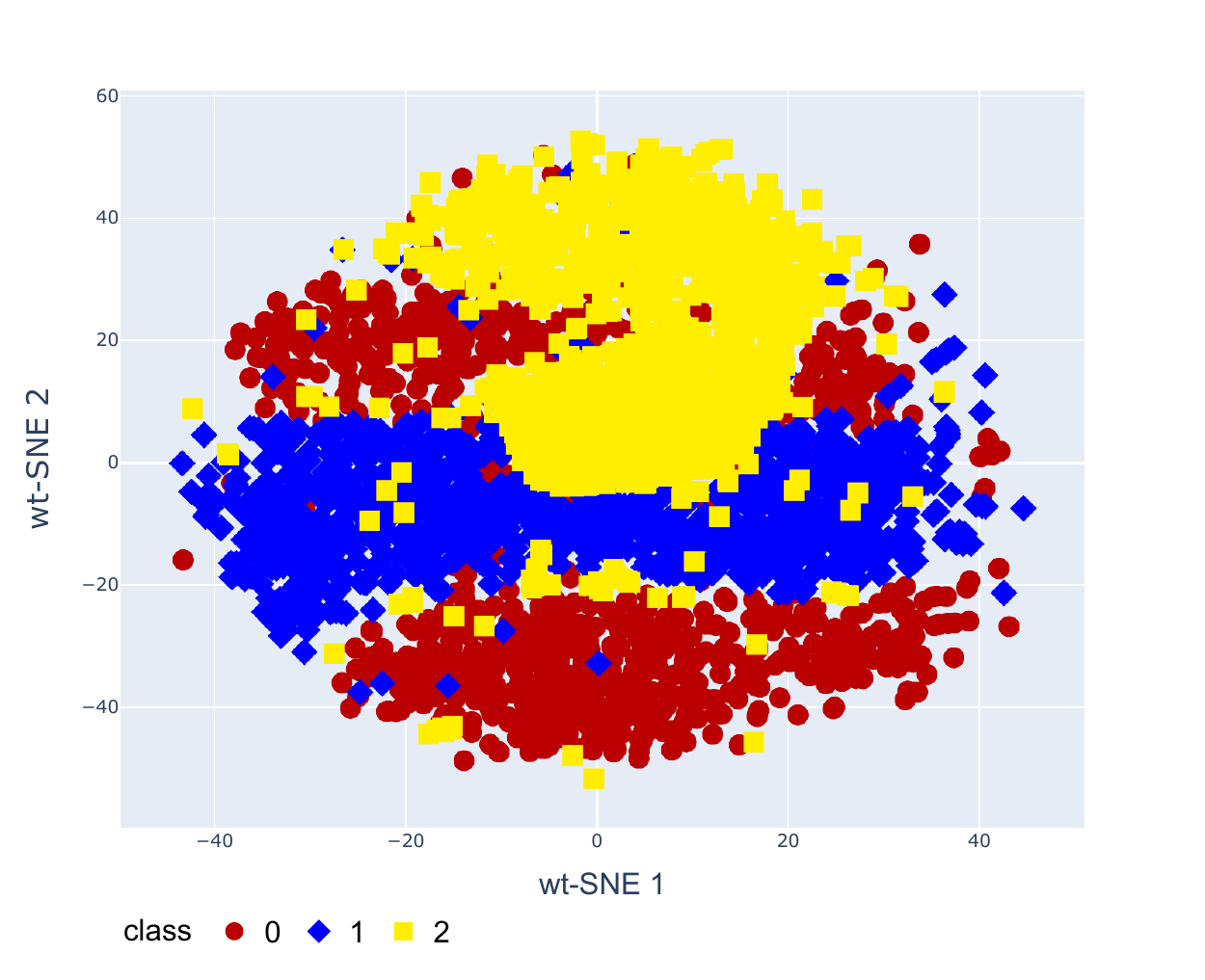}
         \caption{Multiple FSL}
         \label{subfig:weighted_tsne_on_comparison_for_synthA_multiplefsl}
     \end{subfigure}
     \begin{subfigure}[b]{0.3285\textwidth}
         \centering
         \includegraphics[width=\textwidth]{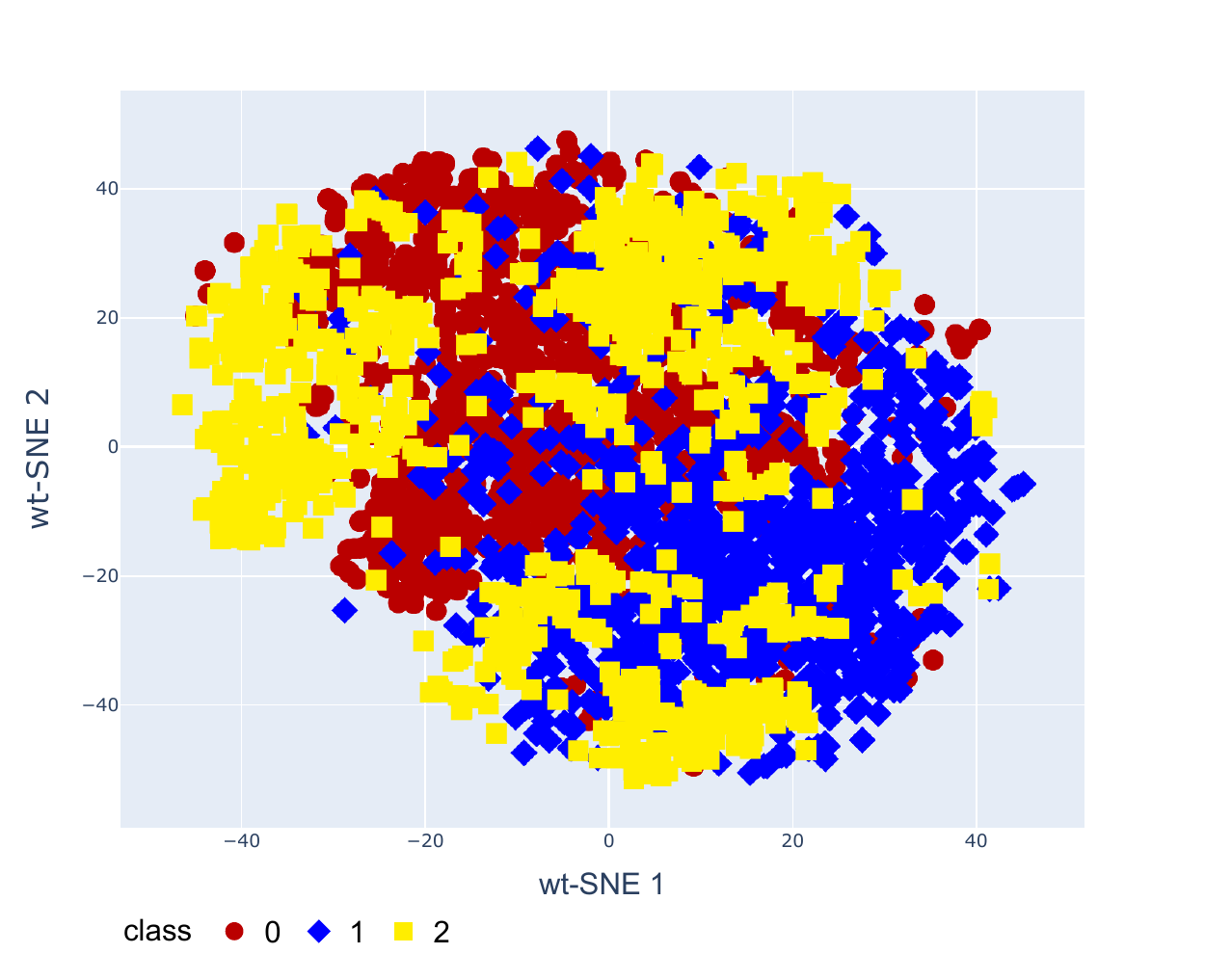}
         \caption{Random Forest}
         \label{subfig:weighted_tsne_on_comparison_for_synthA_random_forest}
     \end{subfigure}
     \begin{subfigure}[b]{0.3285\textwidth}
         \centering
         \includegraphics[width=\textwidth]{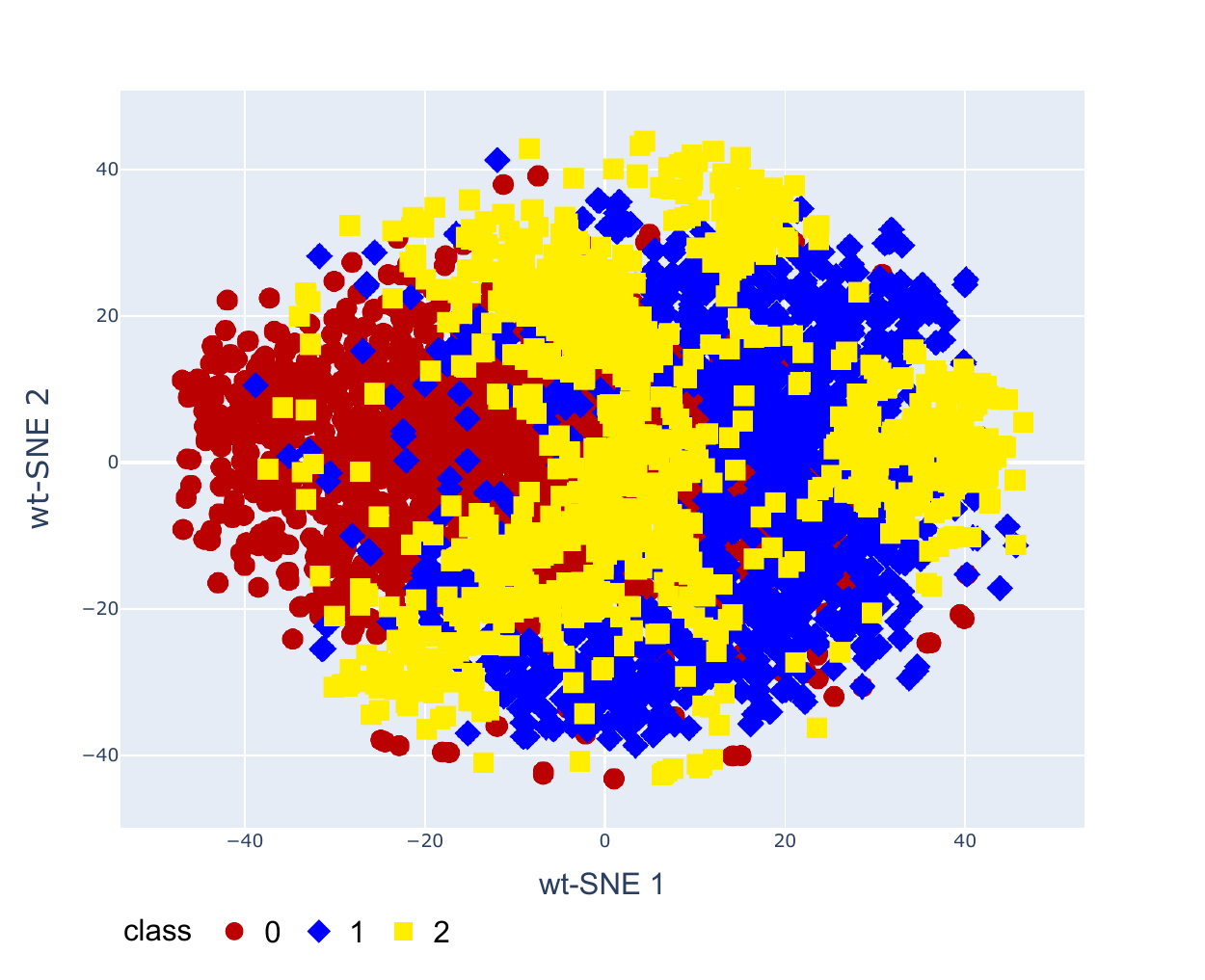}
         \caption{ReliefF}
         \label{subfig:weighted_tsne_on_comparison_for_synthA_relieff}
     \end{subfigure}
\end{figure}

\begin{figure}[!h]
    \caption[
    SynthB weighted t-SNE visualization using the feature weights attributed to label 1.
    The first figure shows the application of t-SNE to the SynthB dataset without feature weights.
    Class 1 samples are mostly on the center-left of the graph, with a minority of samples distributed on all other regions.
    All samples are grouped in the center of the graph forming a circular region.
    The second figure uses the LASSO weights, the circular structure remains, but it becomes divided into  four mostly distinct regions: class 0 is distributed in two main clusters on the top-left and bottom-right, class 1 is mainly concentrated on the bottom-left and class 2 on a region from the center to the top-right of the graph.
    All classes were improved by the LASSO weights attributed to label 1, however class 1 had a higher separation from the others.
    The third figure uses the LIME weights, the circular structure remains with a higher distortion, and internally divided into tree mostly distinct regions: class 0 is distributed in the left with a high overlap with the class 2 region which is distributed from the center to the left of the circular region, and class 1 is mainly concentrated on the right, being the one with the bigger distinction between the tree classes.
    LIME improved mainly the class 1 samples, separating them from the others as expected for label-specific feature weights.
    The fourth figure uses the DeepSHAP feature weights, which internally divides the circular region into two mostly distinct regions: class 1 samples are mostly on the left of the graph and class 0 and 2 samples are overlapping each other from the center to the right.
    LIME improved mainly the class 1 samples, separating them from the others as expected for label-specific feature weights.
    The fifth figure uses the Multiple FSL feature weights, and has similar results.
    Class 1 samples are grouped on a distinct region at the bottom-right of the circle and class 0 and 2 are completely overlapping each other from the center to the top-left of the graph.
    Between the figures, Multiple FSL had the higher overlap between class 0 and 2 samples only.
    ]
    {\textbf{SynthB weighted t-SNE visualization using the feature weights attributed to label 1.}
        \autoref{subfig:weighted_tsne_on_comparison_for_synthB_no_weight} presents a t-SNE visualization of the data without feature weights. 
        In contrast other figures employ weighted t-SNE, utilizing the generated weights from each algorithm specifically related to label 1 (blue diamonds).
     }
     \centering
     \label{fig:weighted_tsne_on_comparison_for_synthB}
     \begin{subfigure}[b]{0.48\textwidth}
         \centering
         \includegraphics[width=\textwidth]{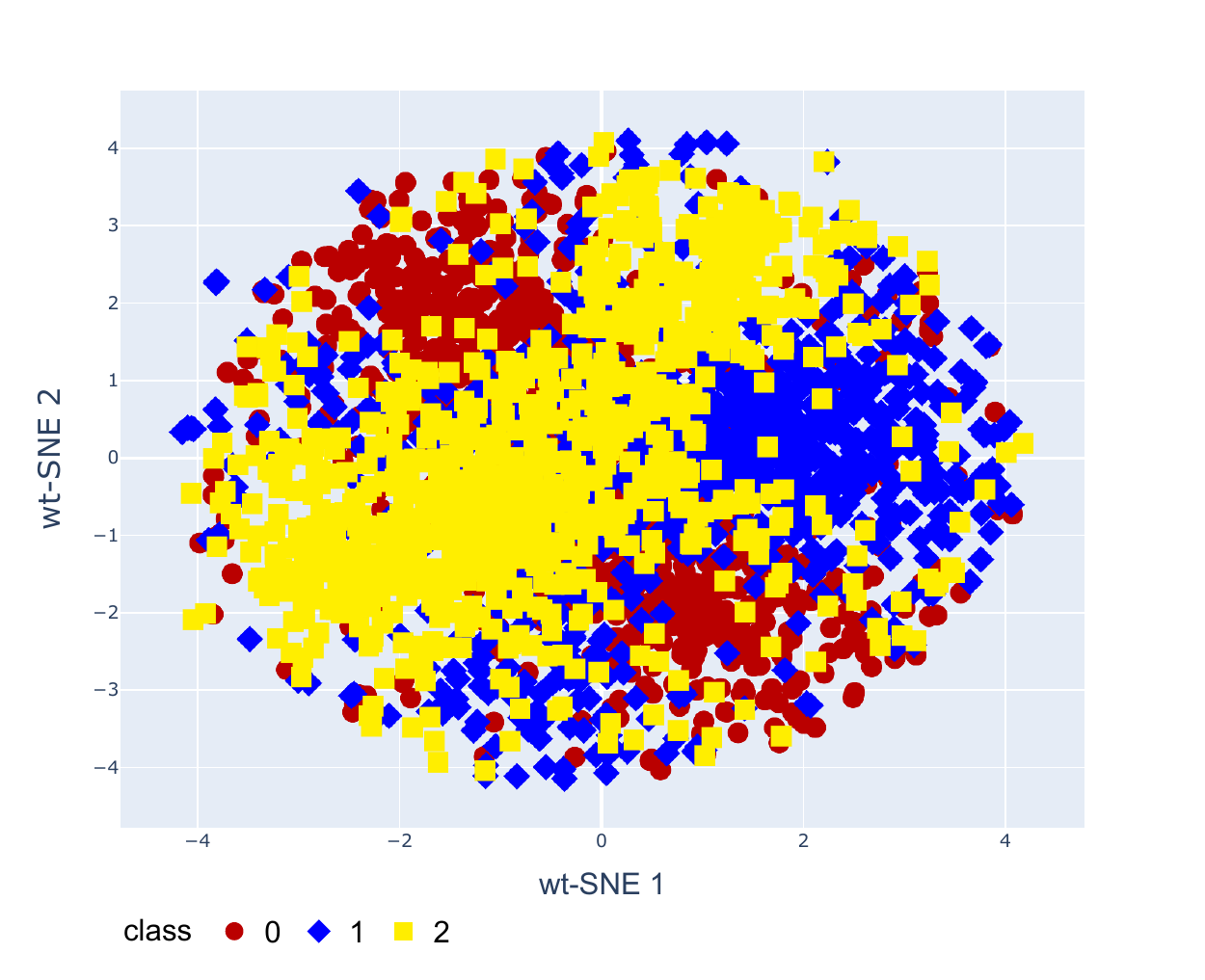}
         \caption{Without weights}
         \label{subfig:weighted_tsne_on_comparison_for_synthB_no_weight}
     \end{subfigure}
     \hfill
     \begin{subfigure}[b]{0.48\textwidth}
         \centering
         \includegraphics[width=\textwidth]{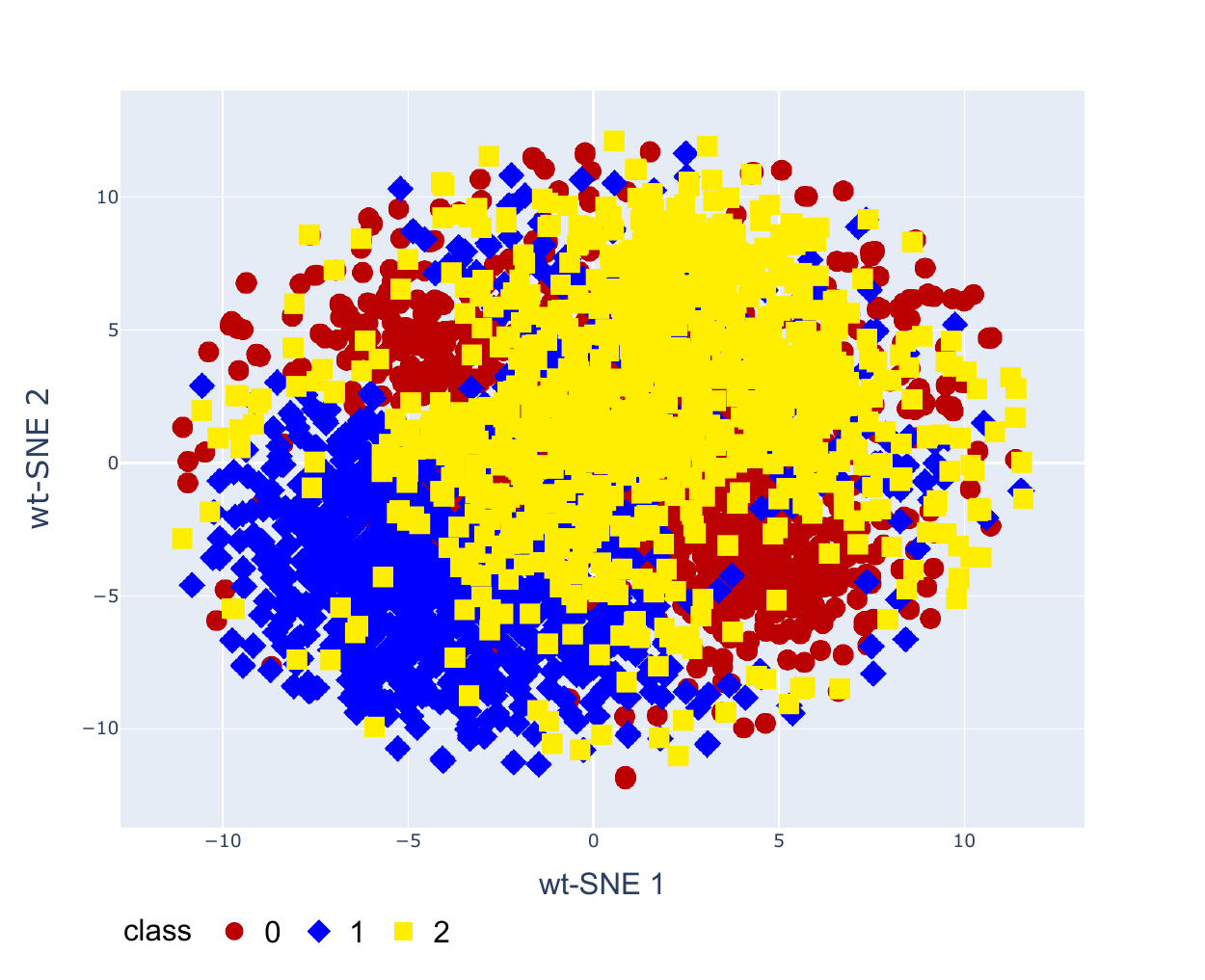}
         \caption{Lasso}
         \label{subfig:weighted_tsne_on_comparison_for_synthB_lasso}
     \end{subfigure}
     \begin{subfigure}[b]{0.48\textwidth}
         \centering
         \includegraphics[width=\textwidth]{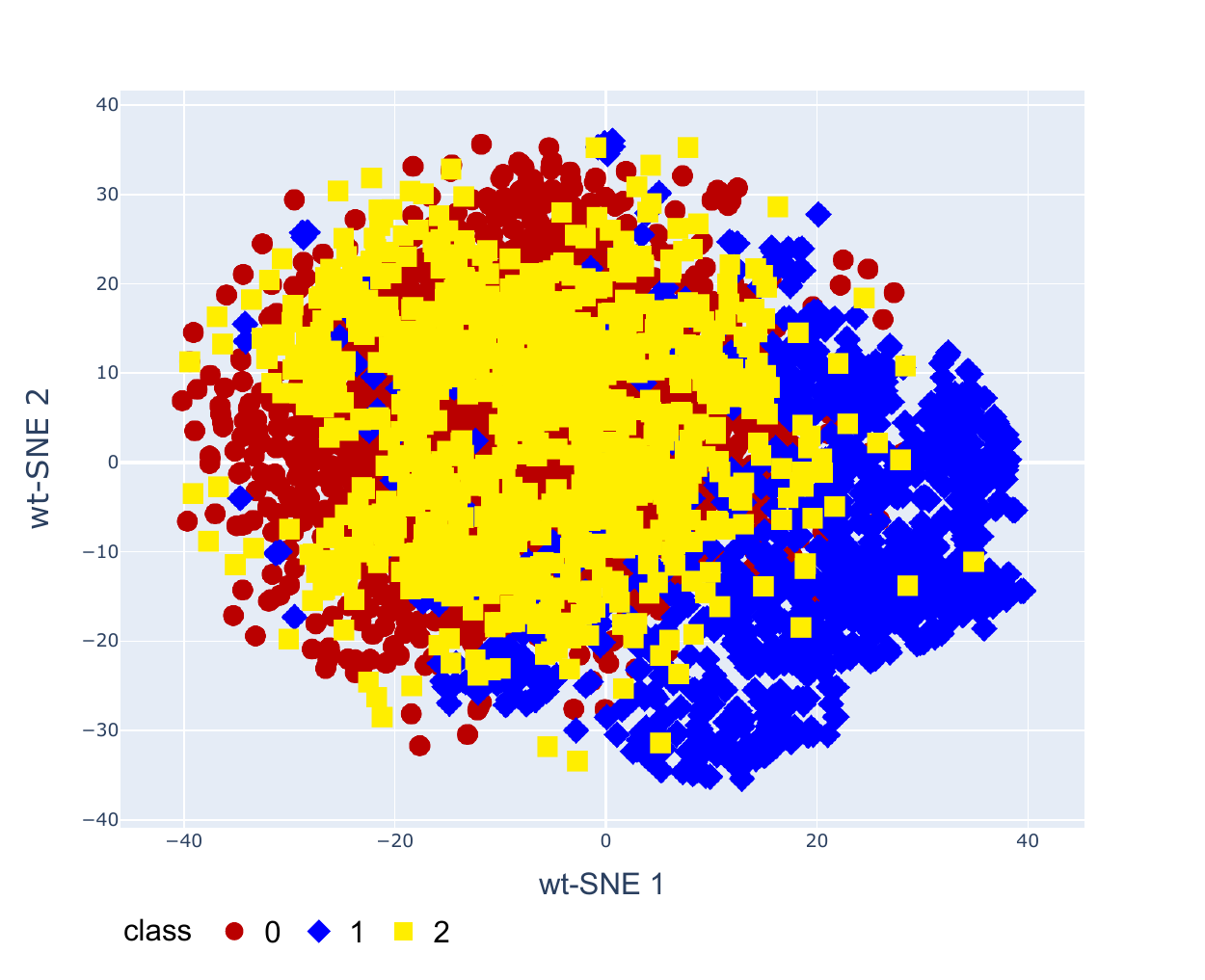}
         \caption{LIME}
         \label{subfig:weighted_tsne_on_comparison_for_synthB_lime}
     \end{subfigure}
     \hfill
     \begin{subfigure}[b]{0.48\textwidth}
         \centering
         \includegraphics[width=\textwidth]{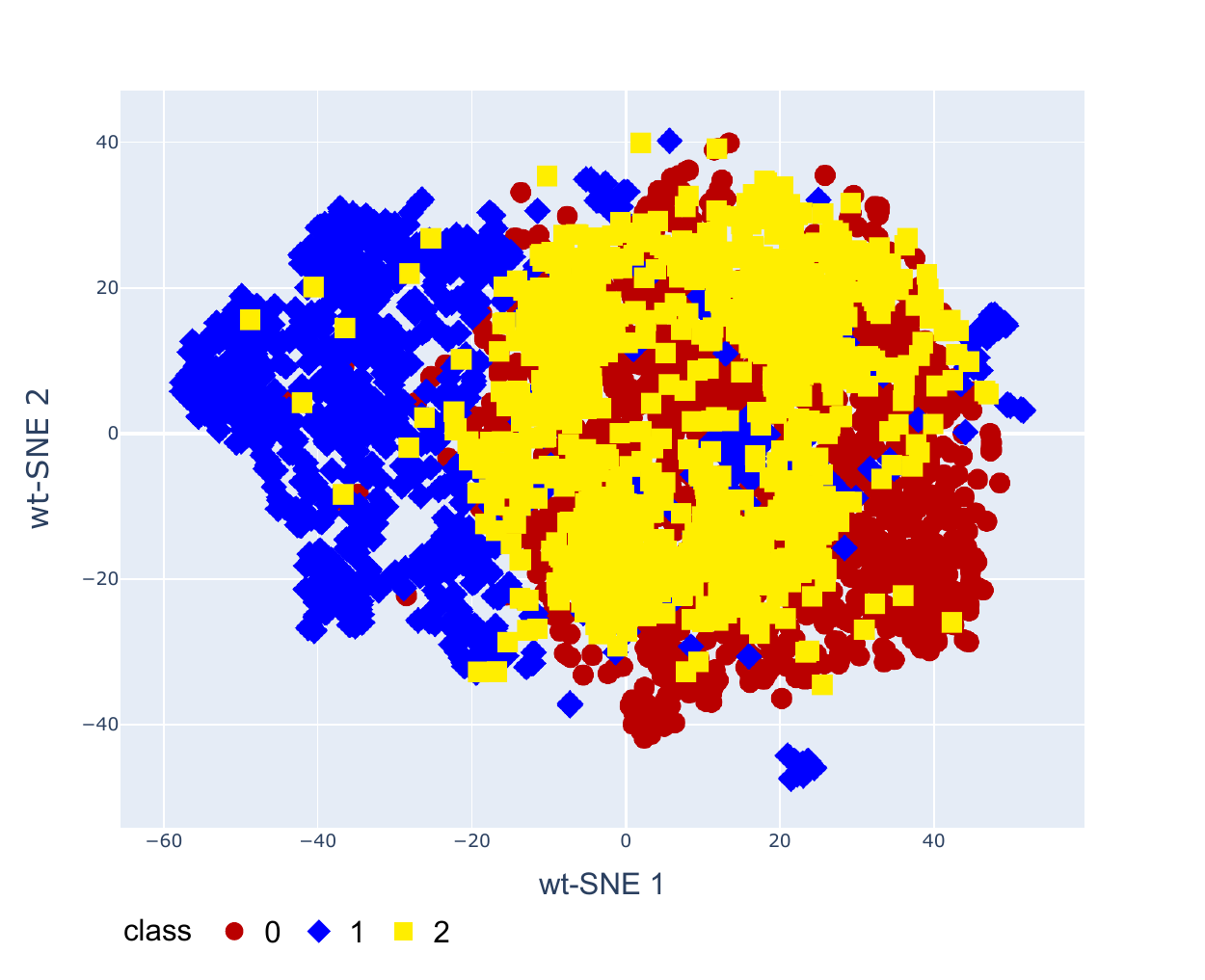}
         \caption{DeepSHAP}
        \label{subfig:weighted_tsne_on_comparison_for_synthB_deepshap}
     \end{subfigure}
     \begin{subfigure}[b]{0.48\textwidth}
         \centering
         \includegraphics[width=\textwidth]{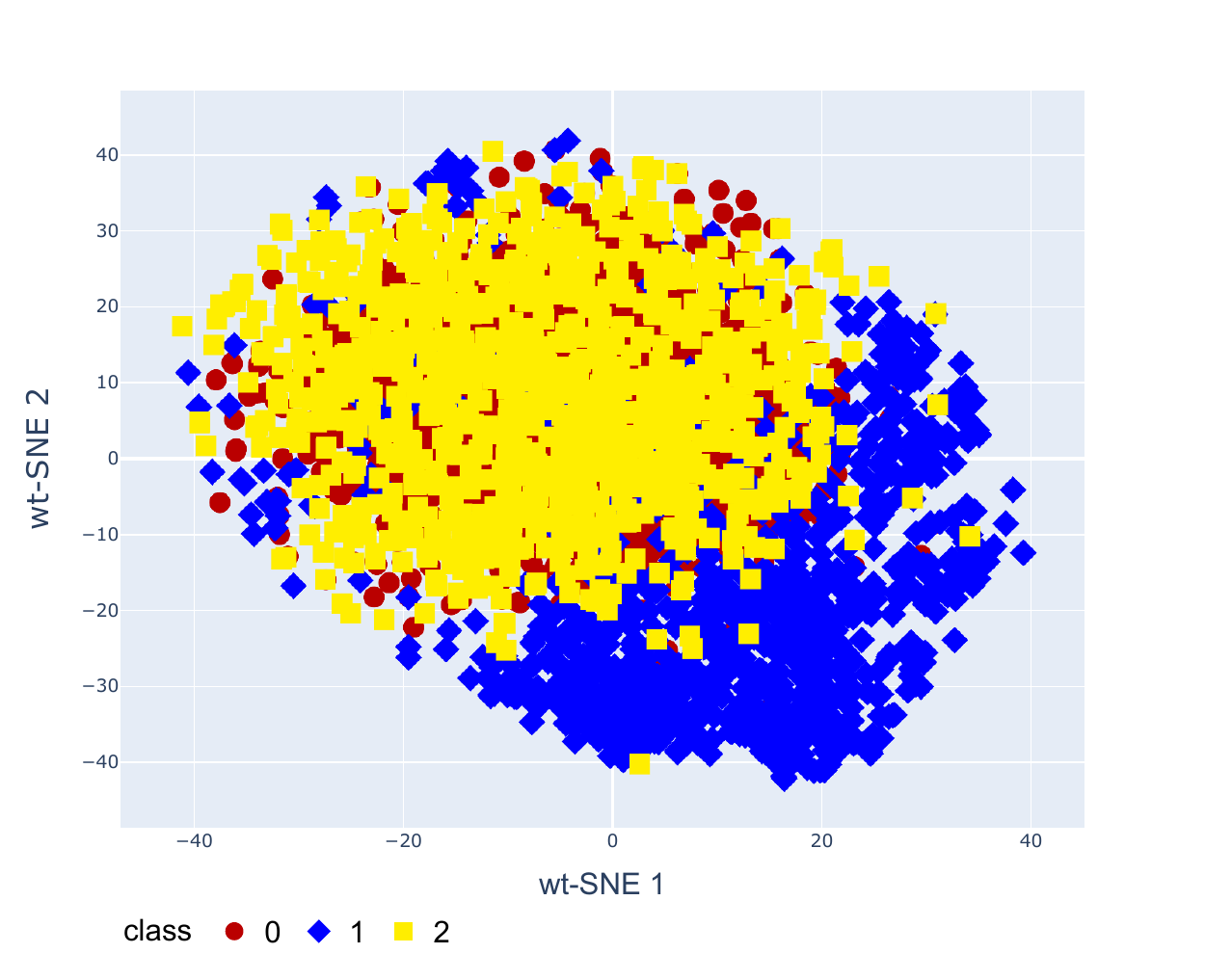}
         \caption{Multiple FSL}
         \label{subfig:weighted_tsne_on_comparison_for_synthB_multiple_fsl}
     \end{subfigure}
\end{figure}

\autoref{fig:weighted_tsne_on_comparison_for_synthA} and \autoref{fig:weighted_tsne_on_comparison_for_synthB} present weighted t-SNE visualizations for Multiple FSL and the comparison algorithms, using general feature weights for the former and label-1-specific weights for the latter.

\autoref{tab:comparative_stability_for_synthB} presents a stability comparison on the SynthB dataset using the $m-1$ approach. 
The results are consistent with the analysis of the SynthA dataset: Kruskal-Wallis demonstrated the highest stability, followed by Random Forest, ReliefF, and Lasso.
Notably, despite relying on neural networks, both DeepSHAP and Multiple FSL exhibited high stability scores.
Specifically, Multiple FSL achieved high Jaccard and Spearman scores, while DeepSHAP attained a slightly higher Pearson score.

\begin{table}[!h]
\centering
\caption[Stability comparison between algorithms feature weights on SynthB dataset.] {\textbf{Stability comparison between algorithms feature weights on SynthB dataset.}
Jaccard similarity measures set overlap, ranging from 0 (no overlap) to 1 (identical sets), being used to assess the similarity between the top-n highest-weighted features.
Spearman accesses ranking similarity, ranging from -1 (perfect inverse correlation) to 1 (perfect correlation). 
Pearson measures linear correlation between the feature weights, ranging from -1 (perfect negative correlation) to 1 (perfect positive correlation) where 0 means no correlation.
Pearson and Spearman was measured using all features (100), while Jaccard was measured using the top-30 features given by the feature weighting algorithm, which is the number of informative features.
In bold is the highest average of each dataset and label.
Best results with statistical significance ($p < 0.01$) according to the Kruskal-Wallis and Dunn’s test are highlighted in orange.
}
\label{tab:comparative_stability_for_synthB}
\begin{tabular}{lrrr}
\hline
Algorithm & Pearson & Spearman & Jaccard \\ 
\hline
\multirow{1}{*}{Decision Tree} 
& 0.728$\pm$.087  & 0.330$\pm$.040 & 0.536$\pm$.044  \\ 
\multirow{1}{*}{Random Forest}  
& \cellcolor{orange!20} 0.989$\pm$.003  & 0.342$\pm$.039 & \cellcolor{orange!20} 0.947$\pm$.026  \\ 
\multirow{1}{*}{Kruskal Wallis}
& \cellcolor{orange!20} \textbf{0.999$\pm$.000}  & \cellcolor{orange!20} \textbf{1.000$\pm$.000} & \cellcolor{orange!20} \textbf{1.000$\pm$.000}  \\ 
\multirow{1}{*}{ReliefF}
&  0.987$\pm$.001  & \cellcolor{orange!20} 0.924$\pm$.008 & \cellcolor{orange!20} 0.976$\pm$.033  \\ 
\multirow{1}{*}{Lasso}
& \cellcolor{orange!20} 0.989$\pm$.001 & \cellcolor{orange!20} 0.840$\pm$.035 & 0.753$\pm$.061   \\ 
\multirow{1}{*}{LIME}
& 0.874$\pm$.026  & 0.162$\pm$.052 & 0.388$\pm$.050  \\ 
\multirow{1}{*}{DeepSHAP}
& 0.920$\pm$.010  & 0.420$\pm$.040 & 0.752$\pm$.046  \\ 
\multirow{1}{*}{Multiple FSL}
& 0.906$\pm$.011  & 0.630$\pm$.032 & 0.876$\pm$.051  \\ 
\hline
\end{tabular}
\end{table}

\begin{figure}[!h]
    \centering
    \caption[
    Colorectal dataset weighted t-SNE visualization using general feature weights.
    The first figure shows the application of t-SNE to the Colorectal dataset without feature weights.
    Class 0 samples are well separated from the others on the bottom of the graph.
    Class 1 samples are overlapping with class 2 and 3 samples from the center to the top of the graph.
    Class 2 samples are mostly from the center to the left of the graph, overlapping with class 1 and 3 samples.
    Class 3 samples are mostly from the center to the top-right of the graph.
    The second figure uses the LASSO general weights.
    Now the class 0 samples are even more separated from the other samples on the top-left of the graph.
    Class 1, 2 and 3 are present from the center to the bottom-right of the graph, being class 1 more to the left, class 2 more to the bottom-right and class 3 more to the top-right.
    The clusters for each class are well defined.
    The third figure uses the LIME general weights.
    It doesn't represent a huge change from the t-SNE without samples, the main difference is a uniform rotation.
    Class 0 samples are now on the bottom-right.
    Class 1 samples are on the top-right, with samples overlapping with class 2 and 3 samples.
    Class 2 samples are mostly on the top-left, with samples overlapping with class 1 and 3 samples.
    And class 3 samples are mainly on the bottom-right of the graph.
    The third figure uses the DeepSHAP general weights.
    As on LIME's graph it did not represent a high improvement when compared with the pure t-SNE without feature weights.
    Now class 0 samples are on the top-left, class 1 on the right, class 2 on the left.
    Class 1, 2 and 3 are stills with a high amount of samples overlapping each other.
    The fourth figure uses the Multiple FSL general weights.
    All classes where improved on this representation.
    Class 0 samples are concentrated on the center-bottom.
    Class 1 on the center-right.
    Class 2 on the center-left.
    Class 3 on the top-left.
    Only a few samples are overlapping with other classes on the center of the graph.
    Visually Multiple FSL and LASSO had the highest improvement in the respective order.
    ]
    {\textbf{Colorectal dataset weighted t-SNE visualization using general feature weights.} 
        \autoref{subfig:comparative_wtsne_model_colorectal_without_weights_general} presents a t-SNE visualization without feature weights.
    }
    \label{fig:comparative_wtsne_model_colorectal_general}
    \begin{subfigure}[b]{0.48\textwidth}
         \centering
         \includegraphics[width=\textwidth]{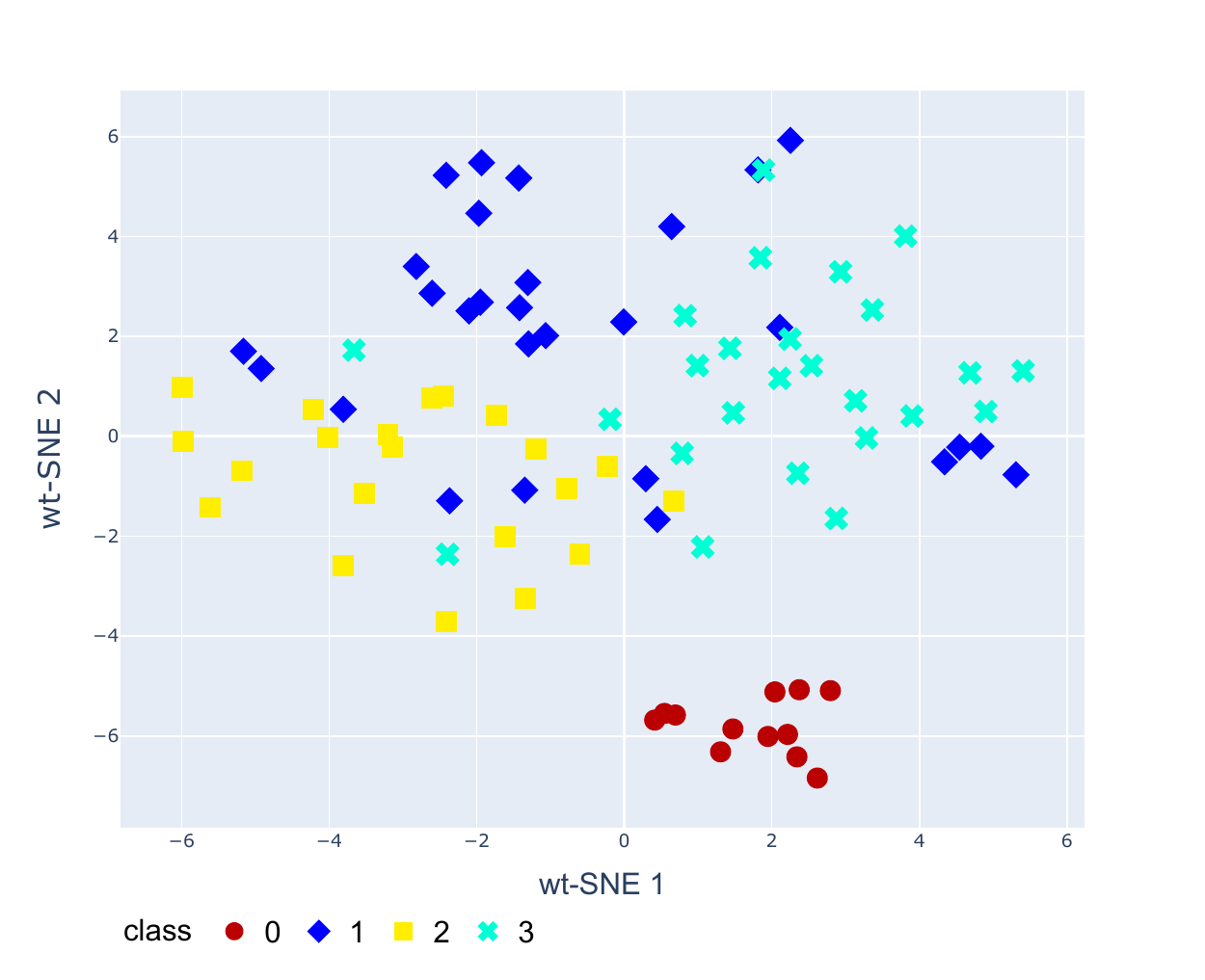}
         \caption{Without weights}
         \label{subfig:comparative_wtsne_model_colorectal_without_weights_general}
     \end{subfigure}
     \hfill
     \begin{subfigure}[b]{0.48\textwidth}
         \centering
         \includegraphics[width=\textwidth]{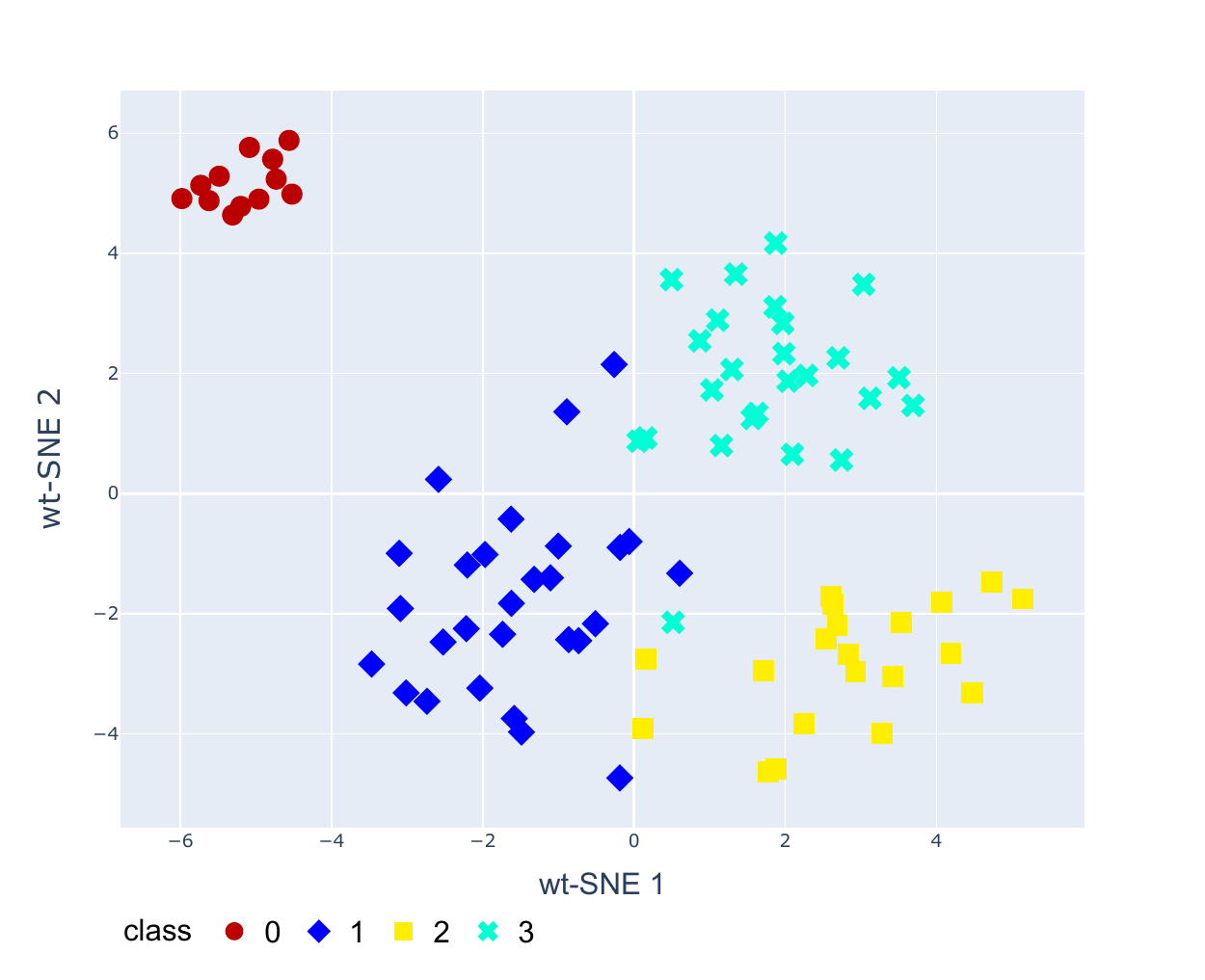}
         \caption{Lasso}
         \label{subfig:comparative_wtsne_model_colorectal_lasso_general}
     \end{subfigure}
     \begin{subfigure}[b]{0.48\textwidth}
         \centering
         \includegraphics[width=\textwidth]{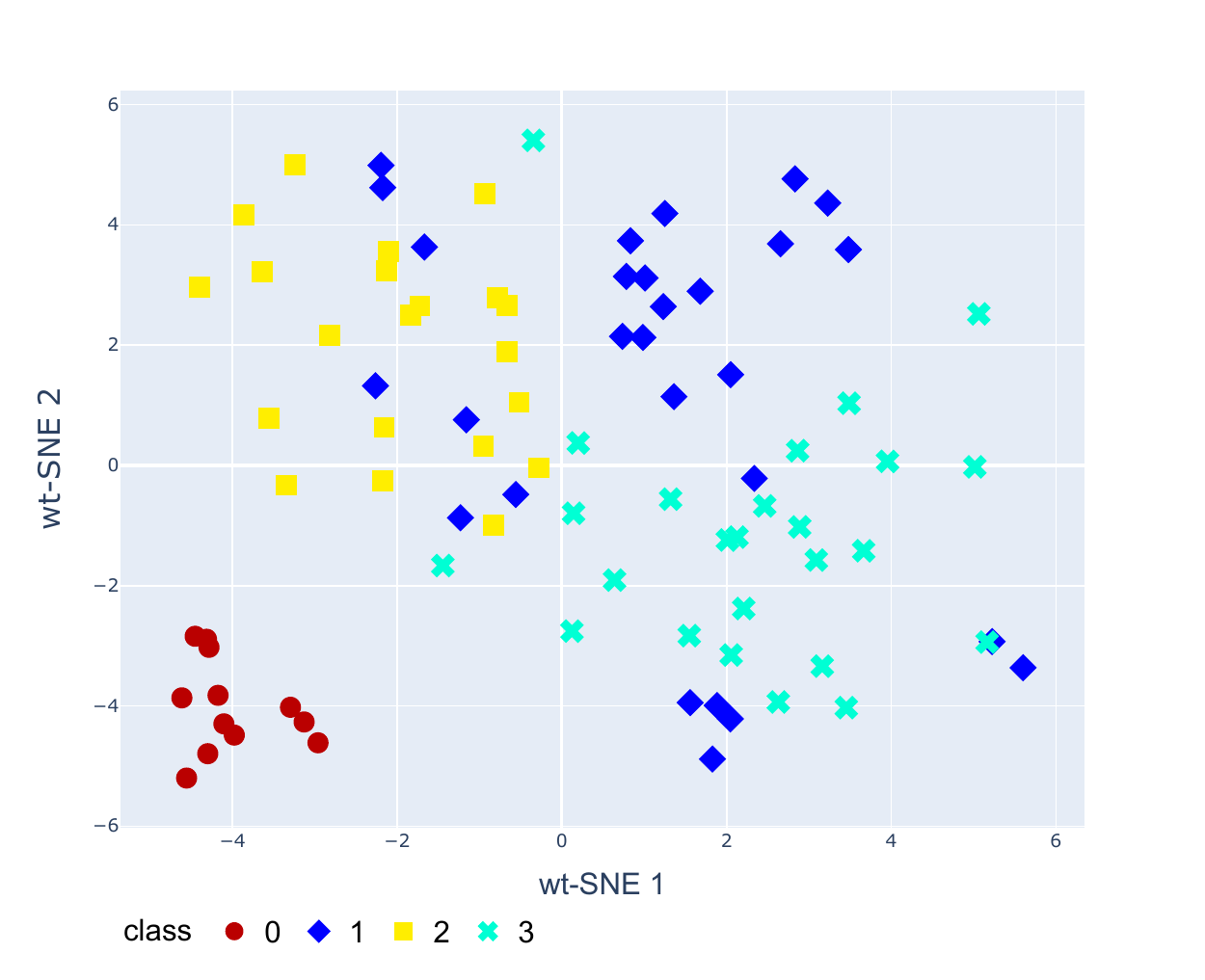}
         \caption{LIME}
         \label{subfig:comparative_wtsne_model_colorectal_lime_general}
     \end{subfigure}
     \hfill
     \begin{subfigure}[b]{0.48\textwidth}
         \centering
         \includegraphics[width=\textwidth]{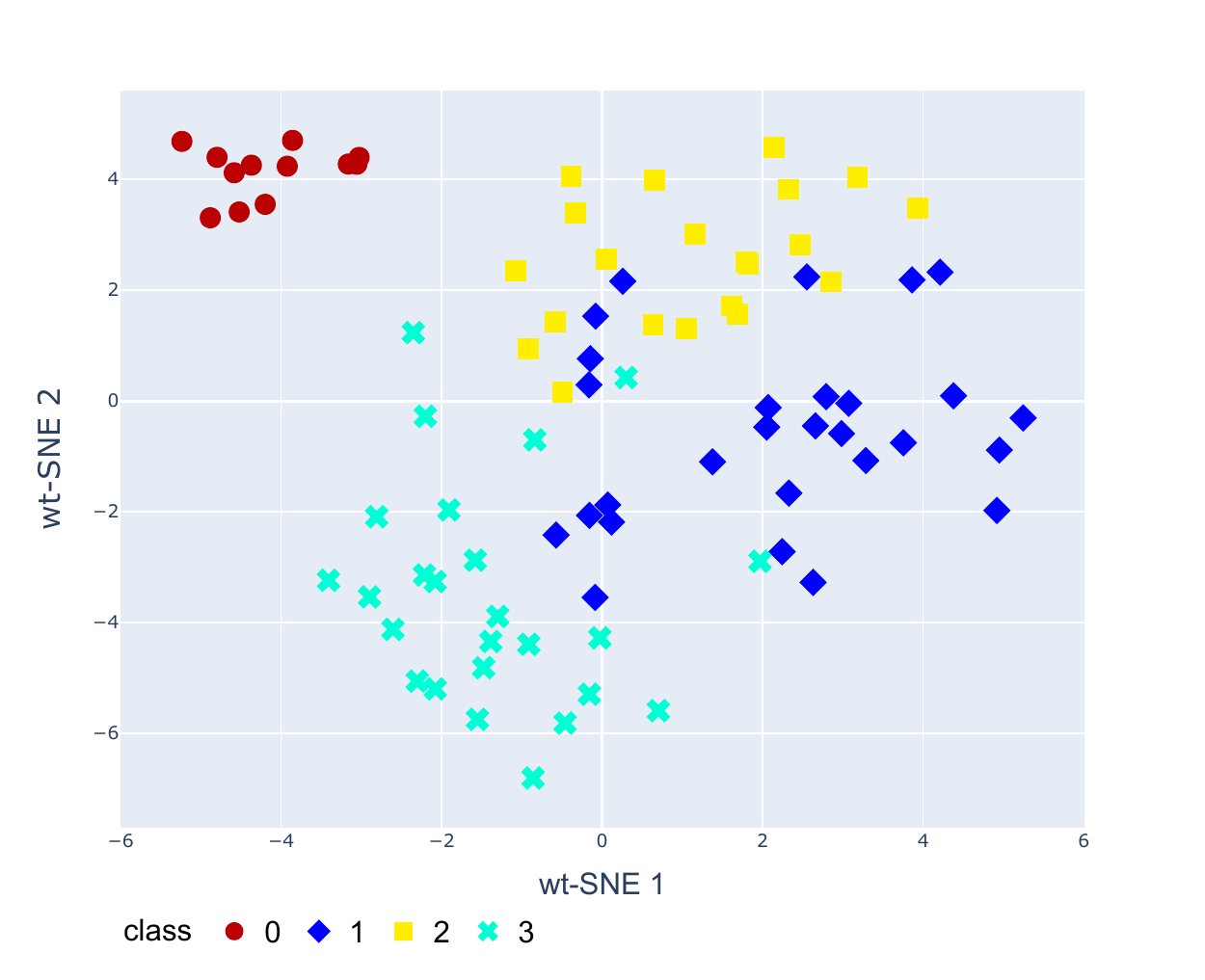}
         \caption{DeepSHAP}
         \label{subfig:comparative_wtsne_model_colorectal_random_deepshap_general}
     \end{subfigure}
     \begin{subfigure}[b]{0.48\textwidth}
         \centering
         \includegraphics[width=\textwidth]{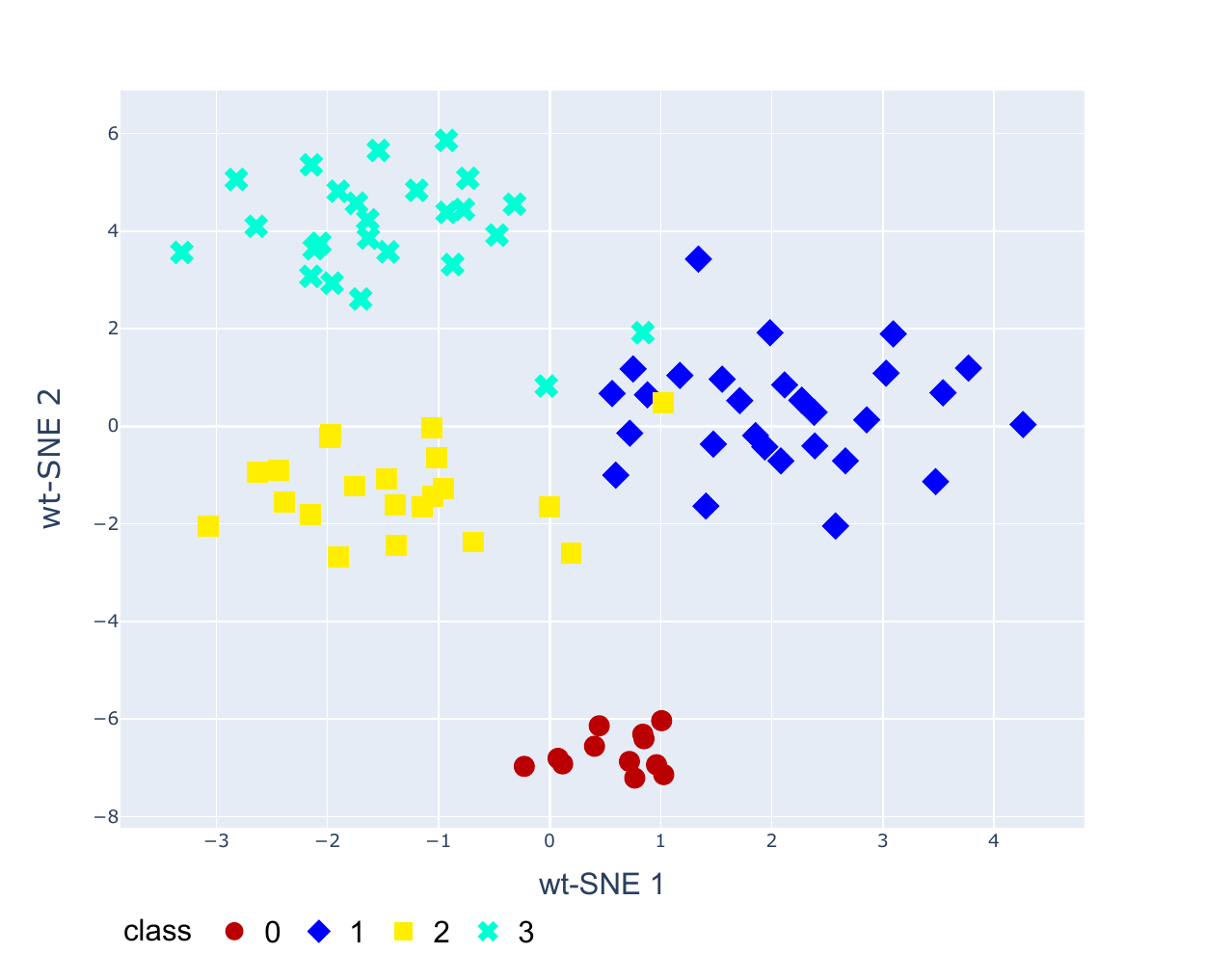}
         \caption{Multiple FSL}
         \label{subfig:comparative_wtsne_model_colorectal_multiple_fsl_general}
     \end{subfigure}
\end{figure}

\autoref{tab:hd_and_2d_silhouette_comparison_colorectal_general_weights} reports the HD and 2D silhouette coefficients for the Colorectal dataset using general feature weights generated by Multiple FSL, DeepSHAP, LIME, and Lasso, complementing the weighted t-SNE visualization in \autoref{fig:comparative_wtsne_model_colorectal_general}.

\begin{table}[!h]
\centering
\caption[Colorectal dataset HD and 2D silhouette coefficients with general feature weights.] {\textbf{Colorectal dataset HD and 2D silhouette coefficients with general feature weights.}
``None'' represents the t-SNE without using weighted features, being a fixed value.
Averages were calculated with the weighted t-SNE results given the feature weights generated by each algorithm through 11 different executions.
For the comparison was used the general feature weights without an specific related label.
In bold is the highest average.
Best results with statistical significance ($p < 0.01$) according to the Kruskal-Wallis and Dunn’s test are highlighted in orange.
 }
\label{tab:hd_and_2d_silhouette_comparison_colorectal_general_weights}
\begin{tabular}{lrr}
\hline
    Weight source               & HD silhouette                                         & 2D silhouette        \\ 
\hline
    None                        & .024                                                & .090       \\
    Lasso                       & \cellcolor{orange!20} \textbf{.187$\pm$.011}      & .487$\pm$.047     \\
    LIME                        & .072$\pm$.001                                     & .207$\pm$.026     \\
    DeepSHAP                    & .096$\pm$.005                                     & .298$\pm$.030     \\
    Multiple FSL                & .159$\pm$.015                                     & \cellcolor{orange!20} \textbf{.561$\pm$.021}     \\
\end{tabular}
\end{table}

\begin{figure}[!h]
    \centering
    \caption[
    SynthB weighted t-SNE visualization using the feature weights attributed to label 1.
    Each one of the four figures represents the label-specific feature weights results for Multiple FSL, LassoNet and TabNet algorithms.
    The first figure shows the application of t-SNE to the SynthB dataset with Multiple FSL weights.
    Class 1 samples are mostly on the left of the graph, with a minority of samples distributed on all other regions.
    All samples are grouped in the center of the graph forming a circular region.
    The second figure uses the LassoNet weights, the samples are now well divided into 4 regions: class 0 is distributed in two main clusters on the right and center-bottom, class 1 is mainly concentrated on the left and class 2 on a region from the bottom-right to the top-right of the graph.
    All classes were improved by the LassoNet weights attributed to label 1, instead of only the target label.
    The third figure uses the TabNet weights, and is internally divided into tree mostly distinct regions: class 0 is distributed in the bottom, class 1 in the top-left and class 2 in the top-right.
    TabNet improved all classes samples, separating them from each other as expected for general feature weights but not for label-specific ones.
    Between the figures, Multiple FSL had the higher overlap between class 0 and 2 samples only. With an improvement of the target label only.
    ]
    {\textbf{Weighted tSNE applied to SynthB using label 1 weights from Multiple FSL, LassoNet and TabNet.} 
        Label 1 samples are presented as blue squares.
    }
    \label{fig:comparative_weighted_tsne_label_1_synth_B_new}
     \begin{subfigure}[b]{0.48\textwidth}
         \centering
         \includegraphics[width=\textwidth]{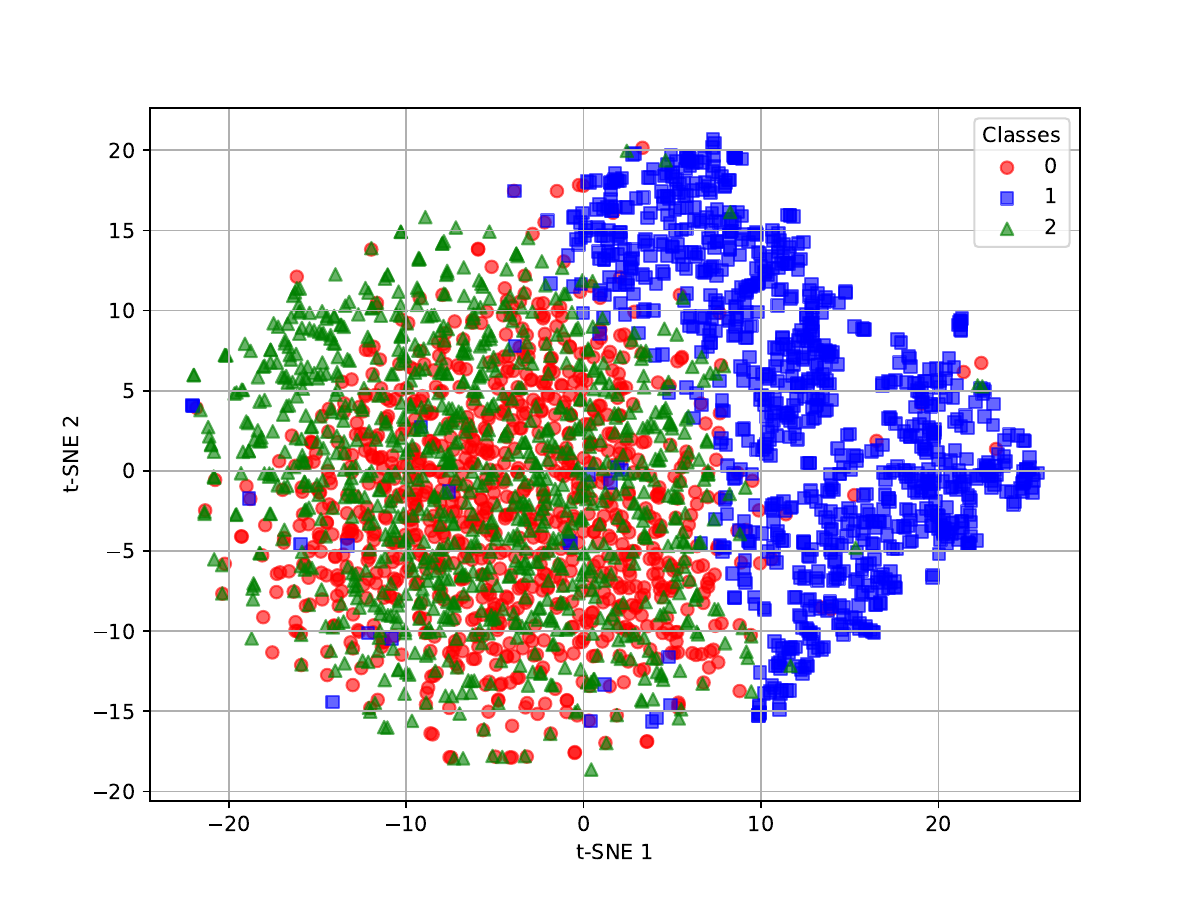}
         \caption{Multiple FSL}
         \label{subfig:comparative_weighted_tsne_label_1_synth_B_new_multiple_fsl}
     \end{subfigure}
     \hfill
     \begin{subfigure}[b]{0.48\textwidth}
         \centering
         \includegraphics[width=\textwidth]{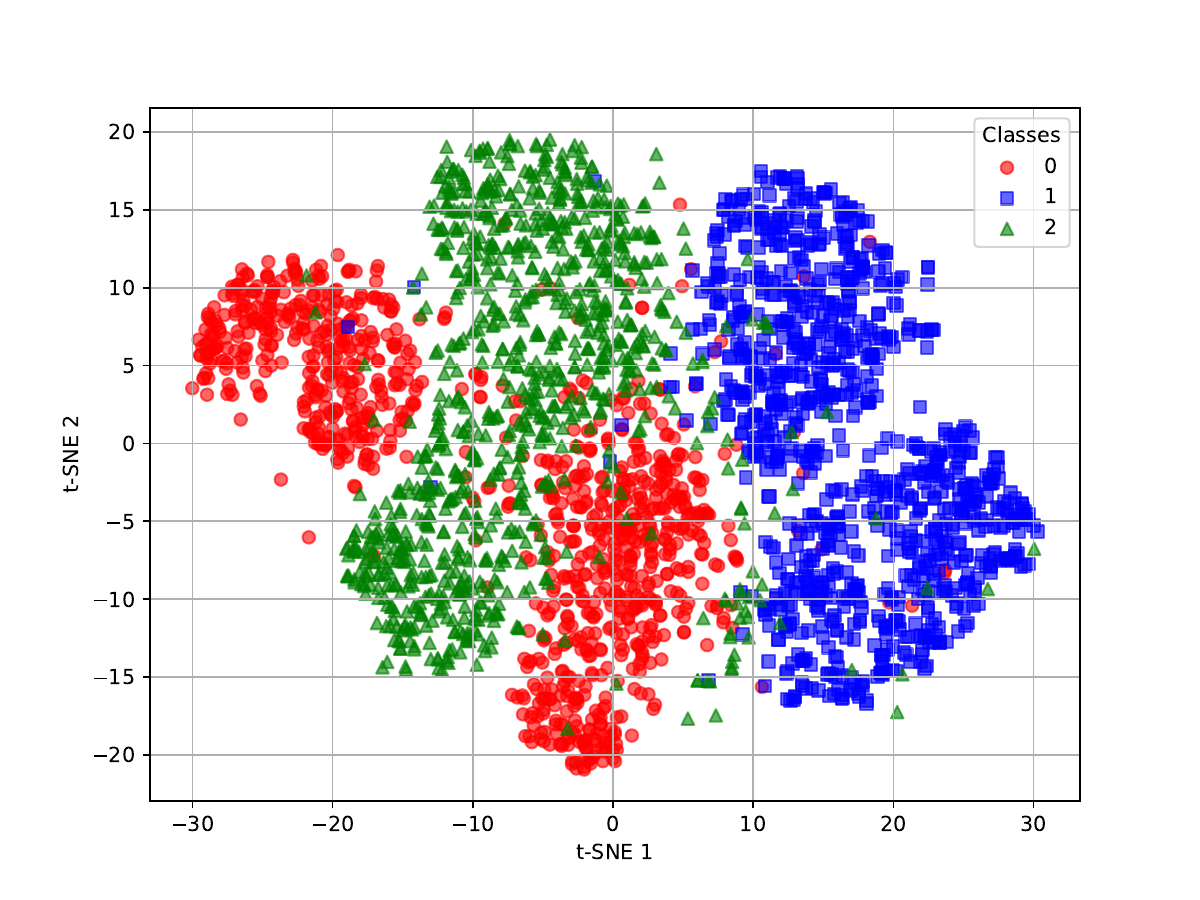}
         \caption{LassoNet}
         \label{subfig:comparative_weighted_tsne_label_1_synth_B_new_lasso_net}
     \end{subfigure}
     \begin{subfigure}[b]{0.48\textwidth}
         \centering
         \includegraphics[width=\textwidth]{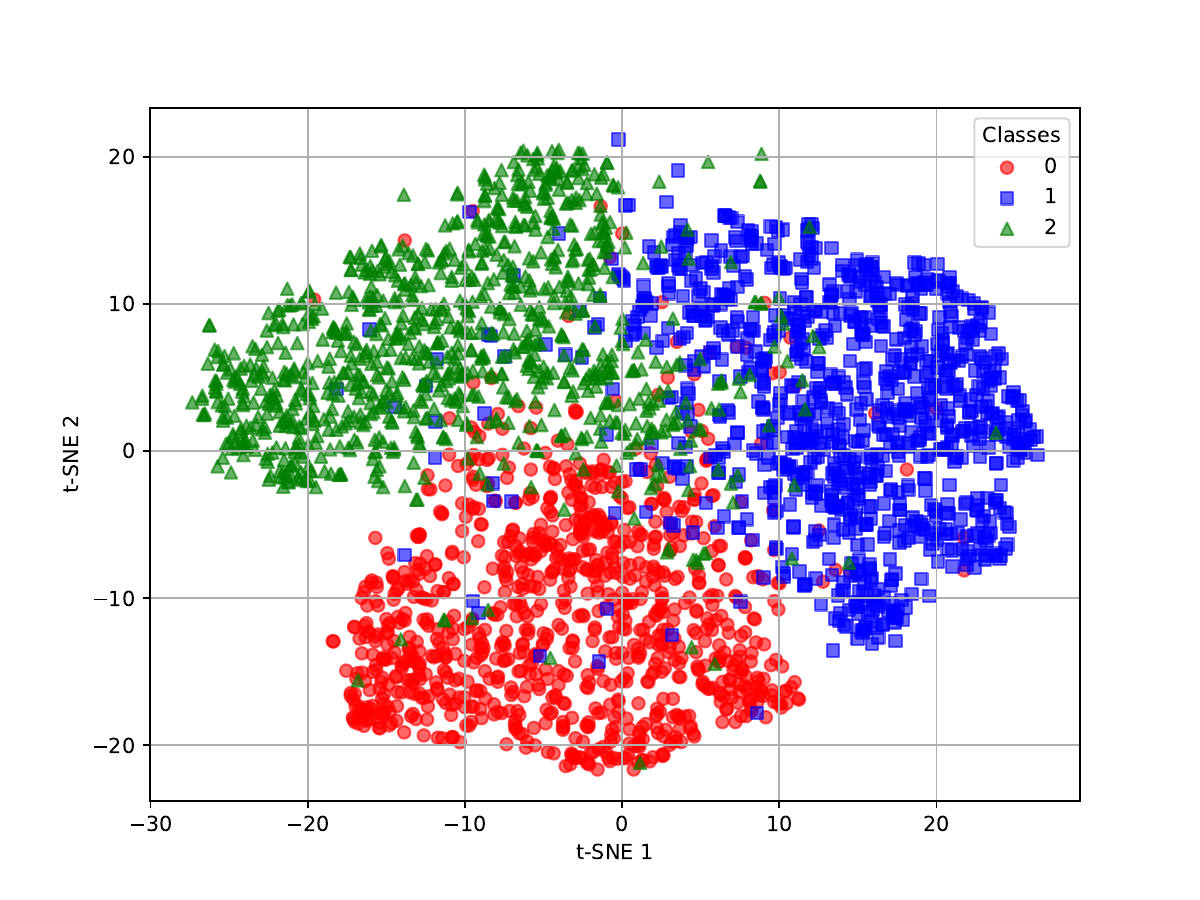}
         \caption{TabNet}
         \label{subfig:comparative_weighted_tsne_label_1_synth_B_new_tabnet}
     \end{subfigure}
\end{figure}

\autoref{fig:comparative_weighted_tsne_label_1_synth_B_new} presents the SynthB weighted tSNE visualization using label 1 from feature weights generated by Multiple FSL, LassoNet and TabNet.
\autoref{tab:label_f1_score_letter_recognition} presents the label-specific F1 scores for the embedded feature weighting techniques capable of detecting label-specific weights.
Multiple FSL achieved the highest scores, followed by LassoNet, and both successfully improved upon the baseline results. 
\autoref{tab:label_f1_score_isolet} extends this analysis to the Isolet dataset. 
Here, TabNet demonstrated superior performance across all labels, followed closely by Multiple FSL. 
Although Multiple FSL yielded slightly lower scores for labels F and S compared to TabNet, it still outperformed the baseline and LassoNet. 
Conversely, Lasso yielded the poorest results on both the Isolet and Letter Recognition datasets.

\begin{table}[t]
\centering
\caption{
\textbf{Label-specific F1-score, mean $\pm$ standard deviation over 11 executions, for Letter Recognition test dataset.}
In bold is the best average per label.
Best results with statistical significance ($p < 0.01$) according to the Kruskal-Wallis and Dunn’s test are highlighted in orange.
}
\label{tab:label_f1_score_letter_recognition}
\resizebox{\textwidth}{!}{%
\begin{tabular}{llllllll}
\toprule
Label & Baseline & Multiple FSL & TabNet & LassoNet & Lasso \\
\midrule
A & 0.989 $\pm$ 0.005 & \cellcolor{orange!20}\textbf{0.993 $\pm$ 0.003} & 0.941 $\pm$ 0.019 & \cellcolor{orange!20}0.988 $\pm$ 0.005 & 0.603 $\pm$ 0.004 \\
B & 0.945 $\pm$ 0.006 & \cellcolor{orange!20}\textbf{0.940 $\pm$ 0.012} & 0.859 $\pm$ 0.025 & \cellcolor{orange!20}0.928 $\pm$ 0.009 & 0.546 $\pm$ 0.011 \\
C & 0.970 $\pm$ 0.010 & \cellcolor{orange!20}\textbf{0.976 $\pm$ 0.005} & 0.908 $\pm$ 0.019 & \cellcolor{orange!20}0.946 $\pm$ 0.010 & 0.492 $\pm$ 0.003 \\
D & 0.959 $\pm$ 0.011 & \cellcolor{orange!20}\textbf{0.957 $\pm$ 0.010} & 0.855 $\pm$ 0.027 & \cellcolor{orange!20}0.935 $\pm$ 0.005 & 0.555 $\pm$ 0.007 \\
E & 0.953 $\pm$ 0.015 & \cellcolor{orange!20}0.950 $\pm$ 0.009 & 0.885 $\pm$ 0.022 & \cellcolor{orange!20}\textbf{0.951 $\pm$ 0.006} & 0.085 $\pm$ 0.023 \\
F & 0.956 $\pm$ 0.008 & \cellcolor{orange!20}\textbf{0.952 $\pm$ 0.010} & 0.858 $\pm$ 0.024 & \cellcolor{orange!20}0.920 $\pm$ 0.019 & 0.475 $\pm$ 0.015 \\
G & 0.955 $\pm$ 0.009 & \cellcolor{orange!20}\textbf{0.951 $\pm$ 0.012} & 0.849 $\pm$ 0.019 & \cellcolor{orange!20}0.935 $\pm$ 0.011 & 0.020 $\pm$ 0.007 \\
H & 0.961 $\pm$ 0.007 & \cellcolor{orange!20}\textbf{0.952 $\pm$ 0.011} & 0.806 $\pm$ 0.035 & \cellcolor{orange!20}0.916 $\pm$ 0.014 & 0.005 $\pm$ 0.008 \\
I & 0.958 $\pm$ 0.006 & \cellcolor{orange!20}0.957 $\pm$ 0.004 & 0.923 $\pm$ 0.022 & \cellcolor{orange!20}\textbf{0.960 $\pm$ 0.011} & 0.748 $\pm$ 0.006 \\
J & 0.961 $\pm$ 0.007 & \cellcolor{orange!20}0.959 $\pm$ 0.009 & 0.895 $\pm$ 0.025 & \cellcolor{orange!20}\textbf{0.972 $\pm$ 0.003} & 0.594 $\pm$ 0.006 \\
K & 0.972 $\pm$ 0.008 & \cellcolor{orange!20}\textbf{0.968 $\pm$ 0.013} & 0.889 $\pm$ 0.021 & \cellcolor{orange!20}0.955 $\pm$ 0.015 & 0.092 $\pm$ 0.029 \\
L & 0.990 $\pm$ 0.006 & \cellcolor{orange!20}\textbf{0.994 $\pm$ 0.005} & 0.932 $\pm$ 0.014 & \cellcolor{orange!20}0.970 $\pm$ 0.007 & 0.655 $\pm$ 0.005 \\
M & 0.984 $\pm$ 0.007 & \cellcolor{orange!20}0.978 $\pm$ 0.006 & 0.934 $\pm$ 0.009 & \cellcolor{orange!20}\textbf{0.980 $\pm$ 0.006} & 0.498 $\pm$ 0.005 \\
N & 0.973 $\pm$ 0.009 & \cellcolor{orange!20}\textbf{0.978 $\pm$ 0.006} & 0.902 $\pm$ 0.022 & \cellcolor{orange!20}0.940 $\pm$ 0.012 & 0.376 $\pm$ 0.010 \\
O & 0.965 $\pm$ 0.008 & \cellcolor{orange!20}\textbf{0.967 $\pm$ 0.011} & 0.871 $\pm$ 0.025 & \cellcolor{orange!20}0.945 $\pm$ 0.008 & 0.053 $\pm$ 0.022 \\
P & 0.954 $\pm$ 0.009 & \cellcolor{orange!20}\textbf{0.955 $\pm$ 0.008} & 0.890 $\pm$ 0.020 & \cellcolor{orange!20}0.947 $\pm$ 0.006 & 0.589 $\pm$ 0.010 \\
Q & 0.967 $\pm$ 0.008 & \cellcolor{orange!20}\textbf{0.967 $\pm$ 0.013} & 0.899 $\pm$ 0.021 & \cellcolor{orange!20}0.957 $\pm$ 0.007 & 0.574 $\pm$ 0.012 \\
R & 0.950 $\pm$ 0.012 & \cellcolor{orange!20}\textbf{0.948 $\pm$ 0.013} & 0.857 $\pm$ 0.018 & \cellcolor{orange!20}0.917 $\pm$ 0.015 & 0.658 $\pm$ 0.006 \\
S & 0.971 $\pm$ 0.006 & \cellcolor{orange!20}\textbf{0.979 $\pm$ 0.008} & 0.868 $\pm$ 0.032 & \cellcolor{orange!20}0.963 $\pm$ 0.009 & 0.239 $\pm$ 0.017 \\
T & 0.980 $\pm$ 0.006 & \cellcolor{orange!20}\textbf{0.970 $\pm$ 0.007} & 0.903 $\pm$ 0.011 & \cellcolor{orange!20}0.961 $\pm$ 0.007 & 0.667 $\pm$ 0.008 \\
U & 0.988 $\pm$ 0.006 & \cellcolor{orange!20}0.979 $\pm$ 0.006 & 0.932 $\pm$ 0.018 & \cellcolor{orange!20}\textbf{0.980 $\pm$ 0.005} & 0.617 $\pm$ 0.011 \\
V & 0.965 $\pm$ 0.007 & \cellcolor{orange!20}0.959 $\pm$ 0.012 & 0.928 $\pm$ 0.014 & \cellcolor{orange!20}\textbf{0.964 $\pm$ 0.008} & 0.613 $\pm$ 0.007 \\
W & 0.977 $\pm$ 0.010 & \cellcolor{orange!20}\textbf{0.976 $\pm$ 0.010} & 0.934 $\pm$ 0.014 & \cellcolor{orange!20}0.968 $\pm$ 0.005 & 0.524 $\pm$ 0.004 \\
X & 0.977 $\pm$ 0.008 & \cellcolor{orange!20}\textbf{0.977 $\pm$ 0.006} & 0.913 $\pm$ 0.012 & \cellcolor{orange!20}0.963 $\pm$ 0.005 & 0.071 $\pm$ 0.030 \\
Y & 0.975 $\pm$ 0.007 & \cellcolor{orange!20}\textbf{0.975 $\pm$ 0.010} & 0.922 $\pm$ 0.022 & \cellcolor{orange!20}0.967 $\pm$ 0.007 & 0.445 $\pm$ 0.023 \\
Z & 0.982 $\pm$ 0.008 & \cellcolor{orange!20}\textbf{0.988 $\pm$ 0.006} & 0.961 $\pm$ 0.013 & \cellcolor{orange!20}0.969 $\pm$ 0.004 & 0.556 $\pm$ 0.011 \\
\bottomrule
\end{tabular}
}
\end{table}

\begin{table}[t]
\centering
\caption{
\textbf{Label-specific F1-score, mean $\pm$ standard deviation over 11 executions, for Isolet test dataset.}
In bold is the best average per label.
Best results with statistical significance ($p < 0.01$) according to the Kruskal-Wallis and Dunn’s test are highlighted in orange.
}
\label{tab:label_f1_score_isolet}
\resizebox{\textwidth}{!}{%
\begin{tabular}{llllllll}
\toprule
Label & Baseline & Multiple FSL & TabNet & LassoNet & Lasso \\
\midrule
A & 0.953 $\pm$ 0.017 & \cellcolor{orange!20}0.964 $\pm$ 0.017 & \cellcolor{orange!20}\textbf{0.974 $\pm$ 0.017} & \cellcolor{orange!20}0.972 $\pm$ 0.012 & 0.752 $\pm$ 0.014 \\
B & 0.764 $\pm$ 0.046 & \cellcolor{orange!20}0.869 $\pm$ 0.027 & \cellcolor{orange!20}\textbf{0.899 $\pm$ 0.030} & \cellcolor{orange!20}0.865 $\pm$ 0.022 & 0.791 $\pm$ 0.017 \\
C & 0.982 $\pm$ 0.010 & \cellcolor{orange!20}\textbf{0.996 $\pm$ 0.006} & \cellcolor{orange!20}0.982 $\pm$ 0.014 & 0.975 $\pm$ 0.008 & 0.882 $\pm$ 0.009 \\
D & 0.836 $\pm$ 0.056 & \cellcolor{orange!20}0.905 $\pm$ 0.025 & \cellcolor{orange!20}\textbf{0.908 $\pm$ 0.023} & \cellcolor{orange!20}0.899 $\pm$ 0.018 & 0.767 $\pm$ 0.012 \\
E & 0.871 $\pm$ 0.048 & \cellcolor{orange!20}0.893 $\pm$ 0.033 & \cellcolor{orange!20}\textbf{0.918 $\pm$ 0.022} & \cellcolor{orange!20}0.913 $\pm$ 0.020 & 0.862 $\pm$ 0.014 \\
F & 0.900 $\pm$ 0.037 & 0.916 $\pm$ 0.021 & \cellcolor{orange!20}0.957 $\pm$ 0.035 & \cellcolor{orange!20}\textbf{0.969 $\pm$ 0.016} & 0.720 $\pm$ 0.020 \\
G & 0.901 $\pm$ 0.059 & \cellcolor{orange!20}0.936 $\pm$ 0.038 & \cellcolor{orange!20}0.959 $\pm$ 0.034 & \cellcolor{orange!20}\textbf{0.960 $\pm$ 0.012} & 0.720 $\pm$ 0.009 \\
H & 0.980 $\pm$ 0.021 & \cellcolor{orange!20}0.989 $\pm$ 0.011 & \cellcolor{orange!20}0.990 $\pm$ 0.017 & \cellcolor{orange!20}\textbf{0.996 $\pm$ 0.006} & 0.765 $\pm$ 0.021 \\
I & 0.961 $\pm$ 0.024 & \cellcolor{orange!20}0.978 $\pm$ 0.014 & \cellcolor{orange!20}\textbf{0.990 $\pm$ 0.008} & \cellcolor{orange!20}0.990 $\pm$ 0.008 & 0.917 $\pm$ 0.011 \\
J & 0.939 $\pm$ 0.033 & \cellcolor{orange!20}\textbf{0.979 $\pm$ 0.010} & \cellcolor{orange!20}0.975 $\pm$ 0.015 & \cellcolor{orange!20}0.973 $\pm$ 0.010 & 0.772 $\pm$ 0.016 \\
K & 0.945 $\pm$ 0.028 & \cellcolor{orange!20}0.972 $\pm$ 0.012 & \cellcolor{orange!20}\textbf{0.983 $\pm$ 0.015} & \cellcolor{orange!20}0.978 $\pm$ 0.017 & 0.686 $\pm$ 0.022 \\
L & 0.919 $\pm$ 0.011 & \cellcolor{orange!20}0.955 $\pm$ 0.015 & \cellcolor{orange!20}\textbf{0.974 $\pm$ 0.011} & 0.951 $\pm$ 0.013 & 0.898 $\pm$ 0.000 \\
M & 0.793 $\pm$ 0.045 & \cellcolor{orange!20}0.878 $\pm$ 0.032 & \cellcolor{orange!20}\textbf{0.917 $\pm$ 0.034} & 0.852 $\pm$ 0.027 & 0.697 $\pm$ 0.022 \\
N & 0.810 $\pm$ 0.039 & \cellcolor{orange!20}0.895 $\pm$ 0.027 & \cellcolor{orange!20}\textbf{0.931 $\pm$ 0.030} & 0.876 $\pm$ 0.023 & 0.692 $\pm$ 0.021 \\
O & 0.909 $\pm$ 0.022 & \cellcolor{orange!20}0.945 $\pm$ 0.019 & \cellcolor{orange!20}\textbf{0.967 $\pm$ 0.009} & 0.935 $\pm$ 0.015 & 0.874 $\pm$ 0.005 \\
P & 0.808 $\pm$ 0.044 & \cellcolor{orange!20}0.870 $\pm$ 0.023 & \cellcolor{orange!20}\textbf{0.881 $\pm$ 0.036} & \cellcolor{orange!20}0.872 $\pm$ 0.020 & 0.667 $\pm$ 0.023 \\
Q & 0.966 $\pm$ 0.010 & \cellcolor{orange!20}0.969 $\pm$ 0.019 & \cellcolor{orange!20}\textbf{0.983 $\pm$ 0.014} & 0.959 $\pm$ 0.008 & 0.705 $\pm$ 0.012 \\
R & 0.967 $\pm$ 0.012 & \cellcolor{orange!20}0.984 $\pm$ 0.008 & \cellcolor{orange!20}\textbf{0.986 $\pm$ 0.010} & \cellcolor{orange!20}0.981 $\pm$ 0.007 & 0.928 $\pm$ 0.010 \\
S & 0.917 $\pm$ 0.050 & 0.938 $\pm$ 0.016 & \cellcolor{orange!20}0.967 $\pm$ 0.030 & \cellcolor{orange!20}\textbf{0.979 $\pm$ 0.011} & 0.821 $\pm$ 0.005 \\
T & 0.809 $\pm$ 0.048 & \cellcolor{orange!20}0.884 $\pm$ 0.023 & \cellcolor{orange!20}\textbf{0.908 $\pm$ 0.021} & \cellcolor{orange!20}0.886 $\pm$ 0.024 & 0.620 $\pm$ 0.015 \\
U & 0.957 $\pm$ 0.022 & \cellcolor{orange!20}0.969 $\pm$ 0.017 & \cellcolor{orange!20}\textbf{0.970 $\pm$ 0.016} & \cellcolor{orange!20}0.960 $\pm$ 0.006 & 0.860 $\pm$ 0.008 \\
V & 0.800 $\pm$ 0.036 & \cellcolor{orange!20}0.890 $\pm$ 0.035 & \cellcolor{orange!20}\textbf{0.896 $\pm$ 0.019} & \cellcolor{orange!20}0.891 $\pm$ 0.025 & 0.718 $\pm$ 0.017 \\
W & 0.957 $\pm$ 0.018 & \cellcolor{orange!20}0.966 $\pm$ 0.021 & \cellcolor{orange!20}\textbf{0.978 $\pm$ 0.016} & \cellcolor{orange!20}0.965 $\pm$ 0.013 & 0.837 $\pm$ 0.008 \\
X & 0.952 $\pm$ 0.015 & \cellcolor{orange!20}0.975 $\pm$ 0.007 & \cellcolor{orange!20}0.974 $\pm$ 0.013 & \cellcolor{orange!20}\textbf{0.983 $\pm$ 0.008} & 0.794 $\pm$ 0.011 \\
Y & 0.986 $\pm$ 0.015 & \cellcolor{orange!20}0.992 $\pm$ 0.007 & \cellcolor{orange!20}0.997 $\pm$ 0.005 & \cellcolor{orange!20}\textbf{1.000 $\pm$ 0.000} & 0.976 $\pm$ 0.004 \\
Z & 0.945 $\pm$ 0.022 & \cellcolor{orange!20}\textbf{0.954 $\pm$ 0.013} & \cellcolor{orange!20}0.950 $\pm$ 0.027 & \cellcolor{orange!20}0.926 $\pm$ 0.014 & 0.766 $\pm$ 0.016 \\
\bottomrule
\end{tabular}
}
\end{table}

\begin{sidewaystable}[h]
\centering
\resizebox{\textwidth}{!}{%
\begin{tabular}{l l >{\raggedright\arraybackslash}p{5cm} >{\raggedright\arraybackslash}p{4cm} >{\raggedright\arraybackslash}p{6cm}}
\toprule
\textbf{Selector} & \textbf{Granularity} & \textbf{Arch. Agnostic} & \textbf{Inference overhead} & \textbf{Training mechanism} \\
\midrule
\multicolumn{5}{l}{\textit{Embedded}} \\
\midrule
Decision Tree & Global          & No (is its own classifier)                 & N/A & Greedy recursive splitting                         \\
Random Forest & Global          & No (is its own classifier)                 & N/A & Bagged greedy tree ensemble                        \\
Lasso         & Label-specific  & No (is its own linear classifier)          & N/A & Coordinate descent ($\ell_1$-regularized)          \\
LassoNet      & Label-specific  & No (fixed skip-connection MLP)             & N/A & Backprop + proximal gradient (path-wise $\ell_1$)  \\
TabNet        & Label-specific  & No (own sequential-attention architecture) & N/A & Backprop (sequential attention masks)              \\
\midrule
\multicolumn{5}{l}{\textit{Filter}} \\
\midrule
Kruskal--Wallis & Global & N/A (produces no classifier) & N/A & Closed-form statistical test per feature  \\
ReliefF         & Global & N/A (produces no classifier) & N/A & Instance-based nearest-neighbor heuristic \\
\midrule
\multicolumn{5}{l}{\textit{Embedded --- FSL family}} \\
\midrule
FSL          & Global         & Partial$^{\ddagger}$ & one extra elementwise multiply                                            & Backprop, joint with classifier ($\ell_1$-regularized gate)                     \\
Listener     & Label-specific & Partial$^{\ddagger}$ & one extra elementwise multiply                                            & Backprop; per-label signal from weight drift while training one label at a time \\
Residual     & Label-specific & Partial$^{\ddagger}$ & one extra linear residual term                                            & Backprop, joint with classifier                                                 \\
Output Aware & Label-specific & Partial$^{\ddagger}$ & one extra elementwise multiply$^{\P}$                                     & Backprop; gate conditioned on the one-hot label at train time                   \\
Multiple FSL & Label-specific & Partial$^{\ddagger}$ & one batched matmul ($n_{\mathrm{labels}} \times n_{\mathrm{features}}$)$^{\|}$ & Backprop, joint with classifier                                        \\
\midrule
\multicolumn{5}{l}{\textit{Post-hoc}} \\
\midrule
LIME     & Label-specific & Yes (any black-box predict function)              & None & Local surrogate linear regression on perturbed samples        \\
DeepSHAP & Label-specific & Partial (restricted to differentiable/neural models)   & None & Gradient-based SHAP approximation (DeepLIFT-style backprop)  \\
SAGE     & Label-specific & Yes (any predict/predict\_proba model)            & None & Marginal-imputation Shapley value estimation (KernelEstimator) \\
\bottomrule
\end{tabular}%
}
\caption{\textbf{Comparison of feature selectors. }
$^{\ddagger}$The FSL family's gate is an input-masking layer applied before an arbitrary downstream differentiable classifier --- the gating mechanism itself is architecture-agnostic in design. 
$^{\S}$Listener trains one shared weight vector but fits it one label at a time, tracking how much it moves during each label's turn as that label's attribution. 
$^{\P}$Output Aware's forward pass branches on train/eval mode: at training time the gate is selected via the batch's one-hot labels; at evaluation/inference it always uses the sum of all label gates, applied as a single element wise multiply, unlike Multiple FSL which never needs a separate forward pass per label at inference time. 
$^{\|}$This selector's forward pass is mathematically identical to running the shared classifier once per label sequentially (same total FLOPs), but stacks every label's input into one batched matrix multiplication instead, which GPUs parallelize far better.
Post-hoc and filter methods incur no inference overhead since feature weighting is decoupled from the prediction model. 
Conversely, FSL and embedded methods integrate weighting directly into the inference process. 
While simple embedded models (e.g., Decision Trees, Random Forest, Lasso) have minimal architectural overhead, complex architectures like LassoNet and TabNet exhibit significantly higher processing times. 
For the FSL family of methods, we specify the computational impact of the feature weighting mechanism, as they are designed to be agnostic to their internal model.
}
\label{tab:selector-comparison}
\end{sidewaystable}

\begin{sidewaystable}[h]
\centering
\renewcommand{\arraystretch}{1.35}
\resizebox{\textwidth}{!}{%
\begin{tabular}{llllllllllll}
\toprule
Dataset & NeuralBaseClassifier & Decision Tree & Random Forest & Lasso & LassoNet & TabNet & FSL & Listener & Residual & Output Aware & Multiple FSL \\
\midrule
XOR & 0.078 $\pm$ 0.051 & 0.439 $\pm$ 0.071 & 0.081 $\pm$ 0.024 & 0.118 $\pm$ 0.028 & \cellcolor{orange!20}\textbf{0.003 $\pm$ 0.001} & 0.067 $\pm$ 0.028 & \cellcolor{orange!20}\textbf{0.003 $\pm$ 0.007} & \cellcolor{orange!20}0.009 $\pm$ 0.007 & 0.069 $\pm$ 0.050 & 0.060 $\pm$ 0.042 & \cellcolor{orange!20}\textbf{0.003 $\pm$ 0.006} \\
SynthA & 0.086 $\pm$ 0.027 & 0.462 $\pm$ 0.023 & 0.276 $\pm$ 0.011 & 0.109 $\pm$ 0.006 & \cellcolor{orange!20}\textbf{0.044 $\pm$ 0.017} & 0.101 $\pm$ 0.025 & \cellcolor{orange!20}0.073 $\pm$ 0.010 & \cellcolor{orange!20}0.069 $\pm$ 0.010 & \cellcolor{orange!20}0.062 $\pm$ 0.010 & \cellcolor{orange!20}0.081 $\pm$ 0.037 & \cellcolor{orange!20}0.062 $\pm$ 0.010 \\
SynthB & 0.090 $\pm$ 0.038 & 0.242 $\pm$ 0.016 & 0.310 $\pm$ 0.008 & 0.158 $\pm$ 0.005 & 0.087 $\pm$ 0.067 & 0.067 $\pm$ 0.018 & \cellcolor{orange!20}0.037 $\pm$ 0.011 & \cellcolor{orange!20}0.038 $\pm$ 0.005 & \cellcolor{orange!20}\textbf{0.025 $\pm$ 0.004} & 0.133 $\pm$ 0.031 & \cellcolor{orange!20}0.029 $\pm$ 0.005 \\
SynthC & \cellcolor{orange!20}0.076 $\pm$ 0.032 & 0.212 $\pm$ 0.016 & 0.297 $\pm$ 0.004 & 0.123 $\pm$ 0.007 & \cellcolor{orange!20}\textbf{0.041 $\pm$ 0.013} & \cellcolor{orange!20}0.054 $\pm$ 0.020 & 0.086 $\pm$ 0.007 & \cellcolor{orange!20}0.074 $\pm$ 0.011 & 0.084 $\pm$ 0.011 & 0.164 $\pm$ 0.029 & \cellcolor{orange!20}0.066 $\pm$ 0.016 \\
Liver & \cellcolor{orange!20}0.094 $\pm$ 0.142 & \cellcolor{orange!20}0.045 $\pm$ 0.084 & 0.242 $\pm$ 0.016 & 0.117 $\pm$ 0.043 & 0.050 $\pm$ 0.074 & 0.111 $\pm$ 0.074 & \cellcolor{orange!20}0.033 $\pm$ 0.080 & \cellcolor{orange!20}0.072 $\pm$ 0.126 & \cellcolor{orange!20}\textbf{0.002 $\pm$ 0.007} & \cellcolor{orange!20}0.070 $\pm$ 0.104 & \cellcolor{orange!20}0.023 $\pm$ 0.053 \\
Colorectal & \cellcolor{orange!20}0.172 $\pm$ 0.080 & 0.399 $\pm$ 0.108 & 0.252 $\pm$ 0.058 & 0.195 $\pm$ 0.059 & 0.190 $\pm$ 0.046 & \cellcolor{orange!20}\textbf{0.109 $\pm$ 0.091} & \cellcolor{orange!20}0.129 $\pm$ 0.064 & \cellcolor{orange!20}0.161 $\pm$ 0.060 & 0.218 $\pm$ 0.102 & \cellcolor{orange!20}0.171 $\pm$ 0.070 & \cellcolor{orange!20}0.136 $\pm$ 0.079 \\
Breast & \cellcolor{orange!20}0.071 $\pm$ 0.032 & 0.173 $\pm$ 0.044 & 0.364 $\pm$ 0.018 & 0.227 $\pm$ 0.032 & 0.138 $\pm$ 0.044 & 0.117 $\pm$ 0.024 & \cellcolor{orange!20}0.034 $\pm$ 0.022 & \cellcolor{orange!20}\textbf{0.027 $\pm$ 0.018} & 0.119 $\pm$ 0.053 & \cellcolor{orange!20}0.054 $\pm$ 0.037 & \cellcolor{orange!20}0.033 $\pm$ 0.030 \\
Isolet & 0.079 $\pm$ 0.019 & 0.199 $\pm$ 0.010 & 0.252 $\pm$ 0.003 & 0.417 $\pm$ 0.004 & \cellcolor{orange!20}0.027 $\pm$ 0.007 & \cellcolor{orange!20}0.023 $\pm$ 0.004 & 0.079 $\pm$ 0.014 & 0.130 $\pm$ 0.011 & \cellcolor{orange!20}0.036 $\pm$ 0.008 & 0.308 $\pm$ 0.071 & \cellcolor{orange!20}\textbf{0.016 $\pm$ 0.005} \\
Letter Recognition & \cellcolor{orange!20}0.010 $\pm$ 0.002 & 0.139 $\pm$ 0.004 & 0.149 $\pm$ 0.002 & 0.287 $\pm$ 0.003 & \cellcolor{orange!20}0.011 $\pm$ 0.002 & \cellcolor{orange!20}0.012 $\pm$ 0.003 & \cellcolor{orange!20}0.011 $\pm$ 0.002 & \cellcolor{orange!20}\textbf{0.009 $\pm$ 0.002} & \cellcolor{orange!20}\textbf{0.009 $\pm$ 0.001} & 0.139 $\pm$ 0.095 & \cellcolor{orange!20}0.011 $\pm$ 0.002 \\
\bottomrule
\end{tabular}%
}
\caption{
\textbf{Expected Calibration Error (ECE) comparison across datasets.}
Lower values indicate better calibration. 
The proposed Multiple FSL consistently achieves highly calibrated probabilities without the need for post-hoc adjustments. 
In bold is the lowest average.
Best results with statistical significance (p < 0.01) according to the Kruskal-Wallis and Dunn’s test are highlighted in orange.
}
\label{tab:expected_calibration_error_complete}
\end{sidewaystable}

\end{document}